\documentclass[nohyperref]{article}

\usepackage{microtype}
\usepackage{graphicx}
\usepackage{graphics}
\usepackage{booktabs} 
\usepackage{multirow}
\usepackage{makecell}
\usepackage{tabularx}

\makeatletter
\def\hlinewd#1{%
\noalign{\ifnum0=`}\fi\hrule \@height #1 %
\futurelet\reserved@a\@xhline}
\makeatother

\usepackage{hyperref}

\usepackage[accepted]{icml2022}

\usepackage{amsmath}
\usepackage{amssymb}
\usepackage{mathtools}
\usepackage{amsthm}

\usepackage{graphicx}
\usepackage{caption}
\usepackage{subcaption}
\usepackage{wrapfig}

\usepackage[capitalize,noabbrev]{cleveref}

\theoremstyle{plain}

\theoremstyle{definition}

\theoremstyle{remark}

\usepackage[textsize=tiny]{todonotes}

\icmltitlerunning{The StarCraft Multi-Agent Challenges Plus}

\begin{document}

\twocolumn[
\icmltitle{The StarCraft Multi-Agent Challenges$^{+}$ : Learning of Multi-Stage Tasks and Environmental Factors without Precise Reward Functions}



\icmlsetsymbol{equal}{*}

\begin{icmlauthorlist}
\icmlauthor{Mingyu Kim}{equal,yyy}
\icmlauthor{Jihwan Oh}{equal,yyy}
\icmlauthor{Yongsik Lee}{yyy}
\icmlauthor{Joonkee Kim}{yyy}
\icmlauthor{Seonghwan Kim}{yyy}
\icmlauthor{Song Chong}{yyy}
\icmlauthor{Se-Young Yun}{yyy}
\end{icmlauthorlist}

\icmlaffiliation{yyy}{KAIST AI, Seoul, South Korea}
\icmlcorrespondingauthor{Se-Young Yun}{yunseyoung@kaist.edu}

\icmlkeywords{Machine Learning, ICML}

\vskip 0.3in
]



\printAffiliationsAndNotice{\icmlEqualContribution} 

\begin{abstract}
 In this paper, we propose a novel benchmark called the StarCraft Multi-Agent Challenges$^{+}$, where agents learn to perform multi-stage tasks and to use environmental factors without precise reward functions. 
The previous challenges (SMAC) recognized as a standard benchmark of Multi-Agent Reinforcement Learning are mainly concerned with ensuring that all agents cooperatively eliminate approaching adversaries only through fine manipulation with obvious reward functions.
 This challenge, on the other hand, is interested in the exploration capability of MARL algorithms to efficiently learn implicit multi-stage tasks and environmental factors as well as micro-control. This study covers both offensive and defensive scenarios. 
 In the offensive scenarios, agents must learn to first find opponents and then eliminate them. The defensive scenarios require agents to use topographic features. For example, agents need to position themselves behind protective structures to make it harder for enemies to attack. 
We investigate MARL algorithms under SMAC$^{+}$ and observe that recent approaches work well in similar settings to the previous challenges, but misbehave in offensive scenarios.
 Additionally, we observe that an enhanced exploration approach has a positive effect on performance but is not able to completely solve all scenarios. This study proposes a new axis of future research. 
\end{abstract}

\section{Introduction}
\label{sec:introduction}

\begin{table*}[!t]
    \caption{List of environmental factors and multi-stage tasks for both SMAC and SMAC$^+$. Both SMAC and SMAC$^+$ employ the same final objective like eliminating all enemies.}
    \label{table:list_sub_tasks}
    \centering
    \resizebox{0.85\textwidth}{!}{%
    \begin{tabular}{ccccc}
    \toprule
    & \multicolumn{2}{c}{SMAC} & \multicolumn{2}{c}{SMAC$^+$} \\
    \cmidrule(r){2-3} \cmidrule(r){4-5}
    & \texttt{2c\_vs\_64zg} & \texttt{corridor}& \texttt{Defense} & \texttt{Offense} \\
    \midrule
    \multirow{3}{*}{\makecell{Environmental  \\ Factors}}       & \multirow{3}{*}{\makecell{Different levels of\\ the terrain}} &  \multirow{3}{*}{\makecell{Limited sight range \\of enemies}} & \multirow{2}{*}{Destroy obstacles hiding enemies}                       & Approach enemies strategically \\
                    &   &   &                                               & Discover a detour$^*$ \\
                    &   &   & Place in less damage zones  & Destroy moving impediments$^*$ \\
    \hline
    \makecell{Multi-Stage \\ Tasks}                &- &\makecell{Avoid enemies first,\\eliminate individually}   &-                                               & \makecell{Identify where enemies locate,\\then exterminate enemies}   \\
    \bottomrule
    \multicolumn{4}{l}{\small $*$ : \texttt{Off\_complicated} scenario} \\
    \end{tabular}
    }%
\end{table*}

The StarCraft Multi-Agent Challenges (SMAC) is recognized as the standard benchmark simulator of Multi-Agent Reinforcement Learning (MARL) studies \citep{samvelyan2019starcraft}. 
Tasks in the SMAC mainly require micro-managed control of agents in defensive situations, where all opponents naturally approach the trained agents. This allows agents to obtain rewards directly. In these environments, the majority of algorithms concentrated on determining the relevance of each agent during training \citep{sunehag2017value, lowe2017multi, rashid2018qmix, hu2021updet, iqbal2021randomized, liu2021coach, sun2021dfac, qiu2021rmix}. 
Some difficult scenarios, such as \texttt{2c\_vs\_64zg} and \texttt{corridor}, on the other hand, require agents to indirectly learn environmental factors, such as exploiting different levels of terrains or discover multi-stage tasks like avoiding rushing enemies first and then eliminating individuals without a specific reward for them. Recently, efficient exploration approaches for MARL algorithms were reported to drastically increase performance in those tough scenarios \cite{sun2021dfac, learning5disentangling}. 
However, those scenarios do not allow quantitative assessment of the algorithm's exploration capabilities, as they do not accurately reflect the difficulty of the task, which depends on the complexity of multi-stage tasks and the significance of environmental factors.

To address this issue, we propose a new class of the StarCraft Multi-Agent Challenges$^{+}$ (SMAC$^{+}$) that encompasses advanced and sophisticated multi-stage tasks, and involves environmental factors agents must learn to accomplish, as seen in \autoref{table:list_sub_tasks}. 
We present three defensive scenarios to encourage agents to employ topographical features such as positioning themselves behind structures to lower the probability of being attacked by enemies. 
In addition, offensive scenarios require agents to initially find the adversaries while also considering topographical obstacles and then rapidly defeating each of the adversarial troops. Like previous challenges, SMAC, in these situations, agents are still rewarded just for eliminating enemies, indicating that they indirectly learn how to do multi-stage tasks and use environmental factors. 
In the experiment results, we compare the performance of 11 MARL algorithms across all scenarios to establish a benchmark. We find that existing approaches perform well in similar settings to the previous challenge, but when environments need to complete sophisticated sub-tasks, most algorithms fail to learn adequately even when the training time is significantly extended. To summarize, we make the following contributions:

\begin{itemize}
    \vspace{-0.02in}    
    \item We propose a novel MARL environment; SMAC$^{+}$. This intends to identify how agents explore to learn sequential completion of multi-stage tasks and environmental factors when reward functions are designed only towards the final objective. 
    \vspace{-0.02in}    
    \item We present an extensive benchmark of MARL algorithms on SMAC$^{+}$. We find that recent MARL algorithms with enhanced exploration demonstrate stable performance in the proposed scenarios, but other baselines cannot be efficiently trained.
    \vspace{-0.02in}    
    \item We suggest the most challenging environments that demand simultaneously learning micro-control and multi-stage tasks. These scenarios are an open-ended problem for efficient exploration toward MARL domains because no algorithm attain satisfactory performance on the challenging scenario.  
\end{itemize}
\vspace{-0.064in}    
\section{The StarCraft Multi-Agent Challenges$^{+}$}
\label{sec:smac+}
We propose a novel multi-agent environment referred to as StarCraft Multi-Agent Challenges$^{+}$ that features a quantitative evaluation of the exploration abilities of MARL algorithms. This challenge offers more advanced and sophisticated environmental factors such as destructible structures that can be used to conceal enemies and terrain features, such as a hill, that may be used to mitigate damages. Also, we newly introduce offensive scenarios that demand sequential completion of multi-stage tasks requiring finding adversaries initially and then eliminating them. Like in SMAC, both defensive and offensive scenarios in SMAC$^+$ employ the reward function proportional to the number of enemies removed. For unit combination, we select units to necessitate environmental factors and completion of multi-stage tasks cooperatively so that the trained agents can validate that they make effective use of these properties. In SMAC$^{+}$, agents must implicitly discover multi-stage tasks and factors by relying on their exploration strategy. Hence, it provides the evaluation of the exploration capability of MARL algorithms by comparing performance in similar but diverse scenarios. We describe the details of our environments in the rest of this section. 

\subsection{General Description of SMAC$^{+}$ : Terrain and Unit Combination}

\begin{figure}[!t]{
    \centering
        \begin{subfigure}{0.30\columnwidth}
            \includegraphics[width=\columnwidth]{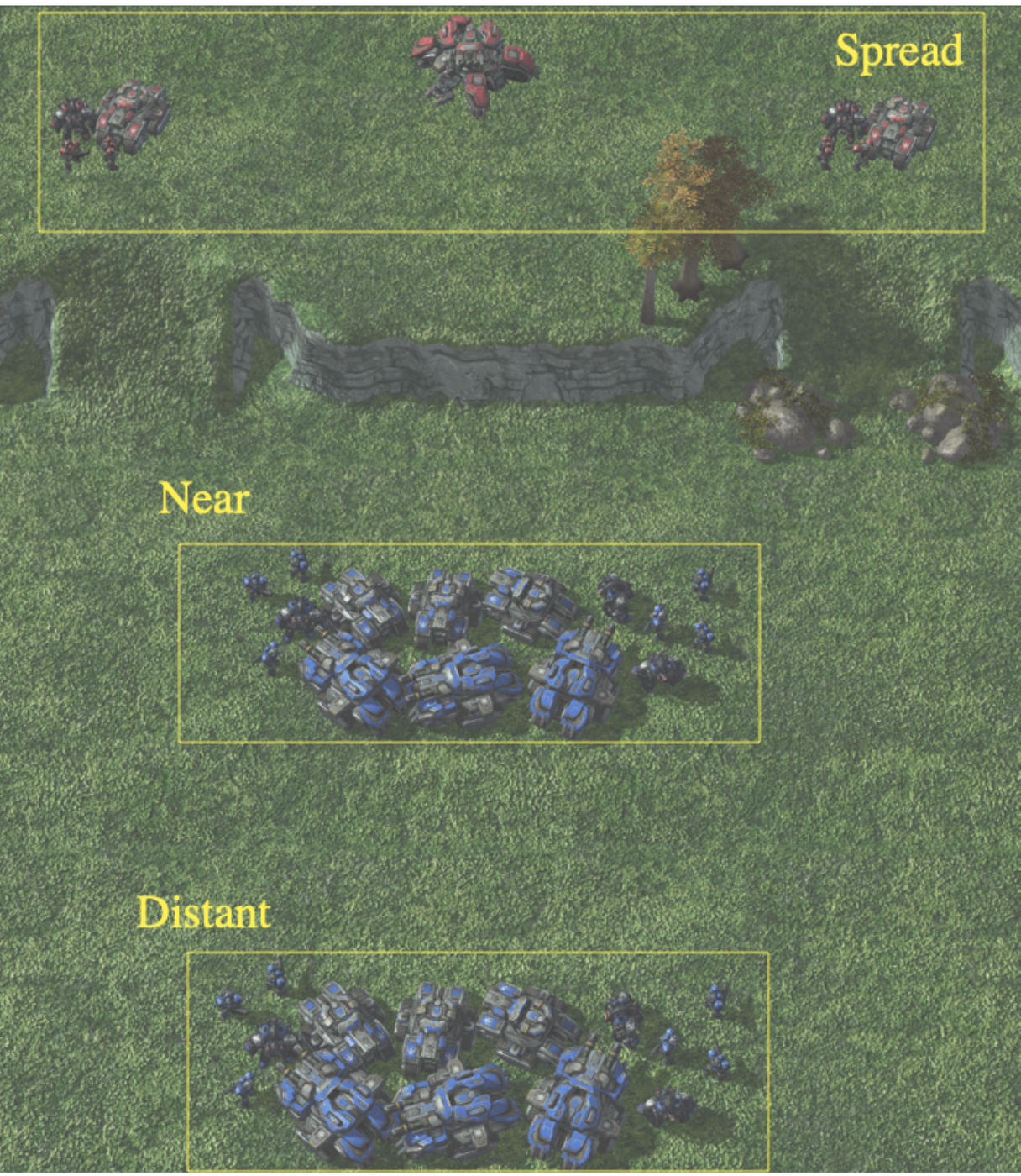}
            \caption{General terrain}
            \label{fig:terrain_feature}
        \end{subfigure}%
        \hspace{0.1cm}
        \begin{subfigure}{0.30\columnwidth}
            \includegraphics[width=\columnwidth]{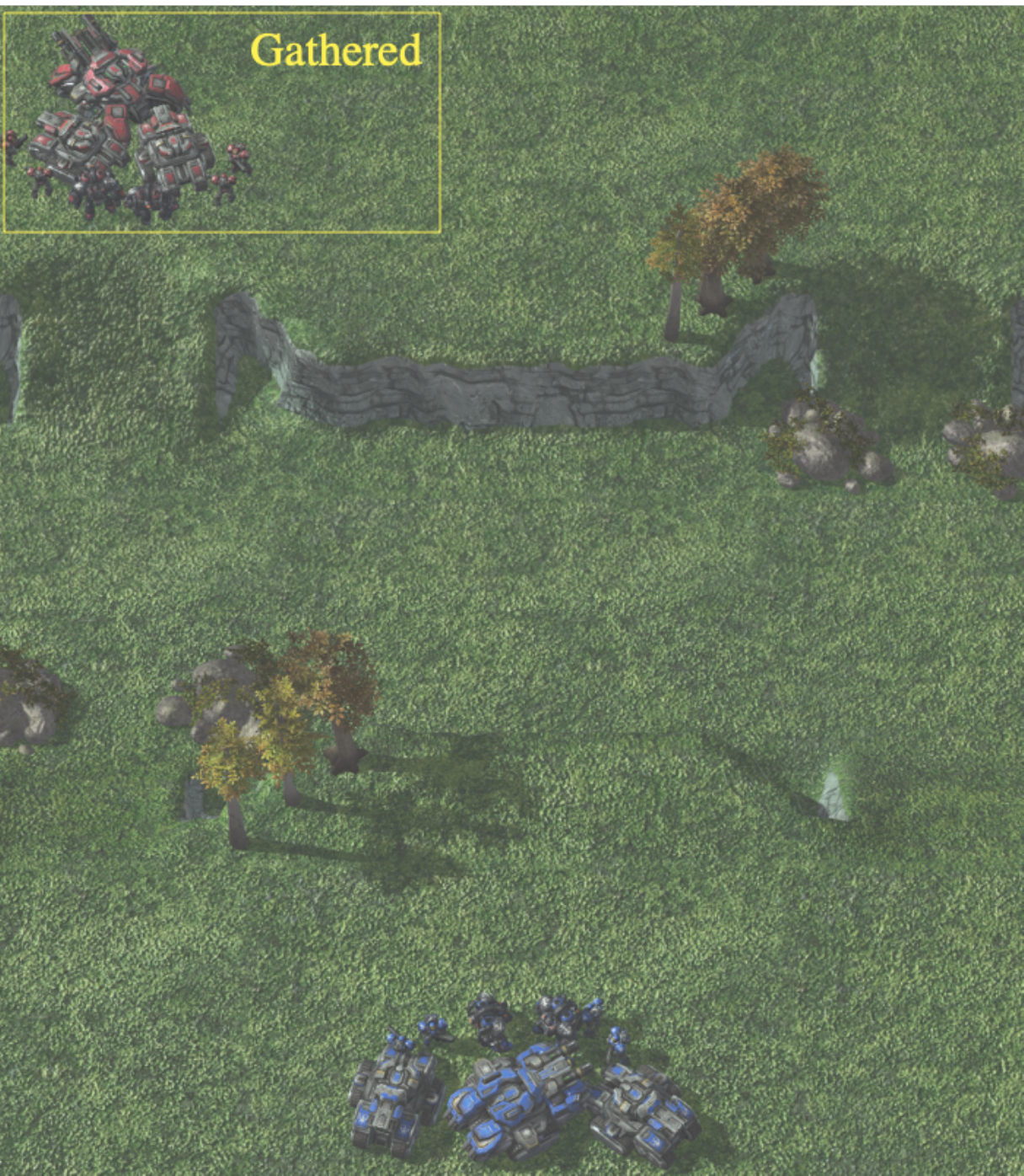}
            \caption{Complex terrain}
            \label{fig:complicated_terrain_feature}
        \end{subfigure}%
        \hspace{0.1cm}
        \begin{subfigure}{0.31\columnwidth}
            \includegraphics[width=\columnwidth]{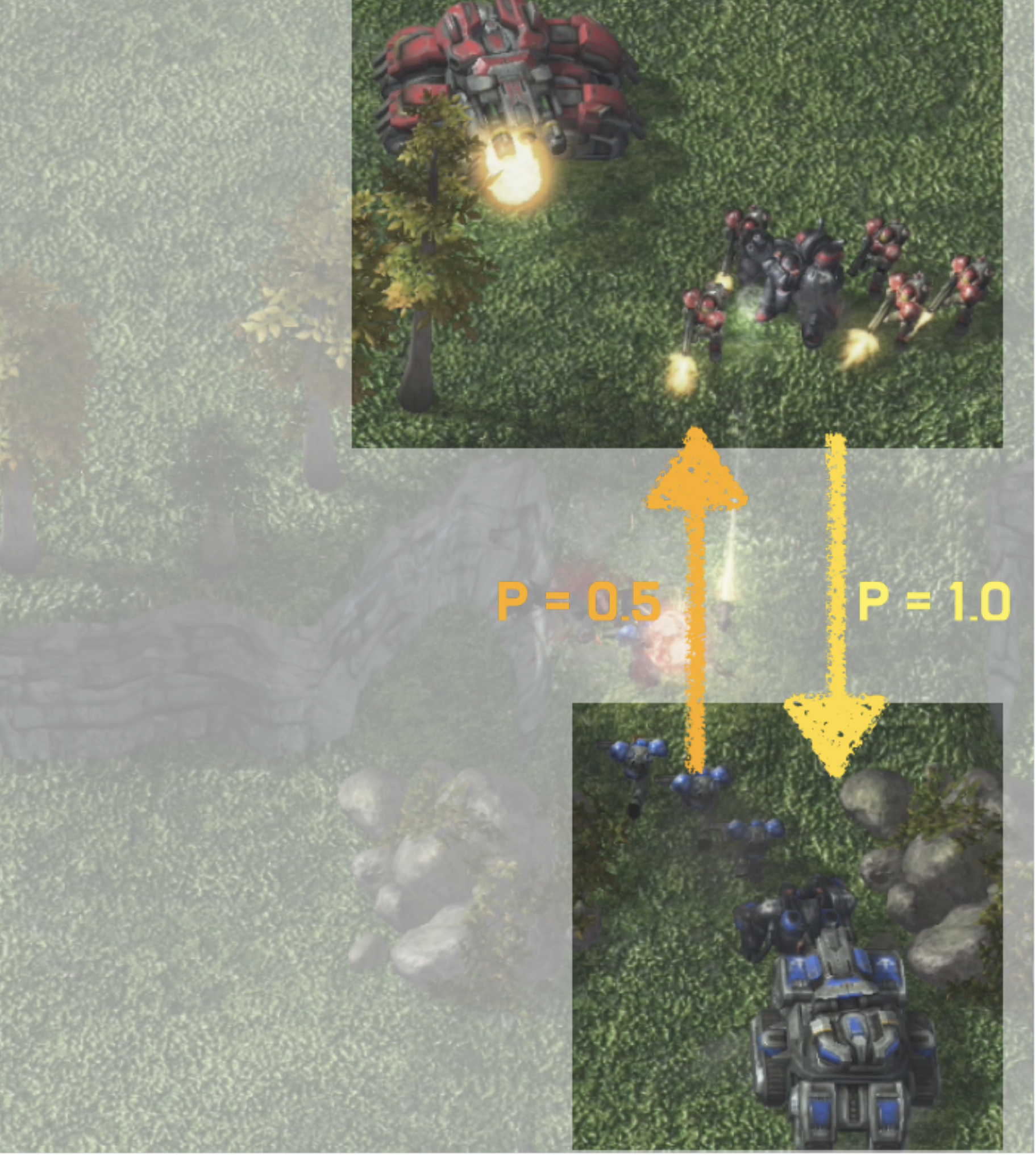}
            \caption{Hill advantage}
            \label{fig:hill_advantage}
        \end{subfigure}%
    \label{overview}
    \caption{Summary of environmental factors in the SMAC$^{+}$. The defense troop is positioned on the hill in preparation for combat, while the offensive troop placed on the lower side goes forward to the defense troops. The yellow line describes an attack by troops on the hill, while the orange line indicates an attack from below the hill.}
    }
\end{figure}

We design topographical features like trees, and stones, which block each unit's sight equally for both allies and opponents. Another topographic feature is the hill, which provides a stochastic environment for damage dealing, as seen in \autoref{fig:hill_advantage}. When an attacker stationed on the hill strikes opponents in the plain, the yellow line indicates deterministic damage dealing with a 100\% probability. On the other hand, the orange line shows probabilistic damage with a 50\% when an attacker positioned on the plain assaults an opponent placed on the hill. This is quite reasonable in the sense that someone in a topographically high position may easily hurt someone, but not vice versa. Hence, exploration of MARL algorithms must encourage agents to find these factors during the training phase, as agents are not explicitly rewarded for using these features. To determine the aforementioned factors, we devise a combination of units that can validate the cooperative decision-making of several agents. The role of each type of agent is presented in \autoref{table:unit}. Each unit in SMAC$^{+}$ has a unique set of attributes, such as shooting range, sight range, and firepower. Thus, it becomes important to determine how to combine and locate agents in battles considering various unit types. We explain more details in \autoref{app:starcraft_details}. 


\begin{table}[!t]
    \caption{List of units and their roles in SMAC$^+$}
    \label{table:unit}
    \centering
    \resizebox{0.99\columnwidth}{!}{%
    \begin{tabular}{cc}
    \toprule
    Unit & Role \\
    \midrule
    Marine   & Intensive firepower / Search for opponent positions \\
    Marauder & Damage absorption from enemy / Enemy's movement restriction \\
    Tank      & Strong firepower against individuals / Limited range of fire \\
    Siege Tank & Firepower against groups / Limited range of sight / Protection from close-range attack \\
    \bottomrule
    \end{tabular}
    }%
\end{table}

\subsection{Defensive Scenarios}
In defensive scenarios, we place allies on the hill and adversaries on the plain. We emphasize the importance of agents defeating adversaries utilizing topographical factors. The defensive scenarios in SMAC$^{+}$ are almost identical to those in SMAC. However, our environment expands the exploration range of allies to scout for the direction of offense by allowing enemies to attack in several directions and adding topographical changes.

\begin{table}[!t]
    \caption{Taxonomy of offensive and defensive scenarios. The term ``Supply difference" refers to the gap in populations between opponents and allies in the StarCraft2. Enemies are scattered in all offensive scenarios such that they cannot fire fiercely, allowing agents to easily defeat them when they engage.}
    \label{table:taxonomy_scenarios}
    \centering
    \resizebox{0.99\columnwidth}{!}{%
    \begin{tabular}{ccc|cc}
    \toprule
                             \multicolumn{3}{c|}{\bf Defensive scenario} & \multicolumn{2}{c}{\bf Offensive scenario} \\
                            \cmidrule(r){1-5} 
                            Scenario & Supply difference     & Opponents approach    &Scenario      &Distance from opponents\\
    \midrule
    \texttt{Def\_infantry}           & -2                    & One-sided    &\texttt{Off\_near}      & Near          \\
    \texttt{Def\_armored}           & -6                    & Two-sided     &\texttt{Off\_distant}      & Distant       \\
    \texttt{Def\_outnumbered}     & -9                    & Two-sided          &\texttt{Off\_complicated}      & Complicated   \\
    \bottomrule
    \end{tabular}
    }%
\end{table}

\subsection{Offensive Scenarios}

\begin{table}[!t]
    \caption{The list of the most challenging offensive scenarios. New scenarios require to simultaneously learn micro-control and multi-stage tasks.}
    \label{table:taxonomy_off_scenarios}
    \centering
    \resizebox{0.99\columnwidth}{!}{%
    \begin{tabular}{cccc}
    \toprule
    Scenario                            &  \makecell{Supply\\ difference}    & \makecell{Distance\\ from opponents}   & \makecell{Opponents\\ formation} \\
    \midrule
    \texttt{Off\_hard}    & 0                     & Complicated               & Spread \\
    \texttt{Off\_superhard}    & 0                     & Complicated               & Gather\\
    \bottomrule
    \end{tabular}
    }%
\end{table}

Offensive scenarios provide learning of multi-stage tasks without direct incentives in MARL challenges. We suggest that agents should accomplish goals incrementally, such as eliminating adversaries after locating them. To observe a clear multi-stage structure, we allocate thirteen supplies to the allies more than the enemies. Hence, as soon as enemies are located, the agents rapidly learn to destroy enemies. As detailed in \autoref{table:list_sub_tasks}, in SMAC$^{+}$, agents will not have a chance to get a reward if they do not encounter adversaries. This is because there are only three circumstances in which agents can get rewards: when agents defeat an adversary, kill an adversary, or inflict harm on an adversary. As a result, the main challenges necessitate not only micro-management, but also exploration to locate enemies. For instance, the agents learn to separate the allied troops, locate the enemies, and effectively use armored troops like a long-ranged siege Tank. We measure the exploration strategy of effectively finding the enemy through this scenario. In this study, we examine the efficiency with which MARL algorithms explore to identify enemies by altering distance from them. In addition, to create more challenging scenarios, we show how enemy formation affects difficulty. 

\section{Experiments}
\label{sec:experiments}

\begin{table*}[!t]
    \caption{Average win-rate (\%) performance of QMIX, DRIMA, COMA and MADDPG. All methods used sequential episodic buffers. Note that MADDPG is only compatible with the sequential episodic buffer.}
    \label{table:final_winrate_episode}
    \centering
    \resizebox{0.80\textwidth}{!}{%
    \begin{tabular}{cccccccc}
    \toprule
        & \multirow{2}{*}{Trial} & \multicolumn{3}{c}{\bf Defensive scenarios} & \multicolumn{3}{c}{\bf Offensive scenarios} \\
    \cmidrule(r){3-8}
        & & \texttt{infantry} & \texttt{armored} & \texttt{outnumbered} & \texttt{near} & \texttt{distant} & \texttt{complicated} \\
    \midrule

\multirow{3}{*}{COMA\cite{foerster2018counterfactual}}   &1  & 75.0 &  0.0 &  0.0 &  0.0 & 0.0 & 0.0 \\
                        &2  & 28.1 &  0.0 &  0.0 &  0.0 & 0.0 & 0.0 \\
                        &3  & 21.9 &  0.0$^{1)}$ &  0.0 &  0.0 & 0.0 & 0.0$^{2)}$ \\
\cmidrule(r){1-8}
\multirow{3}{*}{QMIX\cite{rashid2018qmix}}   &1  & 100 &  100 &  3.1 &  0.0 & 0.0 & 100 \\
                        &2  & 93.8 &  0.0 &  0.0 &  100.0 & 100.0 & 87.5 \\
                        &3  & 96.9 &  0.0 &  0.0 &  90.6 & 93.8 & 0.0$^{3)}$ \\
\cmidrule(r){1-8}
\multirow{3}{*}{MADDPG\cite{lowe2017multi}} &1  & 100 &  96.9 &  81.3 & 0.0 & 90.6 & 0.0 \\
                        &2  & 100 &  84.4 &  81.3 &  75.0 &  0.0 & 75.0 \\
                        &3  & 100 &  90.6 &  71.9 &  100.0 & 0.0 & 0.0 \\
\cmidrule(r){1-8}
\multirow{3}{*}{DRIMA\cite{learning5disentangling}}  &1  & 100 &  100 &  100 & 93.8 & 100$^{4)}$ & 96.9 \\
                        &2  & 100 &  96.9 &  96.9 &  93.8 & 100 & 100 \\
                        &3  & 100 &  100 &  100 & 100 & 100 & 96.9 \\
\hline\hline
\multicolumn{2}{c}{The total number of win-rate $\ge 80\%$} &9 &7 &5 &6 &6 &5\\ 
    \bottomrule
    \multicolumn{8}{l}{1) Takes total cumulative 3.29 million episode steps during training} \\
    \multicolumn{8}{l}{2) Takes total cumulative 4.21 million episode steps during training} \\
    \multicolumn{8}{l}{3) Takes total cumulative 4.59 million episode steps during training} \\
    \multicolumn{8}{l}{4) Takes total cumulative 2.53 million episode steps during training} \\
    \end{tabular}
     }%
\end{table*}

In this section, we describe experimental results to answer three key questions: \emph{1) Do the proposed scenarios provide a quantitative assessment of exploration capability by varying complexity of multi-stage tasks and the significance of environmental factors? 2) Do existing MARL algorithms efficiently utilize exploration to discover environmental factors and the completion of multi-stage tasks? 3) Can these algorithms reliably perform well on SMAC$^+$ despite repeated training?} 

To demonstrate the need for assessment of exploration capabilities, we choose eleven algorithms of MARL algorithms classified into three categories; policy gradient algorithms, typical value-based algorithms, and distributional value-based algorithms. First, as an initial study of the MARL domain, policy gradient algorithms such as COMA \cite{foerster2018counterfactual}, MASAC \cite{haarnoja2018soft}, MADDPG \cite{lowe2017multi} are considered. The typical value-based algorithm including IQL \cite{tan1993multi}, VDN \cite{sunehag2017value}, QMIX \cite{rashid2018qmix} and QTRAN \cite{son2019qtran} are chosen as baselines. Last but not least, we choose DIQL, DMIX, DDN \cite{sun2021dfac} and DRIMA \cite{learning5disentangling} as distributional value-based algorithms that recently reported high performance owing to the effective exploration of difficult scenarios in SMAC.

\begin{figure}{
    \centering
        \begin{subfigure}{0.45\columnwidth}
        \includegraphics[width=\columnwidth]{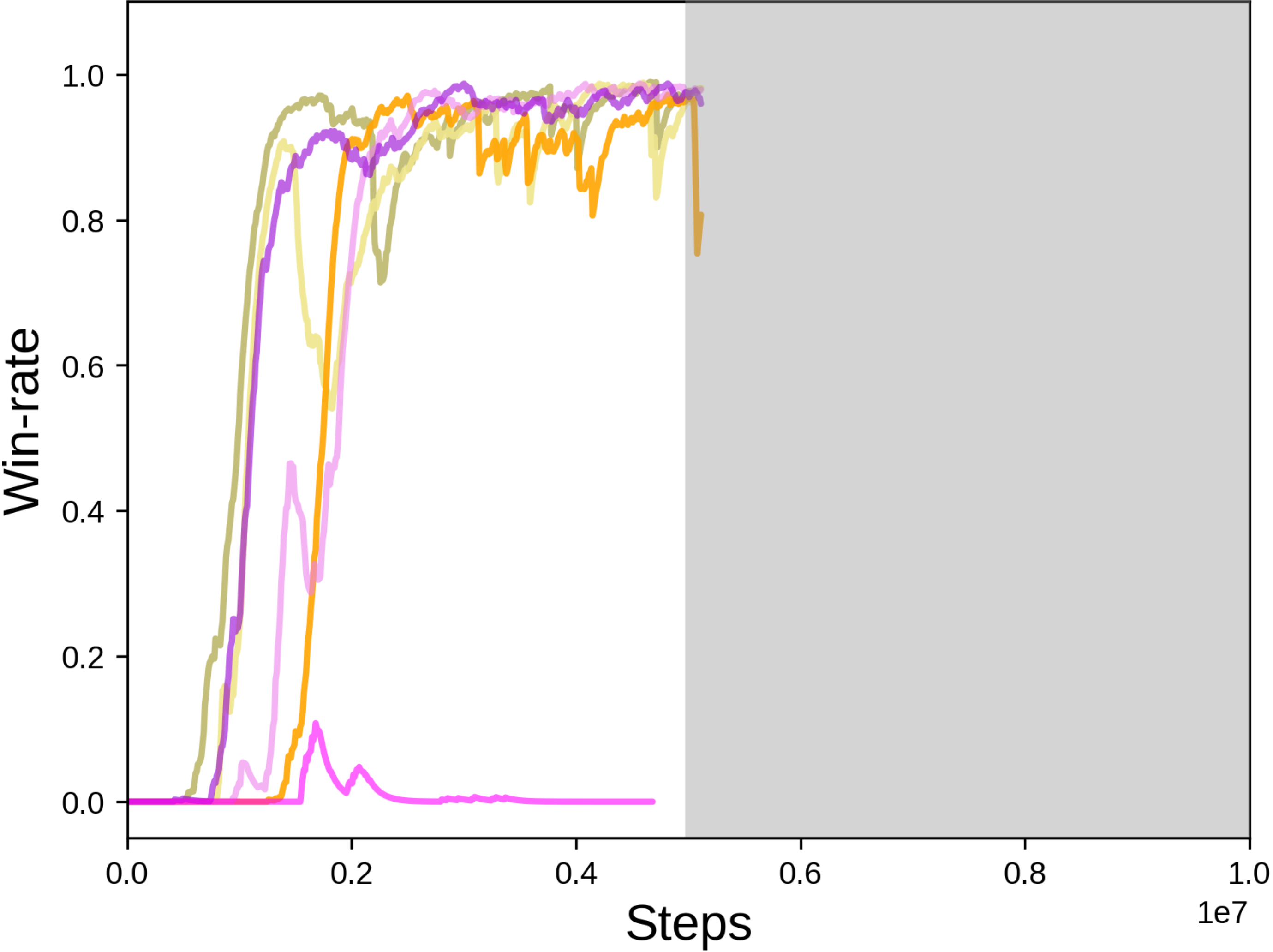}
        \caption{Sequential episodic buffer}
        \label{fig:parallel_off_superhard_qmix}
        \end{subfigure}
        \begin{subfigure}{0.45\columnwidth}
        \includegraphics[width=\columnwidth]{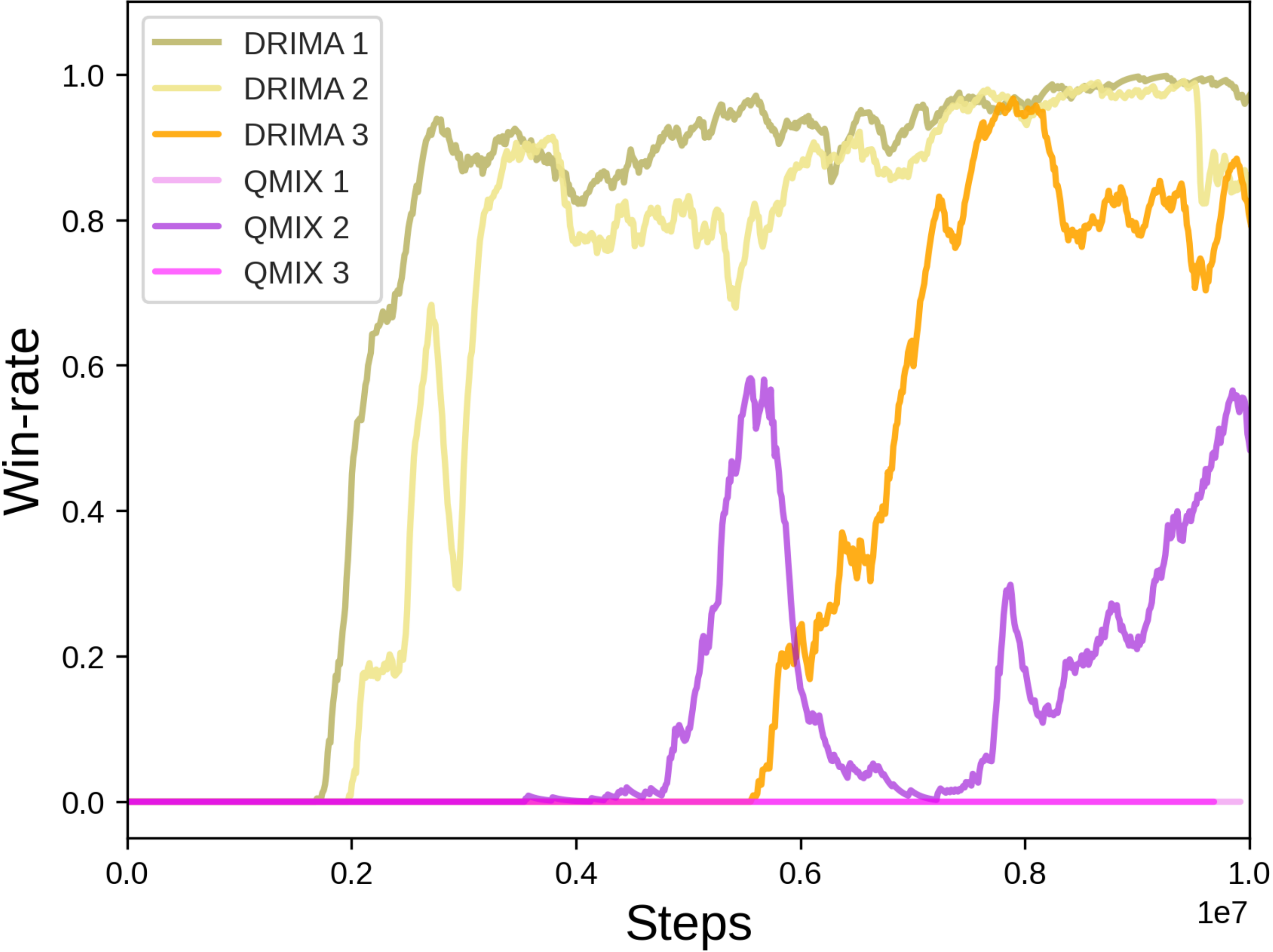}
        \caption{Parallel episodic buffer}
        \label{fig:parallel_off_superhard_drima}
        \end{subfigure}
    \caption{Average win-rates of QMIX and DRIMA according to the cumulative episodic steps during training. Two baselines are respectively trained three times. (a) learning curves of the sequential episodic setting for 5 million steps. (b) learning curves of parallel setting for 10 million steps. The horizontal axis means the total number of cumulative episode steps during the training, and the vertical axis is the win-rate.}
    \label{fig:off_results}
    }
\end{figure}

In order to look into exploration capability, all baselines are trained in this experiment until the total number of cumulative episode steps respectively reaches five million steps for a sequential episodic setting. We respectively train each baseline three times and report average win-rates as the evaluation metric by conducting 32 test-runs in the episodic setting. Initially, we demonstrate the SMAC$^+$ benchmark utilizing sequential episodic buffers. We report experimental results by choosing a representative algorithm from each category because of the massive training time; MADDPG \cite{lowe2017multi}, COMA \cite{foerster2018counterfactual}, QMIX \cite{rashid2018qmix} and DRIMA \cite{learning5disentangling}.
We also conducted on sequential episodic buffer setting and note that the tendency of experimental results remains unchanged as shown in \autoref{fig:off_results}, allowing us to analyze the exploration capabilities of MARL algorithms based on results of the parallel episodic buffer. More training information on sequential episodic buffer, details about algorithms and experiment results are respectively documented at \autoref{app:training}, \autoref{app:algorithms}, \autoref{app:detailed_experiments} and \autoref{app:Ablation_study}. 

\begin{table}[!t]
    \caption{Average win-rate (\%) performance on more challenging offensive scenarios. These scenarios require to address multi-stage tasks and micro-control at once.}
    \label{table:DRIMA_additional_results}
    \centering
    \resizebox{0.99\columnwidth}{!}{%
    \begin{tabular}{cccccc}
    \toprule
            & \multirow{3}{*}{Trial} & \multicolumn{2}{c}{\bf Episodic buffer} & \multicolumn{2}{c}{\bf Parallel buffer} \\
            \cmidrule(r){3-4} \cmidrule(r){5-6}
            &                   & \texttt{Off\_hard}  & \texttt{Off\_superhard} & \texttt{Off\_hard} & \texttt{Off\_superhard} \\
    \midrule
    \multirow{3}{*}{DRIMA\cite{learning5disentangling}}        
    &1  & 96.9 & 15.6 &  100 & 0.0 \\
    &2  & 93.8 & 3.1 &  80.0 &  0.0\\
    &3  & 93.8 & 15.6  &  0.0 &  0.0\\
    \bottomrule
    \end{tabular}
     }%
\end{table}

\subsection{Benchmark on Sequential Episodic Buffer}
We first look into defensive scenario experiments on SMAC$^+$ to test whether MARL algorithms not only adequately employ environmental factors but also learn micro-controls. As seen in \autoref{table:final_winrate_episode}, we find that supply difference and opponents' approach manipulate the difficulty, as the majority of baselines perform worse in response to those variants. In terms of algorithmic performance, we observe COMA and QMIX drastically degrade, but MADDPG gradually degrades. This fact reveals that MADDPG enables agents to effectively learn micro-control. However, among baselines, DRIMA achieves the highest score and retains performance even when the supply difference significantly increases. This is due to the fact that DRIMA efficiently explores micro-control but also environmental factors. This finding indicates that effective exploration uncovers intrinsic environmental factors. Regarding to offensive scenarios, we notice considerable performance differences of each baseline. Overall, even if an algorithm attains high scores at a trial, with exception of DRIMA, it is not guaranteed to train reliably in other trials. As mentioned, offensive scenarios do not require as much high micro-control as defensive scenarios, instead, it is important to locate enemies without direct incentives, such that when agents find enemies during training, the win-rate metric immediately goes to a high score. However, the finding enemies during training is decided by random actions drawn by $\epsilon$-greedy or probabilistic policy, resulting in considerable variance in test outcome. In contrast, we see a perfect convergence of DRIMA in all offensive scenarios by employing its efficient exploration. 


\subsection{Evaluation on Challenging Scenarios}
As previously stated, we identify that DRIMA reliably solves all offensive scenarios. To provide open-ended problems for the MARL domain, we suggest more challenging scenarios as shown in \autoref{table:taxonomy_off_scenarios}. In these scenarios, the agents are required to simultaneously learn completion of multi-stage tasks and micro-control during training. DRIMA produces remarkable performance in past experiments, so that we provide the results acquired by DRIMA to verify that the proposed scenarios are sufficiently challenging. We use the same experimental condition as in the earlier experiments. As you can see \autoref{table:DRIMA_additional_results}, DRIMA still solves \texttt{Off\_hard}, although its performance on \texttt{Off\_superhard} is negligible. We argue that this scenario requires more sophisticated fine manipulation compared to other offensive scenarios. This is due to the fact that not only the strength of allies is identical to that of opponents, but also \textit{Gathered} enables opponents to intensively strike allies at once. This indicates the necessity of more efficient exploration strategies for the completion of multi-stage tasks and micro-control.

\section{Conclusion}
\label{sec:conclusion}

We propose SMAC$^{+}$, a suite of environments to learn multi-stage tasks and environmental factors without specific rewards. We develop new MARL environments based on the previous challenge \cite{samvelyan2019starcraft}. This point allows us to ensure that all baselines are completely compatible with our environments. Consequently, we evaluate a total eleven MARL algorithms on both defensive and offensive scenarios, and their experimental results show that an efficient exploration strategy is required to learn multi-stage tasks and environmental factors. We hope this work serves as a valuable benchmark to evaluate the exploration capabilities of MARL algorithms and give guidance for future research.

\section*{Acknowledgements}
\label{sec:acknowledgements}

This work was conducted by Center for Applied Research in Artificial Intelligence (CARAI) grant funded by DAPA and ADD (UD190031RD). 


\appendix
\setcounter{equation}{0}
\renewcommand{\theequation}{\Alph{section}.\arabic{equation}}
\setcounter{table}{0}
\renewcommand{\thetable}{\Alph{section}.\arabic{table}}
\setcounter{figure}{0}
\renewcommand{\thefigure}{\Alph{section}.\arabic{figure}}


\bibliography{icml2022}
\bibliographystyle{icml2022}

\newpage
\appendix
\onecolumn
\section*{Author Statement} The authors bear all responsibility in case of violation of rights, etc., and confirmation of the data license.

\section*{Accessibility} SMAC$^{+}$ environment is publicly available at the Github repository: \url{https://github.com/osilab-kaist/smac_plus} and we will maintain the released code for long-term accessibility.

\section*{License} The original SMAC environment and PyMARL code follow the MIT license and Apache 2.0 license respectively. The proposed SMAC$^{+}$ environment and the modified PyMARL code are also released under the MIT license and Apache 2.0 license each. The details of licenses can be found in our repository.

\section*{Reproducibility} The \path{README} file in the repository serves as a guide for installation and training. Several \path{yaml} files in \path{pymarl/src/config} directory contains the parameters we used for the paper. Further training details can be found in \autoref{app:training}.
\section{Specification of StarCraft Multi-Agent Challenge$^{+}$}
\label{app:starcraft_details}

\subsection{Environment}
\paragraph{State and Observation features} 
The features of original SMAC environment \cite{samvelyan2019starcraft} are described in from \autoref{smac:global_state} to \autoref{smac:observation_feature_1}. $ut$ means the number of different unit types in the given scenarios. Each agent has 6 moving actions and attacks a particular enemy, so the the dimension of action space is 6 plus the number of enemies. The global state information is provided during the centralized training phase while agents must utilize its own local observation during the decentralized execution phase. 

\begin{table}[!h]
\caption{Global state information in SMAC}
\label{smac:global_state}
\centering
\resizebox{0.99\columnwidth}{!}{%
\footnotesize
\begin{tabular}{c|c|c p{0.5\linewidth}p{0.5\linewidth}}
\toprule
 \textbf{Feature} & \textbf{Description} & \textbf{Dimension}  \\ 
\midrule
Ally & \makecell{\{health, cooldown, relative x, \\relative y, unit type, last action\}} & \makecell{The number of allies\\ $\times(4+ut+6+$number of enemies$)$} \\
Enemy & \{health, relative x, relative y, unit type\} & The number of enemies $\times (3+ut)$ \\
\hline
\hline
Total & \{Ally $+$ Enemy\} & \makecell{\{The number of allies $\times (4+ut+6+$number of enemies$)$ $+$ \\ The number of enemies $\times (3+ut)$\}} \\
\bottomrule
\end{tabular}
}
\end{table}

\begin{table}[!h]
\caption{Observation information in SMAC}
\label{smac:observation_feature_1}
\centering
\resizebox{0.99\columnwidth}{!}{%
\footnotesize
\begin{tabular}{c|c|c p{0.5\linewidth}p{0.5\linewidth}}
\toprule
 \textbf{Feature} & \textbf{Description} & \textbf{Dimension}  \\ 
\midrule
Move & basic movement & 4 \\
Own & \{health, unit type\} & $1+ut$ \\
Ally & \{visibility, health, distance, relative x, relative y, unit type\} & The number of allies $\times (5+ut)$ \\
Enemy & \{visibility, health, distance, relative x, relative y, unit type\} & The number of enemies $\times (5+ut)$ \\
\hline
\hline
Total & \{Move, Own, Ally, Enemy\} & \makecell{\{4 + ($1+ut$) + (The number of allies $\times (5+ut)$) \\ (The number of enemies $\times (5+ut)$)\}} \\ 
\bottomrule
\end{tabular}
}
\end{table}

The observation features in SMAC$^+$ are listed in \autoref{smacplus:observation_feature_2}. The StarCraft2 in-game properties intrinsically provide additional information of terrain levels, such as pathing grid and terrain height. We consider terrain features like hill and entrance, so that agents must distinguish opponents are located higher or not in these scenarios. Furthermore, all agents can receive the last actions of all units within its sight range as a part of observations. These features are also provided in SMAC but not included in default setting. In addition, we incorporate the coordinate information perpendicular to the terrain surface and the agent can have its own coordinate values. The unit type represents a total of eight entities in this scenario; Marine, Marauder, General Tank (which cannot switch to siege mode), Siege Tank in general mode, Siege Tank in siege mode (because siege tank may pick their mode) and three neutral building. The dimension of the last action is specifically described in \autoref{action_space}. Overall, an agent's observation space is composed of \{\texttt{movement feature, agent's own feature, enemies feature, allies feature, neutral buildings' feature}\} which are all visible when objects are within the agent's sight range. As the global state information, SMAC$^+$ provides two options. We might use same global information of SMAC, or a concatenation of all agents' observations. We empirically found that a concatenation of all observations gave better performance across SMAC$^+$ scenarios, hence we use concatenated observations as the global state information in default.

\begin{table}[!h]
\caption{Observation information in SMAC$^{+}$}
\label{smacplus:observation_feature_2}
\centering
\resizebox{0.99\columnwidth}{!}{%
\footnotesize
\begin{tabular}{c|c|c p{0.5\linewidth}p{0.5\linewidth}}
\toprule
 \textbf{Feature} & \textbf{Description} & \textbf{Dimension}  \\ 
\midrule
Move & \{basic movement, pathing grid value\} & $4+8$ \\
Own & \{health, x, y, z, unit type\} & $4+ut$ \\
Ally & \makecell{\{visibility, health, distance, \\ relative x, relative y, unit type, last action \}}& \makecell{The number of allies $\times$ \\ $(14+7+$ The number of enemies and neutrals$)$} \\
Enemy & \makecell{\{visibility, health, distance, \\ relative x, relative y, relative z, unit type\}} & The number of enemies $\times$ 14 \\
Neutral & \makecell{\{visibility, health, distance, \\ relative x, relative y, relative z, unit type\}} & The number of neutrals $\times$ 14 \\
\hline
\hline
Total & \makecell{\{Move, Own, Ally, \\Enemy, Neutral building \}} & \makecell{\{($4+8$) + ($4+ut$) + (The number of allies $\times$ \\ (14+7+ The number of enemies and neutrals)) + \\ 14 $\cdot$ (The number of enemies + The number of neutrals)\}} \\
\bottomrule
\end{tabular}
}
\end{table}

\paragraph{Action space} The total action set in SMAC$^+$ consists of basic action, number of enemies, and number of neutral buildings as shown in \autoref{action_space}. The basic action is almost identical to SMAC, but we additionally consider an units skill. The unit skill is designed for a general tank to change to siege mode and vise versa. If others units choose unit skill action, it regards as stop action. We also add the number of neutral buildings into the action set because those can be removed by agents if needed. Therefore, the action space of SMAC$^+$ is much larger than SMAC.

\begin{table}[!h]
\caption{Action space in SMAC$^{+}$}
\label{action_space}
\centering
\footnotesize
\begin{tabular}{c|c|c p{0.5\linewidth}p{0.5\linewidth}}
\toprule
\textbf{Basic action} & \textbf{Attack enemies} & \textbf{Attack neutral buildings} \\ 
\midrule
North, South, East, West, No-op, Stop, Skill & Enemy 1, $\cdots$ Enemy n & Builidng 1, $\cdots$, Building n\\
\bottomrule
\end{tabular}
\end{table}

\paragraph{Unit features} All units sight range and shooting range are different as shown Table \autoref{range}. The fire power has two types. Basically, every unit has its own fire power. Except marines, the remaining units give a special damage according to unit types. Tank has a machinery and heavy armor character, Marauder has a heavy armor character and Marine has nothing special. When seeing the Table \autoref{range}, Marauder has a enhanced fire power on units who have machinery attributes. Similarly, Tank has a enhanced fire power on units who have heavy armor attributes. We reduced fire power of Siege Tank because of the property of long-ranged and splash damage for game balance. So according to the scenario and training step, usage of Siege Tank might be different.

\begin{table}[!h]
\caption{Shooting, Sight range \& Fire power in SMAC$^{+}$. Fire powers are categorized into two types. The left one is default fire power and the right one is enhanced fire power according to the opponent's characteristic.}
\label{range}
\centering
\begin{tabular}{c|cccc p{0.5\linewidth}p{0.5\linewidth}}
\toprule
\textbf{Range} & \textbf{Marine} & \textbf{Marauder} & \textbf{Tank} & \textbf{Siege Tank} \\ 
\midrule
\textbf{Shooting} & 6 & 7 & 8 & 17 \\
\textbf{Sight} & 9 & 9 & 9 & 9 \\
\textbf{Fire power} & 6 & 10 / 30 & 15 / 25 & 5 / 10 \\

\bottomrule
\end{tabular}
\end{table}

\begin{figure}[!ht]{
    \centering
        \begin{subfigure}{0.32\columnwidth}
            \includegraphics[width=\columnwidth]{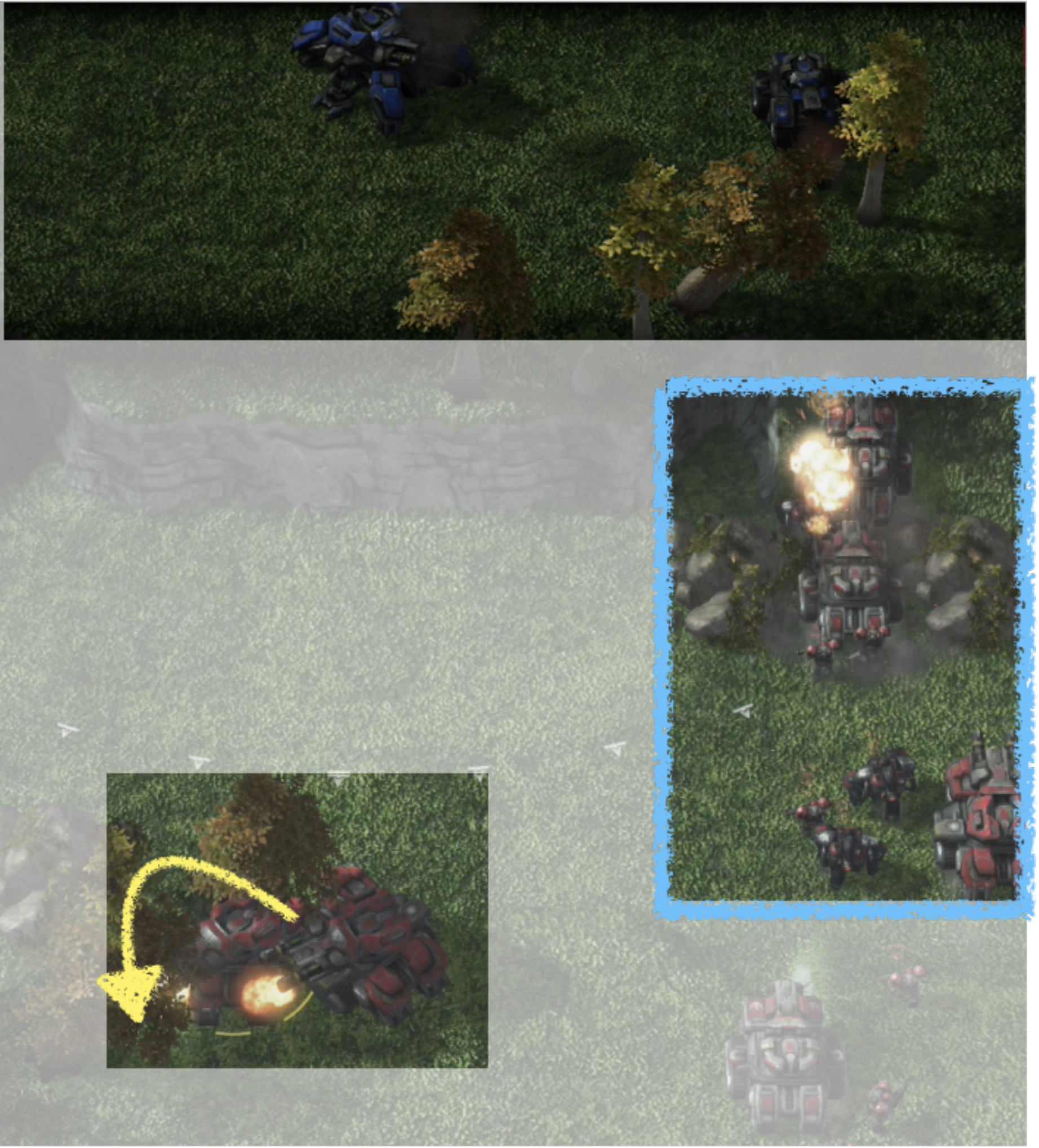}
            \caption{Initial sight of siege tanks}
            \label{fig:initial_sight}
        \end{subfigure}%
        \begin{subfigure}{0.32\columnwidth}
            \includegraphics[width=\columnwidth]{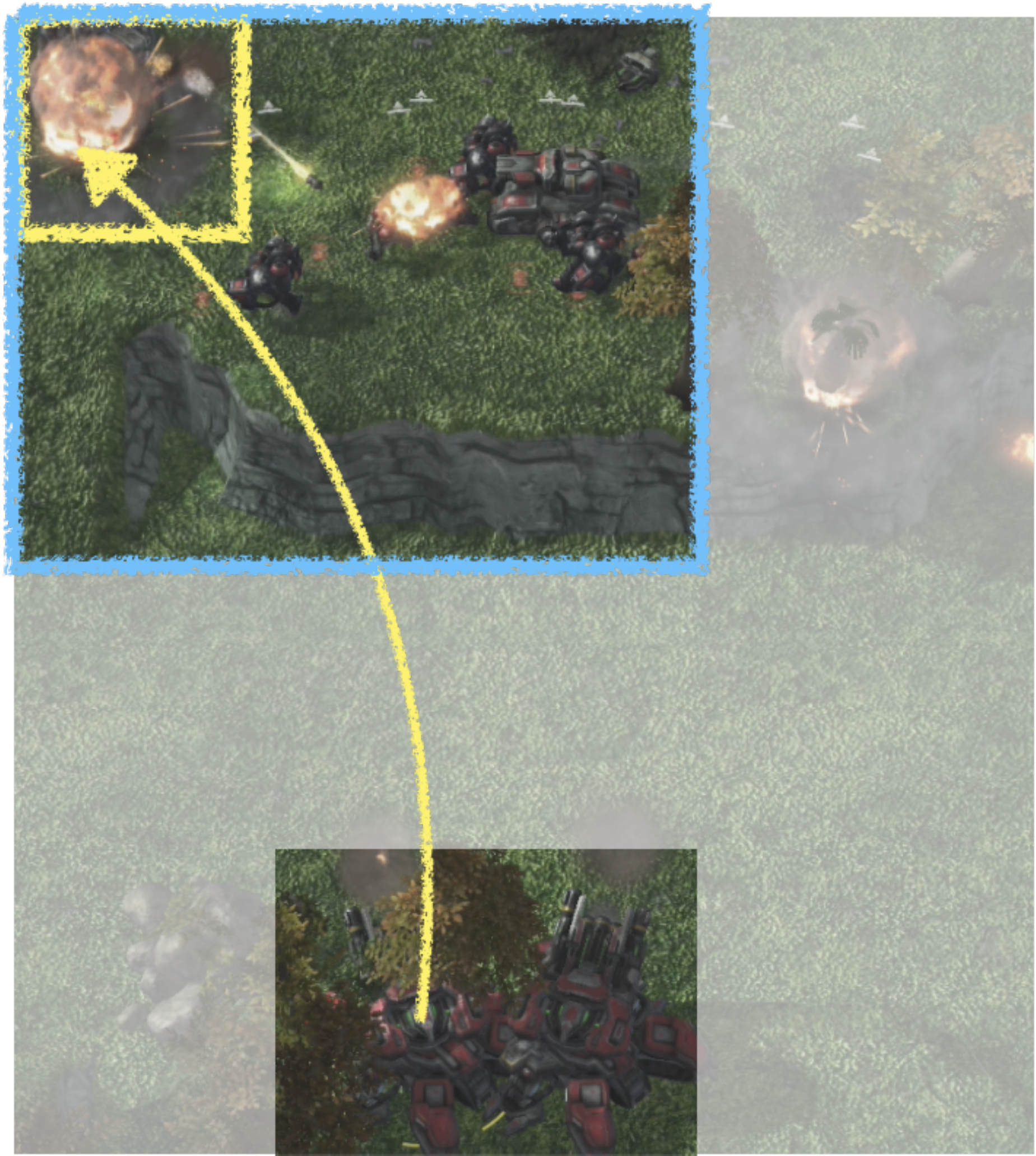}
            \caption{Distant attack by communication}
            \label{fig:attack_beyond_sight}
        \end{subfigure}%
    \caption{The explanation of environmental factors in the SMAC$^{+}$. The yellow line describes range attack by armored troops. The blue rectangle represents the alliance sights between the agent and the alliance.}
    \label{fig:app_communication}
    }
\end{figure}

\paragraph{Communication among agents} 
In this paper, we study a new type option of MARL referred as communicating with all agents. It means all agents share partial information of observations for improving cooperation. For example, all units can access opponent's position when one of allies observe even though opponents are located outside of the unit's sight range. This assumption still assures a decentralized evaluation because none of agents utilize absolute state information and only share a small fraction of observations throughout the test time. We argue that this assumption more accurately reflects the actual challenge because real-world communication technology is being rapidly developed. Specifically, the communication option allows agents to share a portion of their observations when they are within certain agents' sight range. When you see \autoref{fig:initial_sight}, due to limited field of view, armored troops that are capable of long-distance assaults, such as tank and siege-armed tank, cannot engage the adversaries. However, the communication option enables armored troops attack distant enemies by utilizing ally sights as seen \autoref{fig:attack_beyond_sight}. These attributes help each agent to move in a more cooperative manner. 

We technically describe a communication option. The original SMAC environment uses a visibility matrix which is similar to adjacency matrix to indicate visible units for each agent. The dimension of the row is the number of the agents while the dimension of the column is the all the units in the environment including the agents. The entity of matrix $(i,j)$ is 1 if unit $j$ is in the sight range of the agent $i$ and 0 otherwise. The agent's observation only receives the features of visible units. To allow communication among the agents in SMAC$^{+}$, we re-design the visibility matrix by setting communication range. If agent $i$ and $j$ is in the communication range of each other, then they share the visible units. For example, if the enemy $k$ is in the sight range of the agent $j$ but not in the agent $i$'s, visibility matrix entity $(i, k)$ is changed from 0 to 1. The agents use this modified visibility matrix to get observation features. The sight range is 9 for all agents, and the communication range is 16 for the Siege Tank and 12 for other units. Researchers can easily turn on or off the communication function by setting \textsf{obs\textunderscore communicate \textunderscore info} = \textsf{True} or \textsf{False} in the environment setting file. The default value is \textsf{True}. Note that this setting does not violate CTDE assumption, as the agent receives additional \textsf{Enemy feat, Ally feat, Neutral building feat} of \autoref{smacplus:observation_feature_2} according to the updated visibility matrix, but the global state or the combination of the entire observations are not accessible during a test phase. \textsf{obs\textunderscore broadcast \textunderscore info} is another communication function that has no limit on communication range. Hence, when we set either one option to \textsf{True}, the communication option is active, but turning off all these functions makes no communication among agents. 

\begin{figure}[!ht]{
    \centering
        \begin{subfigure}{0.28\columnwidth}
            \includegraphics[width=\columnwidth]{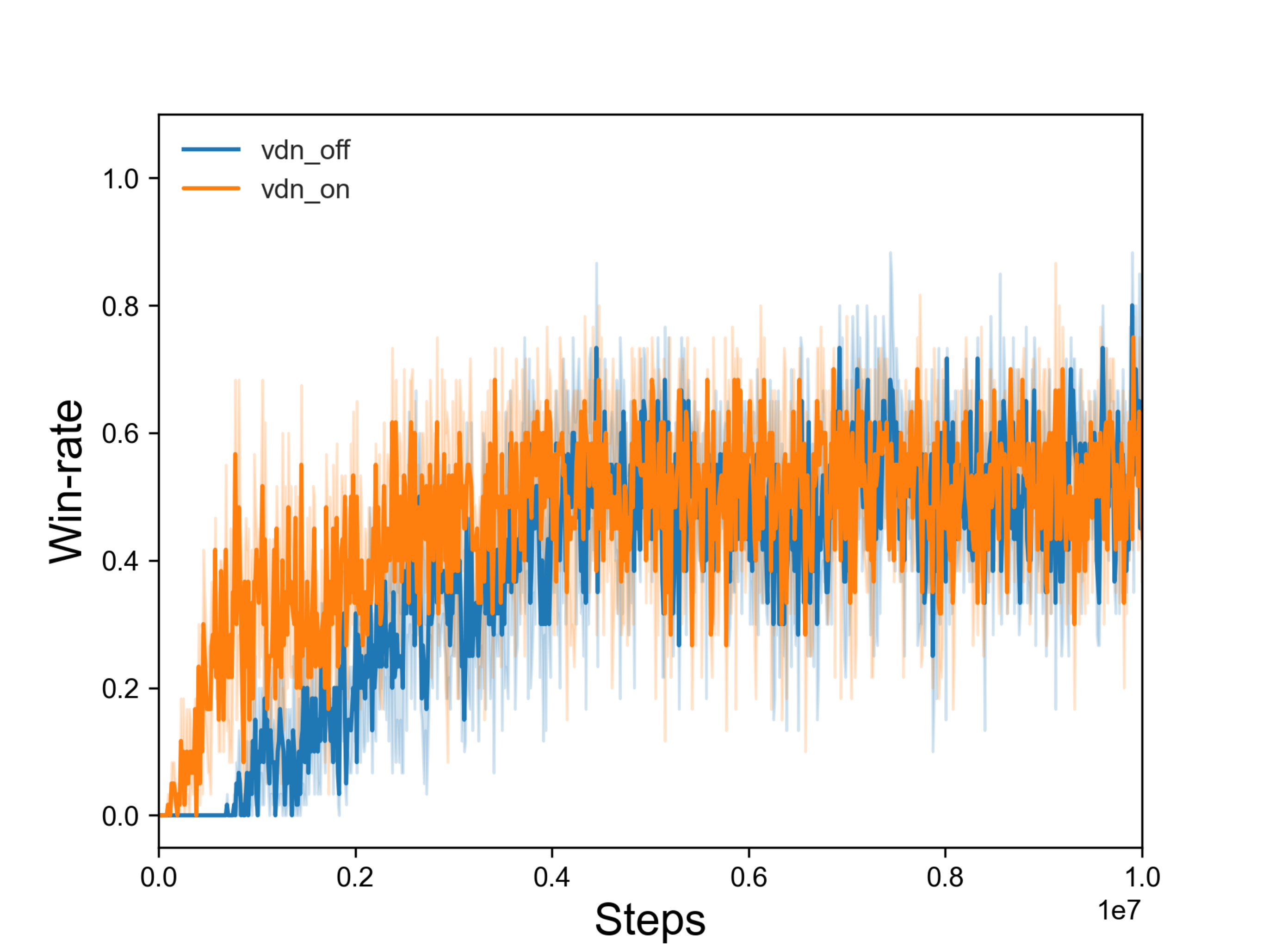}
            \caption{Training curve}
            \label{fig:com_training}
        \end{subfigure}%
        \hspace{0.05cm}
        \begin{subfigure}{0.28\columnwidth}
            \includegraphics[width=\columnwidth]{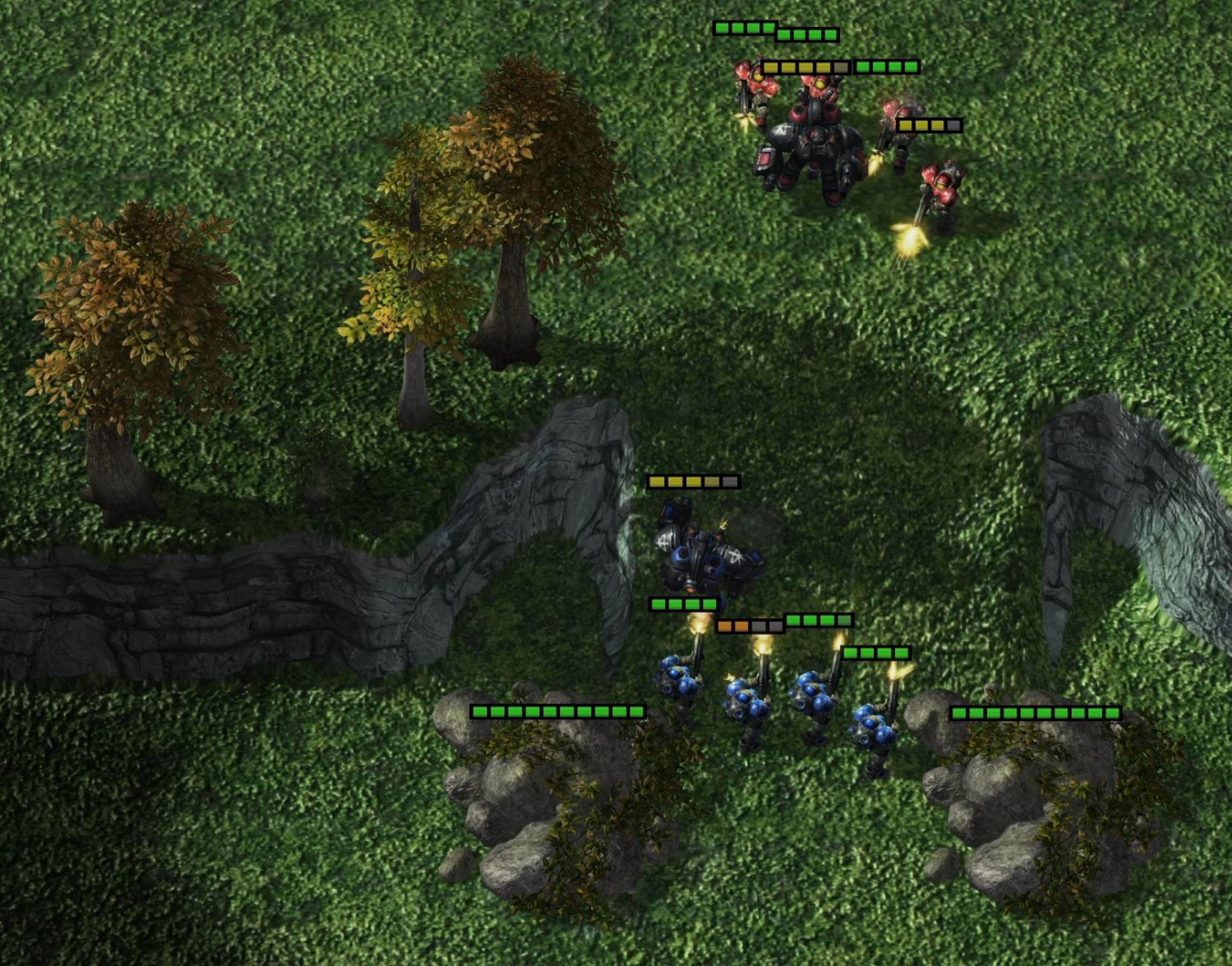}
            \caption{No communication}
            \label{fig:com_off}
        \end{subfigure}%
        \hspace{0.05cm}
        \begin{subfigure}{0.28\columnwidth}
            \includegraphics[width=\columnwidth]{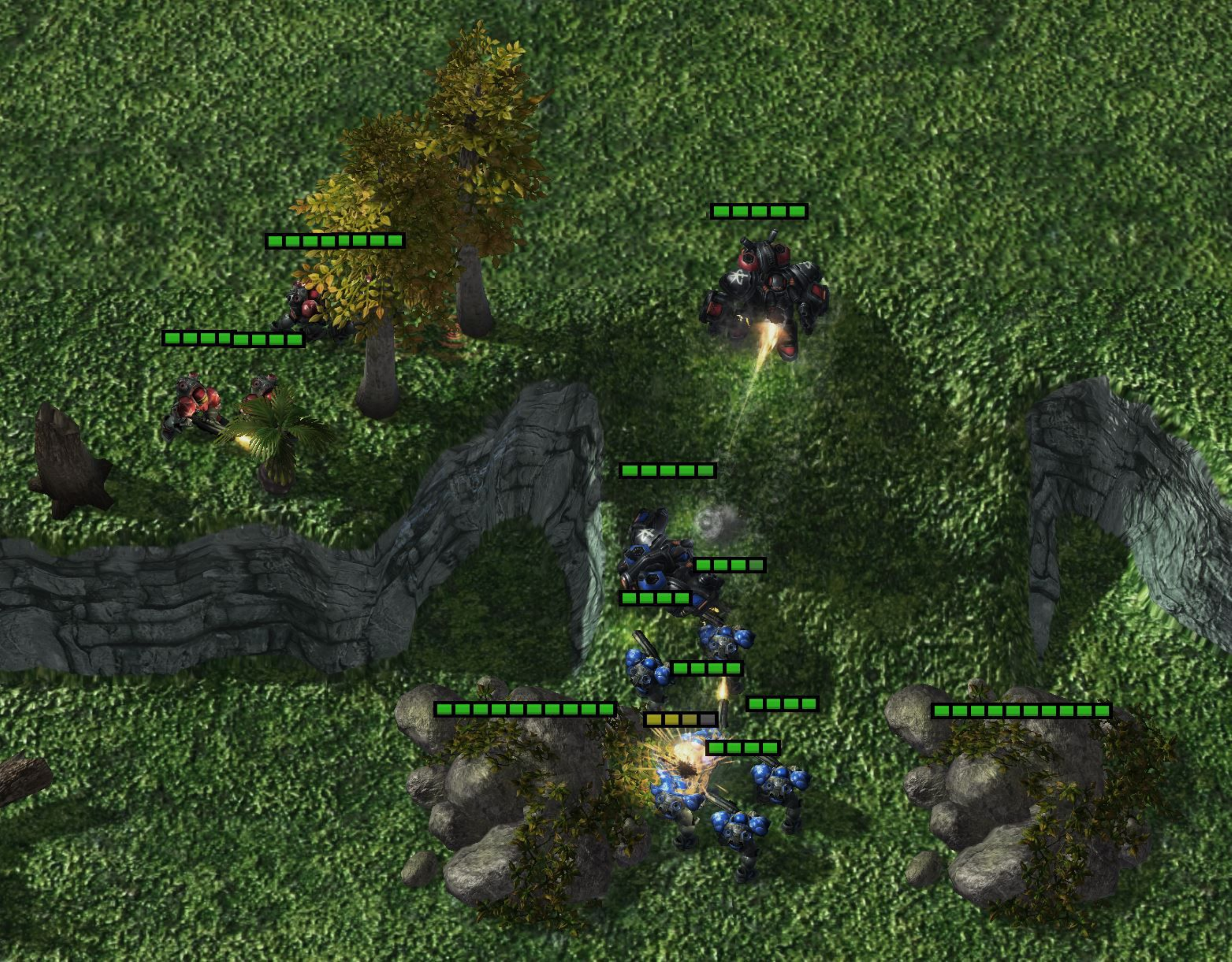}
            \caption{Inter-agent communication}
            \label{fig:com_on}
        \end{subfigure}%
    \caption{Case study of communication option. We present the results of \texttt{def\_infantry} by VDN and its associated test trial.}
    \label{fig:com}   
    }
\end{figure}

\autoref{fig:com_training} shows the training curve of VDN algorithm in the \texttt{def\_infantry} with and without the communication option. Even though there is no significant performance gap, the learned policies are quite different. When the agents are not allowed to communicate, as shown in  \autoref{fig:com_off} they show standing alongside at the entrance of the hill and directly engage the enemies. Even though the agents can benefit the stochastic damage dealing from feature of hill, not utilizing the feature makes it hard to win due to the supply difference between enemies and allies. In contrast, in \autoref{fig:com_on}, the Marauder distracts the enemies while the Marines hide behind the neutral buildings which block the sight of the enemies. This implies that the agents learned to utilize the sight-blocking capabilities of the neutral building blocks with the communication option. The possible reason is that the agents can behave apart from one another due to the observation sharing. It would be an interesting challenge to design an algorithm to learn such polices without communication.

\subsection{The Detail Information of Scenarios}
\begin{figure}[!h]{
    \centering
        \begin{subfigure}{0.35\columnwidth}
            \includegraphics[width=\columnwidth]{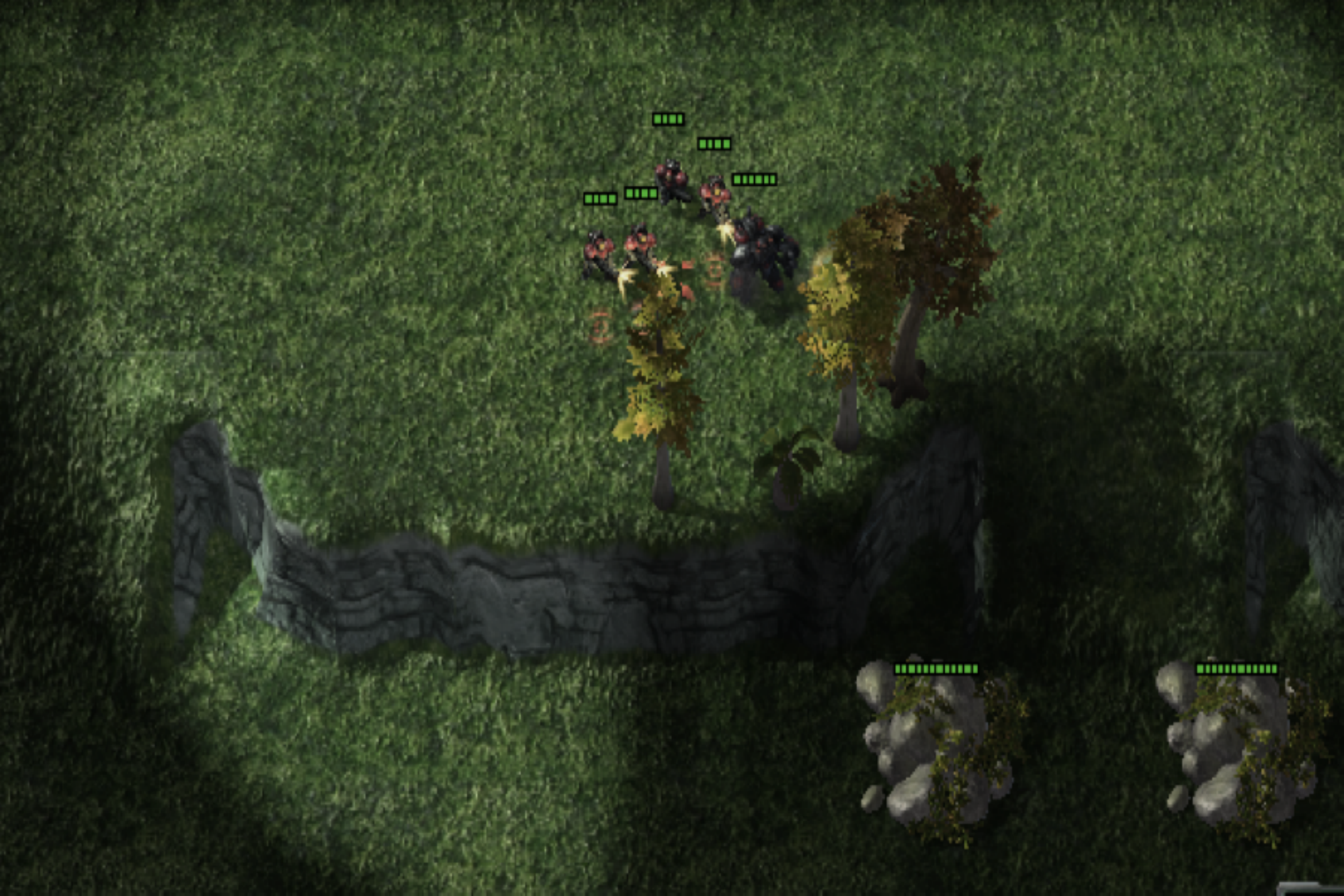}
            \caption{Removing buildings}
            \label{fig:app_2_def_removing_buildings}
        \end{subfigure}%
        \hspace{0.03cm}
        \begin{subfigure}{0.35\columnwidth}
            \includegraphics[width=\columnwidth]{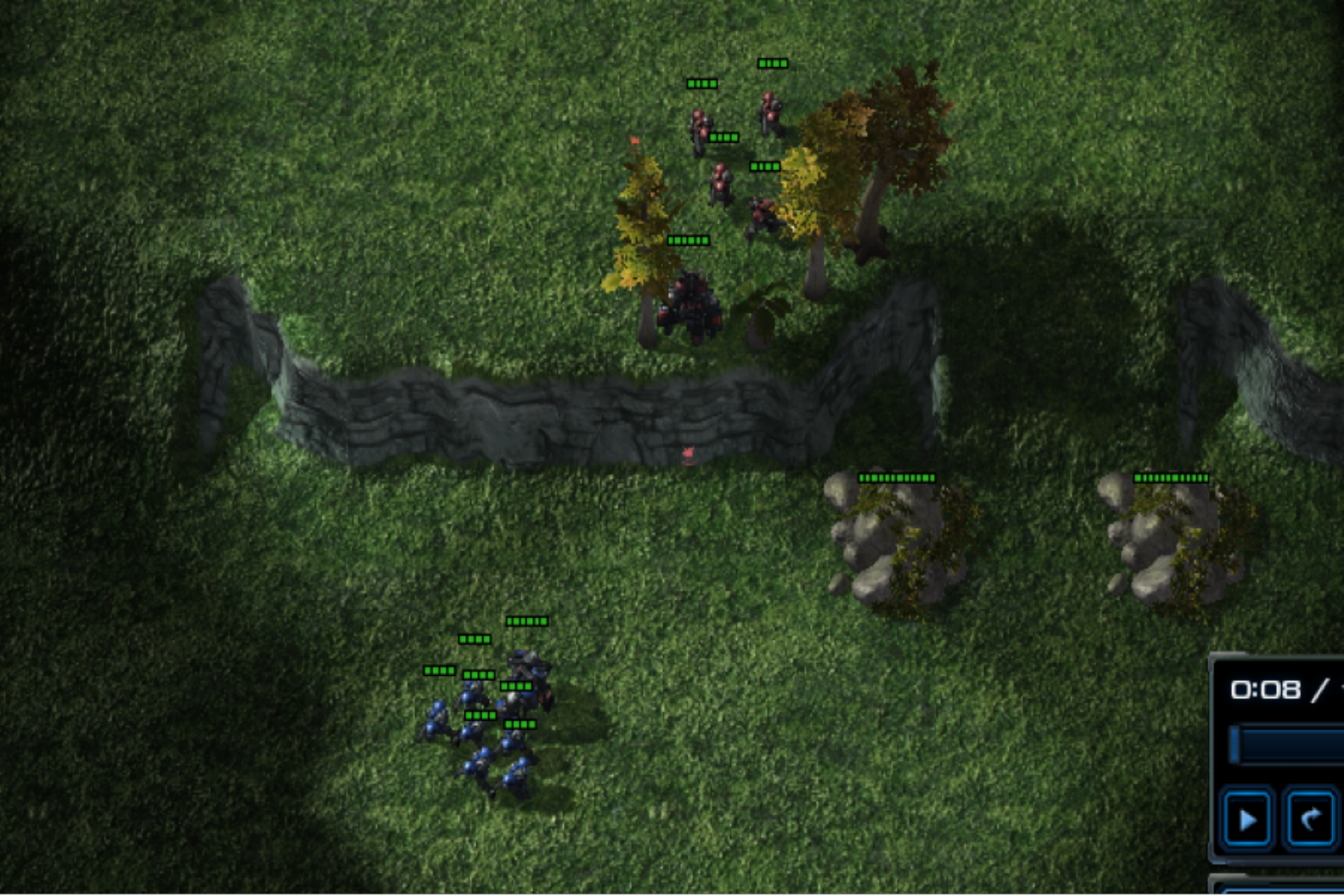}   
            \caption{Scouting}
            \label{fig:app_2_def_scouting}
        \end{subfigure}%
        
        \begin{subfigure}{0.35\columnwidth}
            \includegraphics[width=\columnwidth]{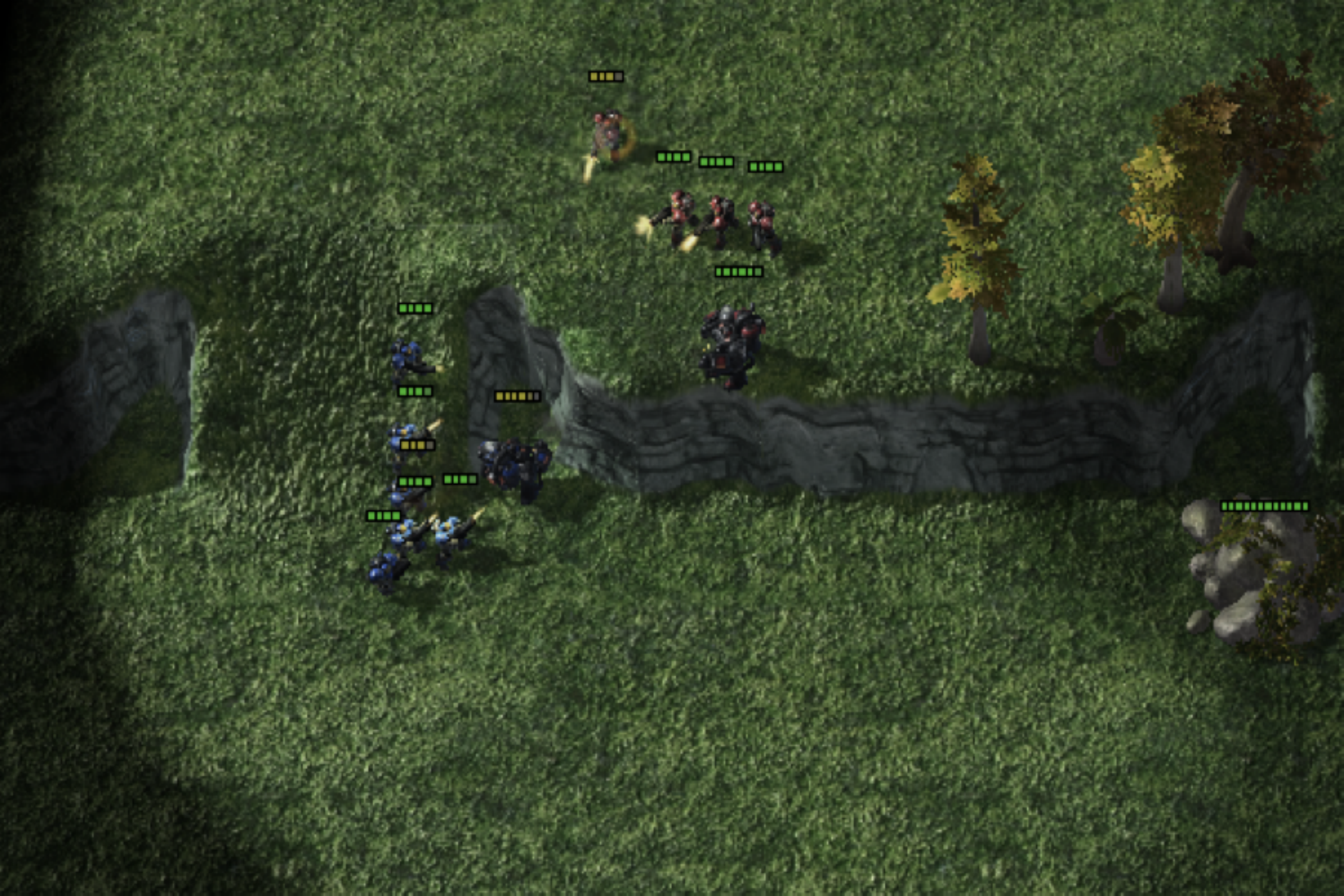}
            \caption{Terrain advantage}
            \label{fig:app_2_def_topographic_advantage}
        \end{subfigure}%
        \hspace{0.03cm}
        \begin{subfigure}{0.35\columnwidth}
            \includegraphics[width=\columnwidth]{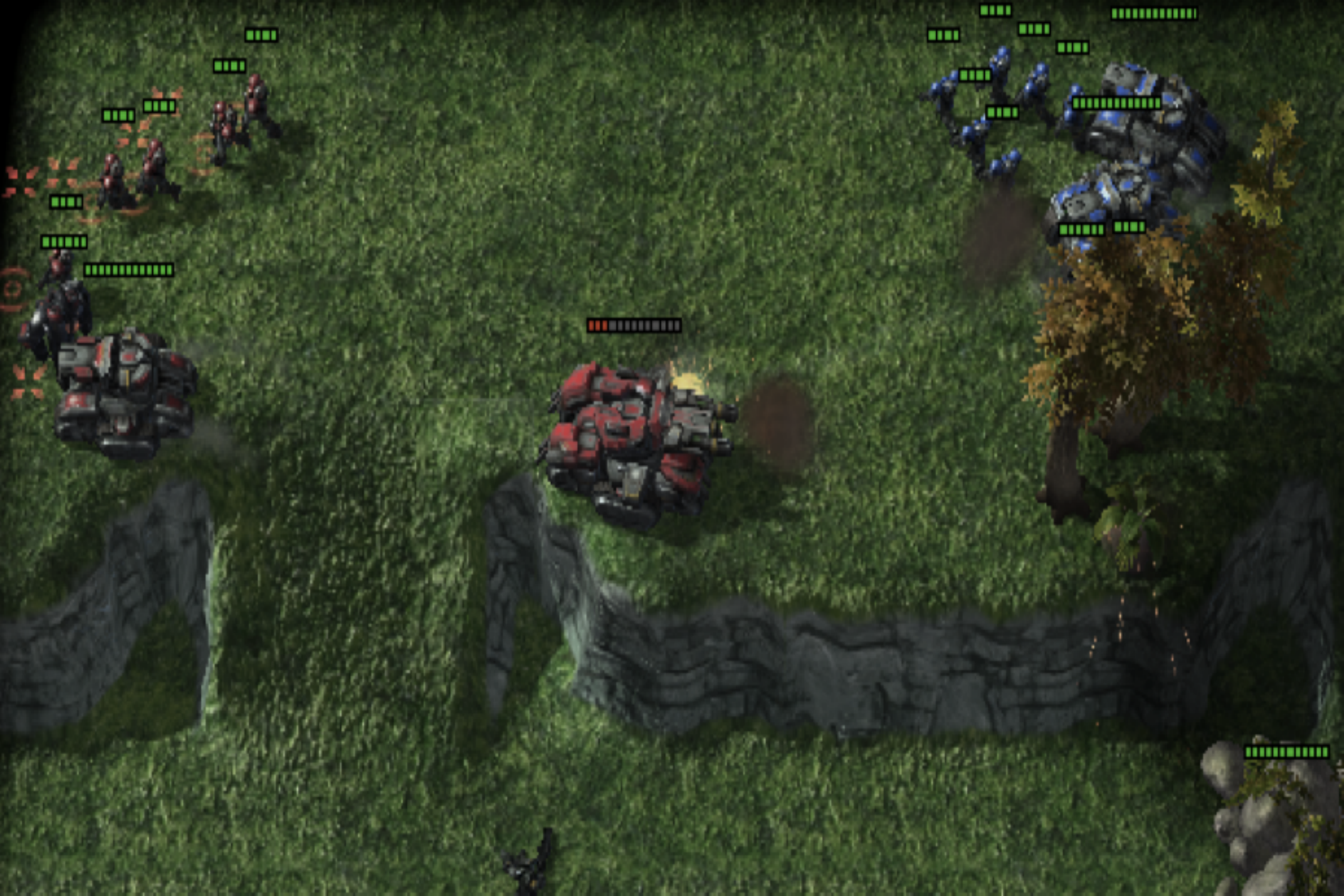}
            \caption{Failed to scouting}
            \label{fig:app_2_def_failed_to_scoutin}
        \end{subfigure}%
    \caption{Sequential screenshots in Defensive Scenarios}
    \label{fig:app_defensive}
    }
\end{figure}

\paragraph{Defensive scenarios} By the extended exploration problem, allies should scout nearby the hill where allies are located to find out the enemies' offense direction as shown in \autoref{fig:app_2_def_scouting}. If trained well, the agents won't try to get out of the hill which make enemies' offense as a stochastic damage. Plus, allies remove the trees (neutral buildings) that block the agents' sight for making it easy to secure original sight range and find the enemies' offense direction which means eliminating the uncertainty which can be found in \autoref{fig:app_2_def_removing_buildings}. After finding the enemies offense direction, allies need to be located at some proper location that provide reduce-able damage from enemies offense shown as \autoref{fig:app_2_def_topographic_advantage} which shows that allies stand on the hill not allowing enemies to enter the hill. If agents are not trained well, they failed to find the enemies offense direction allowing occupation of the allies' respawn place to enemies.

\begin{figure}[!h]{
    \centering    
        \begin{subfigure}{0.35\columnwidth}
            \includegraphics[width=\columnwidth]{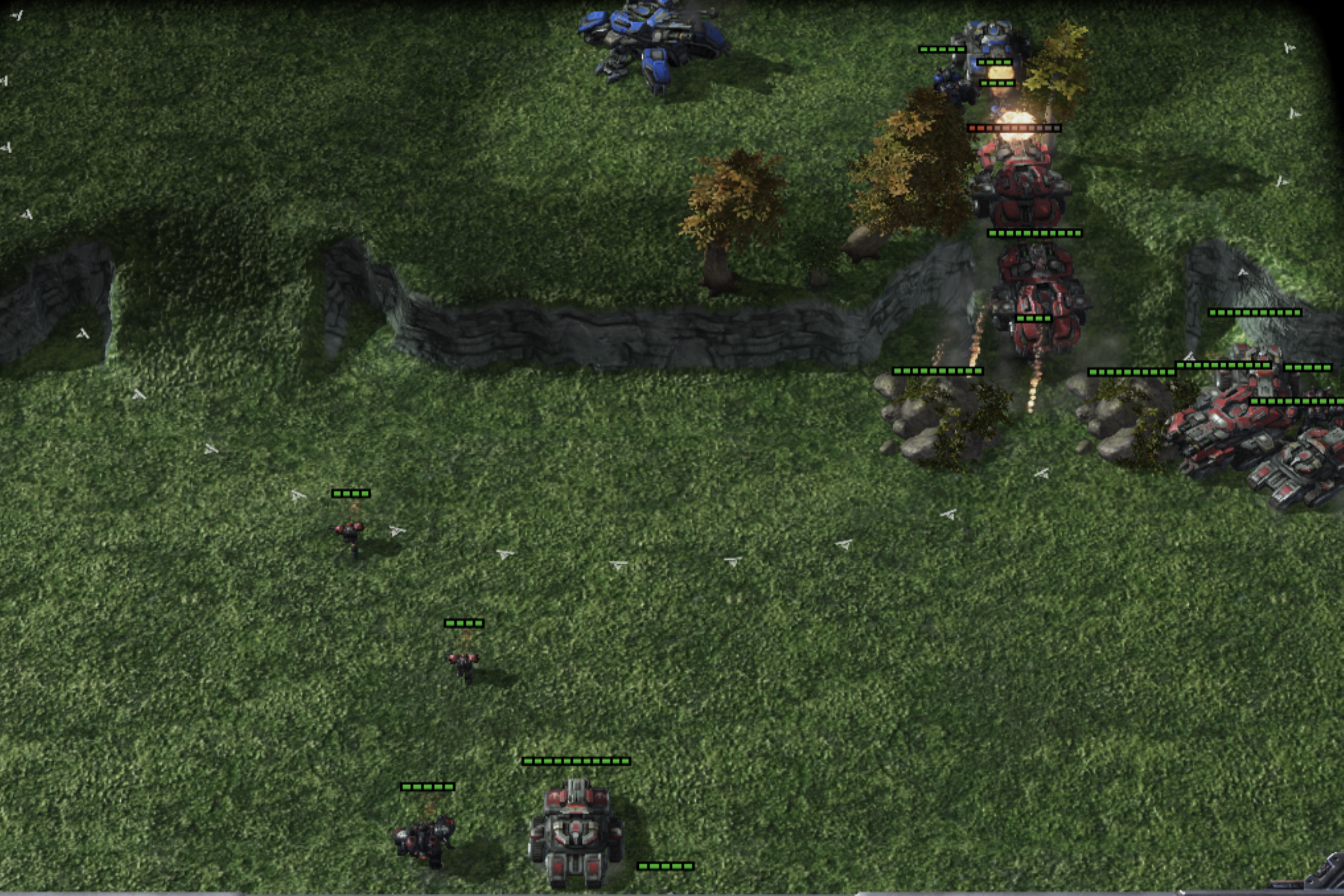}
            \caption{Finding enemies}
            \label{fig:app_2_off_finding_enemies}
        \end{subfigure}%
        \hspace{0.03cm}
        \begin{subfigure}{0.35\columnwidth}
            \includegraphics[width=\columnwidth]{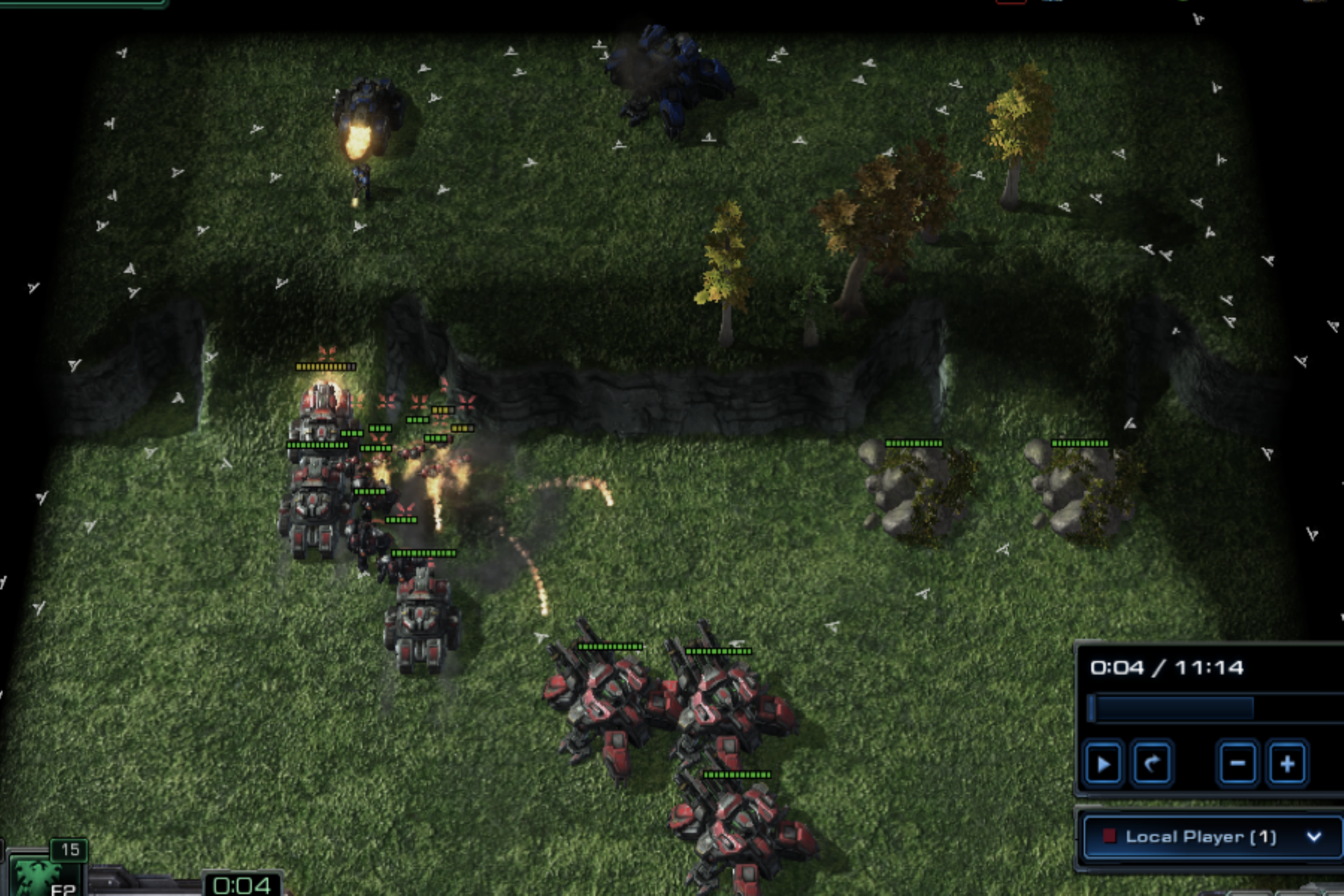}
            \caption{Before communication}
            \label{fig:app_2_off_before_communication}
        \end{subfigure}%
        
        \begin{subfigure}{0.35\columnwidth}
            \includegraphics[width=\columnwidth]{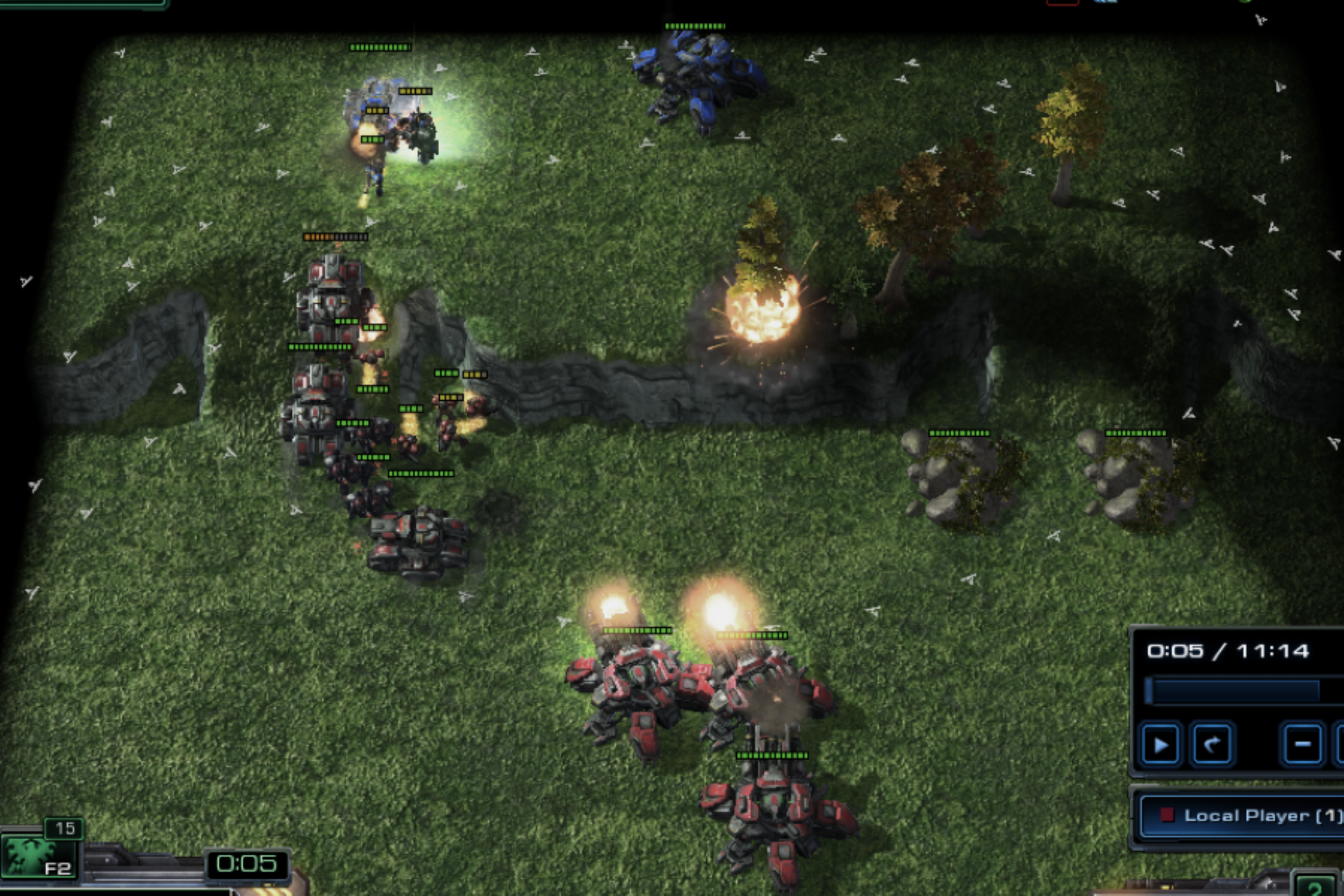}
            \caption{After communication}
            \label{fig:app_2_off_after_communication}
        \end{subfigure}%
        \hspace{0.03cm}
        \begin{subfigure}{0.35\columnwidth}
            \includegraphics[width=\columnwidth]{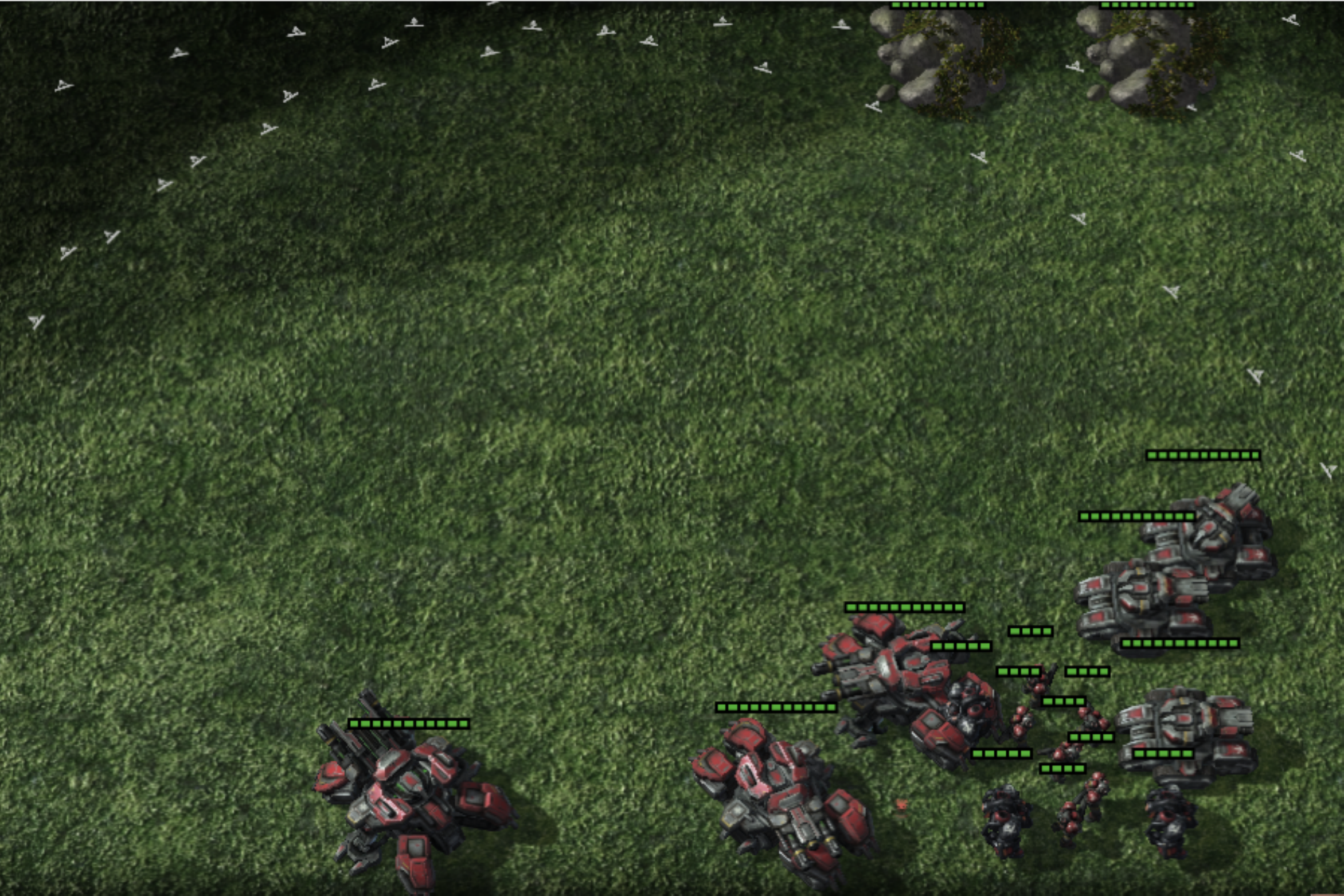}
            \caption{Failed to scouting}
            \label{fig:app_2_off_failed_to_scouting}
        \end{subfigure}%
    \caption{Sequential screenshots in Offensive Scenarios}
    \label{fig:app_2_offensive}
    }
\end{figure}

\paragraph{Offensive scenarios} In the offensive scenarios in SMAC$^{+}$, allies must find the place where the enemies are located. However, finding enemies far away from the allies is not an easy problem. Because the action space in SMAC$^{+}$ becomes much larger than the original SMAC \cite{samvelyan2019starcraft} environment caused by the neutral buildings. Therefore, it is difficult for agents to find enemies at the beginning of learning. To find the enemies, allies should endure the damage from the enemies like shown as \autoref{fig:app_2_off_finding_enemies}. When finding the location and deployment of enemies, allies begin to attack the enemies with proper strategy if agents trained well as shown in \autoref{fig:app_2_off_before_communication} and \autoref{fig:app_2_off_after_communication}. But if agents are not trained well in finding the enemies, agents get overfitted to the bad samples resulting the bad behavior as shown in \autoref{fig:app_2_off_failed_to_scouting}. Offensive scenario is hard to solve for agents because of the sparse rewards problem and multi-goal objectives, we set the number of ally units quite larger than defensive scenarios' enemies' number for observing the training development. We can reduce the number of ally units for future work, if proper algorithms are developed.

\begin{table*}[!h]
\caption{SMAC$^{+}$ defensive scenarios. All units are only from Terran race for the reflecting realistic part of the world. SG, Mar and M mean each Siege Tank, Marauder, Marine}
\label{def_scenarios}
\centering
\begin{tabular}{c|c|c p{0.25\linewidth}p{0.5\linewidth}p{0.25\linewidth}}
\toprule
\multicolumn{3}{c}{\textbf{Defense}} \\
\midrule
\textbf{Scenario} & \textbf{Ally Units} & \textbf{Enemy Units} \\ 
\hline
\texttt{Infantry} & 1 Mar \& 4 M & 1 Mar \& 6 M \\
\texttt{Armored} & 1 SG Tank, 1 Tank, 1 Mar \& 5 M & 2 Tank, 2 Mar \& 9 M  \\
\texttt{Outnumbered} & 1 SG Tank, 1 Tank, 1 Mar \& 5 M & 2 Tank, 3 Mar \& 10 M \\
\bottomrule
\end{tabular}
\end{table*}

\begin{table*}[!h]
\caption{SMAC$^{+}$ offensive scenarios. }
\label{off_scenarios}
\centering
\begin{tabular}{c|c|c|c p{0.25\linewidth}p{0.25\linewidth}p{0.25\linewidth}p{0.25\linewidth}}
\toprule
\multicolumn{4}{c}{\textbf{Offense}} \\
\midrule
\textbf{Scenario} & \textbf{Ally Units} & \textbf{Enemy Units} & \textbf{Distance \& formation} \\ 
\hline
\texttt{Near} & \multirow{1}{*}{\{3 SG Tank,} & \multirow{1}{*}{\{1 SG Tank,} & Near \& Spread \\
\texttt{Distant} & \multirow{1}{*}{3 Tank,} & \multirow{1}{*}{2 Tank,} &  Distant \& Spread \\
\texttt{Complicated} & \multirow{1}{*}{3 Mar \& 4 M\}} & \multirow{1}{*}{2 Mar \& 4 M\}} & Complicated \& Spread \\
\cmidrule(r){1-4}
\texttt{Hard} & \multirow{1}{*}{\{1 SG Tank, 2 Tank,} & \multirow{1}{*}{\{1 SG Tank, 2 Tank,} & Complicated \& Spread \\
\texttt{Superhard} & \multirow{1}{*}{2 Mar \& 4 M\}} & \multirow{1}{*}{2 Mar \& 4 M\}}& Complicated \& Gathered \\ 
\bottomrule
\end{tabular}
\end{table*}

\section{Training Details}
\label{app:training}

\paragraph{Training Protocol}
We evaluate the following eleven MARL algorithms using the CTDE paradigm. For the value-based category, widely-adopted baselines employing value function factorization (IQL, VDN \cite{sunehag2017value}, QMIX \cite{rashid2018qmix}, QTRAN \cite{son2019qtran}) and more recent state-of-the-art algorithms (DIQL, DMIX, DDN \cite{sun2021dfac}, DRIMA \cite{learning5disentangling}) are selected. COMA \cite{foerster2018counterfactual}, MASAC \cite{haarnoja2018soft}, MADDPG \cite{lowe2017multi} are chosen for the policy-based category by general usage in MARL domain. 
Additionally, we report the four most effective baselines based on the results of parallel training in \autoref{table:final_winrate}. For the fair comparison with the MADDPG algorithm \cite{lowe2017multi} which is only compatible to the episodic setting, we respectively select the best performing algorithms in value-based, distribution-based and policy-based algorithms such as QMIX \cite{rashid2018qmix}, DRIMA \cite{learning5disentangling} and COMA \cite{foerster2018counterfactual} and train those algorithms using the episodic buffer setting. 
We run each algorithm for a total of ten million timesteps with 3 different random seeds for parallel training using 20 runners and 5 million timestep for episodic training. The trained model is tested at every ten thousand timesteps during 32 episodes for episodic training and 20 episodes for parallel training. As a evaluation metric, the percentage of winning episodes referred as to win-rate is employed. In these experiments, we set all rewards to positive values.

\paragraph{Hyperparameters} We describe about hyperparameters we used. Hyperparmeters are almost same with experiments conducted in the other papers for the fairness except training steps. Throughout the training, we anneal $\epsilon$ from 1.0 to 0.05 over 50000 training steps and fix the $\epsilon$ during the rest of the training time. We fix $\gamma = 0.99$. Replay buffer is capable for containing most recent 5000 episodes and we sample 32 size of batch from the replay buffer randomly. We update target network with current network every 200 time steps. We have 10050000 steps for each training. The details about hyperparameters are on \autoref{hyperparameter}.

\begin{table}[!h]
\caption{Hyperparameters of Network}
\label{hyperparameter}
\centering
\begin{tabular}{c|cc p{0.5\linewidth}p{0.5\linewidth}}
\toprule
\textbf{Hyperparameter} & \textbf{Value} & \textbf{Description} \\ 
\midrule
Training step  & 10050000 & how many steps we trained the model \\
Discount factor & 0.99 & how we estimate the future rewards \\
Learning rate & $5 \times 10^{-4}$ & learning rate by RMSProp optimizer \\
Target update period & 200 & update period of target network \\
Replay buffer size & 5000 & maximum container size of the past samples \\
Batch size & 32 & number of samples for each update \\
Batch size run & 20 & number of parallel simulator  \\
$\epsilon$ & 1.0 to 0.05 & $\epsilon$-greedy exploration over 50000 training steps\\
Number of sampling $\tau$ & 32 & number of quantile fraction samples in DFAC\\
\bottomrule
\end{tabular}
\end{table}

\paragraph{Computational Cost}

We use a machine containing 512GB memory, 4 GPUs(GeForce RTX 3080) with 10240MB memory each and AMD EPYC 7543 Processor with 32 cores. On the basis of an offensive scenario, SMAC$^+$ requires about 70-80GB of main memory and 3000-7000 MB of GPU capacity per scenario with parallel 20 simulators settings, and we are able to obtain results within 12-24 hours. Meanwhile, when we train for total 5 million cumulative episode steps via the sequential episodic buffer, it takes at least 24 hours and grows up to 48 hours. The detailed information of computation cost is listed in \autoref{app_tab_training_time}.

\begin{table}[!ht]
    \caption{Approximate training hours of SMAC$^{+}$. Results are evaluated with sequential episodic buffers during 5 million training timesteps and parallel episodic buffers during 10 million training timesteps.}
    \label{app_tab_training_time}
    \centering
    \resizebox{0.99\columnwidth}{!}{%
    \begin{tabular}{cccccccccc}
    \toprule
        & \multicolumn{3}{c}{\bf Defensive scenarios} & \multicolumn{5}{c}{\bf Offensive scenarios} \\
    \cmidrule(r){2-4} \cmidrule(r){5-9} 
        & \texttt{infantry} & \texttt{armored} & \texttt{outnumbered} & \texttt{near} & \texttt{distant} & \texttt{complicated} & \texttt{hard} & \texttt{superhard} \\
    \midrule
    Episode & 23 & 35 & 36 & 36 & 38 & 46 & 44 & 47  \\
    Parallel & 5 & 8 & 8 & 13 & 13 & 16 & 9 & 9 \\
    \bottomrule
    \end{tabular}
    }%
\end{table}
\section{Algorithms}
\label{app:algorithms}

In this section, we describe how each algorithm works. We divided algorithms in detail taxonomies as Value-based, Policy-based, and Distribution-based in this section. Value-based and Distribution-based algorithms showed better performance so far than policy based algorithms. We think that this was induced by a sample efficiency and suitable action space for discrete action choice which is advantages of value-based algorithms. We will describe about Distribution-based algorithms which got popular from \cite{bellemare2017distributional} that achieved state-of-the-art performance in Atari and Mujoco environments and also in SMAC's challenging hard scenario.

\paragraph{Partially Observable MDP} In Markov Decision Process (MDP), agents can observe all the environments like \cite{hausknecht2015deep} if we don't use screen flickering technique where environment is stochastic. But in real world, MDP is not achievable generally. Instead, Partially Observable MDP (POMDP) is observable in the real world rather than MDP. For example, human being can observe his around where he is located but cannot observe where he is not located. So, POMDP is an MDP that agents can observe things only in their sight and agents decide their actions based on their observations. This is called Decentralized-POMDP (Dec-POMDP). In MARL settings, we consider simulator's environment as a Dec-POMDP setting. 

\paragraph{Notation} Formally, Dec-POMDP is given with a tuple G = $\langle {S, U, P, \it{r}, Z, O, \it{n}, \gamma} \rangle$. s $\in$ S is the true state that the environment provides. $\textit{a}$ $\in$ A $\equiv$ \{1, ..., \textit{n}\} is an agent that chooses an action \textit{u}$^{a}$ $\in$ U which forms a joint action space \textbf{u} $\in$ \textbf{U} where $\textit{n}$ is the number of agents. P($s\prime$ $\mid$ s, \textbf{u}) : S $\times$  \textbf{U} $\times$ S $\to$ [0, 1] is a transition probability function. All agents in Dec-POMDP receive shared reward, so the reward function is $\textit{r}(s, \textbf{u}) : S \times \textbf{U} \to \mathbb{R}$. Observation function is $O(s, a):S \times A \to Z$ that determines agents' observation $z^{a} \in Z$. $\gamma \in [0, 1)$ is a discount factor for the reward.

\begin{figure}[!h]{
    \centering
        \begin{subfigure}{0.30\columnwidth}
            \includegraphics[width=\columnwidth]{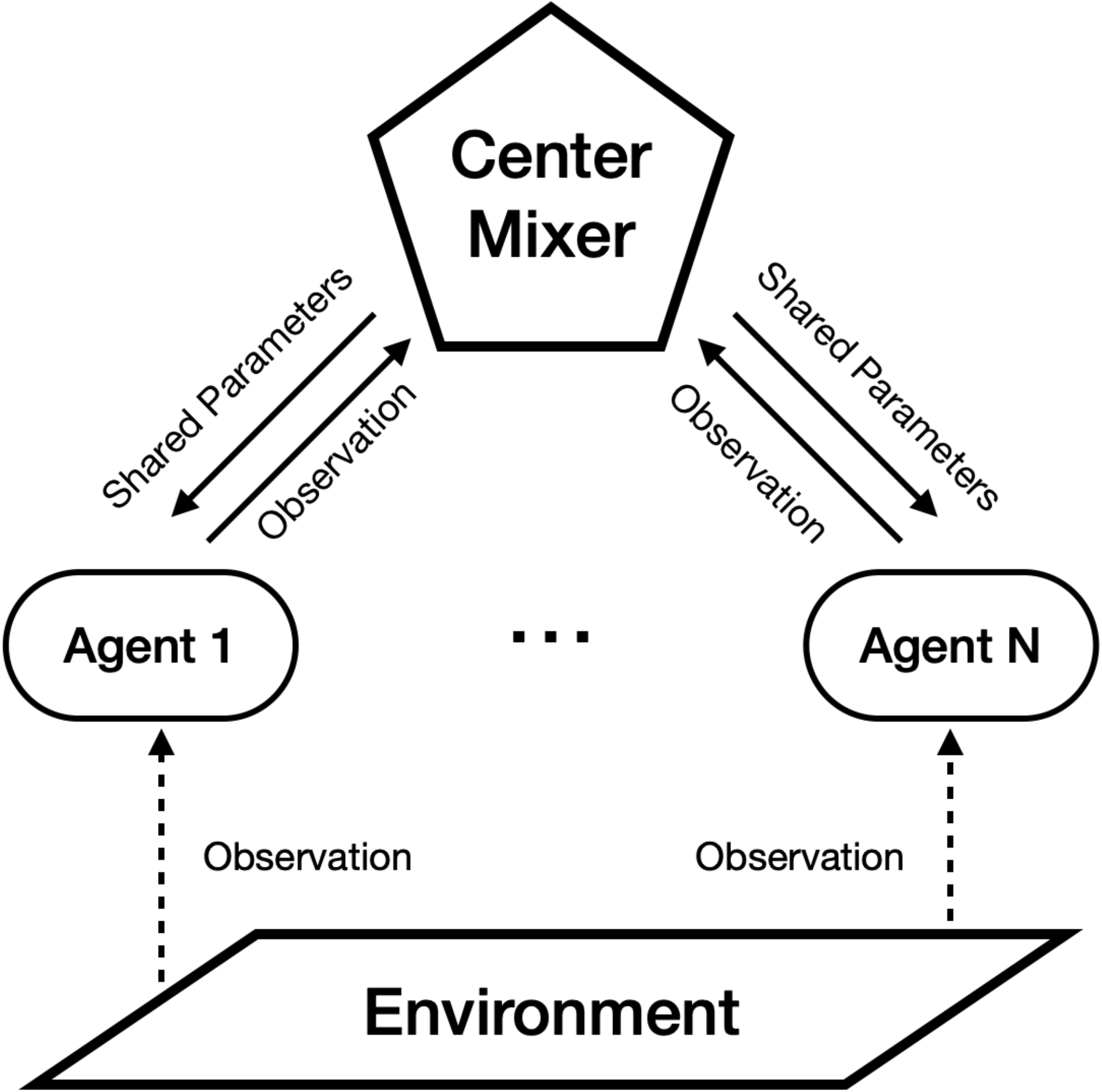}
            \caption{Centralized training}
            \label{fig:centralized training}
        \end{subfigure}%
        \hspace{0.03cm}
        \begin{subfigure}{0.30\columnwidth}
            \vspace{+0.4cm}
            \includegraphics[width=\columnwidth]{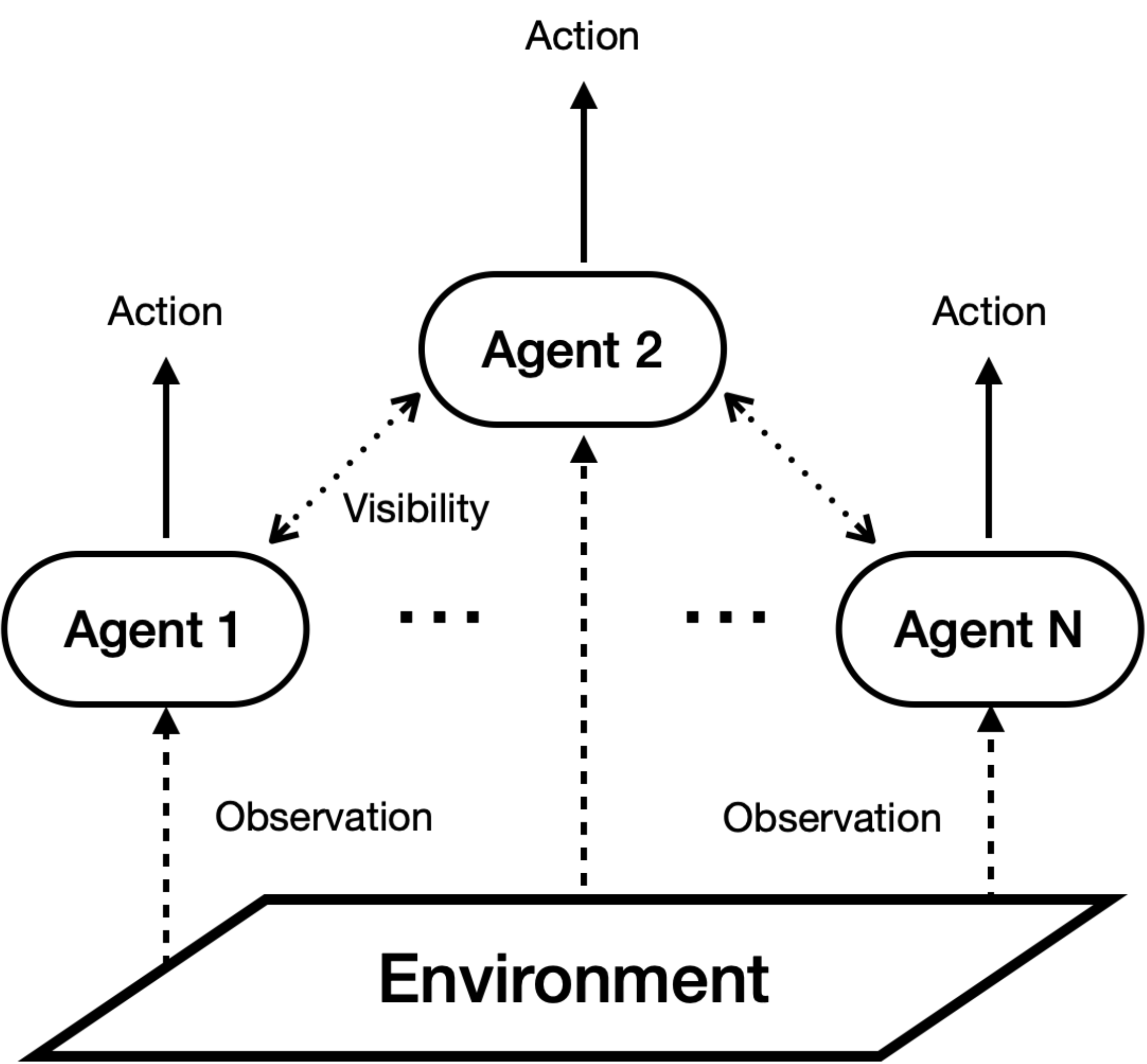}
            \caption{Decentralized execution}
            \label{fig:decentralized execution}
        \end{subfigure}%
    \caption{CTDE Framework}
    \label{fig:ctde}
    }
\end{figure}

\paragraph{Centralized Training \& Decentralized Execution}
In early stage of MARL, decentralized training \& decentralized execution (DTDE) is the main framework of training model like training single agent RL model. But it was very hard to train model with the DTDE framework in multi-agent setting, because of the more enhanced randomness by other agents' action selection, Partially Observable MDP setting (POMDP), and insufficient information depending only on one's observation during training the model's parameters. So, now centralized training \& decentralized execution (CTDE) \citep{oliehoek2008optimal} framework is highly employed in Multi-Agent Reinforcement Learning (MARL) domain because of the advantage of informative training data like global state or gathering all of the observation of the agents. Most of the recent MARL algorithms have developed on the basis of CTDE learning framework which has access on all information at training step but has access on individual information only at execution step as shown in \autoref{fig:ctde}. Gathering all agents' information at the center enables learning in POMDP and complex setting. We describe the development of the algorithms based on CTDE from VDN\citep{sunehag2017value} to DFAC\citep{sun2021dfac} which uses distributional RL algorithm.

\subsection{Value-based}
Value-based algorithms were developed for environments that require discrete action space. So, SMAC environment had started with value-based algorithms naturally and showed better performance than policy-based algorithms. Value-based algorithms use n-step (1-step in general) temporal difference error (TD-error) for updating critic parameters. Basically these algorithms follow below equation as a basis of loss function:
\begin{equation}
    \begin{aligned}
    \label{eqn:td-error}
        \delta_{td-error} = (R + \gamma * \mathrm{max}_{a '}Q_{\theta^{-}}(s ', a ') - Q_{\theta}(s, a))^{2}
    \end{aligned}
\end{equation}
where $s$ and $a$ represent state and action. Prime $'$ means next step and $\theta$, $\theta^{-}$ means parameters of behavior and target network each. In general, state and action  are denoted as $s$, $a$, but from now we will use the notation of $\mathrm{h}$ and $\mathrm{u}$ for state and action which are also utilized as a method of notation state and action in RL domains. Before entering to the value-based algorithms, we have to know about the condition which is needed for stable learning and avoiding lazy agents that is called IGM(Individual Global Max) condition which means individual maxiaml Q-values consist of total (or global) maximal Q-values. It can be represented as follows:
\begin{equation}
    \begin{aligned}
    \label{igm}
        \mathrm{arg\,max}_{u}Q_{total}(\textbf{h, u}) = \begin{pmatrix}
        \mathrm{arg\,max_{u_{1}}} Q_{1}(\mathrm{h_{1}, u_{1}})\\
        \vdots \\
        \mathrm{arg\,max_{u_{N}}} Q_{N}(\mathrm{h_{N}, u_{N}})\\
        \end{pmatrix}    
    \end{aligned}
\end{equation}
where h and u represents history of observation and action each. For CTDE framework, value-based algorithms are consist of utility function and mixer. Utility function receive individual observation every time step and outputs action which is comprised of DRQN\citep{hausknecht2015deep} for POMDP setting that take history of observation. And mixer takes all agents' individual Q-values and outputs joint Q-values for calculating temporal difference error for centralized training. For decentralized execution, mixer is not used but only utility function is used. Most of the utility function in various algorithms is almost same, but architecture of mixer is highly related to the training model's architecture. 

\paragraph{IQL} 
Value-based algorithms for MARL start from decentralized training \& decentralized execution (DTDE) setting which means that train agents independently.  Independent Q-Learning (IQL) trains agents and execute action in a decentralized manner as the same way of training single agent not caring about whether there are other agents near itself or not. IQL has only utility function for training and execution and no mixer for centralized training. So generally, agents do not share parameters contrary to the current CTDE based algorithm that share parameters among all agents. For updating IQL model's parameters, temporal difference (TD) error is calculated individually and in this point IQL violates Markov Decision Process (MDP) assumption which is needed for converging Q learning. This is because environment appears to be non-stationary for each agent for other agents' actions. So, IQL shows not good performance in SMAC\citep{samvelyan2019starcraft} and SMAC+ scenarios but sometimes show good performance in specific scenarios like \textit{2s vs 1sc} and \textit{bane vs bane} that obstinately don't demand cooperative strategy between agents.

\paragraph{VDN} 
Value Decomposition Network\citep{sunehag2017value} (VDN) is a simple methods for cooperative MARL algorithm that adapt CTDE training framework first time to overcome IQL's learning shortcoming that only use individual observation for training the model. VDN just sum up the individual action-state value (which is called additivity), and use it as a joint Q-value which is utilized for calculating joint TD-error that is needed for centralized training. So the mixer in VDN is summation function of individual Q-values. The joint Q-value can be acquired as follows which satisfy the IGM condition by additivity:
\begin{equation}
    \begin{aligned}
    \label{eqn:vdn}
        Q_{joint}\left (\bf{h, u} \right) &= \sum_{i=1}^{N}Q_{i}\left (h_{i}, u_{i} \right)\\
    \end{aligned}
\end{equation}
where Q and N means Q-value and number of agents. In this algorithm all agents share parameters and the performance and the performance start to increase substantially compared to IQL algorithm.

\paragraph{QMIX} 
VDN\citep{sunehag2017value} has limitation on expressing complexity of the centralized joint Q-value that can ignore the additional global state or gathered observational information by just summing up the all individual Q-values. QMIX \citep{rashid2018qmix} develop the architecture of VDN which utilizes additivity for making $Q_{joint}$ satisfying IGM condition. QMIX adapt mixing network as a mixer for weighted linear summation of individual Q-values rather than simple linear summation like VDN where the parameters for mixing network are made by hypernetwork. And the IGM condition is satisfied with the monotonocity that make the parameters of mixing network positive values and enforce the joint-Q-value as monotonic in the per agent Q-values, which can be represented as:
 \begin{equation}
    \begin{aligned}
    \label{eqn:qmix_igm}
        \frac{\partial Q_{joint}}{\partial Q_{i}} \geq 0, \forall i \in \mathbb{N} 
    \end{aligned}
\end{equation}
which enable tractable optimization of the joint Q-value in off-policy learning. To make joint Q-value using mixing network, algorithm utilizes value factorization like VDN, meanwhile it uses a mixing network to compute the total value function. In the \autoref{eqn:qmix}, $\psi$ means a linear combination function by a mixing network which can be denoted as follows:. 
 \begin{equation}
    \begin{aligned}
    \label{eqn:qmix}
     Q_{total}\left (\bf{h, u} \right) = \psi \big( Q_{1} \left(\mathrm{h}_{1}, \mathrm{u}_{1}\right), \dots, Q_{n}\left(\mathrm{h}_{N}, \mathrm{u}_{N} \right)\big)
    \end{aligned}
\end{equation}
By the mixing network and monotonicity, QMIX outperformed other algorithms like IQL, VDN, COMA etc. QMIX algorithms has become the most popular algorithms in MARL and still lots of variants of QMIX have continued to emerge.

\paragraph{QTRAN} 
Former VDN and QMIX algorithms can handle the MARL issue as a way of being trapped in a constrained structures which are called additivitiy and monotonicity. QTRAN tried to relax the constraints on architecture's structure for abundant expressiveness by transforming the $Q_{joint}$ into an easily factorizable value. Mitigating the constraints, QTRAN proved guarantee for general factorization better than VDN and QMIX. The relaxed condition can be represented as:
\begin{equation}
    \begin{aligned}
    \label{eqn:qtran1}
        \sum_{i=1}^{N}Q_{i}\left (\mathrm{h}_{i}, \mathrm{u}_{i} \right) - Q_{joint}\left (\bf{h}, \textbf{u} \right) + V_{joint}(\bf{h}) &=
        \begin{cases}
            0 & \bf{u} = \Bar{\bf{u}} \\
            \ge 0 & \bf{u} \neq \Bar{\bf{u}}\\
        \end{cases}      
    \end{aligned}
\end{equation}
where, 
\begin{equation}
    \begin{aligned}
    \label{eqn:qtran2}
    V_{joint}\left(\bf{h} \right) &= \mathrm{max_{\textbf{u}}}Q_{joint}\left (\bf{h}, \textbf{u} \right) - \sum_{i=1}^{N}Q_{i}\left (\mathrm{h}_{i}, \mathrm{\bar{u}}_{i} \right)
    \end{aligned}
\end{equation}
where $\Bar{u}$ means optimal action and $V(\bf{h}_{joint})$ is a discrepancy that occurs from the Partially Observable MDP (POMDP) environment which can corrects the discrepancy between centralized $Q_{joint}$ and the sum of individual $Q$ which is $\sum_{i=1}^{N}Q_{i}\left (h_{i}, u_{i} \right)$. Despite of the expressive power that relaxed constraints on architecture's structure, QTRAN shows poor performance in many complex scenarios.

\subsection{Distribution-based}
These days policy-based distributional RL algorithms are emerging in MARL domain but by the the fact that SMAC demands discrete control optimization, usually distributional RL (DRL) for SMAC is based on the value-based RL algorithm so far. DRL algorithm has been developing substantially with great attention since \citep{bellemare2017distributional}, \cite{bellemare2017distributional}, \cite{dabney2018distributional} suggested new concept of value-based RL which outputs a distribution of return per action. \cite{bellemare2017distributional} fixed possible return value and made model to predict probability of returns with projected KL-divergence loss function. \cite{dabney2018distributional} approximate quantile regression as a output of return distribution per action and fixed probability of each returns with $\frac{1}{N}$ where N means number of returns per action and made model to predict value of returns. And \cite{dabney2018distributional} utilize Wasserstein metric as a loss function measuring distance of TD-error between $Q(s_{t}, a_{t})$ and $R(s_{t+1}, a_{t+1}) +  \gamma* max_{a_{t+1}}Q(s_{t+1}, a_{t+1})$. \cite{dabney2018implicit} samples quantile fractions uniformly in $[0, 1]$ which were fixed in \cite{dabney2018distributional}. And also model outputs value of returns corresponds to quantile fractions embeded with cosine function. \cite{yang2019fully} approximates both of probability and value of returns contrary to former algorithms that only approximate probability or value of returns. So, DRL algorithms develop following the sequence of C51\cite{bellemare2017distributional}$\longrightarrow$QR-DQN\cite{dabney2018distributional}$\longrightarrow$IQN\cite{dabney2018implicit}$\longrightarrow$FQF\cite{yang2019fully}.
\paragraph{DFAC} 
DFAC\citep{sun2021dfac} is a first algorithm that combines distributional RL and multi-agent RL which is based on the IQN\cite{dabney2018implicit} algorithm. Especially, IQN\citep{dabney2018implicit} was used for distributional output sampling quantile fractions from $\mathrm{U}[0, 1]$ and approximating return values with quantile regression. By mean-shape decomposition, authors integrated distributional perspective in the multi-agent setting not violating the IGM condition which can be written as:
\begin{equation}
    \begin{aligned}
    \label{dfac_igm}
        \mathrm{arg\,max}_{\textbf{u}}\mathbb{E}[Z_{joint}(\textbf{h, u})] = \begin{pmatrix}
        \mathrm{arg\,max_{u_{1}}} \mathbb{E}[Z_{1}(\mathrm{h_{1}, u_{1}})]\\
        \vdots \\
        \mathrm{arg\,max_{u_{N}}} \mathbb{E}[Z_{N}(\mathrm{h_{N}, u_{N}})]\\
        \end{pmatrix}    
    \end{aligned}
\end{equation}
that can be satisfied by the following DFAC Theorem (mean-shape decomposition):
\begin{equation}
    \begin{aligned}
    \label{eqn:mean-shape}
        Z_{joint}\left (\bf{h}, \textbf{u} \right) &= \mathbb{E}[Z_{joint}\left (\textbf{h}, \textbf{u})\right] + Z_{joint}\left (\textbf{h}, \textbf{u} \right) - \mathbb{E}[Z_{joint}\left (\textbf{h}, \textbf{u})\right] \\
        &= Z_{mean}\left (\textbf{h}, \textbf{u}\right) + Z_{shape}\left (\textbf{h}, \textbf{u}\right) \\
        &= \psi ( Q_{1} \left (\mathrm{h}_{1}, \mathrm{u}_{1}\right), \dots, Q_{N}\left(\mathrm{h}_{N}, \mathrm{u}_{N} \right) )\\
        &+ \Phi ( Z_{1} \left (\mathrm{h}_{1}, \mathrm{u}_{1}\right), \dots, Z_{N}\left(\mathrm{h}_{N}, \mathrm{u}_{N} \right) )
    \end{aligned}
\end{equation}
which is proved to satisfy IGM condition. DFAC\cite{sun2021dfac} shows outperforming performance than any other algorithms especially in \textit{Hard} scenarios. Also this algorithm can be adapted to IQL, VDN\cite{sunehag2017value}, QMIX\cite{rashid2018qmix}. So the DFAC algorithm's variants are named as DIQL, DDN, DMIX that were used as our baselines.

\paragraph{DRIMA} 
DFAC only considers a simple risk source but DRIMA\citep{learning5disentangling} considers separating risk sources into agent-wise risk and environment-wise risk which makes another hyperparameter contrary to DFAC algorithm that has a hyperparameter in risk setting, agent-wise risk. Environment-wise risk can be interpreted as transition stochasticity and agent-wise risk can be seen as the randomness induced by the other agents' action which can't be modeled by environment MDP (Markov Decision Process). In distributional multi-agent reinforcement learning algorithms(\citep{sun2021dfac}, \citep{qiu2021rmix}), models take risk level as an input to the agent utility function that output a distribution of return per an action which can be considered as a randomness by agents. But in DRIMA, agent receives agent-wise risk $w_{agt}$ and in the process of making joint distribution of returns, joint action-value network takes $w_{env}$ as input where agent utility function and joint action-value network composes of hierarchical architecture that resembles with QTRAN\citep{son2019qtran} structure.

Network architecture of DRIMA consists of agent-wise utility function, true action-value network, and transformed action-value network. Agent-wise utility function is structured by DRQN\citep{hausknecht2015deep} taking $w_{agt}$ as a input. True action-value network approximates true distribution of returns with additional representation power that receives environment-wise risk $w_{env}$, state $s$ and utility functions' outputs $\mathbb{Z}_{i}$. Transformed action-value network is similar to the QMIX's\citep{rashid2018qmix} mixing network as follows:
\begin{equation}
    \label{eqn:qtran}
    \resizebox{0.92\columnwidth}{!}{$
        \begin{aligned}
        Z_{trans}\left(s, \bf{\tau}, \bf{u}, w_{agt}\right) &= f_{mix}\left(z_{1}\left(\tau_{1}, u_{1}, w_{agt}\right), \dots, z_{N}\left(\tau_{N}, u_{N}, w_{agt}\right);\theta_{mix}(s, w_{agt})\right)
        \end{aligned}
    $}
\end{equation}

where not considering environment-wise risk and $\theta_{mix}(s, w_{agt})$ consists of a non-negative values obtained from hypernetwork like QMIX. DRIMA outperforms other algorithms especially in offense scenarios and in our experiments, we follow the default risk setting, agent-wise risk to be seeking and environment-wise risk to be averse.

\subsection{Policy-based}
\label{app:policy_algorithm}
Policy-based algorithm directly optimize the parameters $\theta$ to approximate the optimal policy $\pi$ which is especially specialized in continuous action control setting like robot system and autonomous vehicle. The simplest way of training policy-based model is independently taking policy gradient per agent. As expected, this works poorly more than IQL algorithm. Therefore, centralized training and decentralized execution is main framework for training models in policy-based algorithms. Policy based algorithms maximizes objective function that take expectation on initial state value as follows which is usually noted as J($\theta$):
\begin{equation}
    \begin{aligned}
    \label{eqn:policy-based objective}
        J(\theta) = \mathbb{E}_{\pi_{\theta}}[V_{0}]
    \end{aligned}
\end{equation}
where the gradient of objective is represented as $\nabla_{\theta}J(\theta) = \mathbb{E}_{s, a}[\nabla_{\theta} \log \pi_{\theta}(a|s)G(s)]$ which update actor's parameter realized in algorithm REINFORCE\citep{sutton1999policy}. In actor-critic algorithm, Return $G(s)$ is substituted by Q-value $Q_{\pi}(s, a)$ represented as same as \autoref{eqn:td-error} or advantage function $A_{\pi}(s, a)$ which demands another parameterized model to approximate critic. In multi-agent setting, we derive gradient of objective as:
\begin{equation}
    \begin{aligned}
    \label{eqn:policy-based gradient}
        \nabla_{\theta_{i}}J(\theta_{i}) = \mathbb{E}_{s_{i}, a_{i}}[\nabla_{\theta_{i}} \log \pi_{i}(a_{i}|s_{i})Q_{i}^{\pi}(\textbf{s}, a_{1}, \dots, a_{N})]
    \end{aligned}
\end{equation}
where $Q_{i}^{\pi}$ is a centralized individual critic that takes as inputs other agents' actions and additional information like global state or gathering all observations of the agents'.
\paragraph{COMA} 
COMA \citep{foerster2018counterfactual} is policy-based method that uses actor-critic algorithm that takes contribution to solving credit assignment problem by using concept of Difference rewards\cite{wolpert2002optimal}. In COMA \citep{foerster2018counterfactual}, a new form of advantage function was proposed to solve the `credit assignment' problem in a multi-agent environment. Original advantage function is calculated through differences between state-value functions and action-state value function, but in COMA \citep{foerster2018counterfactual}, all other agents' actions are fixed and use the average value of the action-state value function for a particular agent's action as follows:
\begin{equation}
    \begin{aligned}
    \label{eqn:coma}
        A^{a}\left (h, u  \right ) = Q \left (h, u \right ) - \sum_{u^{'a}} \pi^{a}\left ( u^{'a} | h^{a} \right ) Q\left ( h, \left ( \textbf{u}^{-a}, u^{'a} \right ) \right )
    \end{aligned}
\end{equation}

which estimate credit $r\left (s, \left (\textbf{u}^{-a}, c^{a} \right) \right)$ where c means default action of agents that is hard to estimate. This is why COMA uses \autoref{eqn:coma} as a estimation. The average of the action-state value function for each agent's action becomes the reference value for that agent's action. It serves to determine how good an agent’s action is compared to the average of action-state value. 
\paragraph{MASAC} 
MASAC\citep{pu2021decomposed} is soft actor critic (SAC)\citep{haarnoja2018soft} algorithm for multi-agent system. SAC is a actor-critic algorithm that maximizes exploration in action space utilizing entropy of distribution of action probability based on the maximum entropy RL theory. Like the original SAC for single agent, MASAC substitute individual Q-value with $Q_{total}$ and calculate TD-error and objective gradient using same value network with QMIX\citep{rashid2018qmix}. So, the TD-error is calculated as follows:
\begin{equation}
    \begin{aligned}
    \label{eqn:masac_critic}
        \delta_{td-error} = \left (R + \gamma \mathrm{min}_{u '}Q_{\theta^{-}}^{target}(h ', u ') - Q_{\theta}^{total}(h, u)\right)^{2}
    \end{aligned}
\end{equation}
which update the critic parameters. And for optimal policy, derived from soft policy iteration, objective is as follows:
\begin{equation}
    \begin{aligned}
    \label{eqn:masac_policy}
        J(\theta) = \mathbb{E}_{D}[\alpha \log \pi(u | h) - Q^{tot}_{\phi^{'}}(h, u)]
    \end{aligned}
\end{equation}
where $\alpha$ controls exploration and exploitation trade-off that if $\alpha$ is near to 1, it means exploration more, on the contrary $\alpha$ is near to 0, it means exploitation more.

\paragraph{MADDPG}
MADDPG\cite{lowe2017multi} has main contribution of reducing uncertainty by taking input as actions by other agents and learns centralized individual critic and decentralized actor for policy that enable both of cooperative and competitive strategy by splitting centralized critic in individual manner. MADDPG\cite{lowe2017multi} has emerged for continuous actions like MASAC\cite{pu2021decomposed} algorithm contrary to COMA\cite{foerster2018counterfactual} algorithm. MADDPG\cite{lowe2017multi} is based on the DDPG\cite{lillicrap2015continuous} algorithm that select deterministic action which is different from general policy-based algorithm that sample actions based on the action's distribution. The MADDPG\cite{lowe2017multi}'s gradient of objective function can be written as a new version of \autoref{eqn:policy-based gradient} adapting deterministic policy $a = \mu(s)$ as follows:
\begin{equation}
    \begin{aligned}
    \label{eqn:policy-based}
        \nabla_{\mu_{i}}J(\mu_{i}) = \mathbb{E}_{h, u \sim D}[\nabla_{\theta_{i}} \mu_{i}(u_{i}|h_{i})\nabla_{u_{i}}Q_{i}^{\mu}(\textbf{h}, u_{1}, \dots, u_{N})|_{u_{i} = \mu_{i}(u_{i})}]
    \end{aligned}
\end{equation}
where replay buffer $D$ contains tuples of $(h, u, r, h')$ that also utilized for critic update like \autoref{eqn:td-error} as:
\begin{equation}
    \begin{aligned}
    \label{eqn:maddpg loss function}
        \mathcal{L}(\theta_{i}) =\mathbb{E}_{u, h, r, h'}[(Q_{i}^{\mu}(\textbf{h}, u_{1}, \dots, u_{N}) - y)^{2}]
    \end{aligned}
\end{equation}
where $y = r_{i} + \gamma Q^{\mu^{'}}_{i}(\textbf{h'}, u_{1}^{'}, \dots, u_{N}^{'})|_{u^{'}_{j} = \mu^{'}_{j}(h_{j})}$. Updating actor parameters only needs for state and action information from replay buffer not next state and next action. From this point, MADDPG takes off-policy algorithm that enable to have replay buffer which makes sample efficient algorithms like value-based. This is why MADDPG showed good performance than general actor-critic algorithm that it has similar algorithm architecture with value-based RL. And because of the nature of the MADDPG algorithm, actor and critic should be updated at every step, making that distributed training (known as a \textit{parallel runner}) in SMAC\cite{samvelyan2019starcraft}) unable. In SMAC\cite{samvelyan2019starcraft}, we can select distributed training with multiple simulator or basic training with single simulator. In this paper, we use only non-distributed setting for MADDPG.

\newpage
\section{Completed Experimental Results of All Possible Cases}
\label{app:detailed_experiments}

We present a part of the experimental results in the manuscript. Here, we show experiment results in all possible cases and report the win-rate performances. As mentioned, due to the constraint on MADDPG that is not compatible with training in the parallel method, we respectively show the experimental results based on both sequential and parallel episodic buffers. Prior to the explanation of all results, we note the update interval difference between the sequential episodic buffer and the parallel episodic buffer as shown in \autoref{tab:comparison_update_interval}.

 \begin{table}[!ht]
 \centering
 \caption{Comparing model update interval between episodic and parallel episodic buffer when training 10 million steps.}
 \label{tab:comparison_update_interval}
 \begin{tabular}{ccc|cc}
 \toprule
 \multirow{2}{*}{} & \multicolumn{2}{c|}{Model update interval} & \multicolumn{2}{c}{The number of model update} \\
 \cline{2-5}
       & Target network  & Behavior network & Target network & Behavior network \\
 \midrule
  Episode & 200 episodes & 1 episode & 420 & 80000 \\
  Parallel & 200 episodes & 20 episodes & 420 & 4240 \\
 \bottomrule
 \end{tabular}
 \end{table}

\begin{figure}[!ht]{
    \centering
        \begin{subfigure}{0.26\columnwidth}
            \includegraphics[width=\columnwidth]{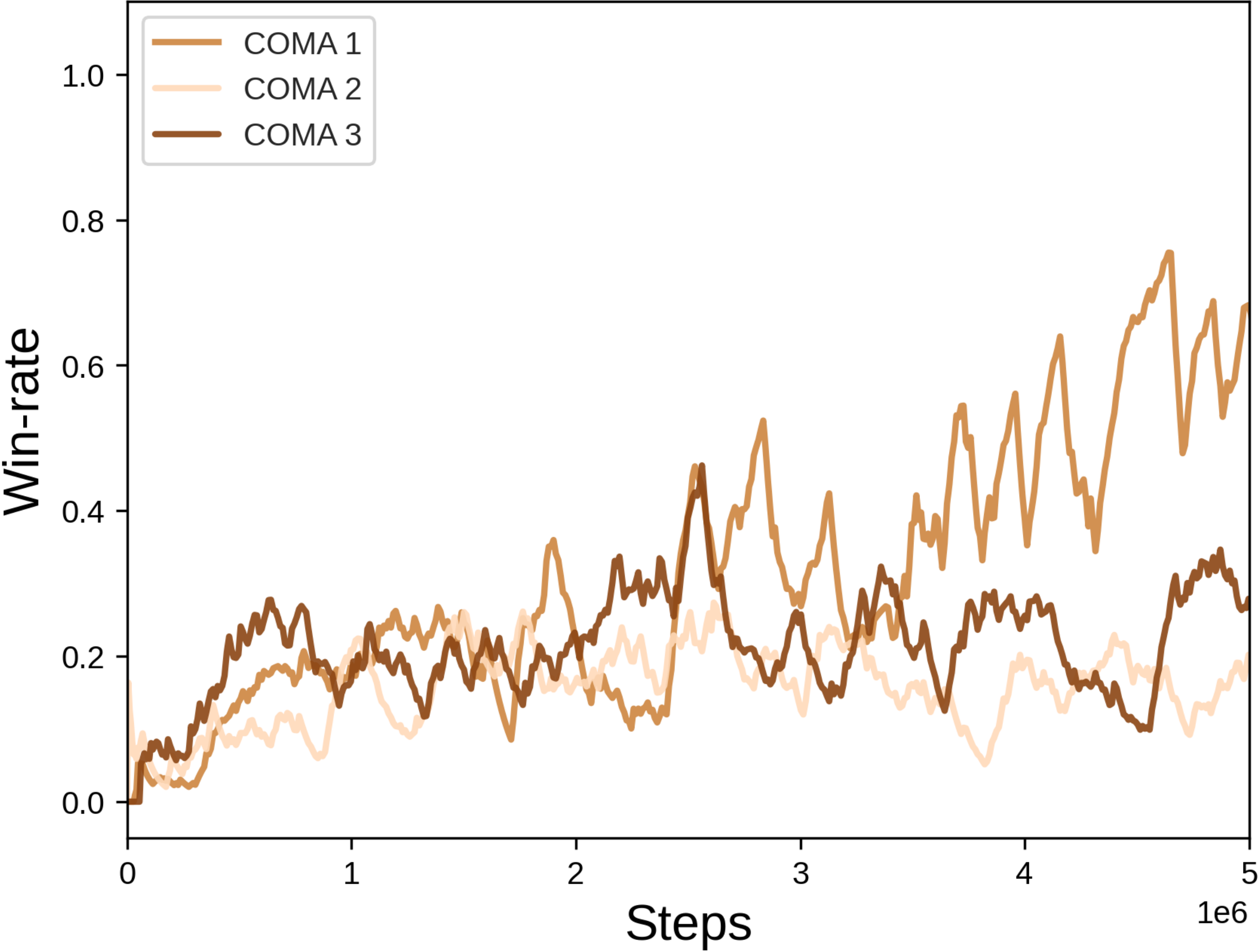}
            \caption{Defense infantry}
            \label{fig:app_coma_episode_def_inf}
        \end{subfigure}%
        \begin{subfigure}{0.26\columnwidth}
            \includegraphics[width=\columnwidth]{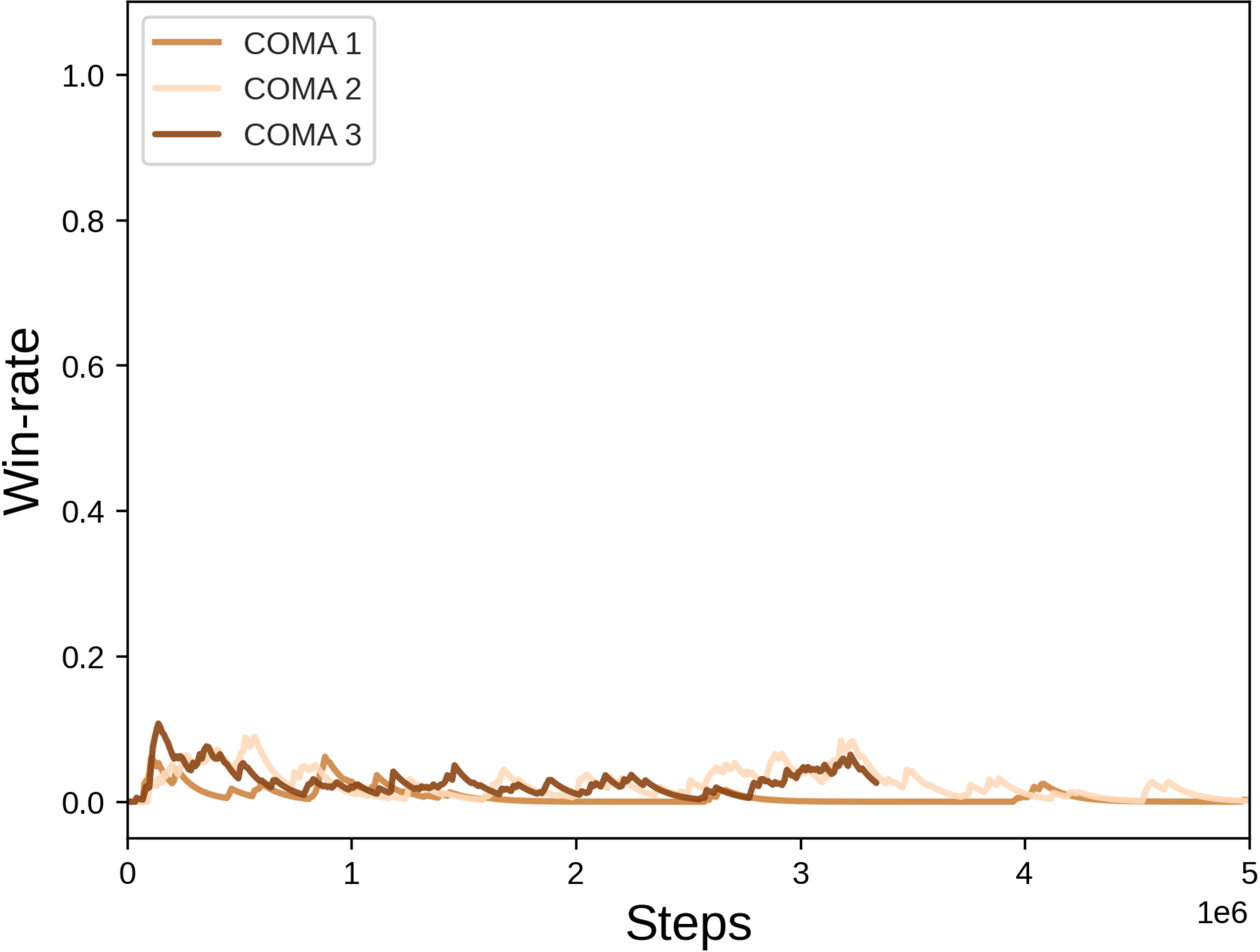}
            \caption{Defense armored}
            \label{fig:app_coma_episode_def_arm}
        \end{subfigure}%
        \begin{subfigure}{0.26\columnwidth}
            \includegraphics[width=\columnwidth]{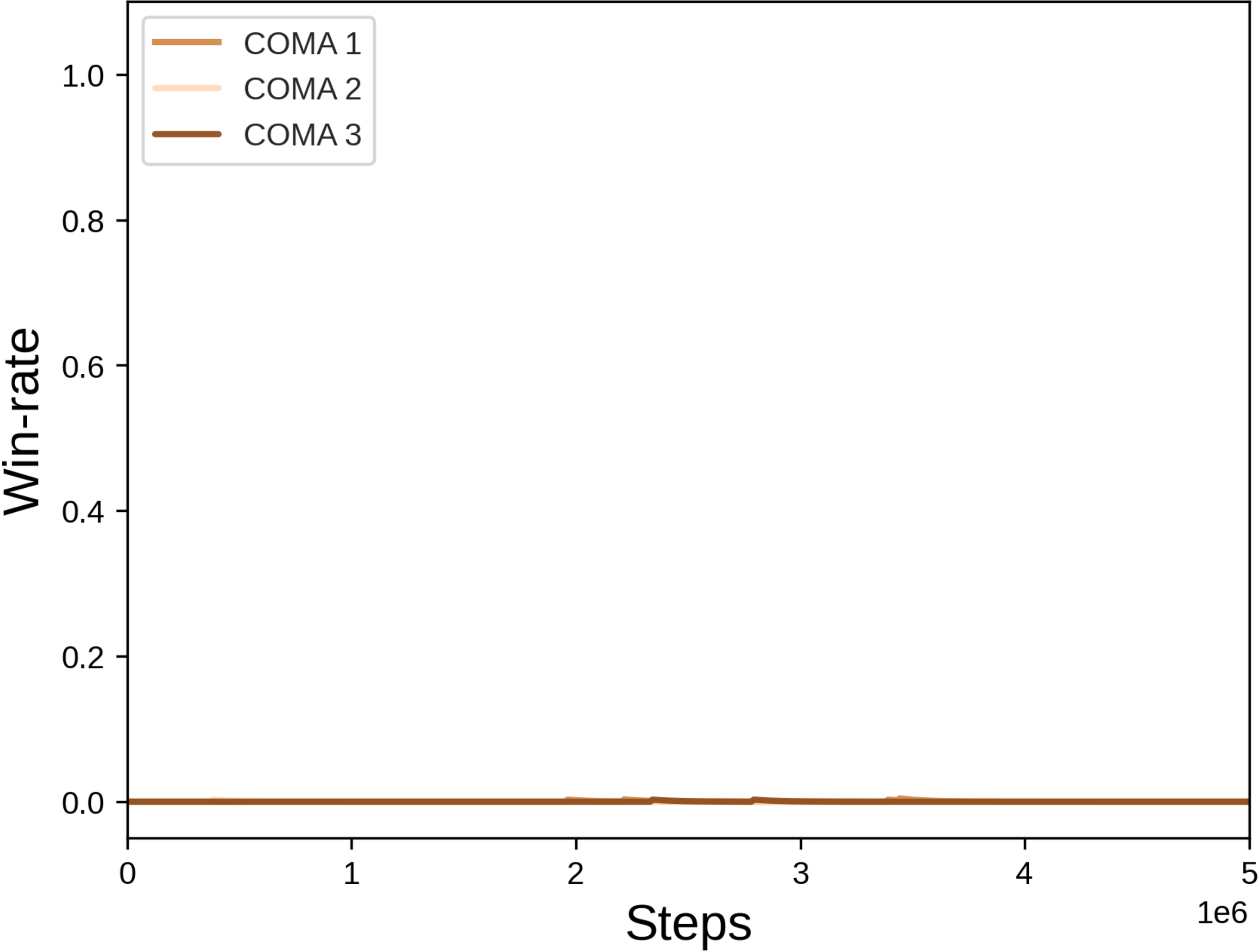}
            \caption{Defense outnumbered}
            \label{fig:app_coma_episode_def_out}
        \end{subfigure}%
        
        \begin{subfigure}{0.26\columnwidth}
            \includegraphics[width=\columnwidth]{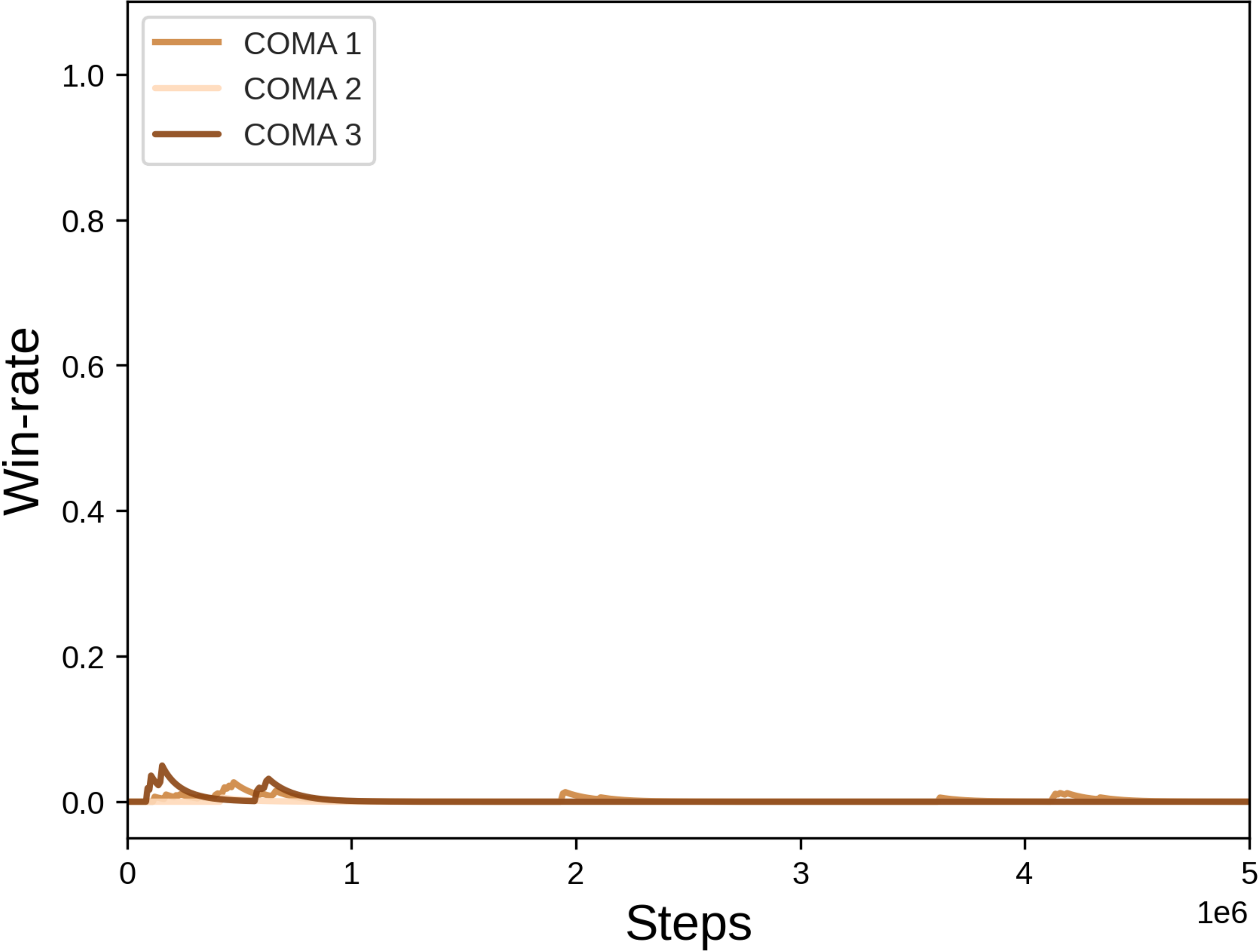}
            \caption{Offense near}
            \label{fig:app_coma_episode_off_near}
        \end{subfigure}%
        \begin{subfigure}{0.26\columnwidth}
            \includegraphics[width=\columnwidth]{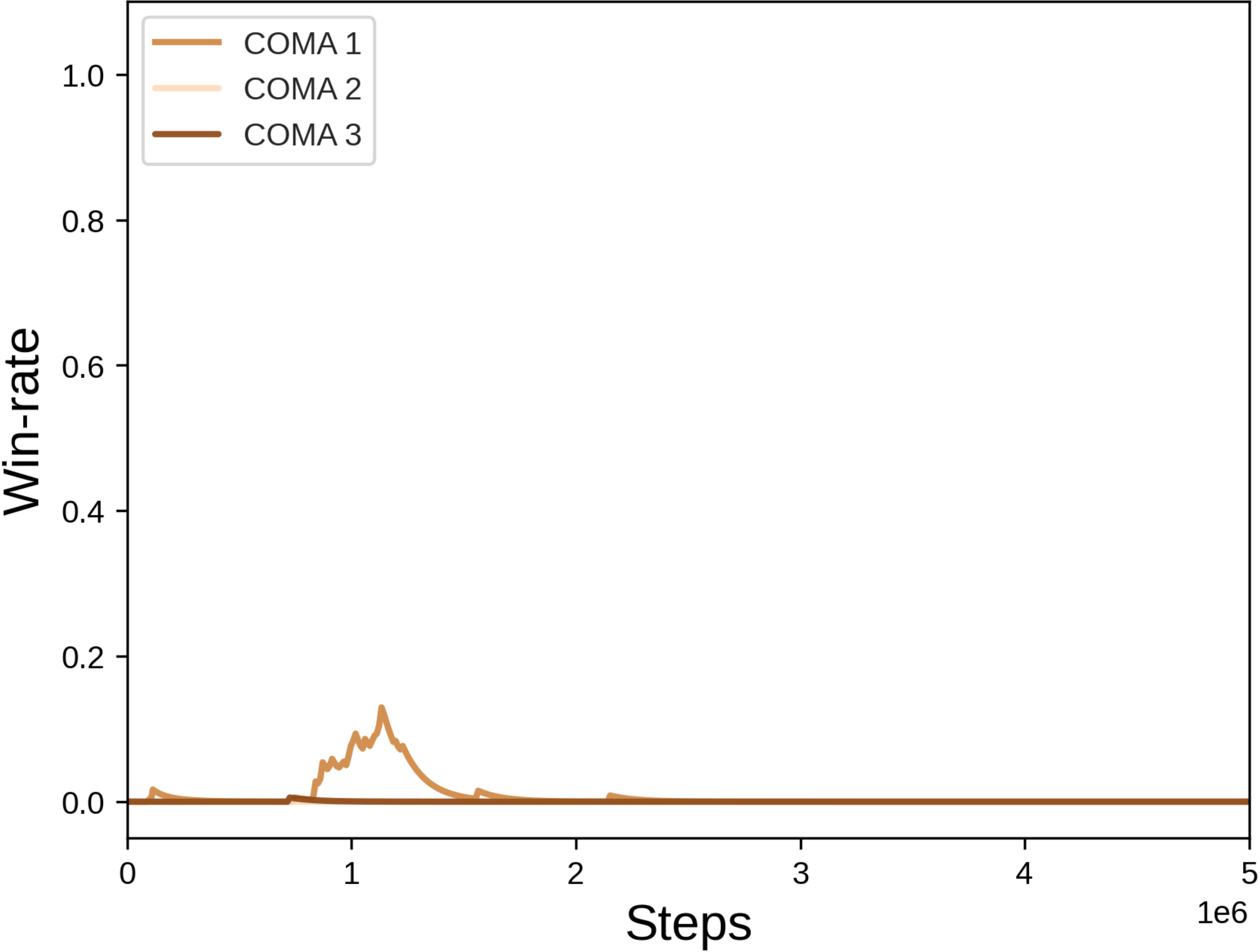}
            \caption{Offense distant}
            \label{fig:app_coma_episode_off_dist}
        \end{subfigure}%
        \begin{subfigure}{0.26\columnwidth}
            \includegraphics[width=\columnwidth]{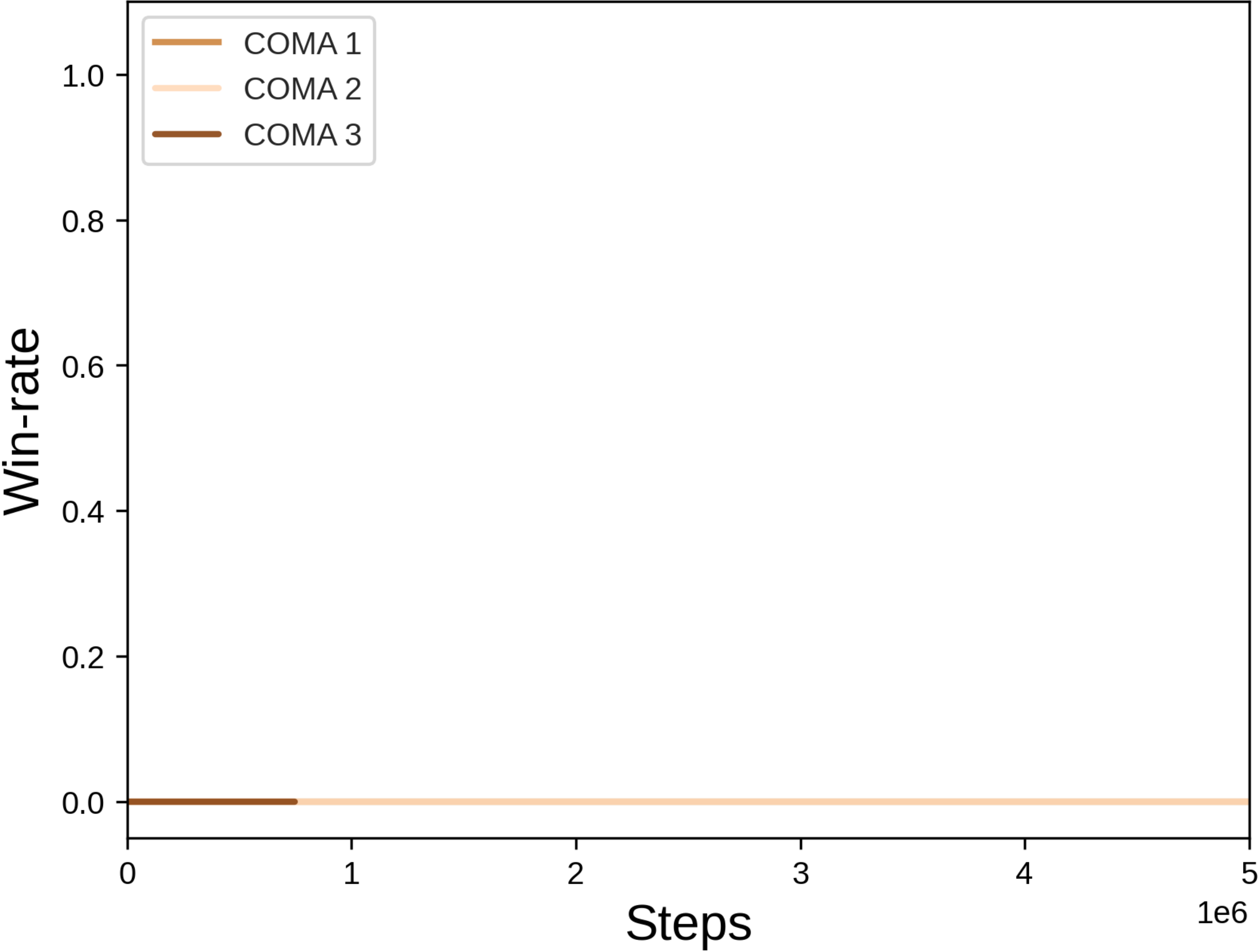}
            \caption{Offense complicated}
            \label{fig:app_coma_episode_off_com}
        \end{subfigure}%
        
        \begin{subfigure}{0.27\columnwidth}
            \includegraphics[width=\columnwidth]{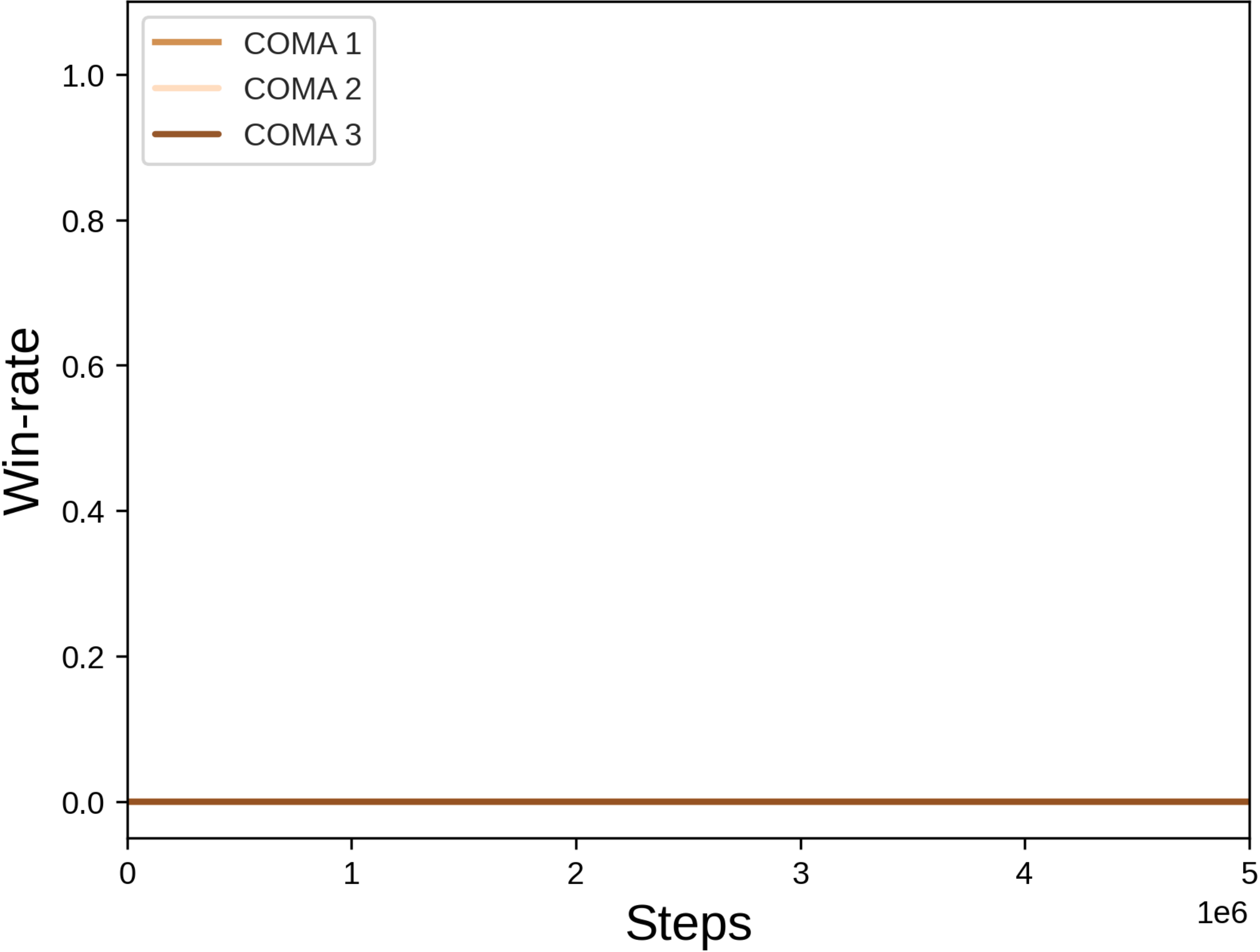}
            \caption{Offense hard}
            \label{fig:app_coma_episode_off_hard}
        \end{subfigure}%
        \begin{subfigure}{0.27\columnwidth}
            \includegraphics[width=\columnwidth]{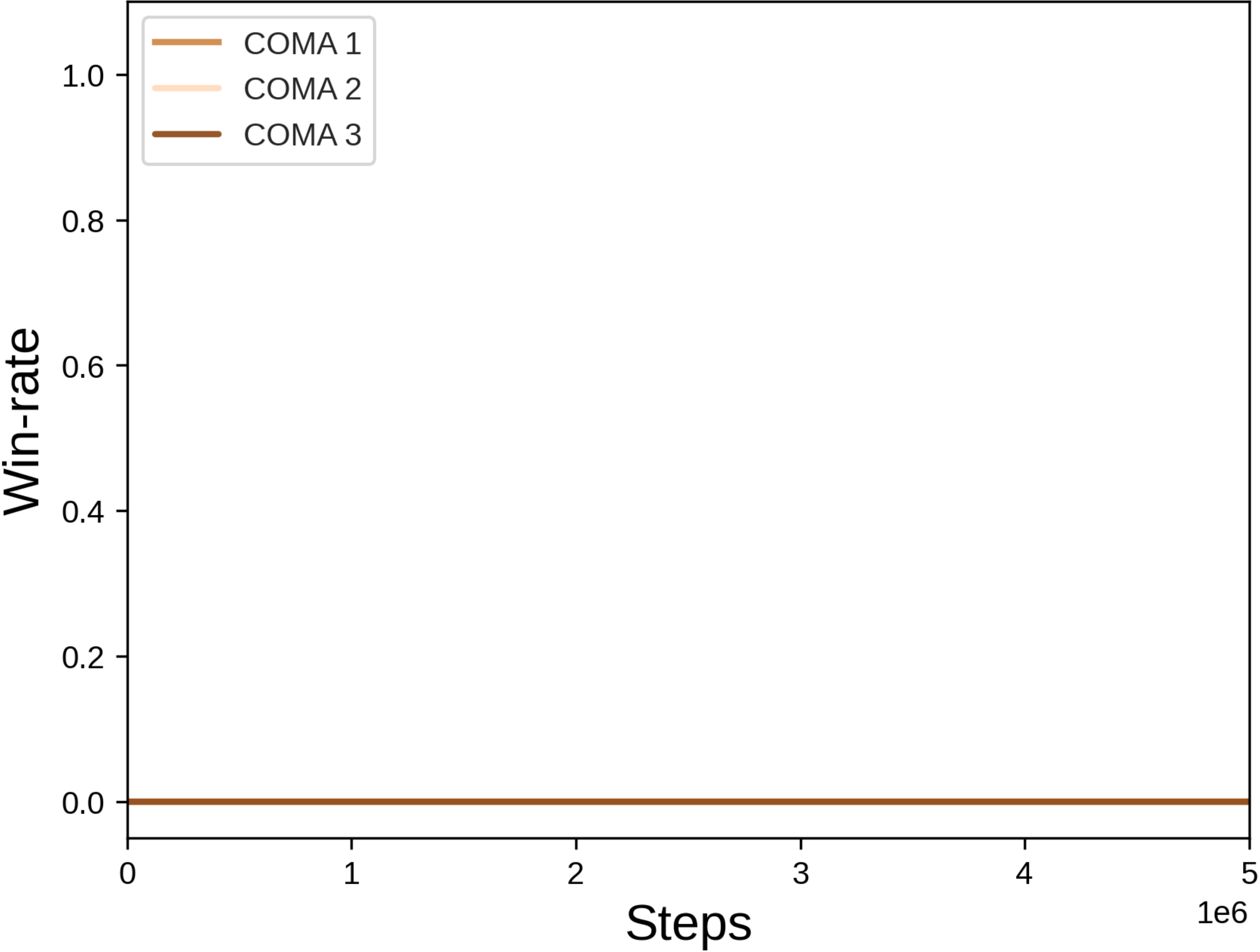}
            \caption{Offense superhard}
            \label{fig:app_coma_episode_off_super}
        \end{subfigure}%
    \caption{COMA trained on the sequential episodic buffer}
    \label{fig:app_coma_episode}
}
\end{figure}

\begin{figure}[!ht]{
    \centering
        \begin{subfigure}{0.26\columnwidth}
            \includegraphics[width=\columnwidth]{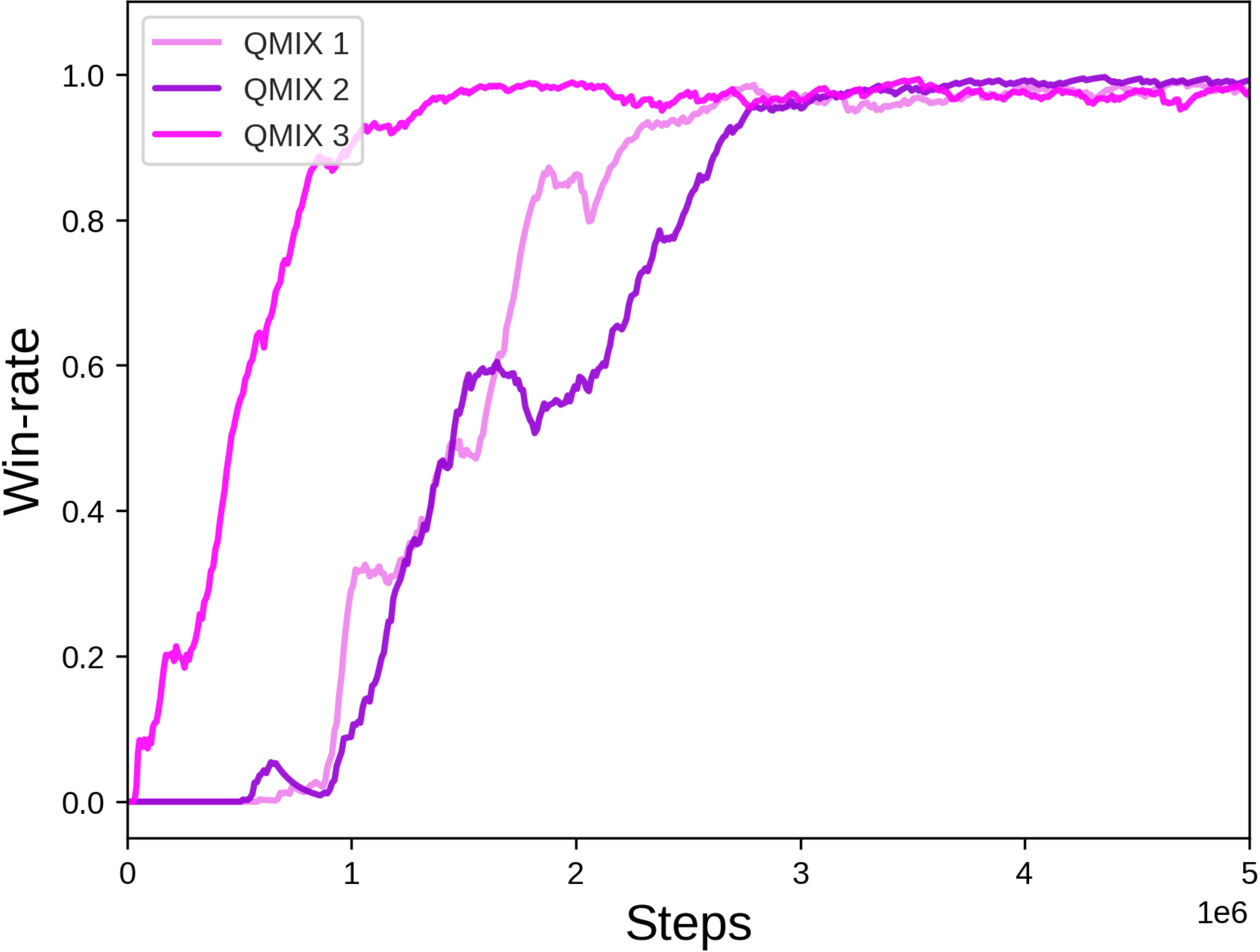}
            \caption{Defense infantry}
            \label{fig:app_qmix_episode_def_inf}
        \end{subfigure}%
        \begin{subfigure}{0.26\columnwidth}
            \includegraphics[width=\columnwidth]{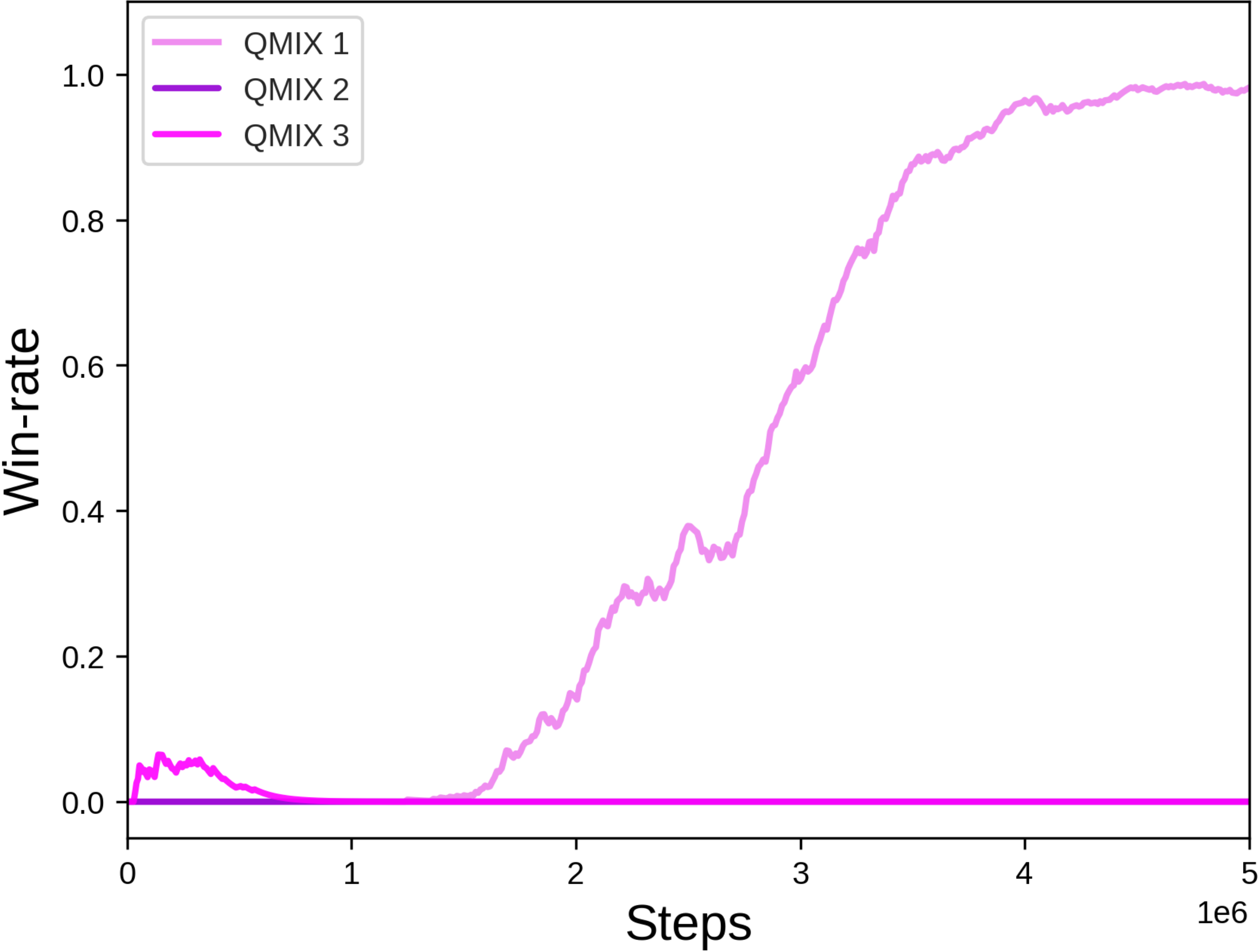}
            \caption{Defense armored}
            \label{fig:app_qmix_episode_def_arm}
        \end{subfigure}%
        \begin{subfigure}{0.26\columnwidth}
            \includegraphics[width=\columnwidth]{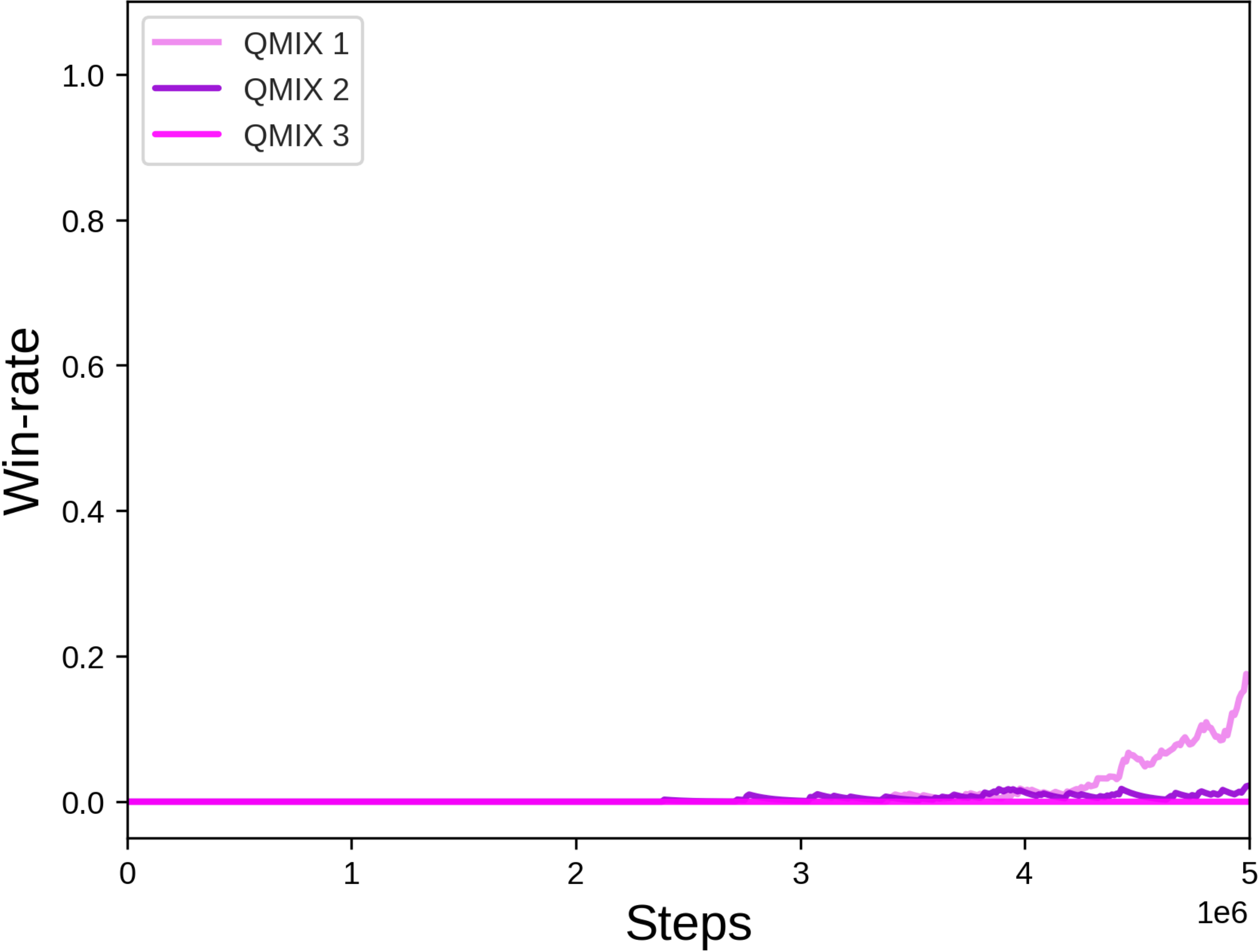}
            \caption{Defense outnumbered}
            \label{fig:app_qmix_episode_out}
        \end{subfigure}%
        
        \begin{subfigure}{0.26\columnwidth}
            \includegraphics[width=\columnwidth]{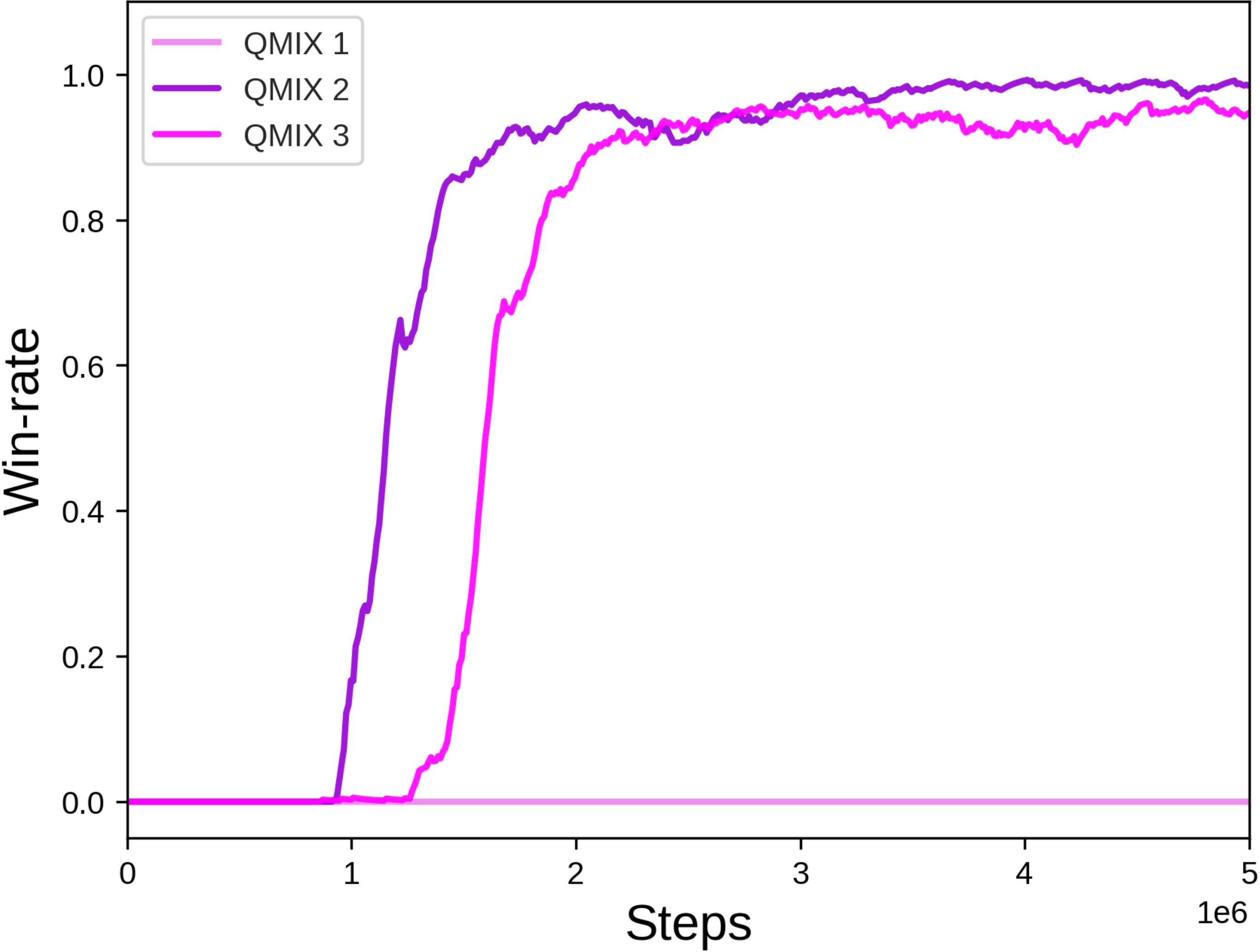}
            \caption{Offense near}
            \label{fig:app_qmix_episode_off_near}
        \end{subfigure}%
        \begin{subfigure}{0.26\columnwidth}
            \includegraphics[width=\columnwidth]{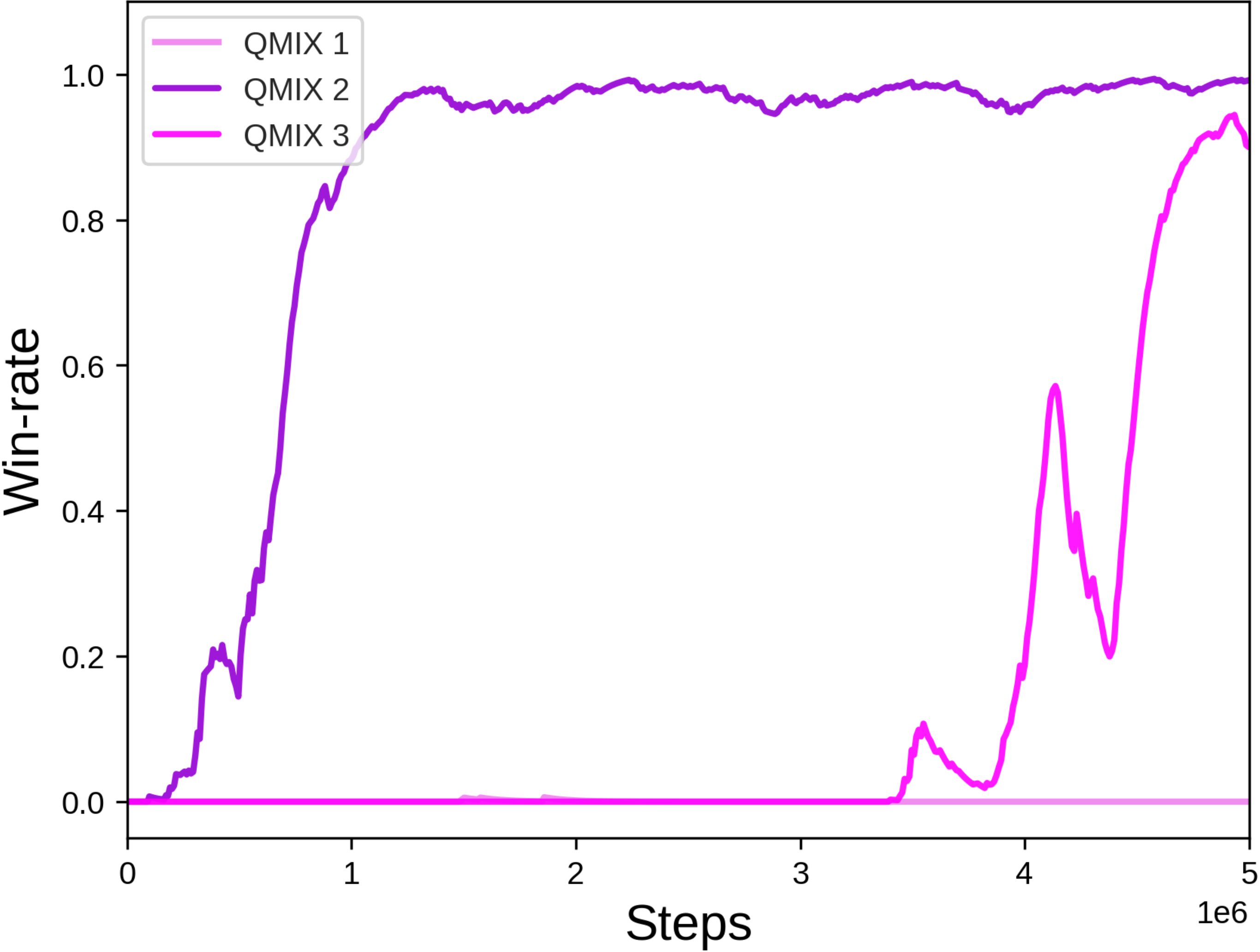}
            \caption{Offense distant}
            \label{fig:app_qmix_episode_dist}
        \end{subfigure}%
        \begin{subfigure}{0.26\columnwidth}
            \includegraphics[width=\columnwidth]{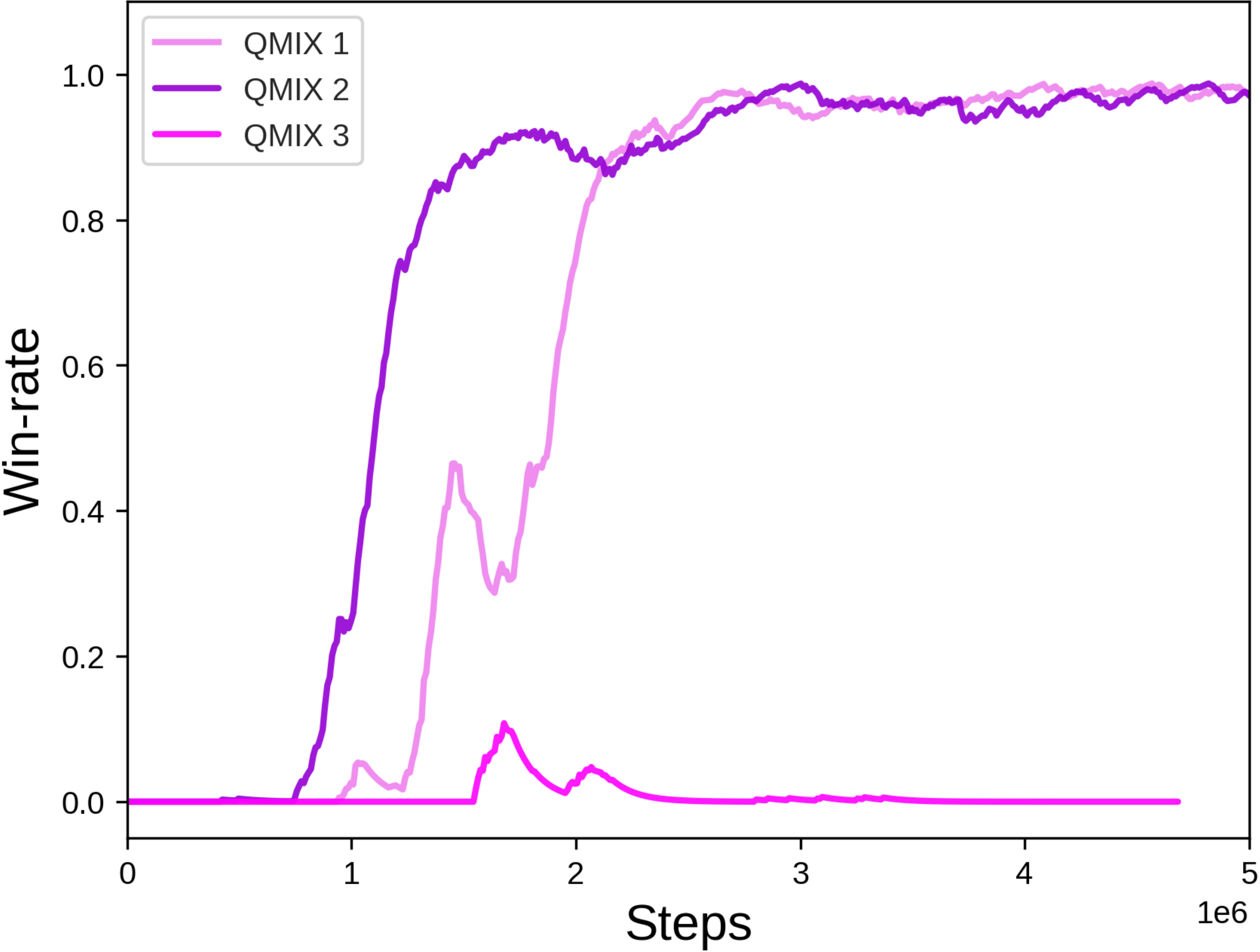}
            \caption{Offense complicated}
            \label{fig:app_qmix_episode_com}
        \end{subfigure}%
        
        \begin{subfigure}{0.27\columnwidth}
            \includegraphics[width=\columnwidth]{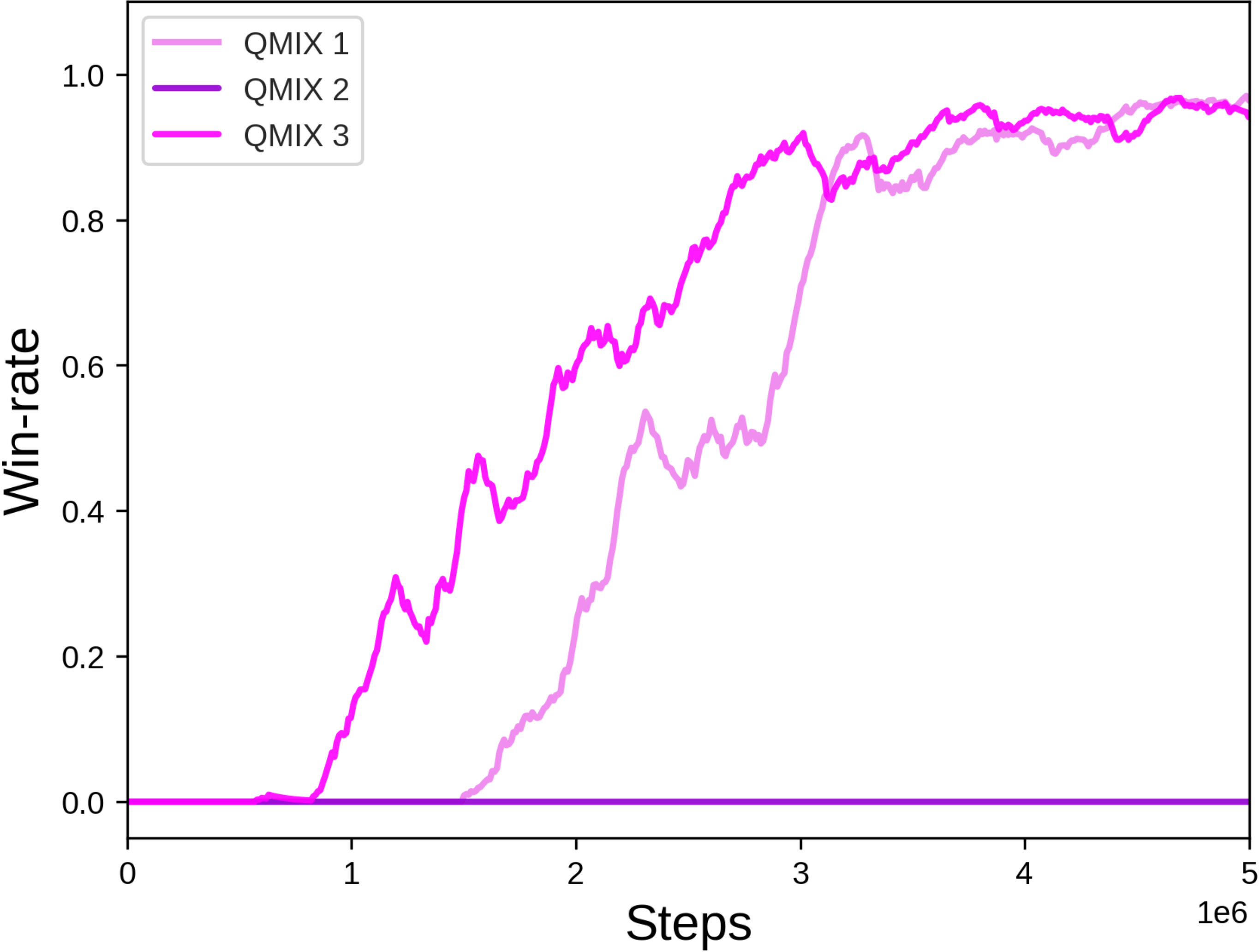}
            \caption{Offense hard}
            \label{fig:app_qmix_episode_off_hard}
        \end{subfigure}%
        \begin{subfigure}{0.27\columnwidth}
            \includegraphics[width=\columnwidth]{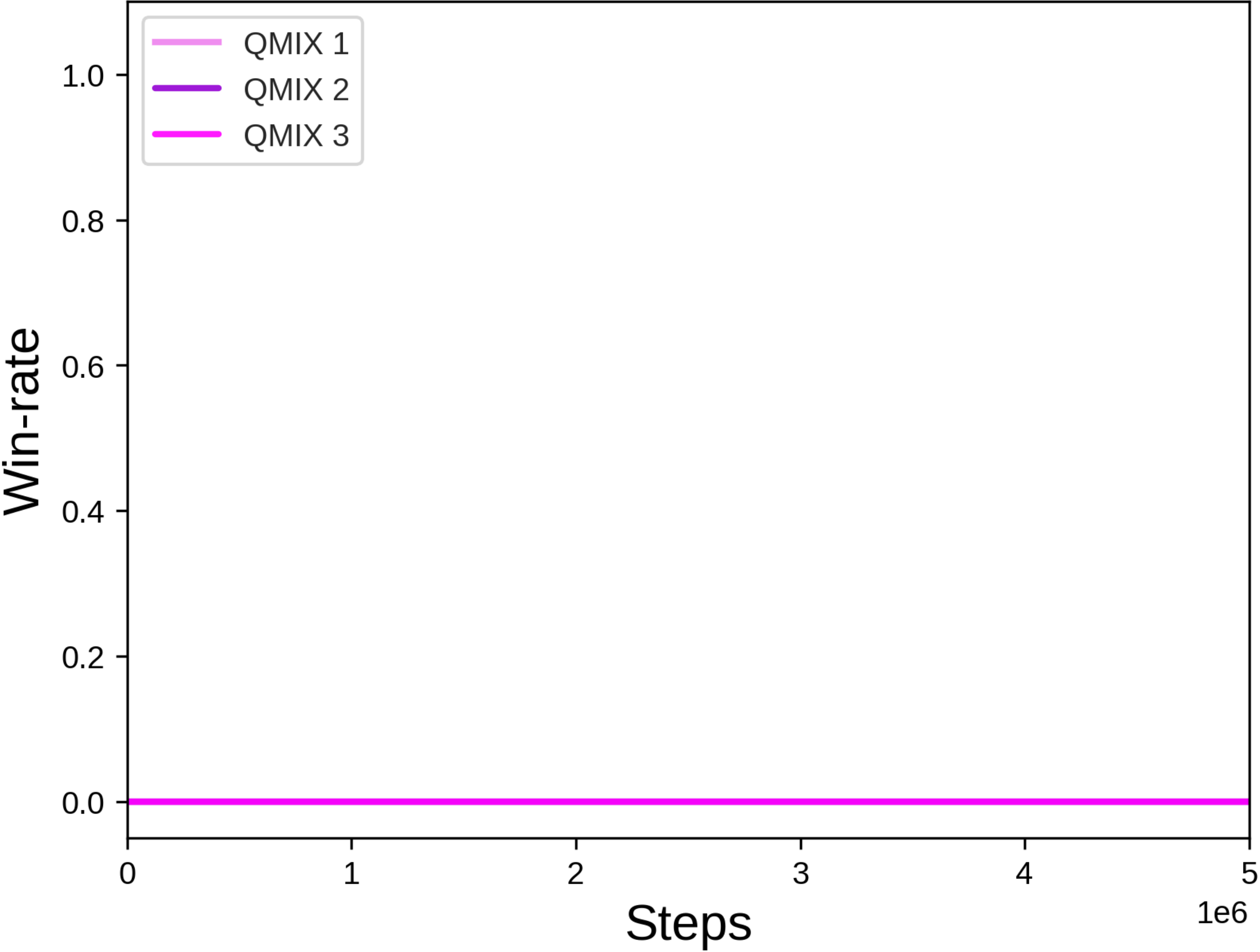}
            \caption{Offense superhard}
            \label{fig:app_qmix_episode_off_super}
        \end{subfigure}%
    \caption{QMIX trained on the sequential episodic buffer}
    \label{fig:app_qmix_episode}
}
\end{figure}

\begin{figure}[!ht]{
    \centering
        \begin{subfigure}{0.26\columnwidth}
            \includegraphics[width=\columnwidth]{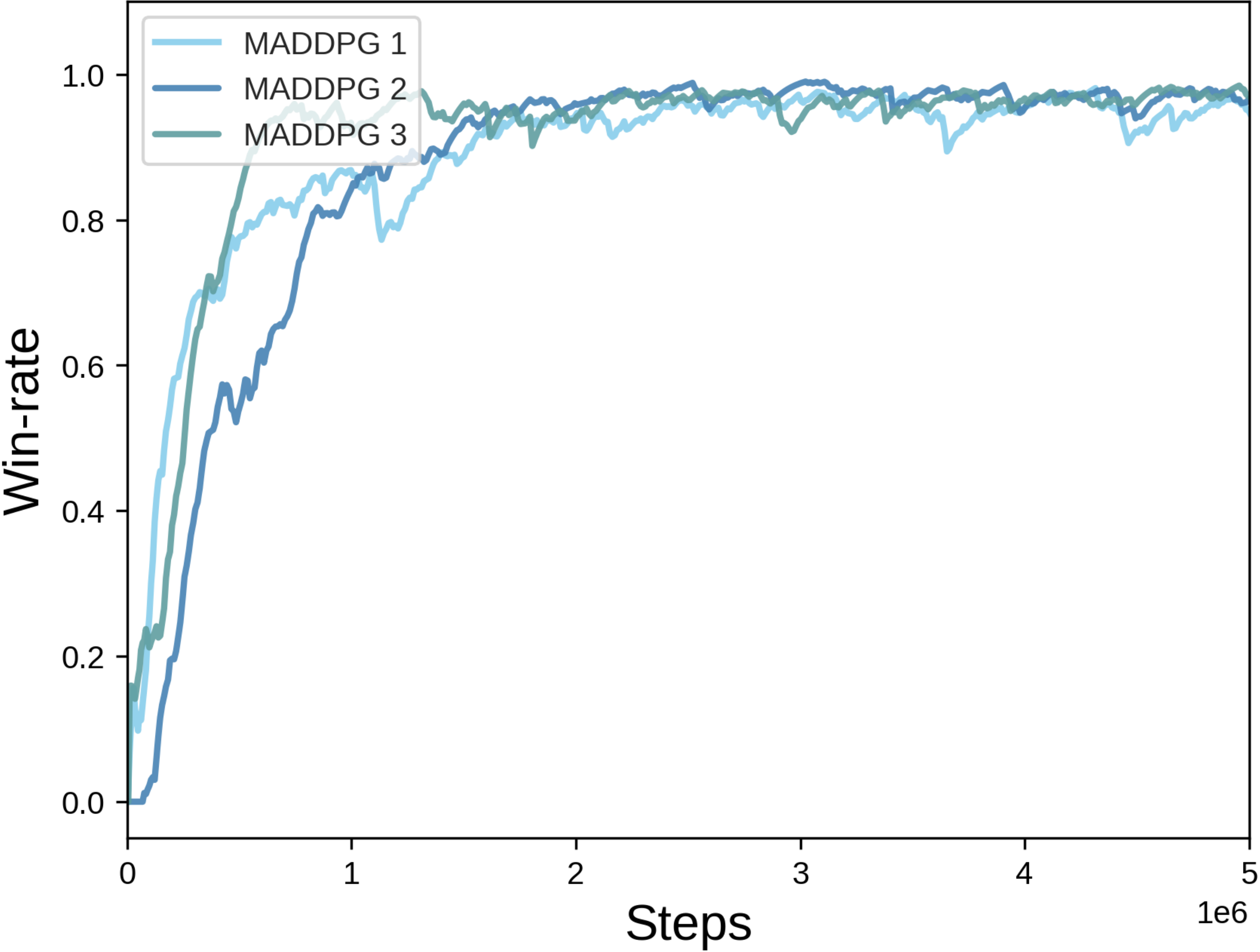}
            \caption{Defense infantry}
            \label{fig:app_maddpg_episode_def_inf}
        \end{subfigure}%
        \begin{subfigure}{0.26\columnwidth}
            \includegraphics[width=\columnwidth]{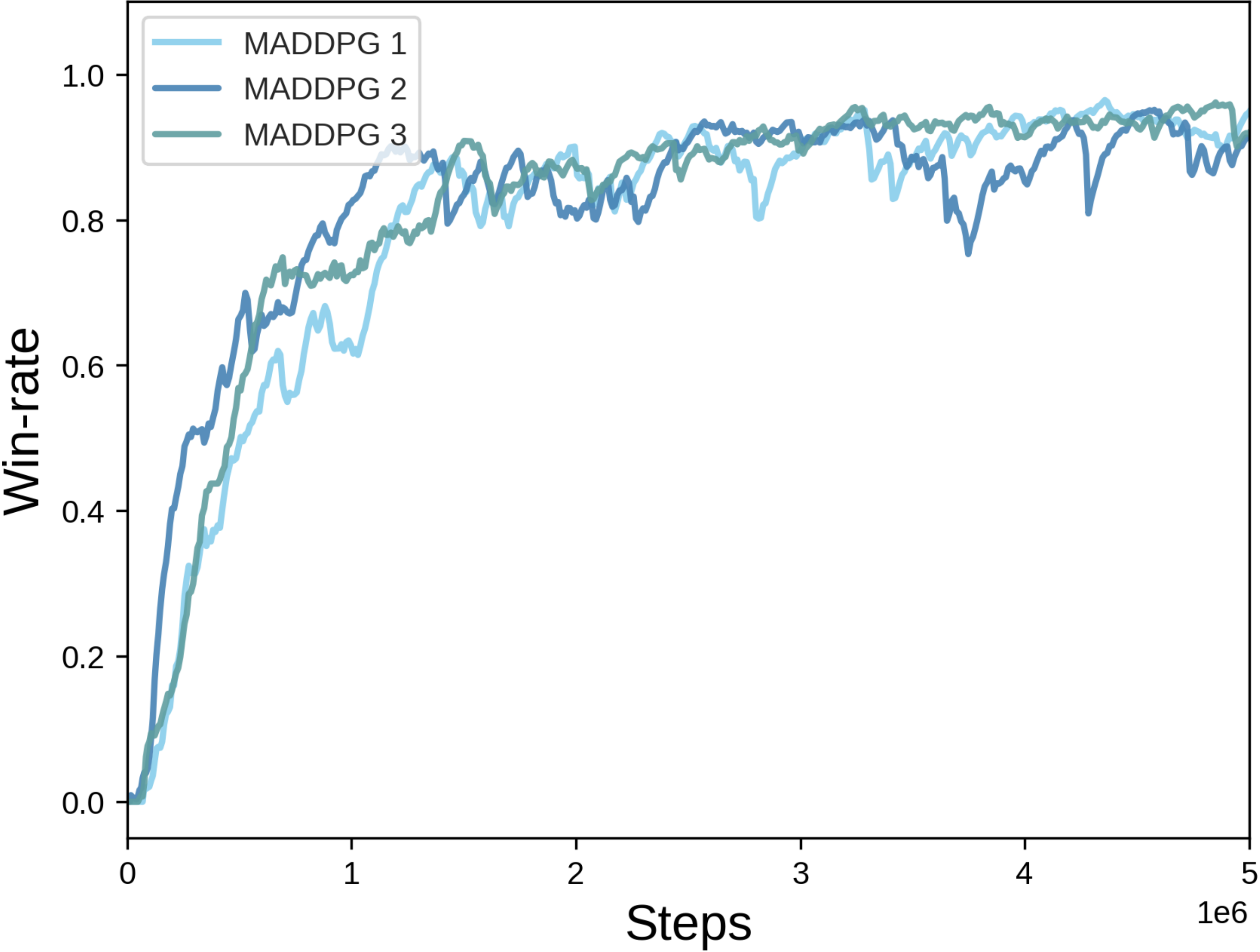}
            \caption{Defense armored}
            \label{fig:app_maddpg_episode_def_arm}
        \end{subfigure}%
        \begin{subfigure}{0.26\columnwidth}
            \includegraphics[width=\columnwidth]{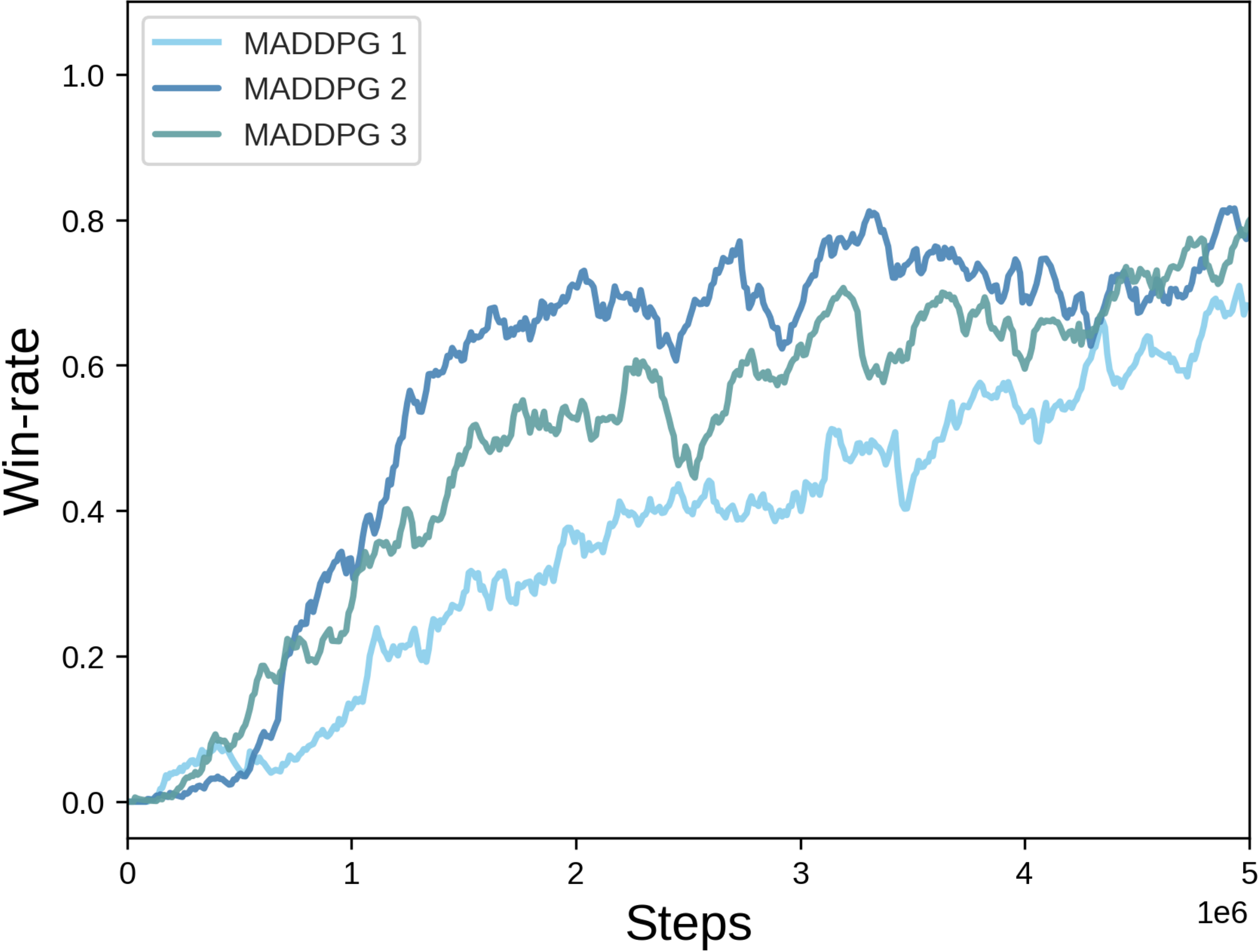}
            \caption{Defense outnumbered}
            \label{fig:app_maddpg_episode_def_out}
        \end{subfigure}%
        
        \begin{subfigure}{0.26\columnwidth}
            \includegraphics[width=\columnwidth]{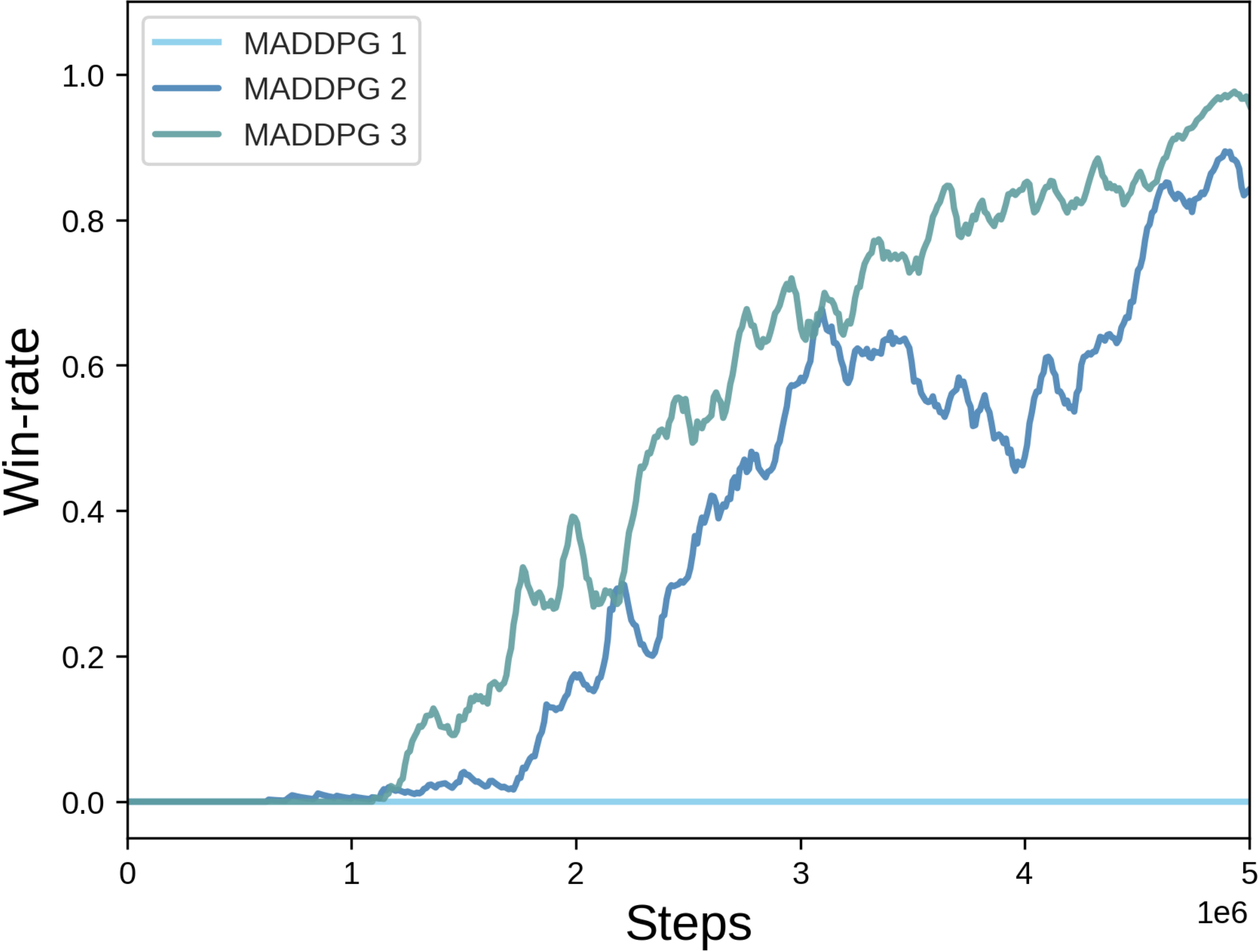}
            \caption{Offense near}
            \label{fig:app_maddpg_episode_off_near}
        \end{subfigure}%
        \begin{subfigure}{0.26\columnwidth}
            \includegraphics[width=\columnwidth]{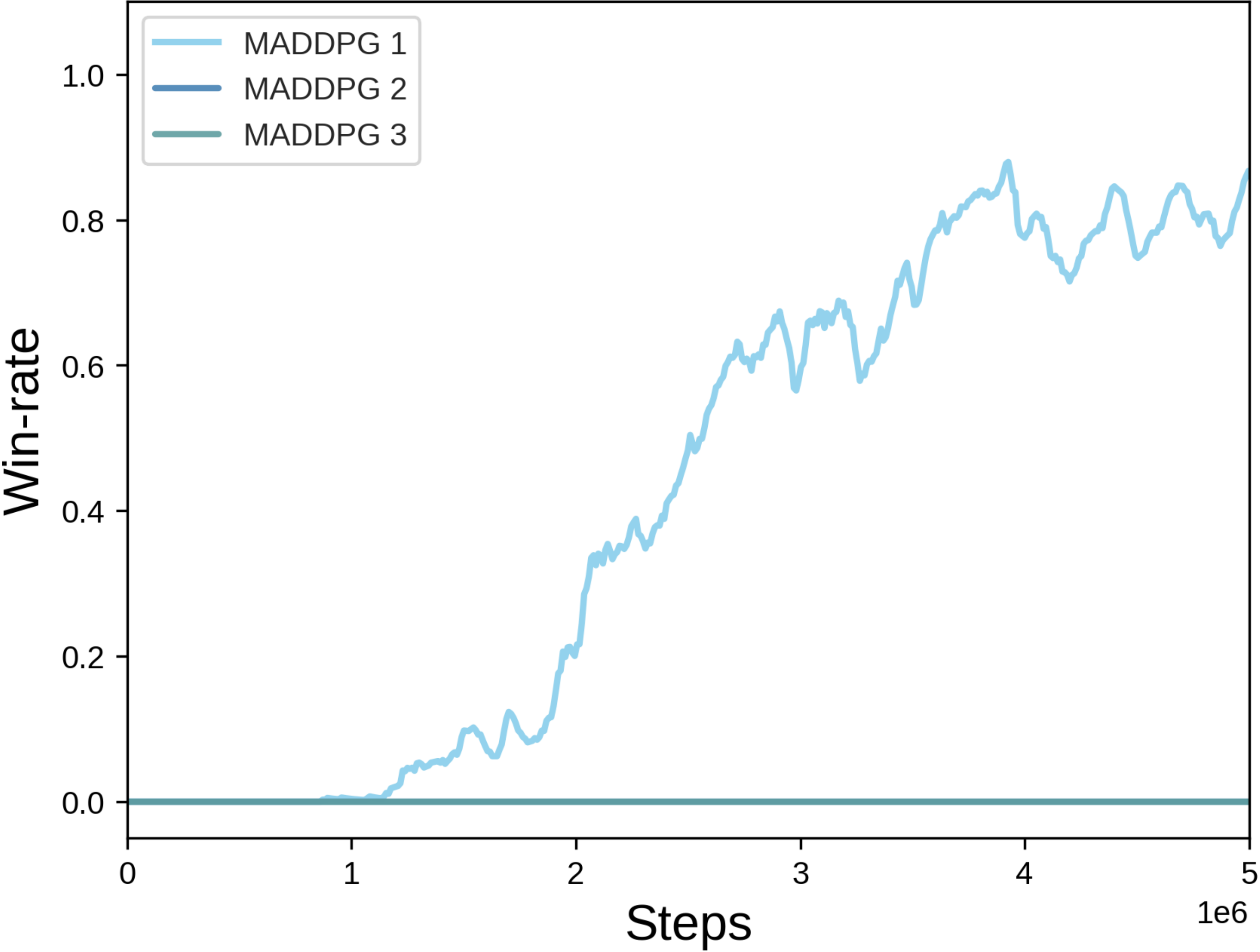}
            \caption{Offense distant}
            \label{fig:app_maddpg_episode_off_dist}
        \end{subfigure}%
        \begin{subfigure}{0.26\columnwidth}
            \includegraphics[width=\columnwidth]{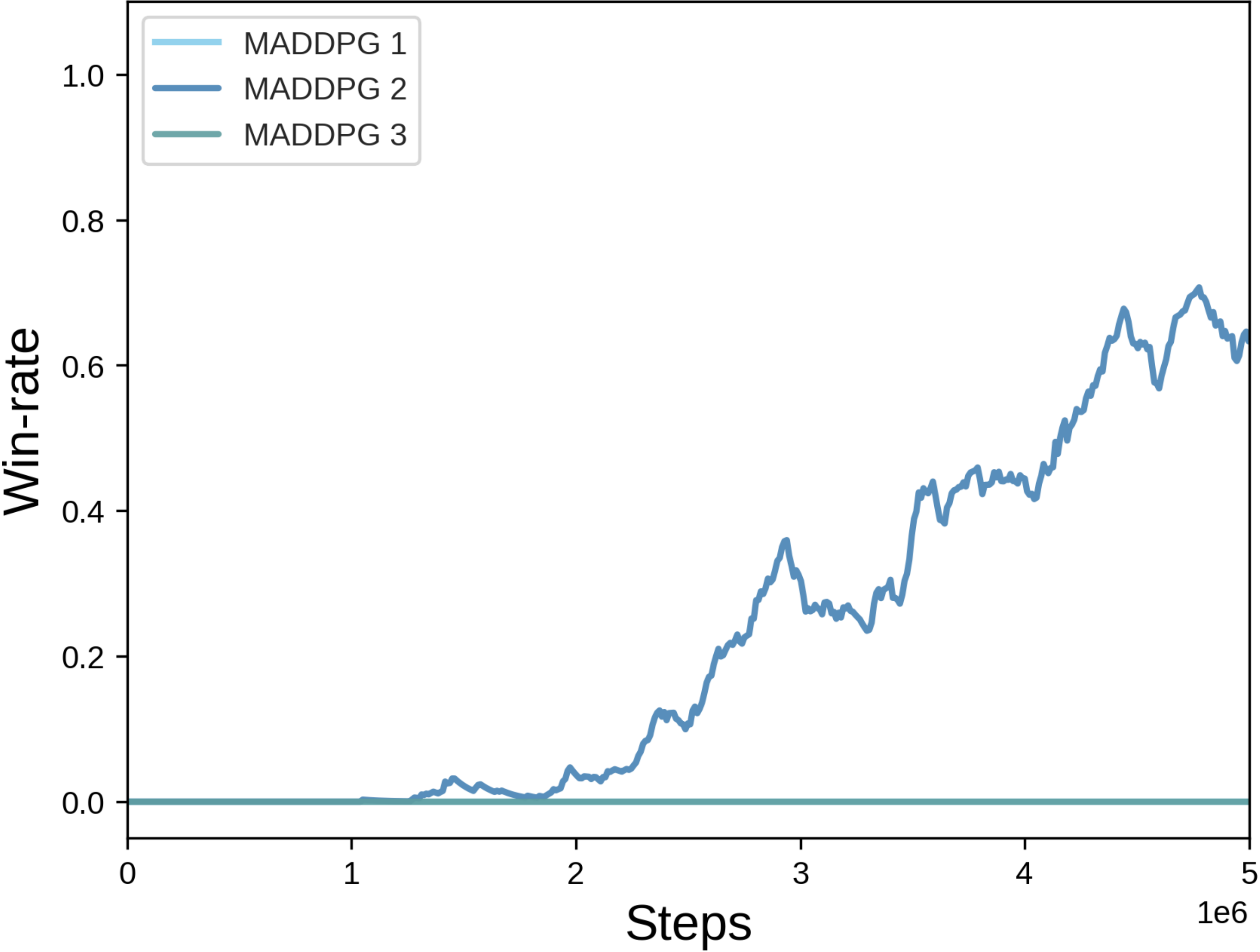}
            \caption{Offense complicated}
            \label{fig:app_maddpg_episode_off_com}
        \end{subfigure}%
        
        \begin{subfigure}{0.27\columnwidth}
            \includegraphics[width=\columnwidth]{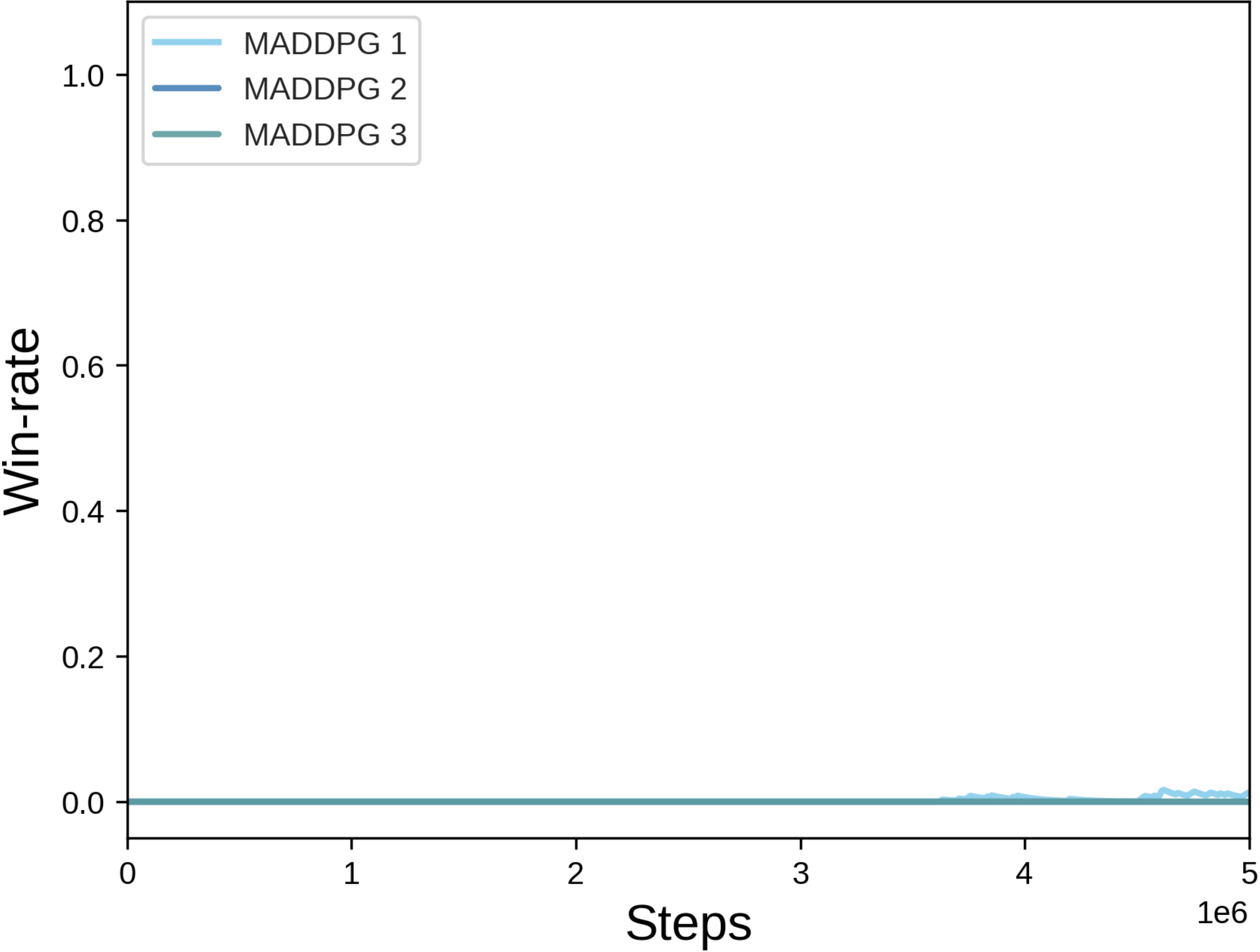}
            \caption{Offense hard}
            \label{fig:app_maddpg_episode_off_hard}
        \end{subfigure}%
        \begin{subfigure}{0.27\columnwidth}
            \includegraphics[width=\columnwidth]{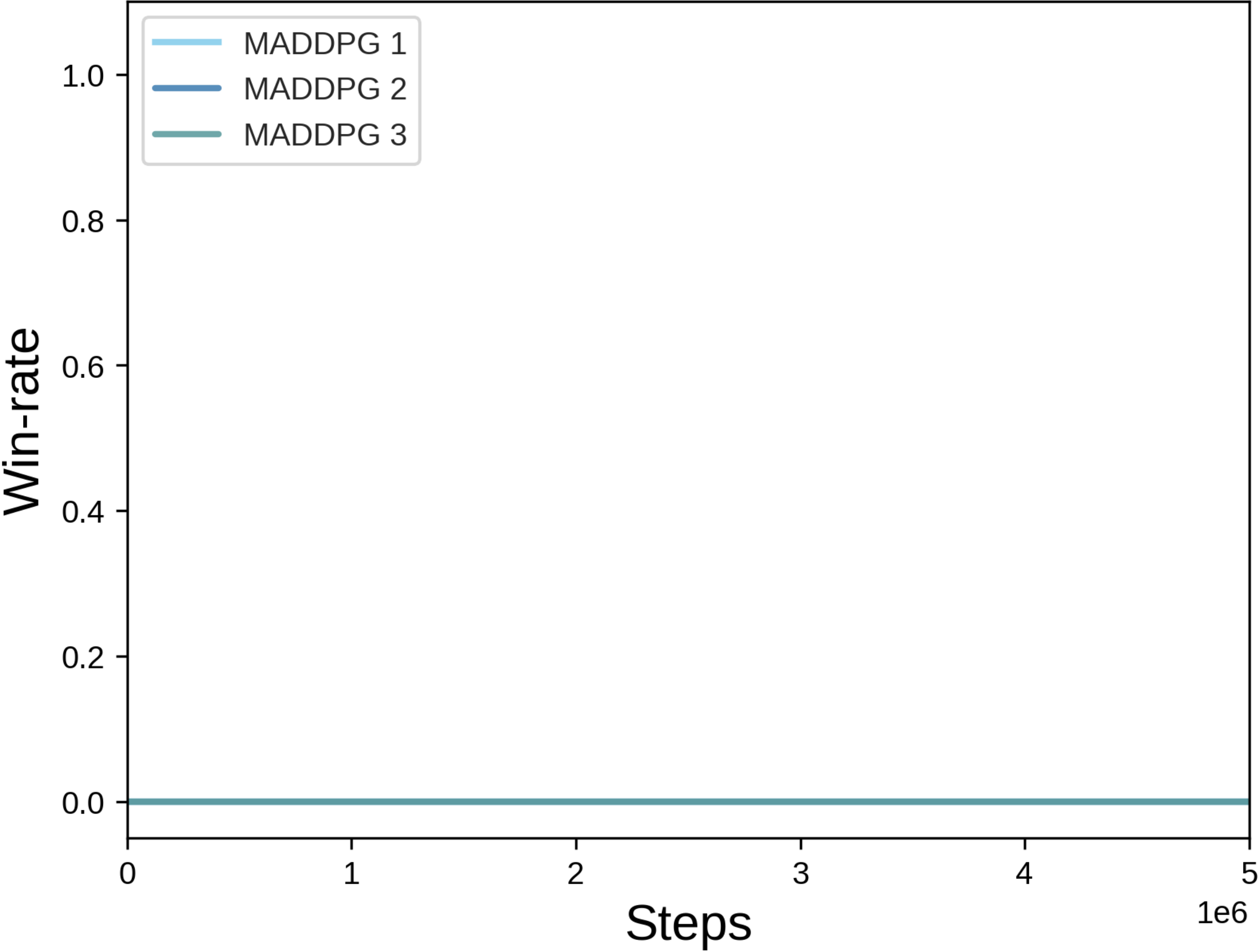}
            \caption{Offense superhard}
            \label{fig:app_maddpg_episode_off_super}
        \end{subfigure}%
    \caption{MADDPG trained on the sequential episodic buffer}
    \label{fig:app_maddpg_episode}
}
\end{figure}

\begin{figure}[!ht]{
    \centering
        \begin{subfigure}{0.26\columnwidth}
            \includegraphics[width=\columnwidth]{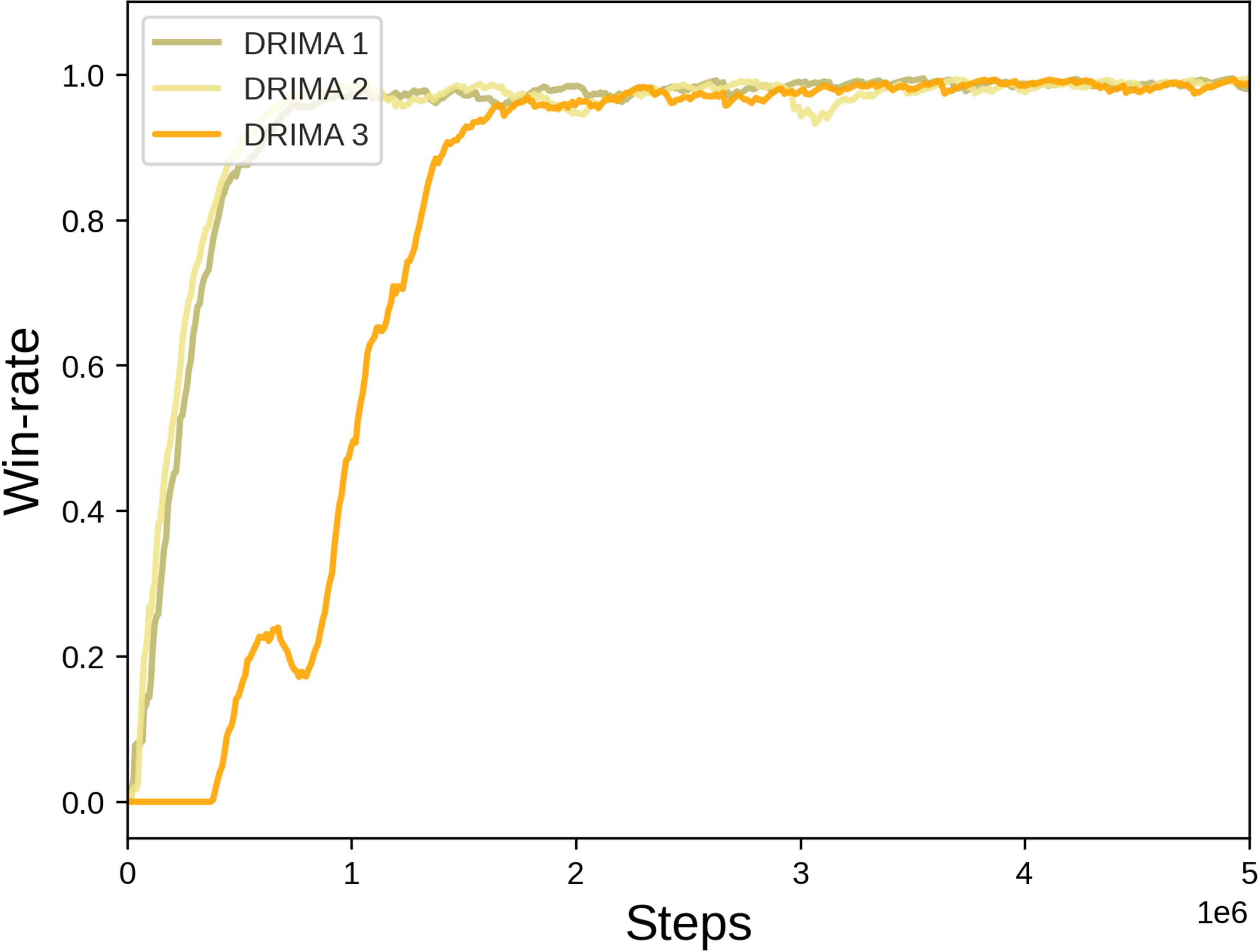}
            \caption{Defense infantry}
            \label{fig:app_drima_episode_def_inf}
        \end{subfigure}%
        \begin{subfigure}{0.26\columnwidth}
            \includegraphics[width=\columnwidth]{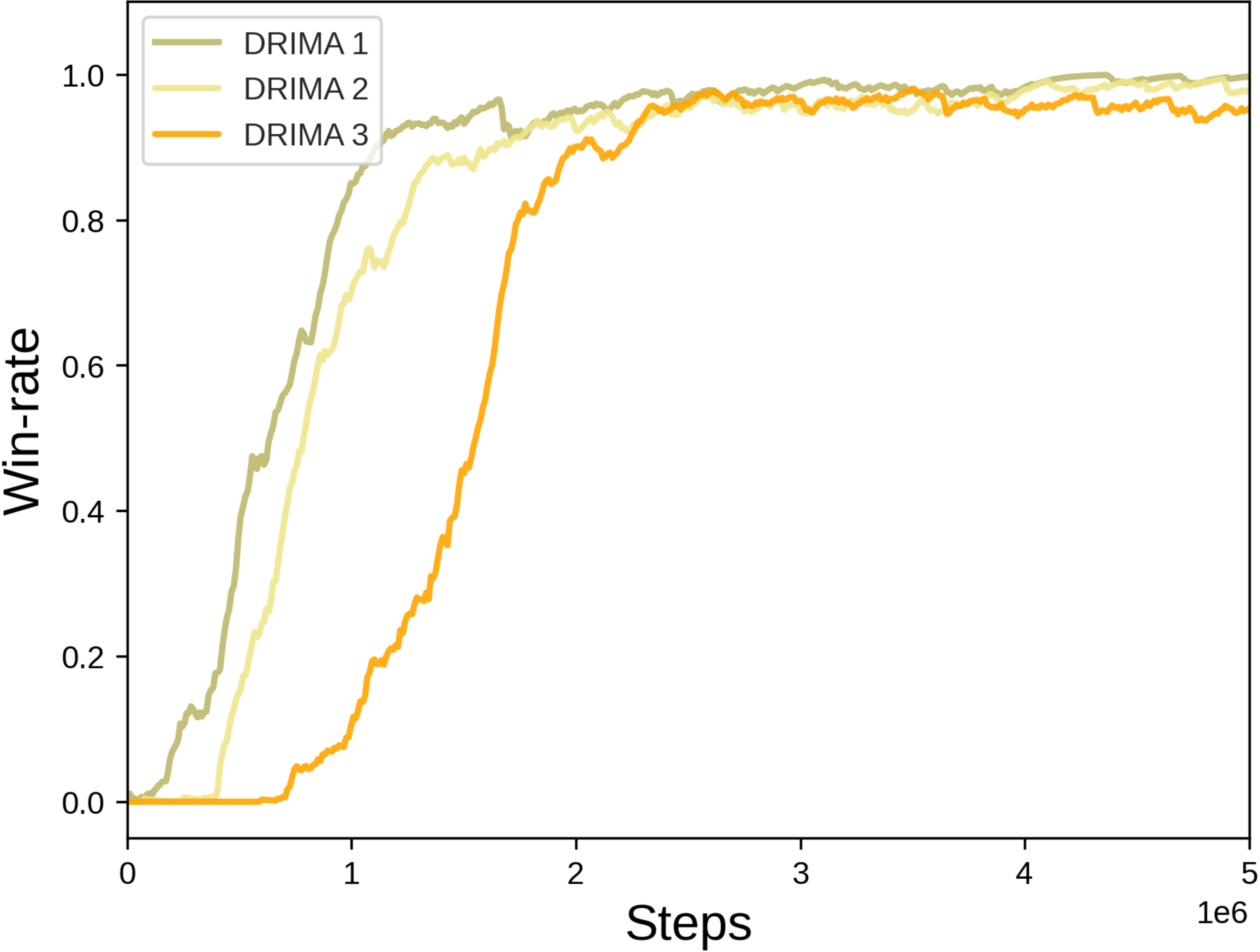}
            \caption{Defense armored}
            \label{fig:app_drima_episode_def_arm}
        \end{subfigure}%
        \begin{subfigure}{0.26\columnwidth}
            \includegraphics[width=\columnwidth]{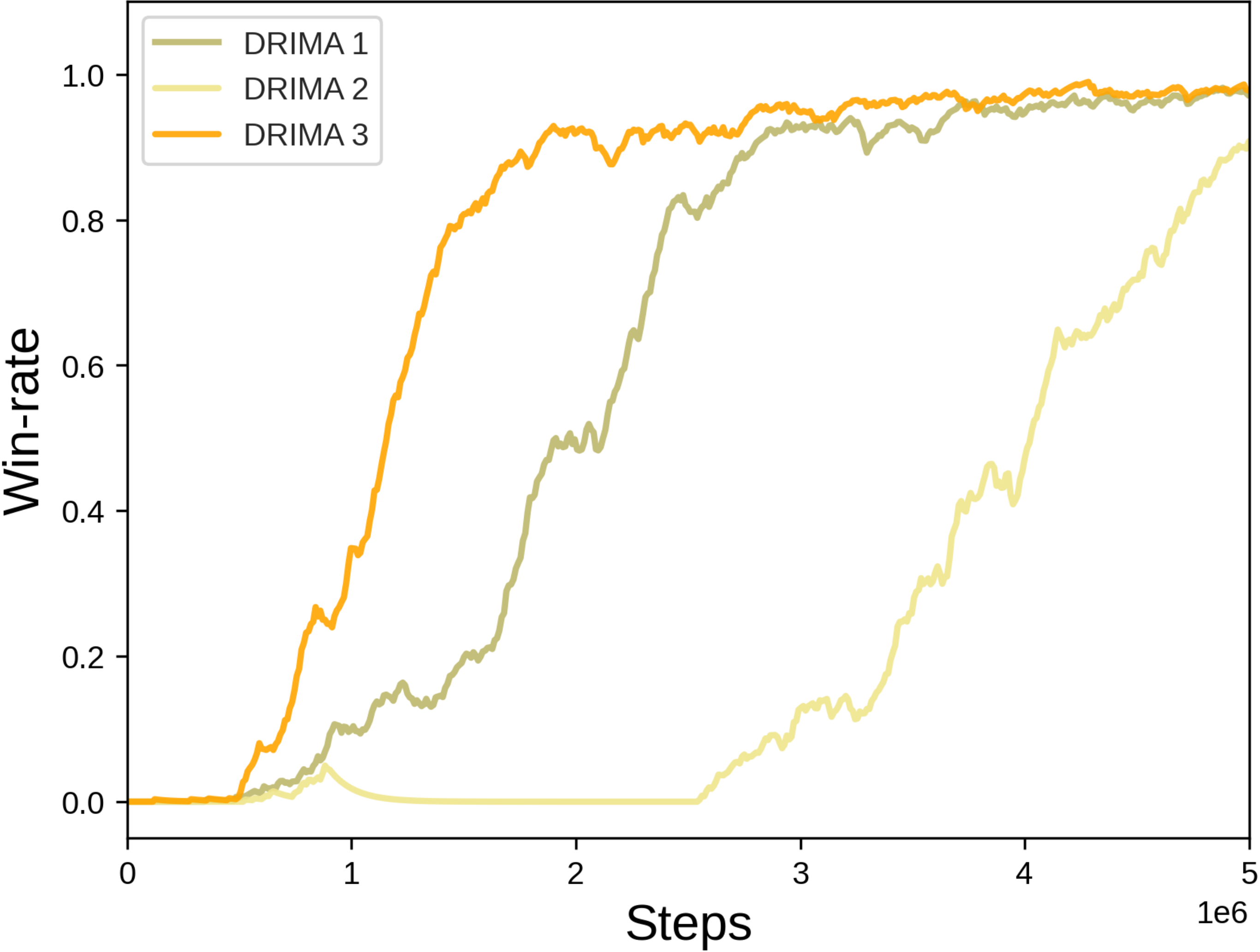}
            \caption{Defense outnumbered}
            \label{fig:app_drima_episode_def_out}
        \end{subfigure}%
        
        \begin{subfigure}{0.26\columnwidth}
            \includegraphics[width=\columnwidth]{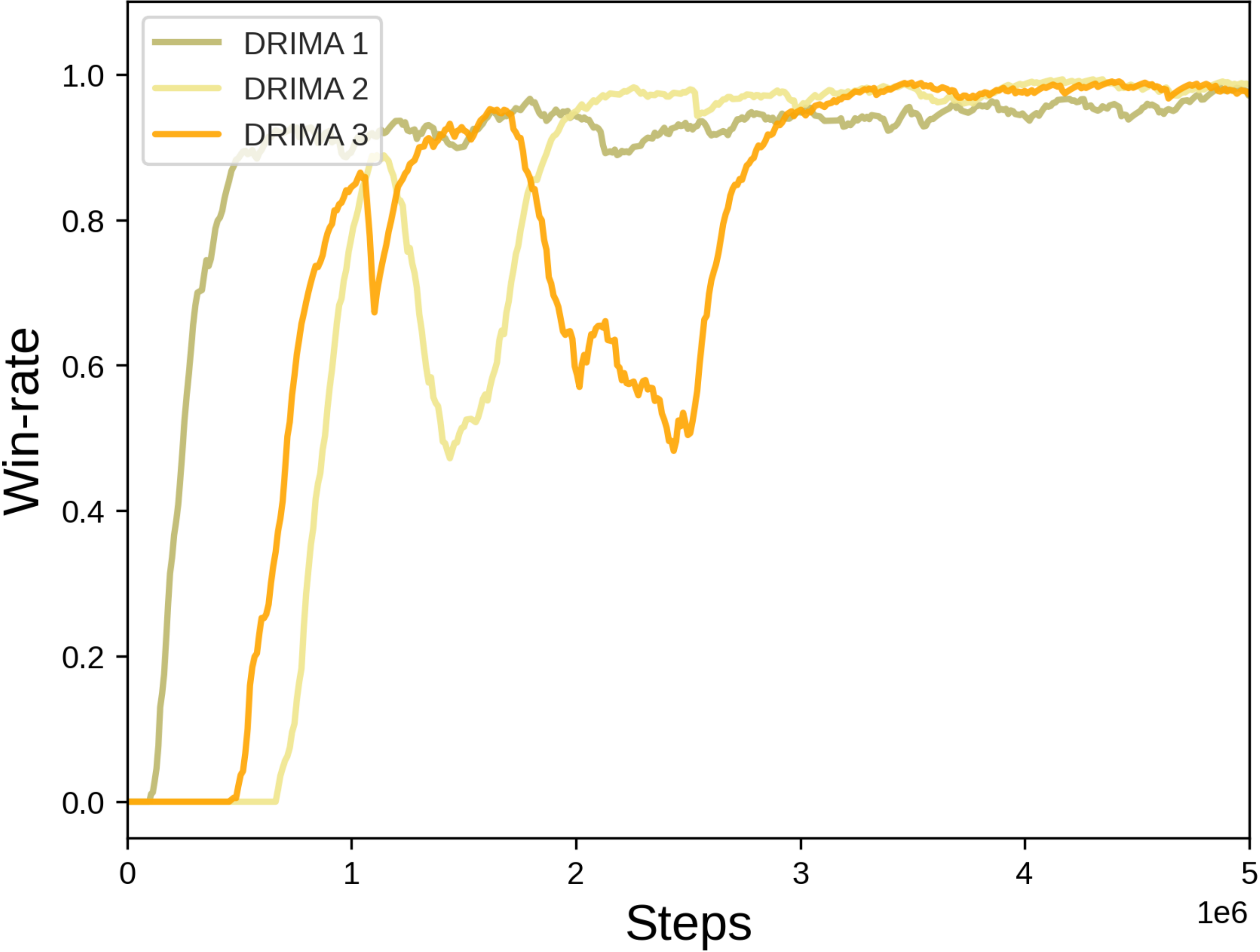}
            \caption{Offense near}
            \label{fig:app_drima_episode_off_near}
        \end{subfigure}%
        \begin{subfigure}{0.26\columnwidth}
            \includegraphics[width=\columnwidth]{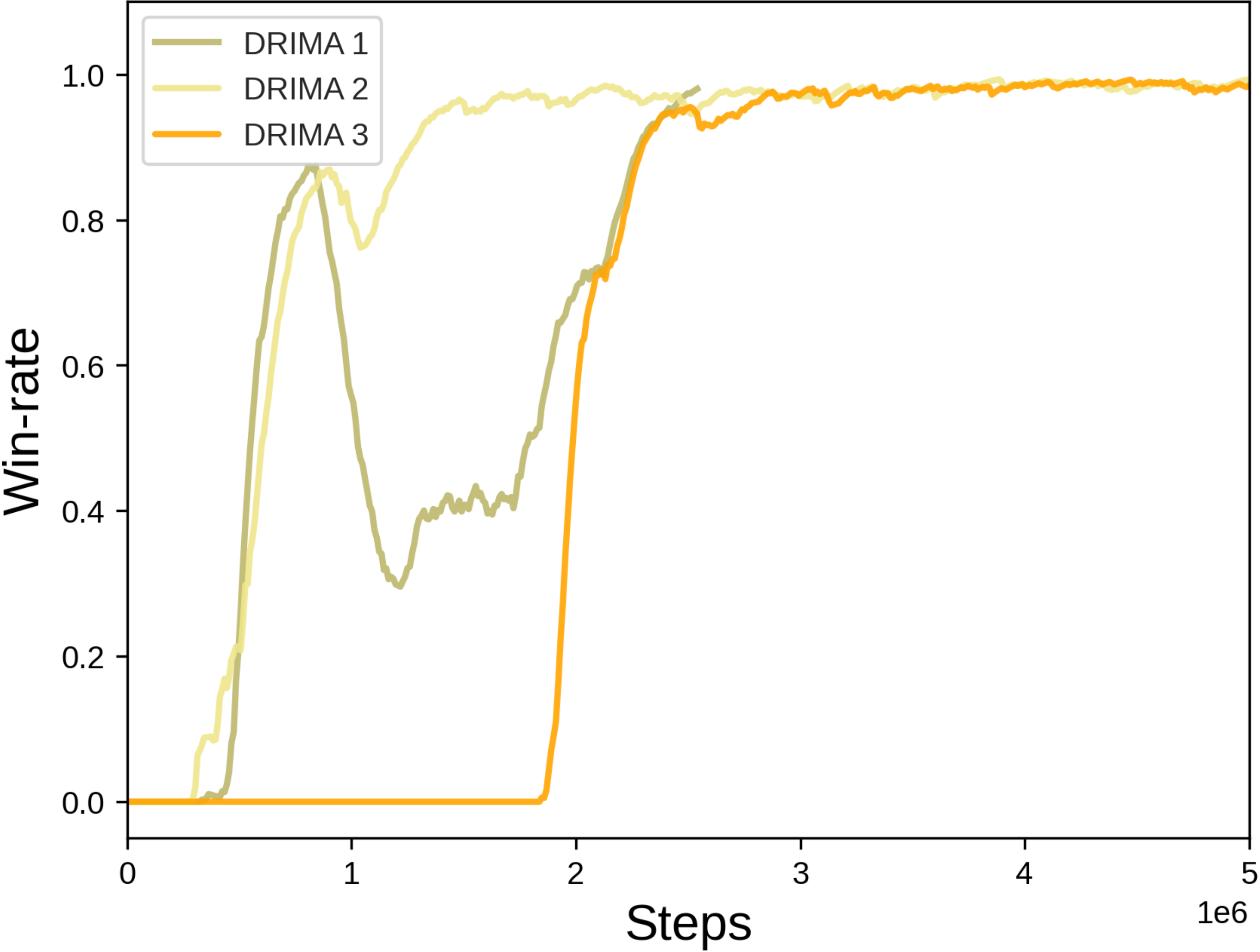}
            \caption{Offense distant}
            \label{fig:app_drima_episode_off_dist}
        \end{subfigure}%
        \begin{subfigure}{0.26\columnwidth}
            \includegraphics[width=\columnwidth]{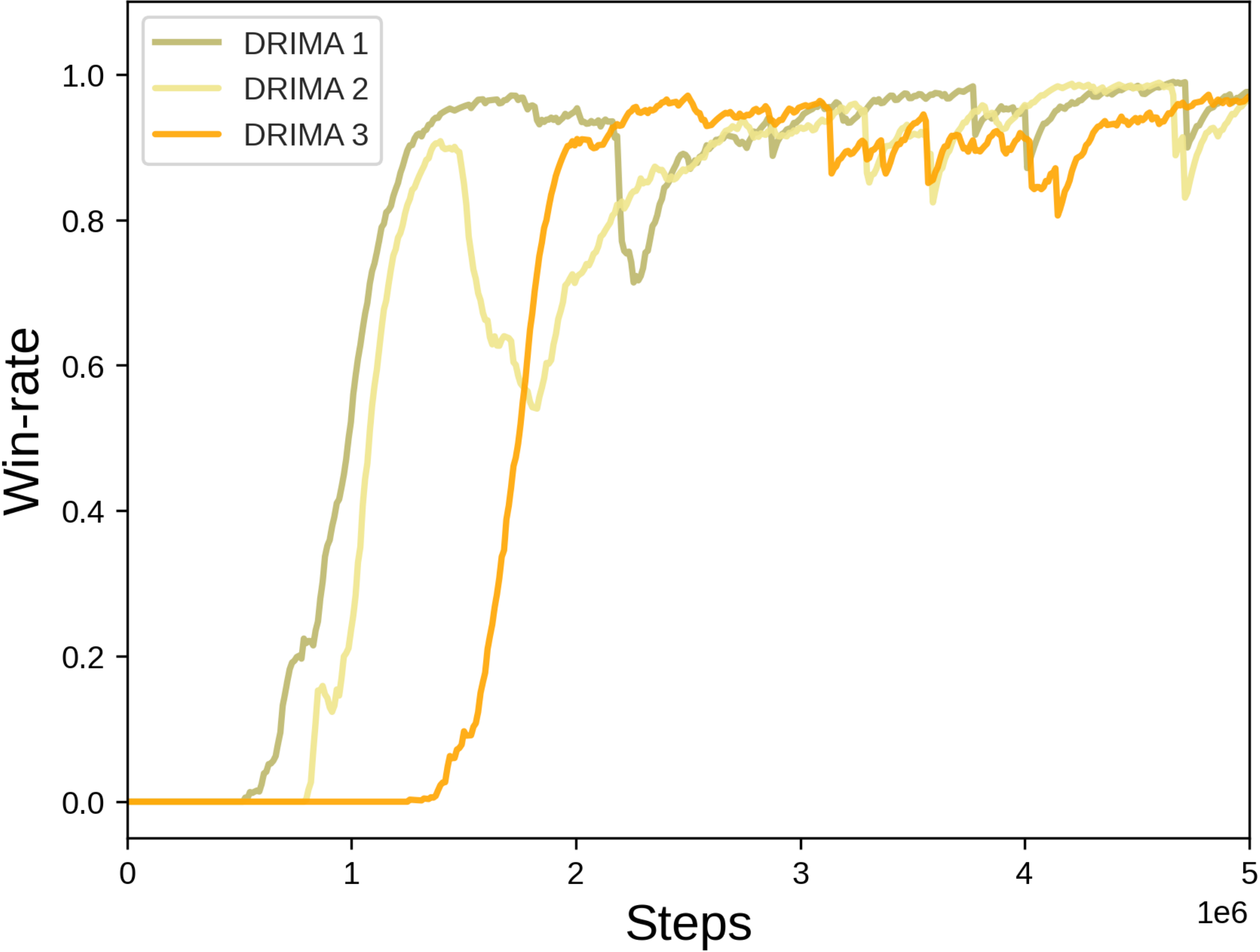}
            \caption{Offense complicated}
            \label{fig:app_drima_episode_off_com}
        \end{subfigure}%
        
        \begin{subfigure}{0.27\columnwidth}
            \includegraphics[width=\columnwidth]{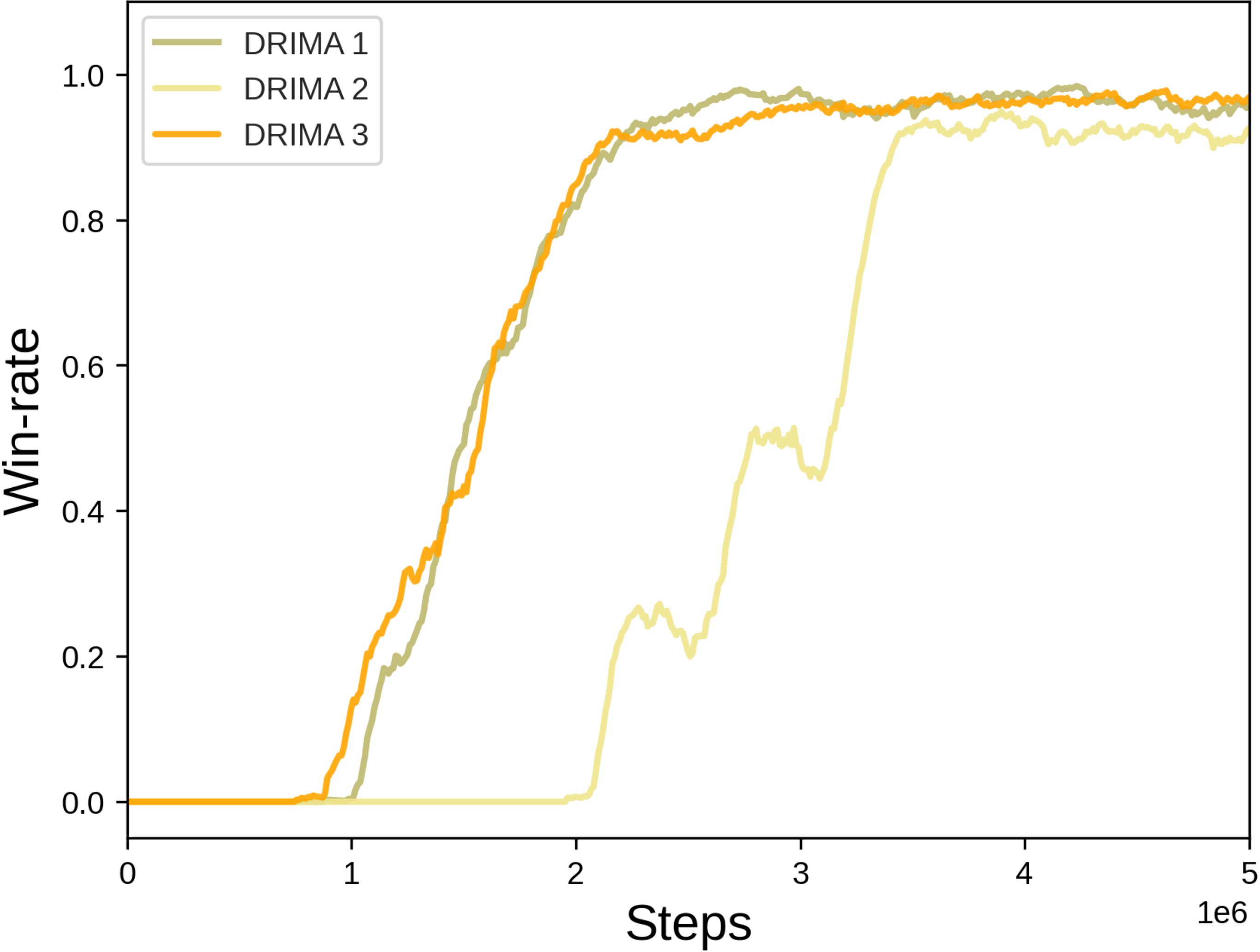}
            \caption{Offense hard}
            \label{fig:app_drima_episode_off_hard}
        \end{subfigure}%
        \begin{subfigure}{0.27\columnwidth}
            \includegraphics[width=\columnwidth]{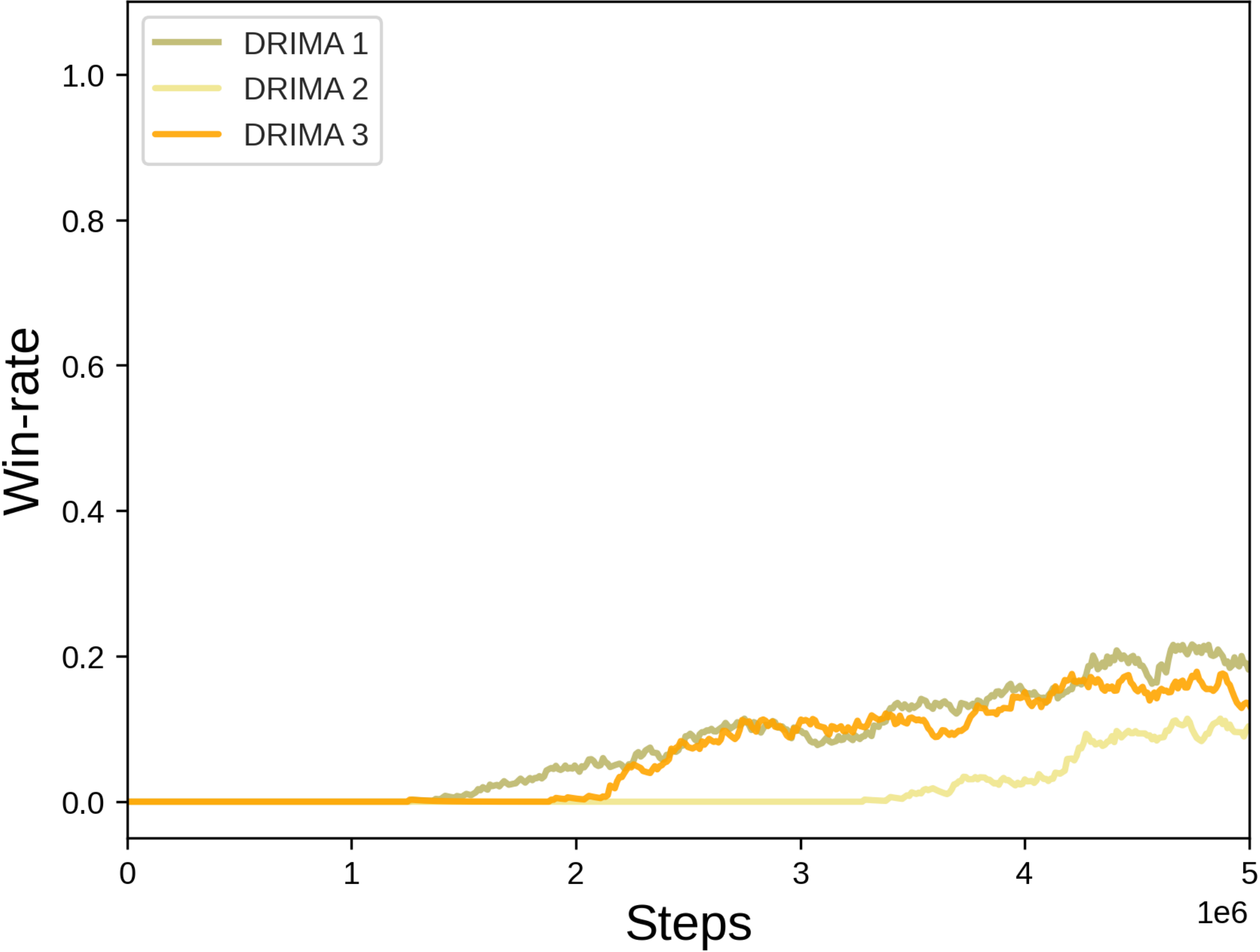}
            \caption{Offense superhard}
            \label{fig:app_drima_episode_off_super}
        \end{subfigure}%
    \caption{DRIMA trained on the sequential episodic buffer}
    \label{fig:app_drima_episode}
}
\end{figure}

\begin{figure}[!ht]{
    \centering
        \begin{subfigure}{0.26\columnwidth}
            \includegraphics[width=\columnwidth]{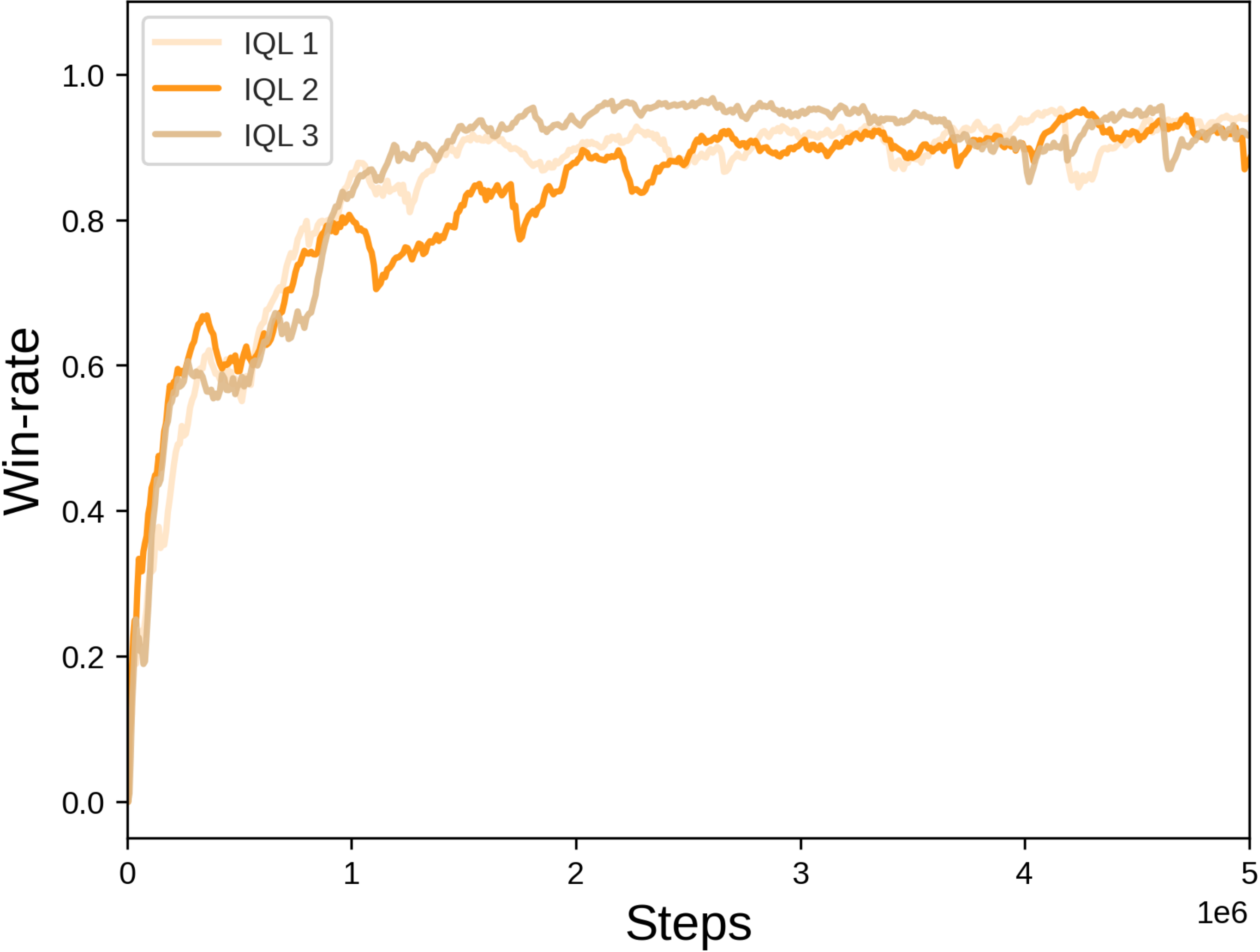}
            \caption{Defense infantry}
            \label{fig:app_iql_episode_def_inf}
        \end{subfigure}%
        \begin{subfigure}{0.26\columnwidth}
            \includegraphics[width=\columnwidth]{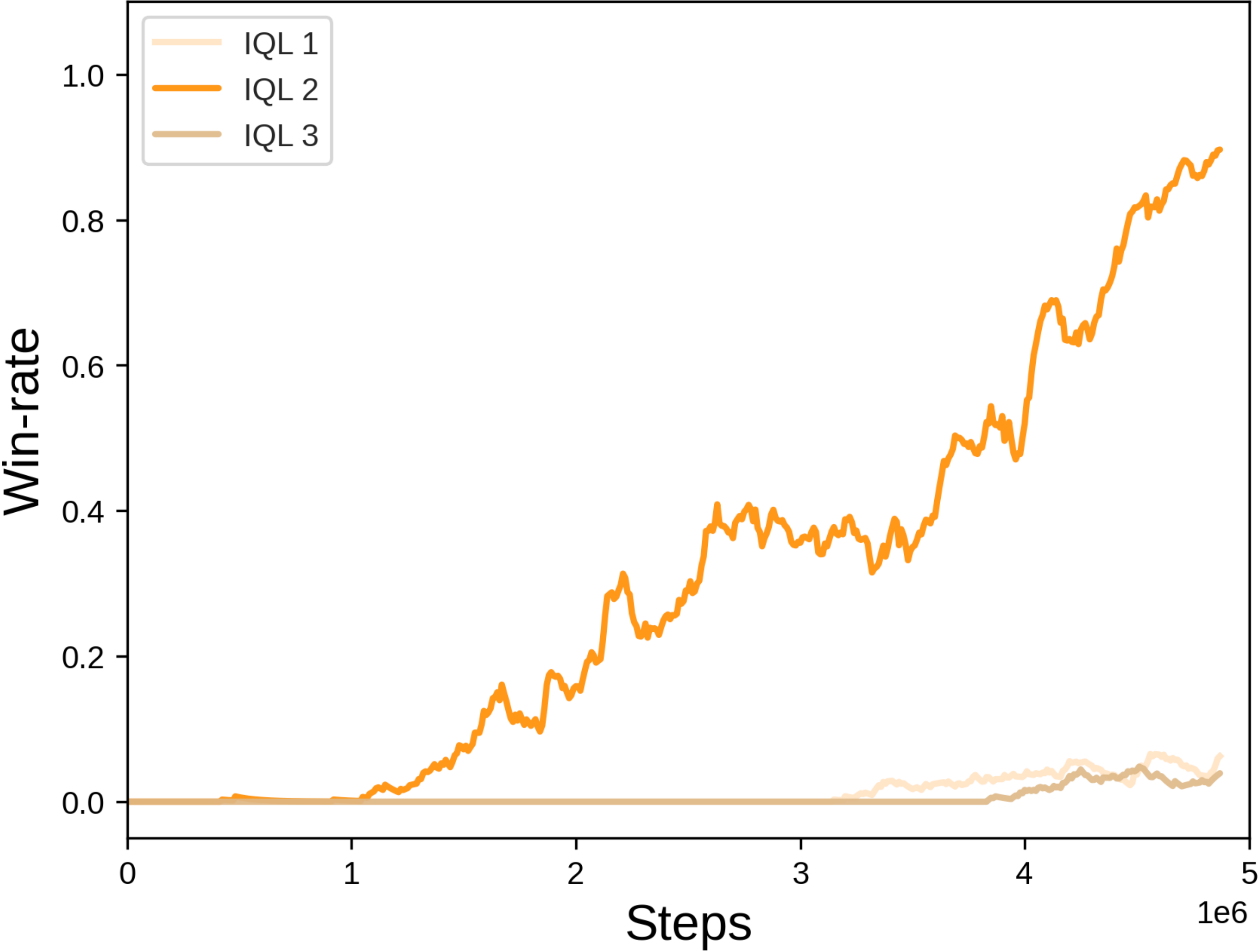}
            \caption{Defense armored}
            \label{fig:app_iql_episode_def_arm}
        \end{subfigure}%
        \begin{subfigure}{0.26\columnwidth}
            \includegraphics[width=\columnwidth]{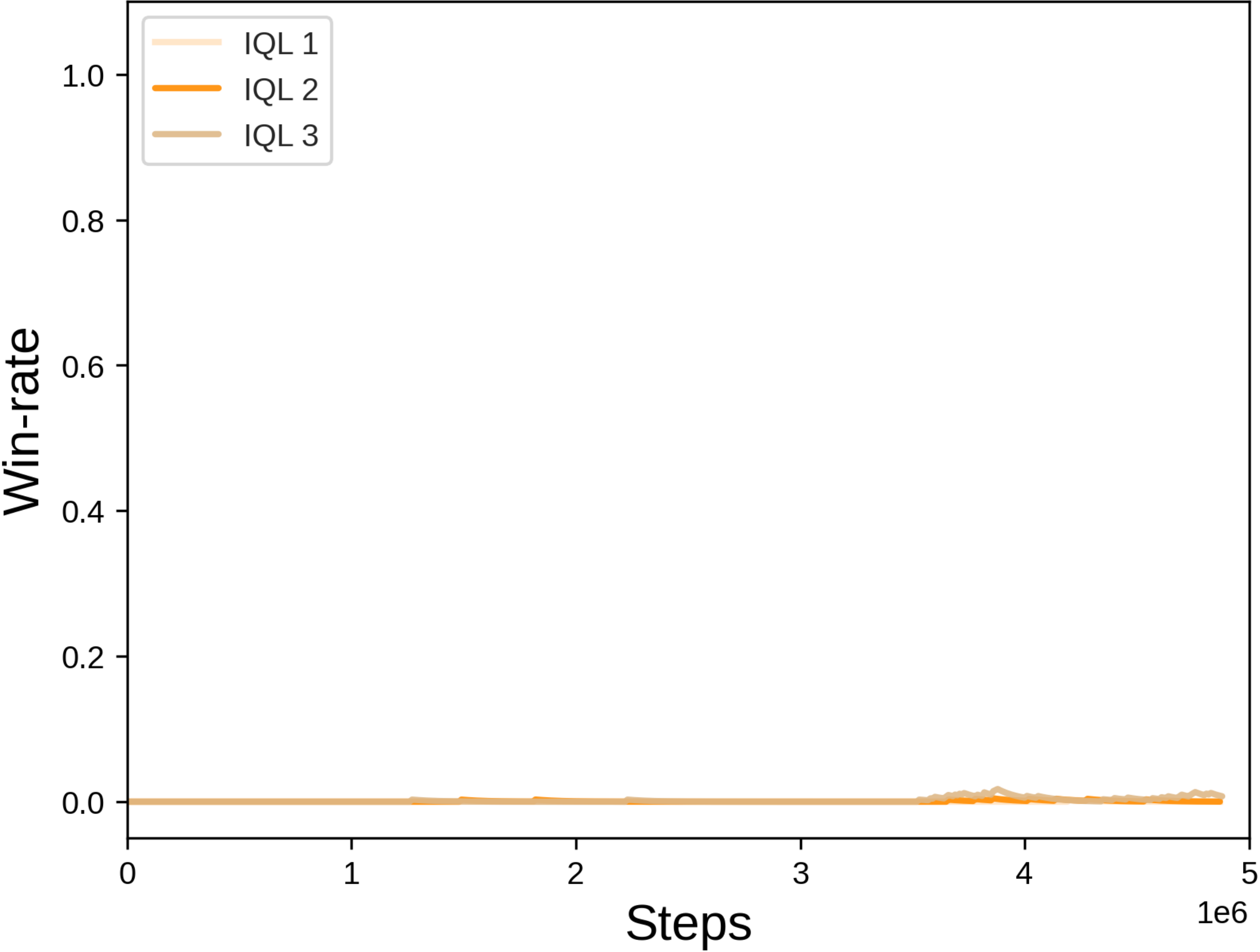}
            \caption{Defense outnumbered}
            \label{fig:app_iql_episode_def_out}
        \end{subfigure}%
    \caption{IQL trained on the sequential episodic buffer}
    \label{fig:app_iql_episode}
}
\end{figure}

\begin{figure}[!ht]{
    \centering
        \begin{subfigure}{0.26\columnwidth}
            \includegraphics[width=\columnwidth]{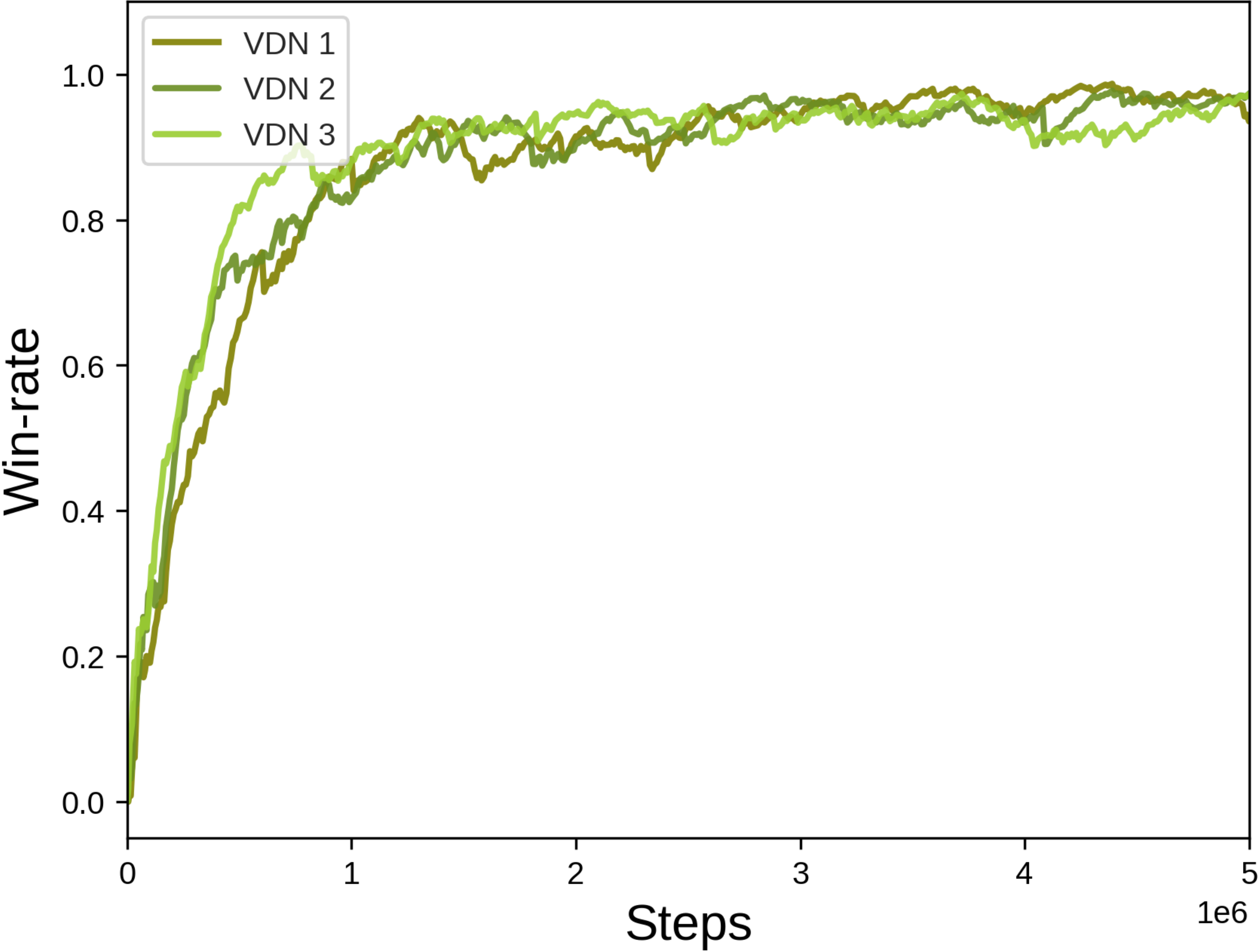}
            \caption{Defense infantry}
            \label{fig:app_vdn_episode_def_inf}
        \end{subfigure}%
        \begin{subfigure}{0.26\columnwidth}
            \includegraphics[width=\columnwidth]{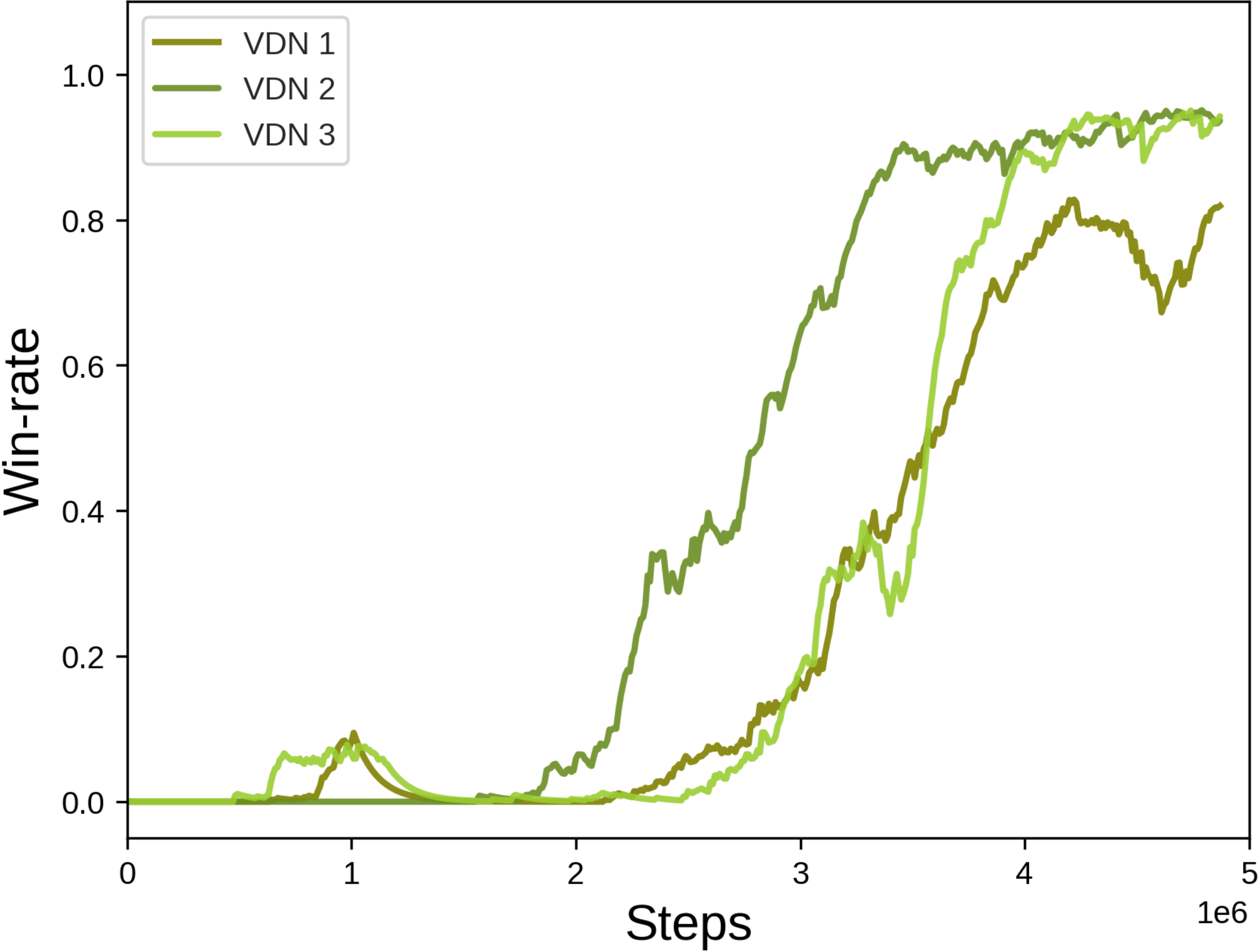}
            \caption{Defense armored}
            \label{fig:app_vdn_episode_def_arm}
        \end{subfigure}%
        \begin{subfigure}{0.26\columnwidth}
            \includegraphics[width=\columnwidth]{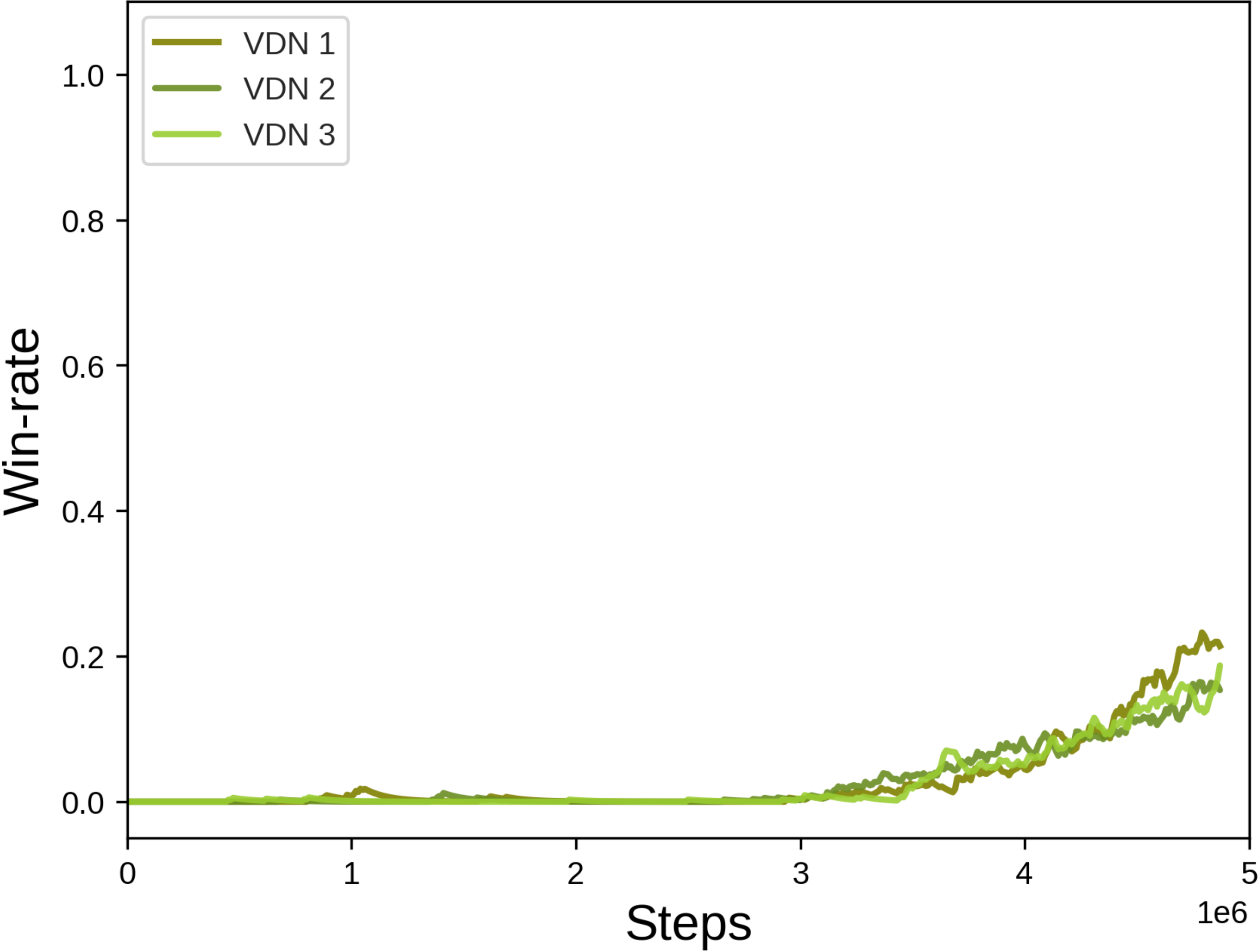}
            \caption{Defense outnumbered}
            \label{fig:app_vdn_episode_out}
        \end{subfigure}%
    \caption{VDN trained on the sequential episodic buffer}
    \label{fig:app_vdn_episode}
}
\end{figure}

\begin{figure}[!ht]{
    \centering
        \begin{subfigure}{0.26\columnwidth}
            \includegraphics[width=\columnwidth]{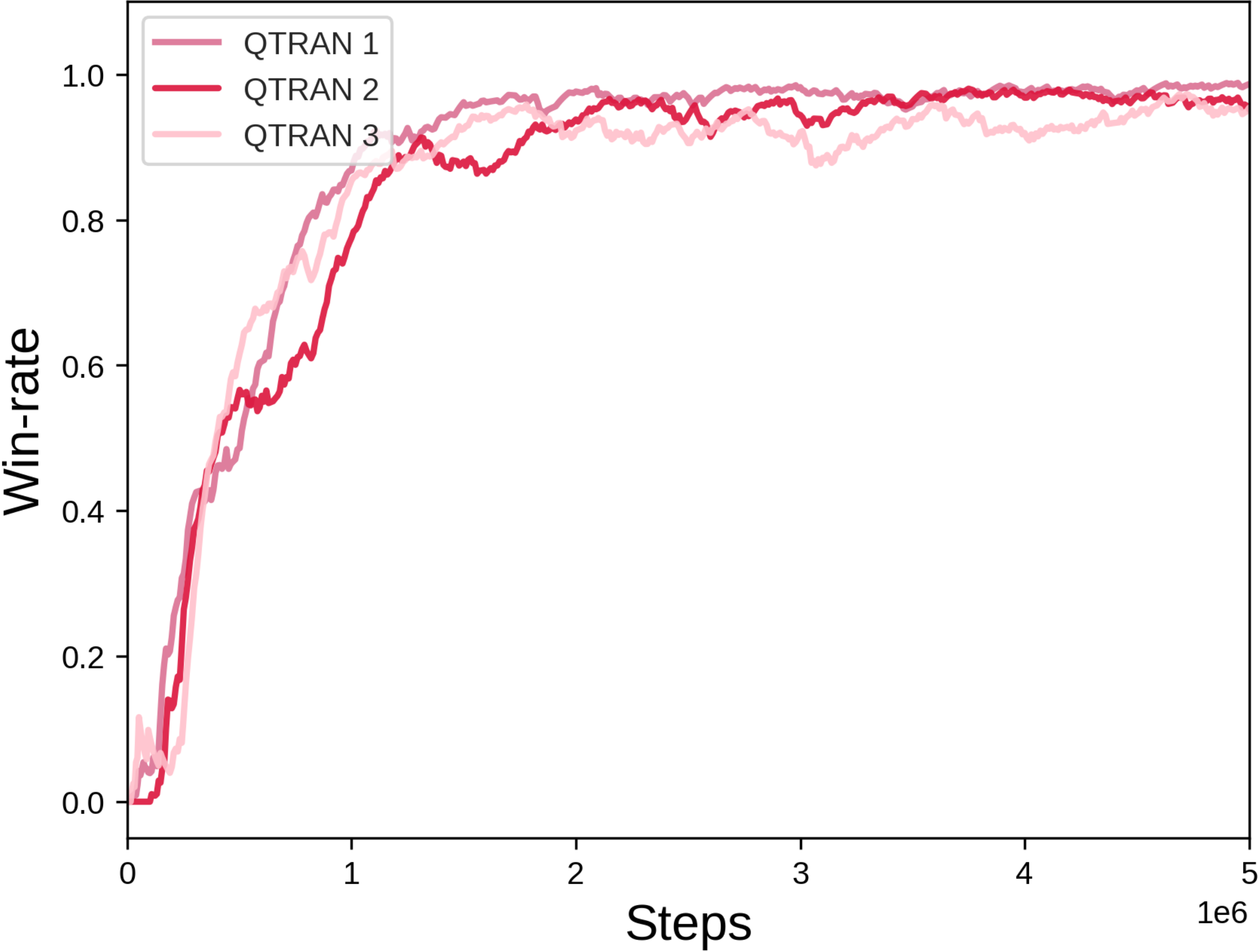}
            \caption{Defense infantry}
            \label{fig:app_qtran_episode_def_inf}
        \end{subfigure}%
        \begin{subfigure}{0.26\columnwidth}
            \includegraphics[width=\columnwidth]{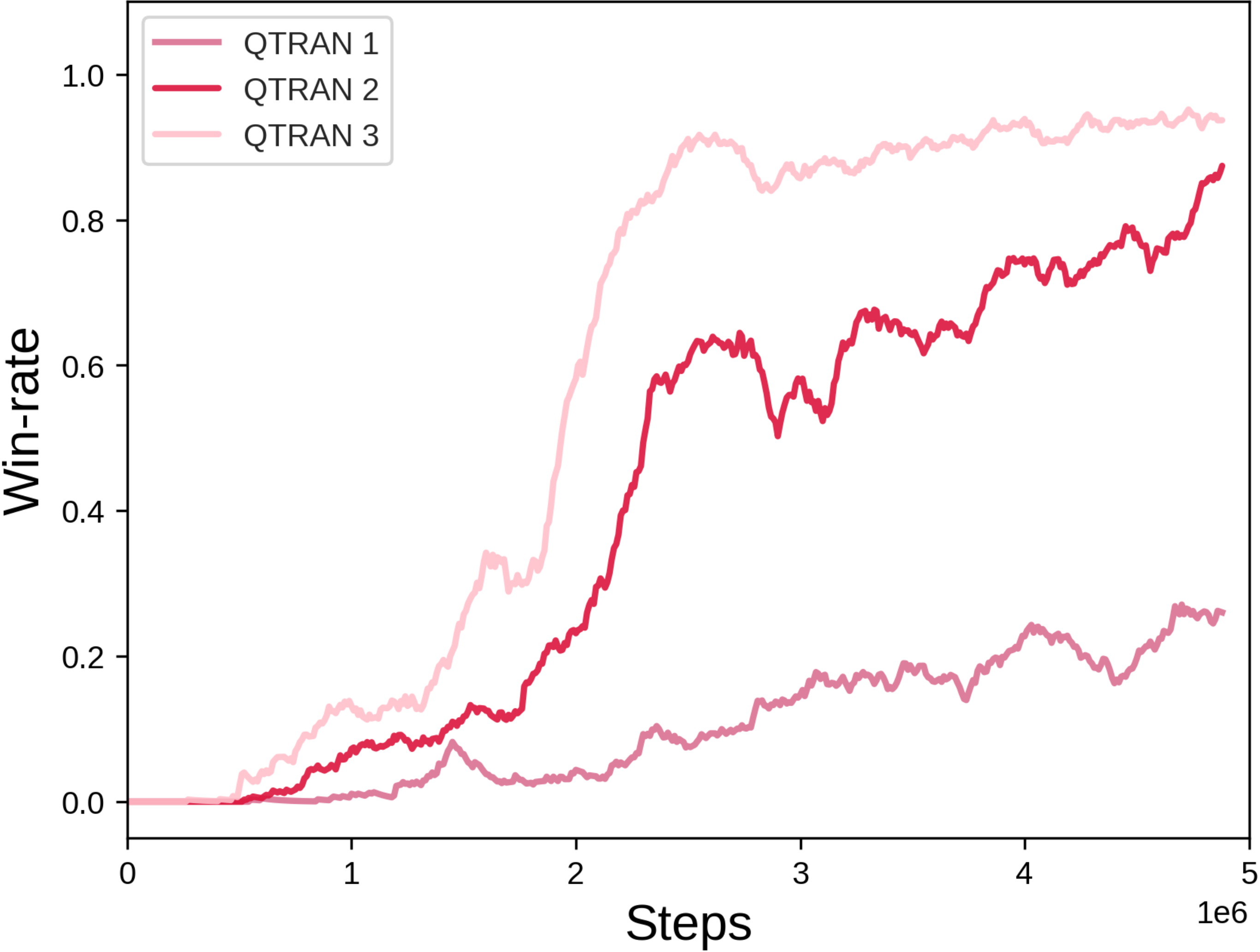}
            \caption{Defense armored}
            \label{fig:app_qtran_episode_def_arm}
        \end{subfigure}%
        \begin{subfigure}{0.26\columnwidth}
            \includegraphics[width=\columnwidth]{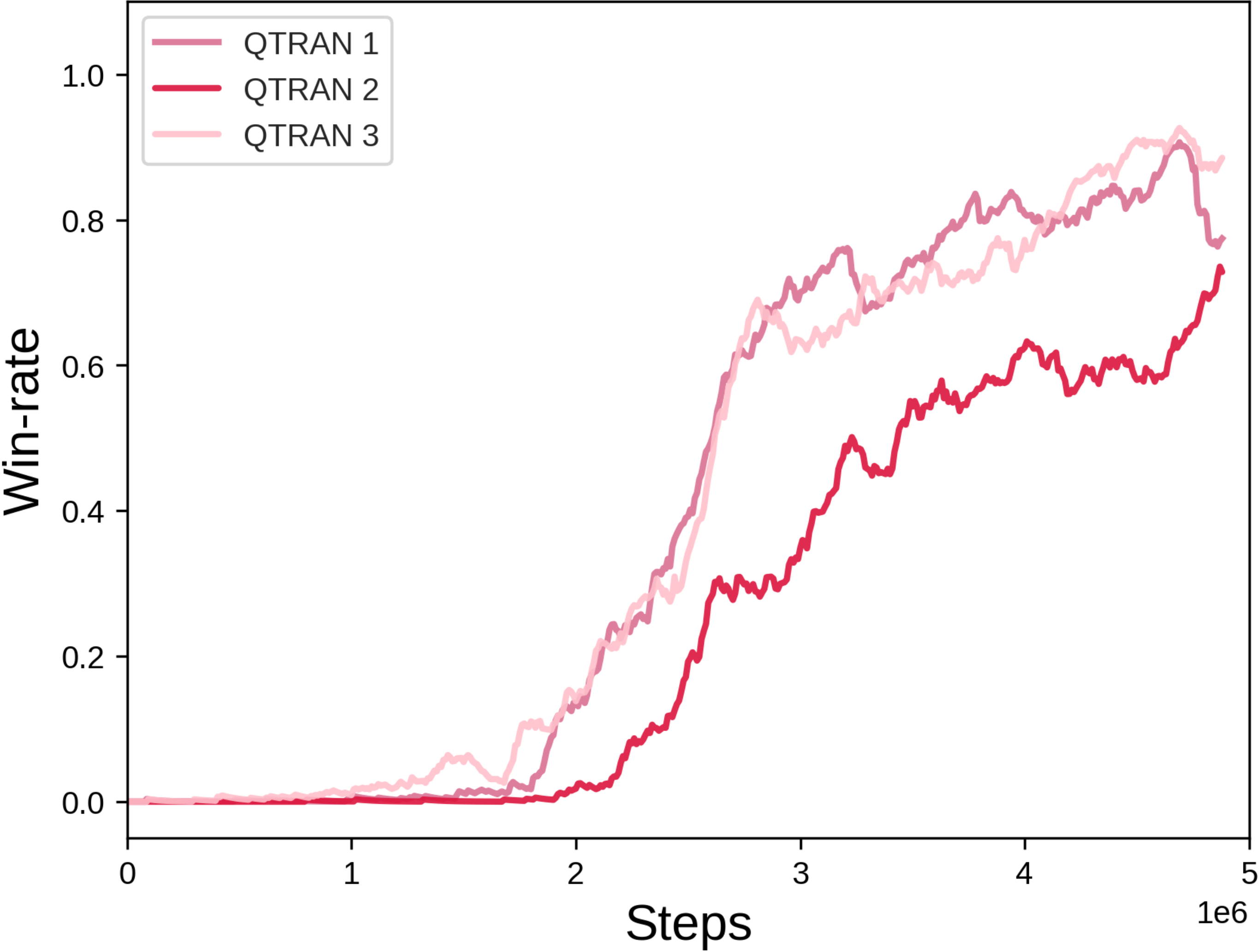}
            \caption{Defense outnumbered}
            \label{fig:app_qtran_episode_def_out}
        \end{subfigure}%
    \caption{QTRAN trained on the sequential episodic buffer}
    \label{fig:app_qtran_episode}
}
\end{figure}

\begin{figure}[!ht]{
    \centering
        \begin{subfigure}{0.26\columnwidth}
            \includegraphics[width=\columnwidth]{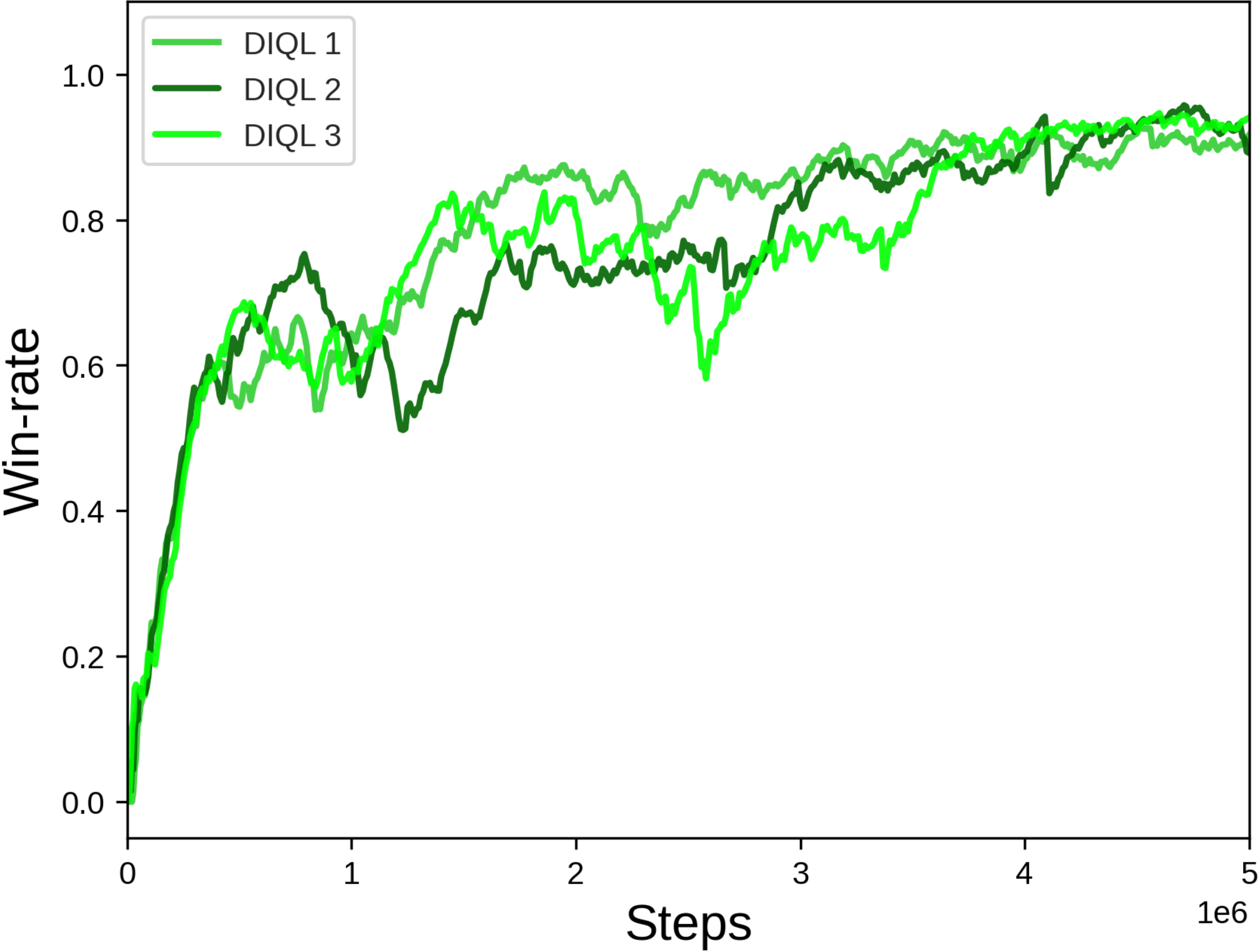}
            \caption{Defense infantry}
            \label{fig:app_diql_episode_def_inf}
        \end{subfigure}%
        \begin{subfigure}{0.26\columnwidth}
            \includegraphics[width=\columnwidth]{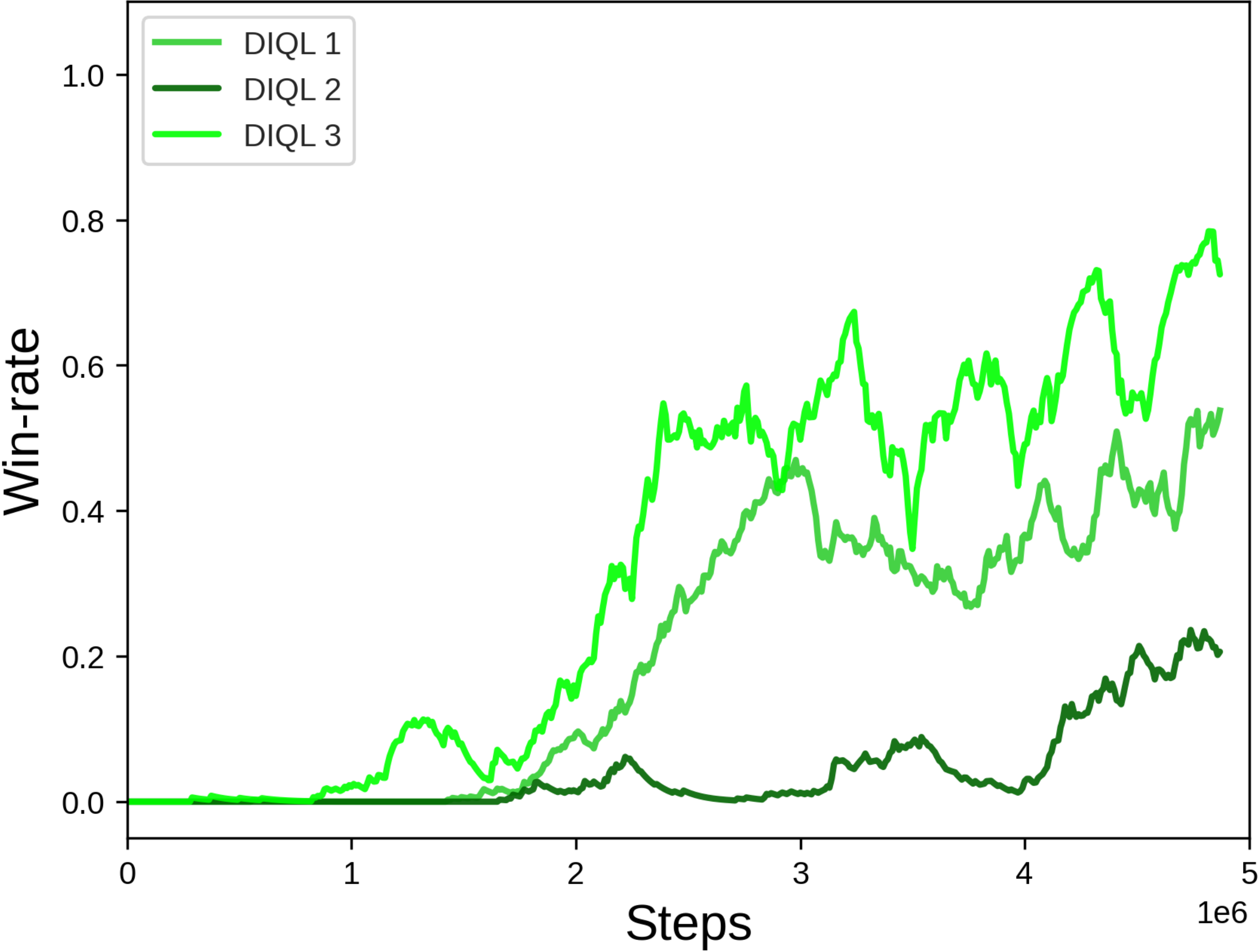}
            \caption{Defense armored}
            \label{fig:app_diql_episode_def_arm}
        \end{subfigure}%
        \begin{subfigure}{0.26\columnwidth}
            \includegraphics[width=\columnwidth]{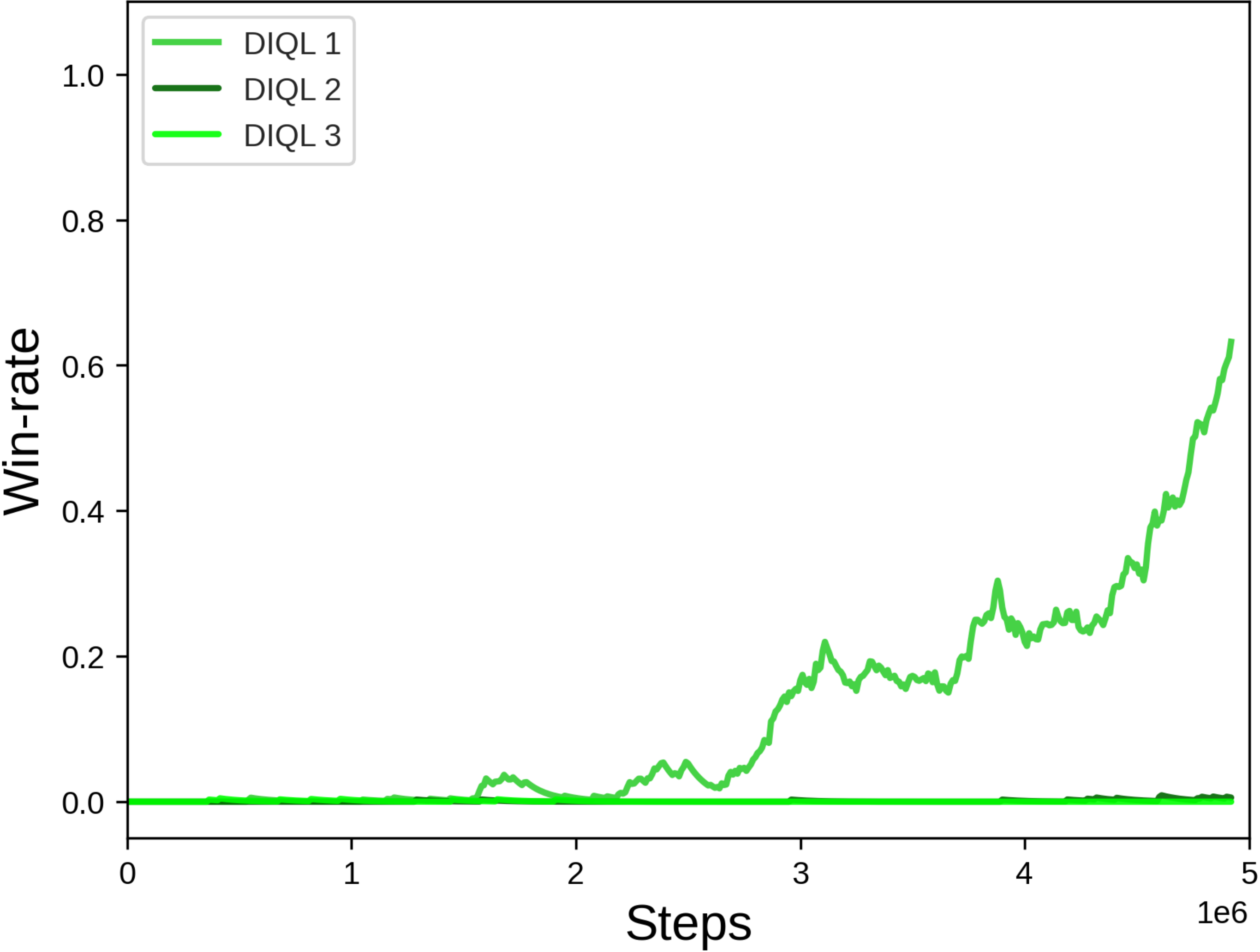}
            \caption{Defense outnumbered}
            \label{fig:app_diql_episode_def_out}
        \end{subfigure}%
    \caption{DIQL trained on the sequential episodic buffer}
    \label{fig:app_diql_episode}
}
\end{figure}

\begin{figure}[!ht]{
    \centering
        \begin{subfigure}{0.26\columnwidth}
            \includegraphics[width=\columnwidth]{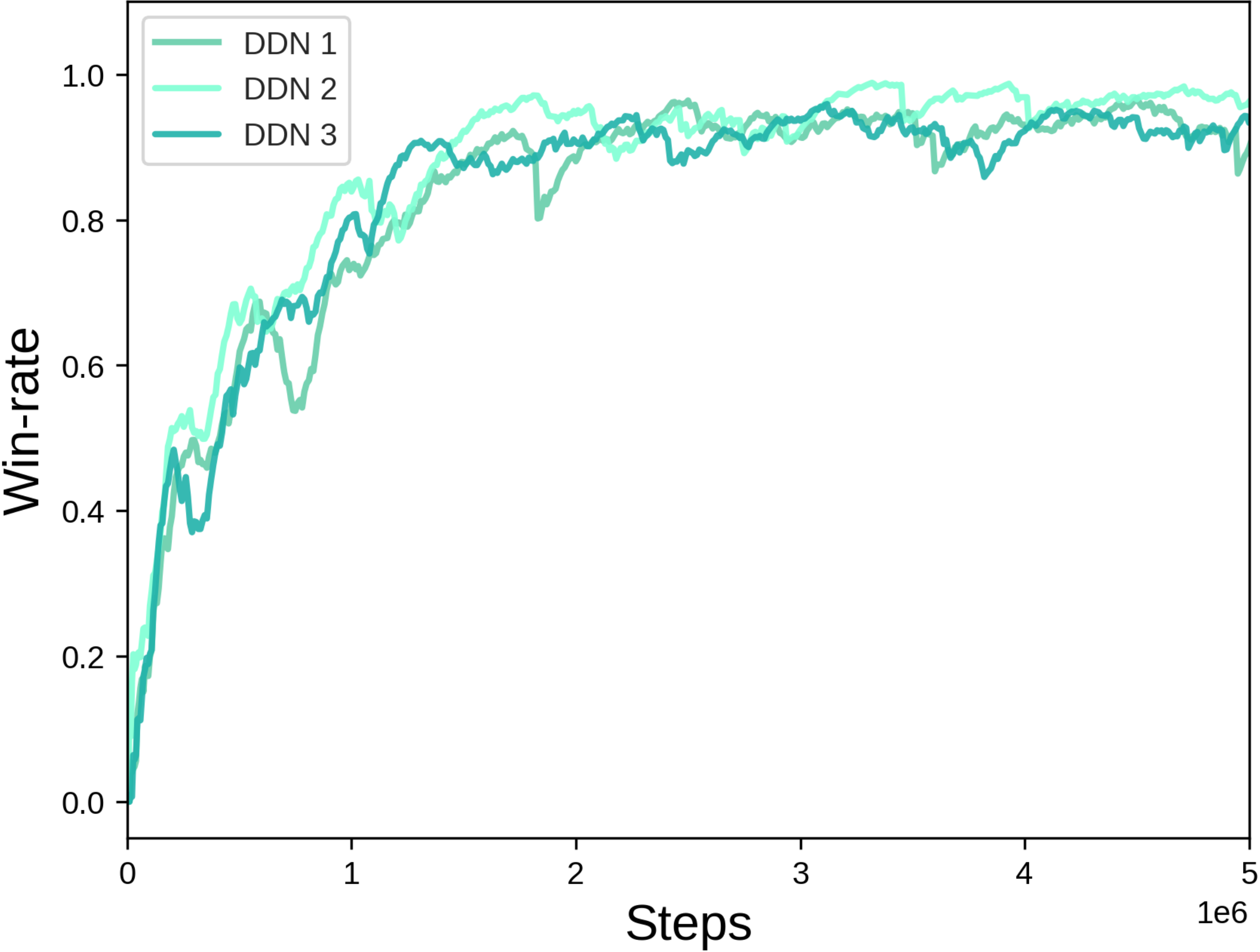}
            \caption{Defense infantry}
            \label{fig:app_ddn_episode_def_inf}
        \end{subfigure}%
        \begin{subfigure}{0.26\columnwidth}
            \includegraphics[width=\columnwidth]{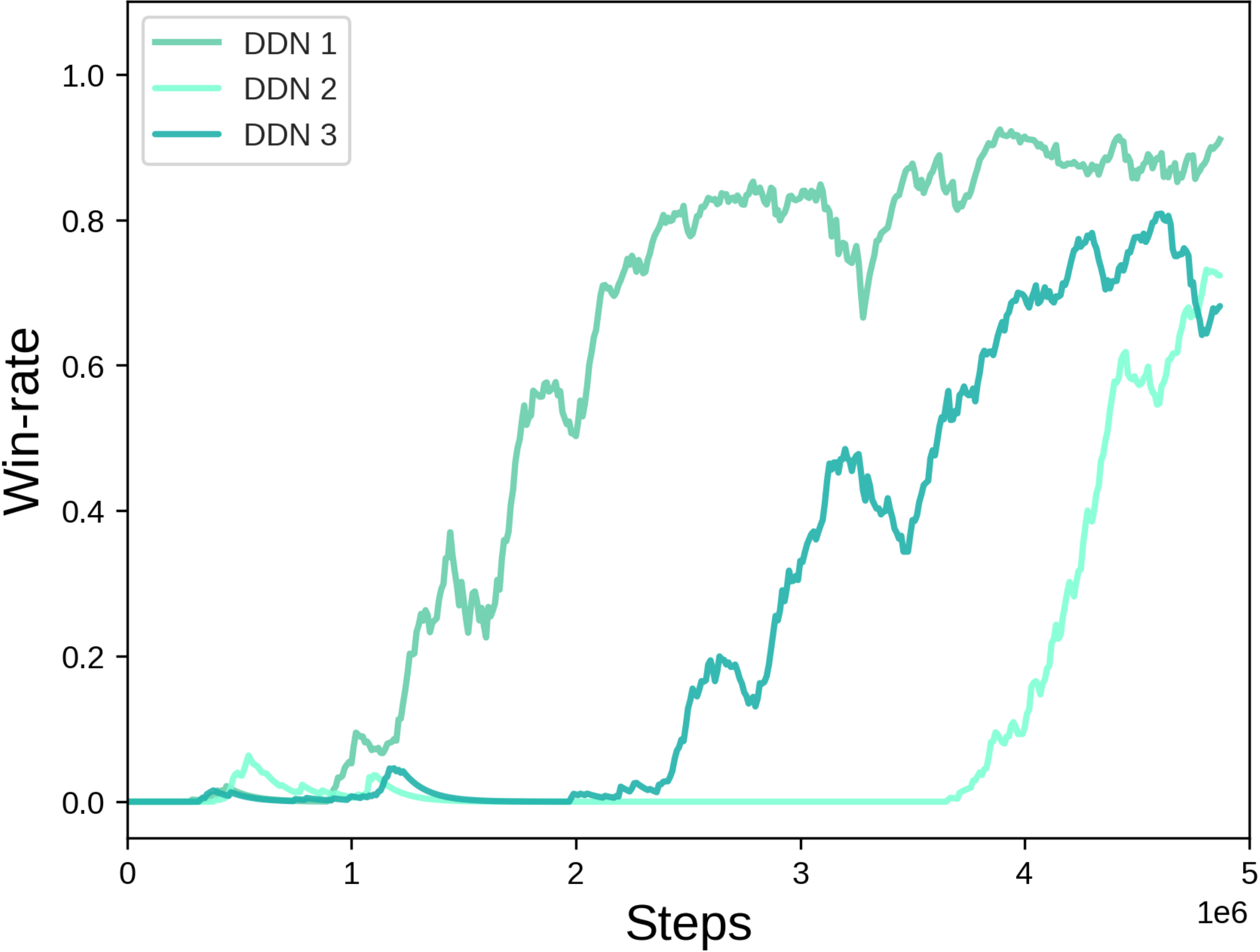}
            \caption{Defense armored}
            \label{fig:app_ddn_episode_def_arm}
        \end{subfigure}%
        \begin{subfigure}{0.26\columnwidth}
            \includegraphics[width=\columnwidth]{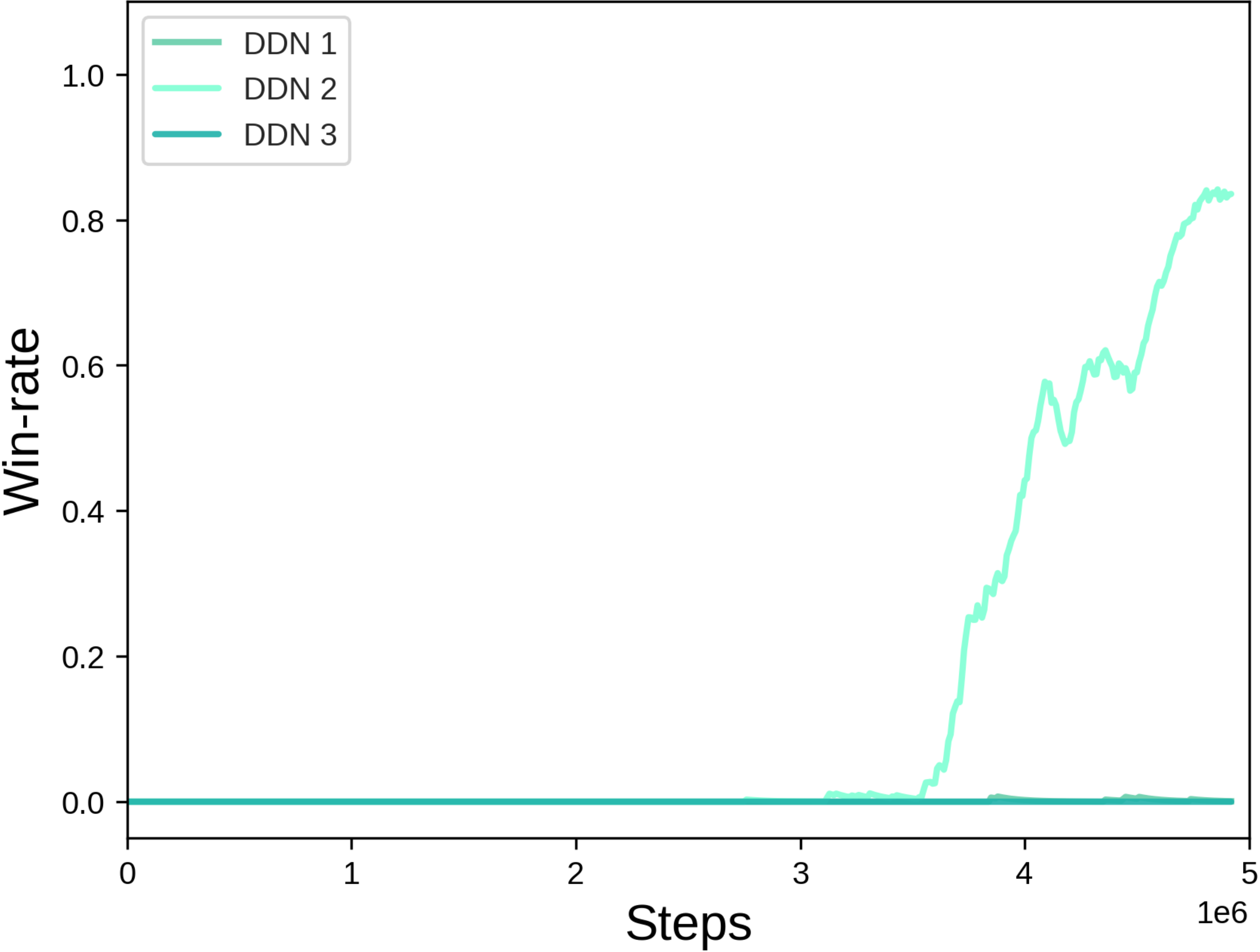}
            \caption{Defense outnumbered}
            \label{fig:app_ddn_episode_def_out}
        \end{subfigure}%
    \caption{DDN trained on the sequential episodic buffer}
    \label{fig:app_ddn_episode}
}
\end{figure}

\begin{figure}[!ht]{
    \centering
        \begin{subfigure}{0.26\columnwidth}
            \includegraphics[width=\columnwidth]{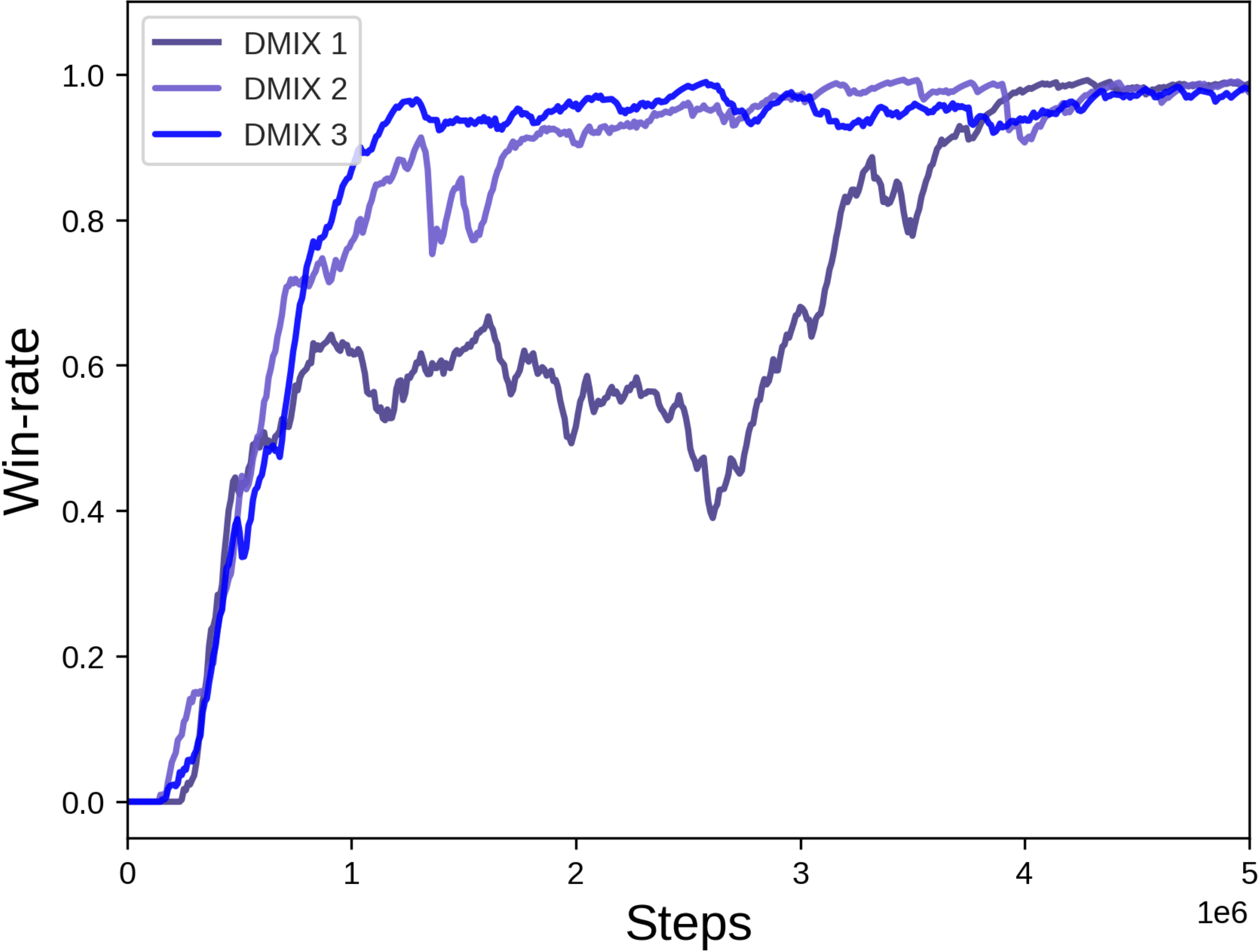}
            \caption{Defense infantry}
            \label{fig:app_dmix_episode_def_inf}
        \end{subfigure}%
        \begin{subfigure}{0.26\columnwidth}
            \includegraphics[width=\columnwidth]{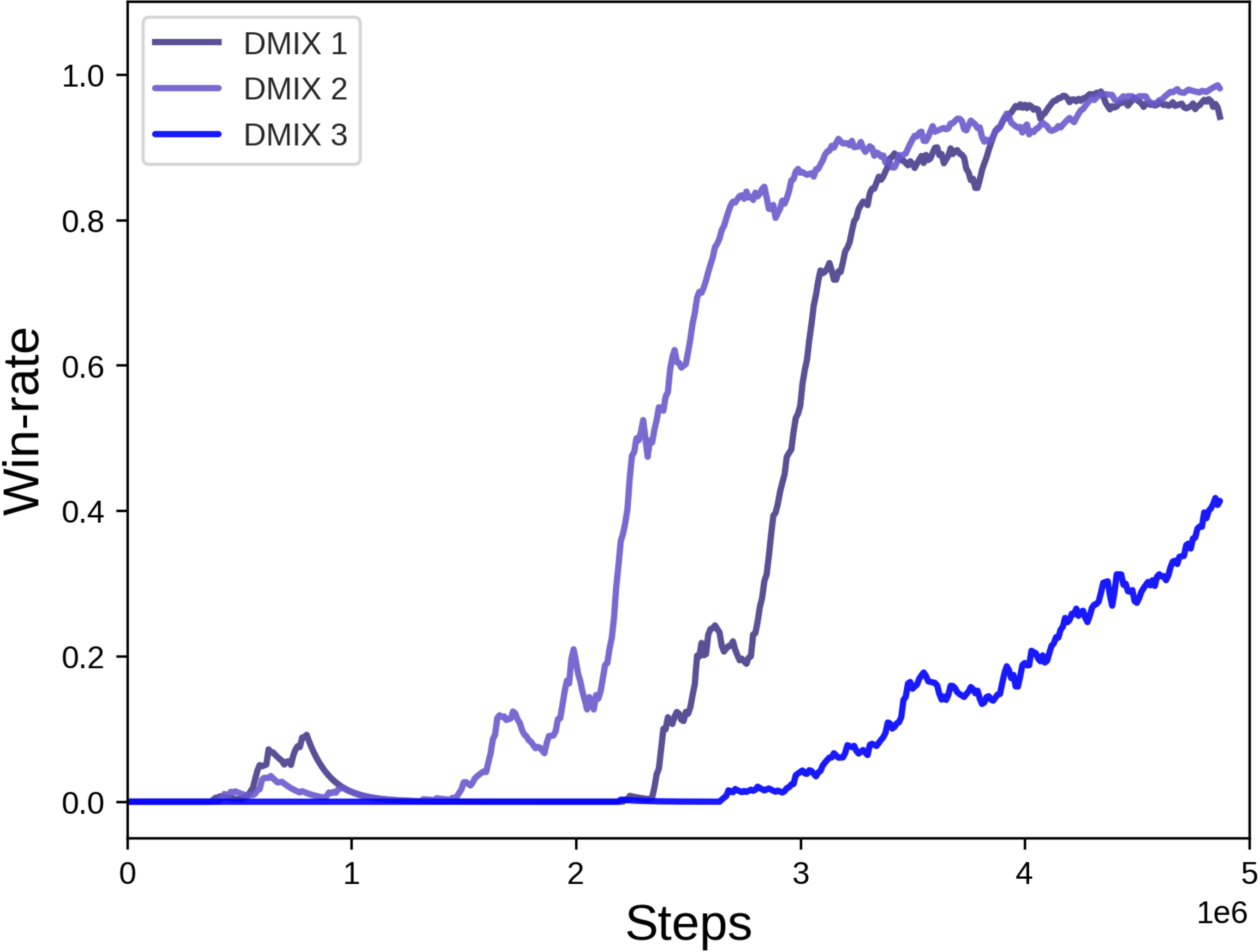}
            \caption{Defense armored}
            \label{fig:app_dmix_episode_def_arm}
        \end{subfigure}%
        \begin{subfigure}{0.26\columnwidth}
            \includegraphics[width=\columnwidth]{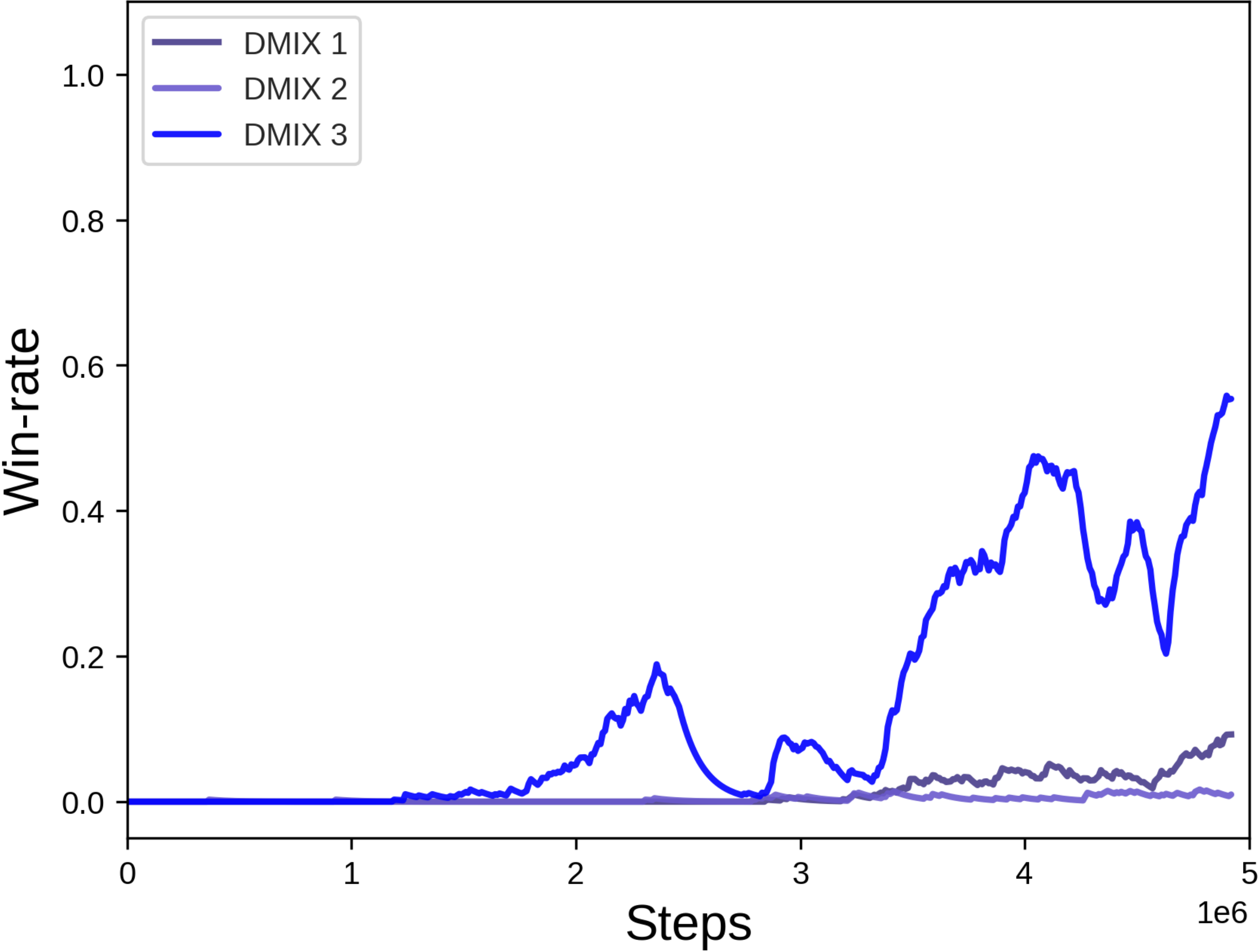}
            \caption{Defense outnumbered}
            \label{fig:app_dmix_episode_def_out}
        \end{subfigure}%
    \caption{DMIX trained on the sequential episodic buffer}
    \label{fig:app_dmix_episode}
}
\end{figure}

\begin{figure}[!ht]{
    \centering
        \begin{subfigure}{0.26\columnwidth}
            \includegraphics[width=\columnwidth]{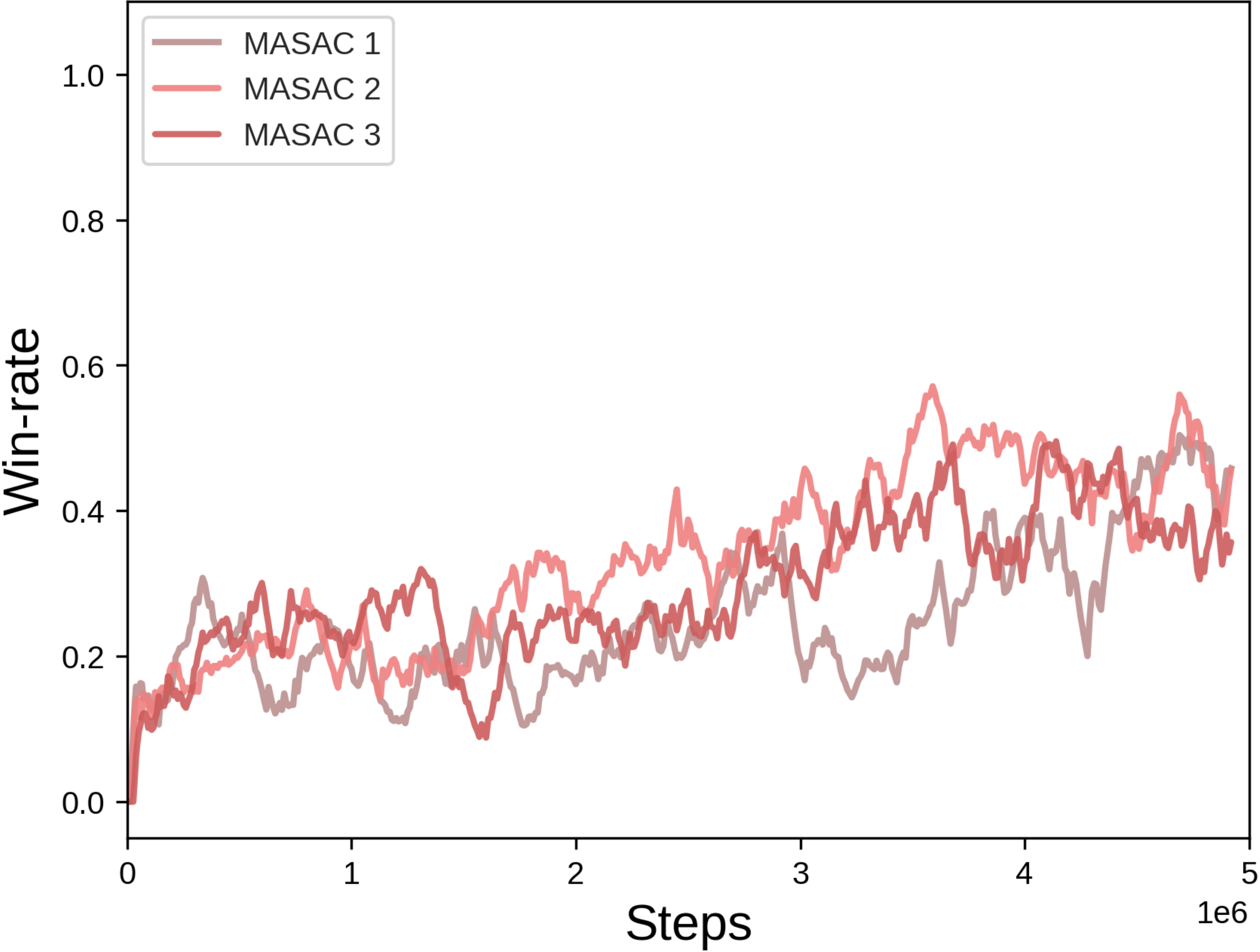}
            \caption{Defense infantry}
            \label{fig:app_masac_episode_def_inf}
        \end{subfigure}%
        \begin{subfigure}{0.26\columnwidth}
            \includegraphics[width=\columnwidth]{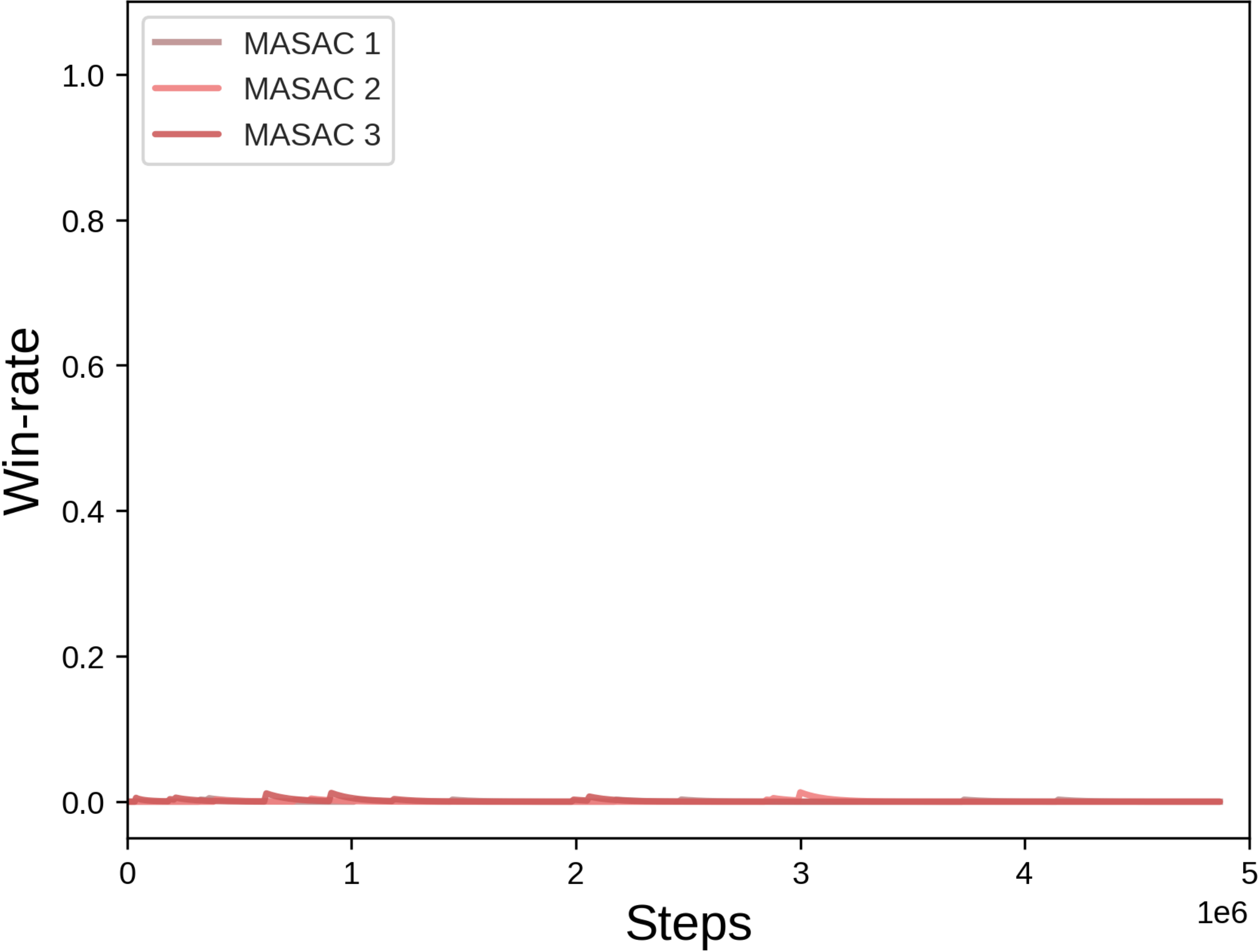}
            \caption{Defense armored}
            \label{fig:app_masac_episode_def_arm}
        \end{subfigure}%
        \begin{subfigure}{0.26\columnwidth}
            \includegraphics[width=\columnwidth]{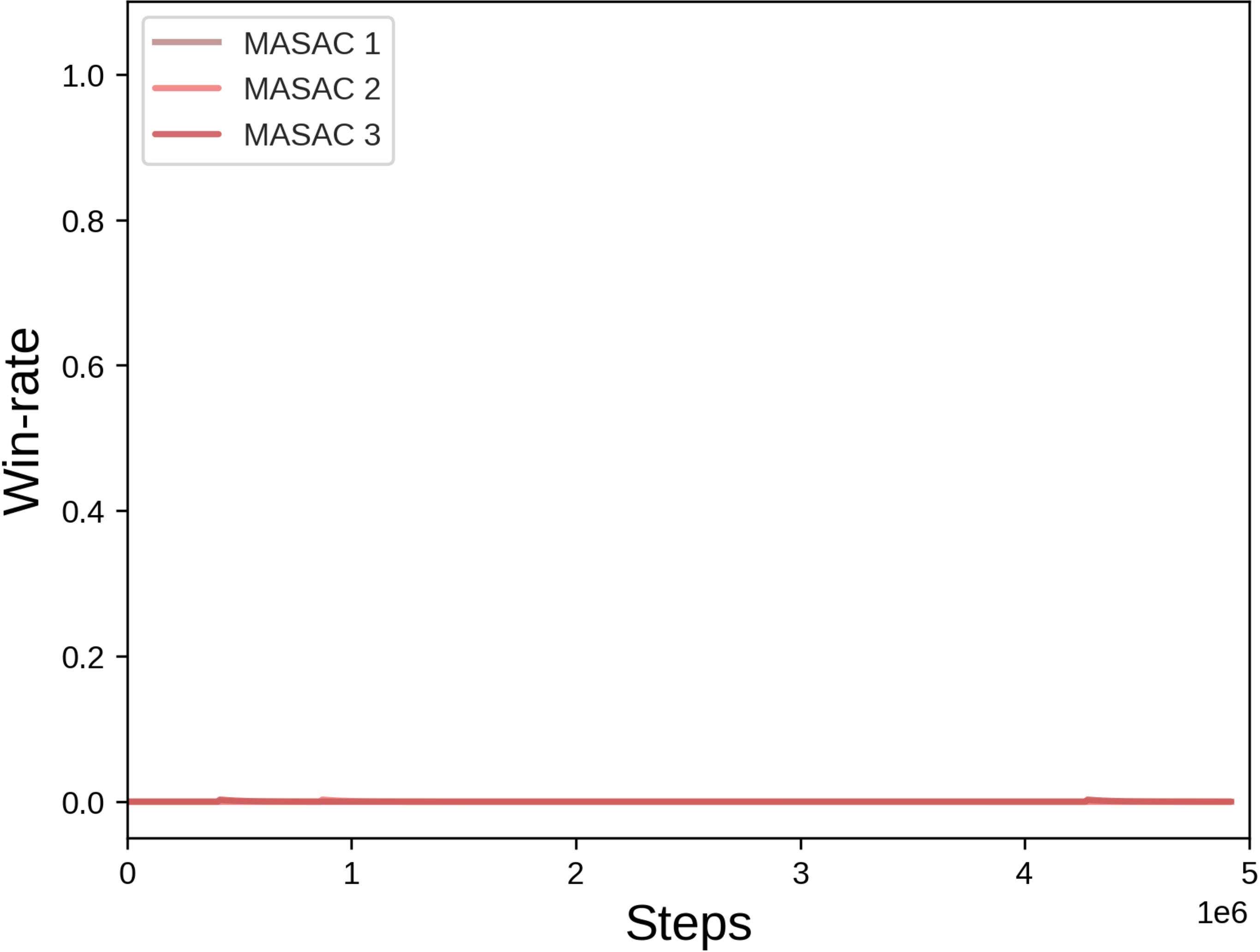}
            \caption{Defense outnumbered}
            \label{fig:app_masac_episode_def_out}
        \end{subfigure}%
    \caption{MASAC trained on the sequential episodic buffer}
    \label{fig:app_masac_episode}
}
\end{figure}

We pick numerous benchmark algorithms based on the selection of traditional value-based and policy-based algorithms as well as contemporary algorithms exhibiting state-of-the-art performance in the MARL domain. Consequently, we chose IQN, VDN, QMIX, and QTRAN for value-based algorithms, COMA, MASAC, and MADDPG for policy-based algorithms, and DMIX, DDN, DIQL, and DRIMA for more modern algorithms. In this part, we discuss the performance of each algorithm trained on the parallel episodic buffer and sequential episodic buffer. As demonstrated in \autoref{table:app_final_winrate_episode}, and \autoref{fig:app_coma_episode} - \ref{fig:app_masac_episode}, for 5 million training time steps, we provide experiment results in an sequential episodic buffer for most of the SMAC$^{+}$ scenarios using the all algorithms. In episodic buffer settings, the performance in defense scenarios tends to decline as the supply differential increases, while the performance in offensive scenarios is marginally getting worse due to the complexity of the path from the starting place of allied troops to enemy units. Whereas many of the scenarios are quite solved by the algorithms, all the algorithms struggle to solve offense hard and superhard scenarios. Only QMIX and DRIMA can solve the offense hard scenario, as shown in \autoref{fig:app_qmix_episode_off_hard} and \autoref{fig:app_drima_episode_off_hard} whereas none of the algorithms can solve offense superhard scenario. We report only defensive scenarios for some algorithms for the time limitation.

\begin{figure}[!ht]{
    \centering
        \begin{subfigure}{0.26\columnwidth}
            \includegraphics[width=\columnwidth]{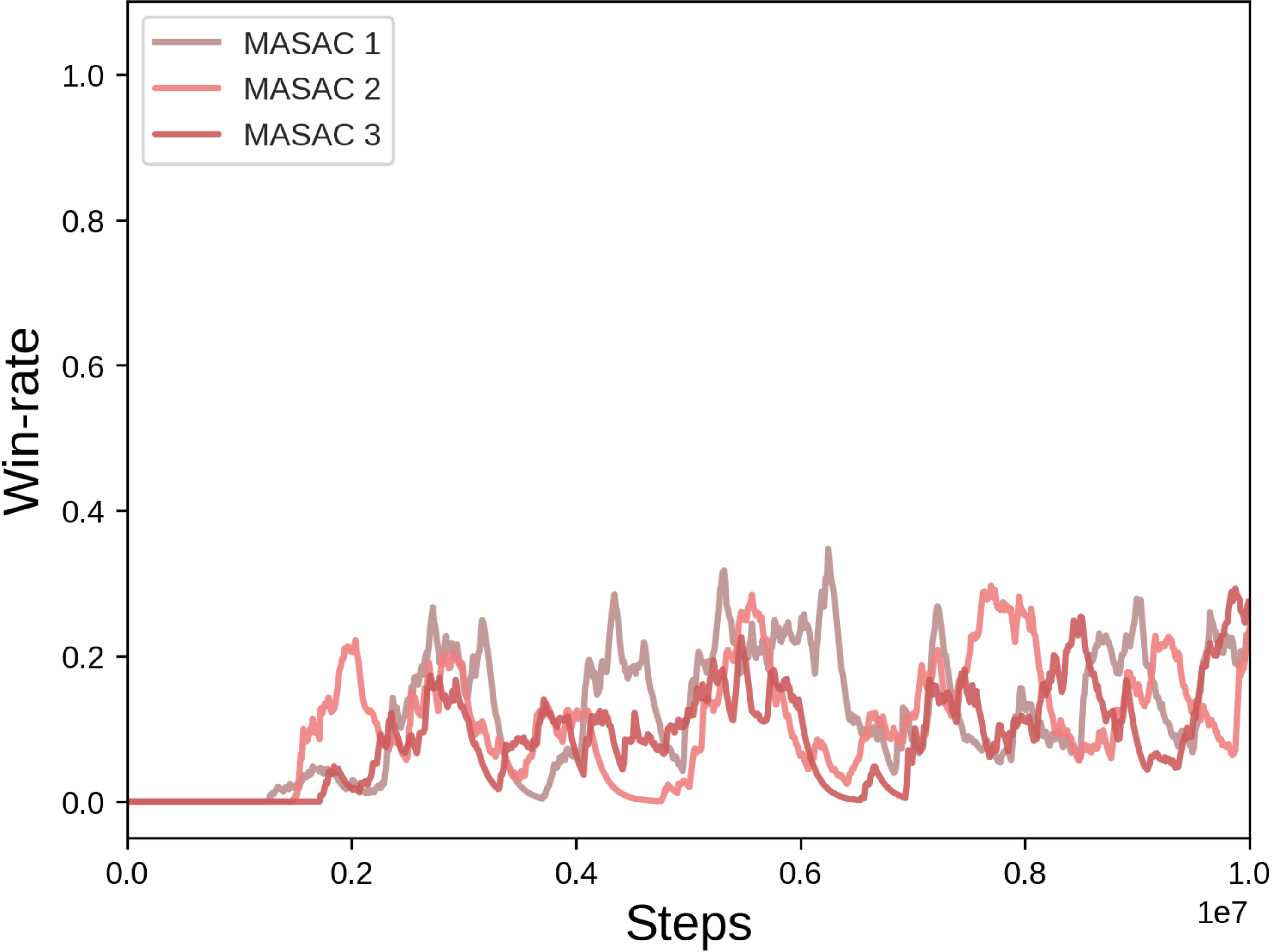}
            \caption{Defense infantry}
            \label{fig:app_masac_parallel_def_inf}
        \end{subfigure}%
        \begin{subfigure}{0.26\columnwidth}
            \includegraphics[width=\columnwidth]{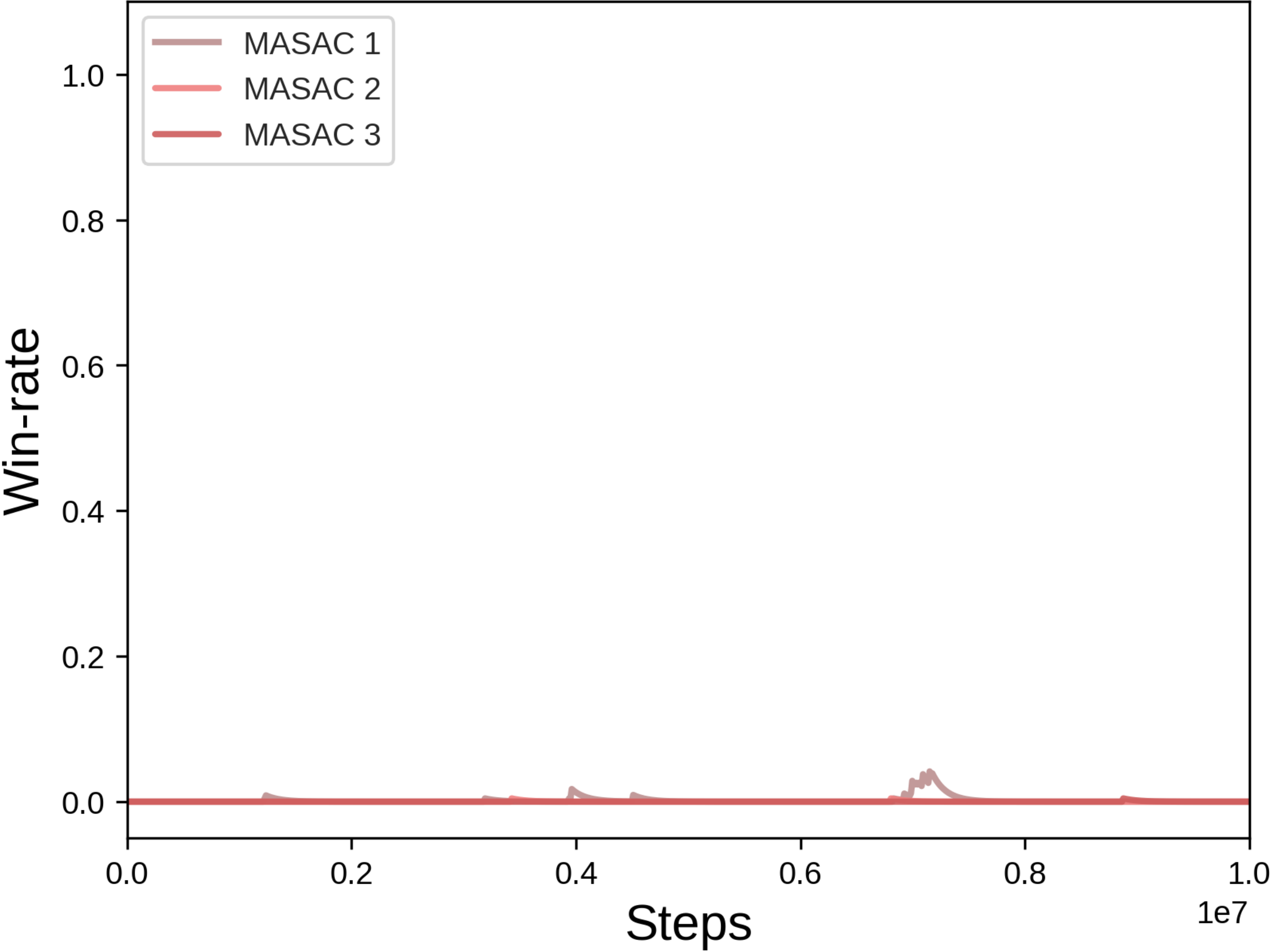}
            \caption{Defense armored}
            \label{fig:app_masac_parallel_def_arm}
        \end{subfigure}%
        \begin{subfigure}{0.26\columnwidth}
            \includegraphics[width=\columnwidth]{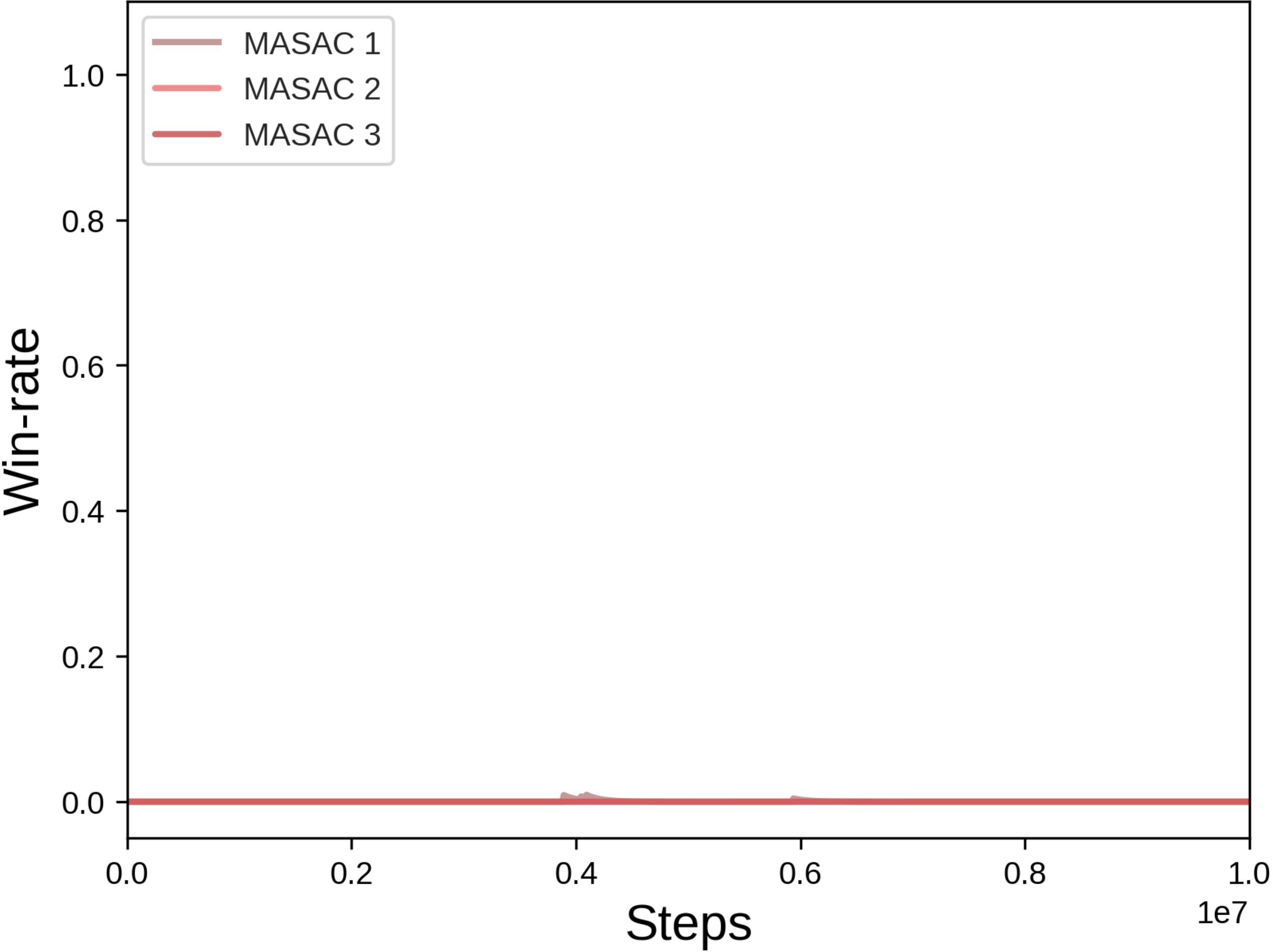}
            \caption{Defense outnumbered}
            \label{fig:app_masac_parallel_def_out}
        \end{subfigure}%
        
        \begin{subfigure}{0.26\columnwidth}
            \includegraphics[width=\columnwidth]{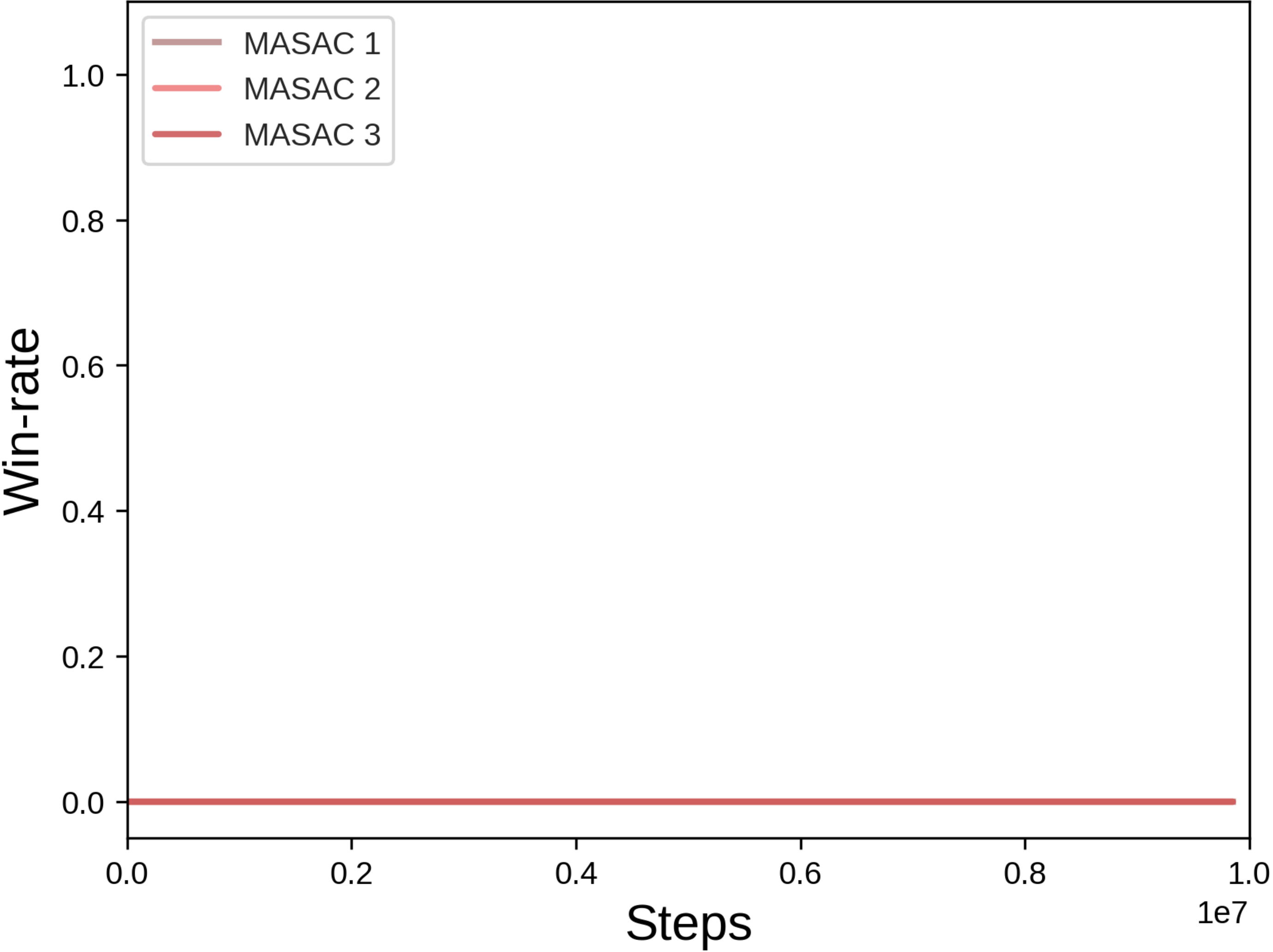}
            \caption{Offense near}
            \label{fig:app_masac_parallel_off_near}
        \end{subfigure}%
        \begin{subfigure}{0.26\columnwidth}
            \includegraphics[width=\columnwidth]{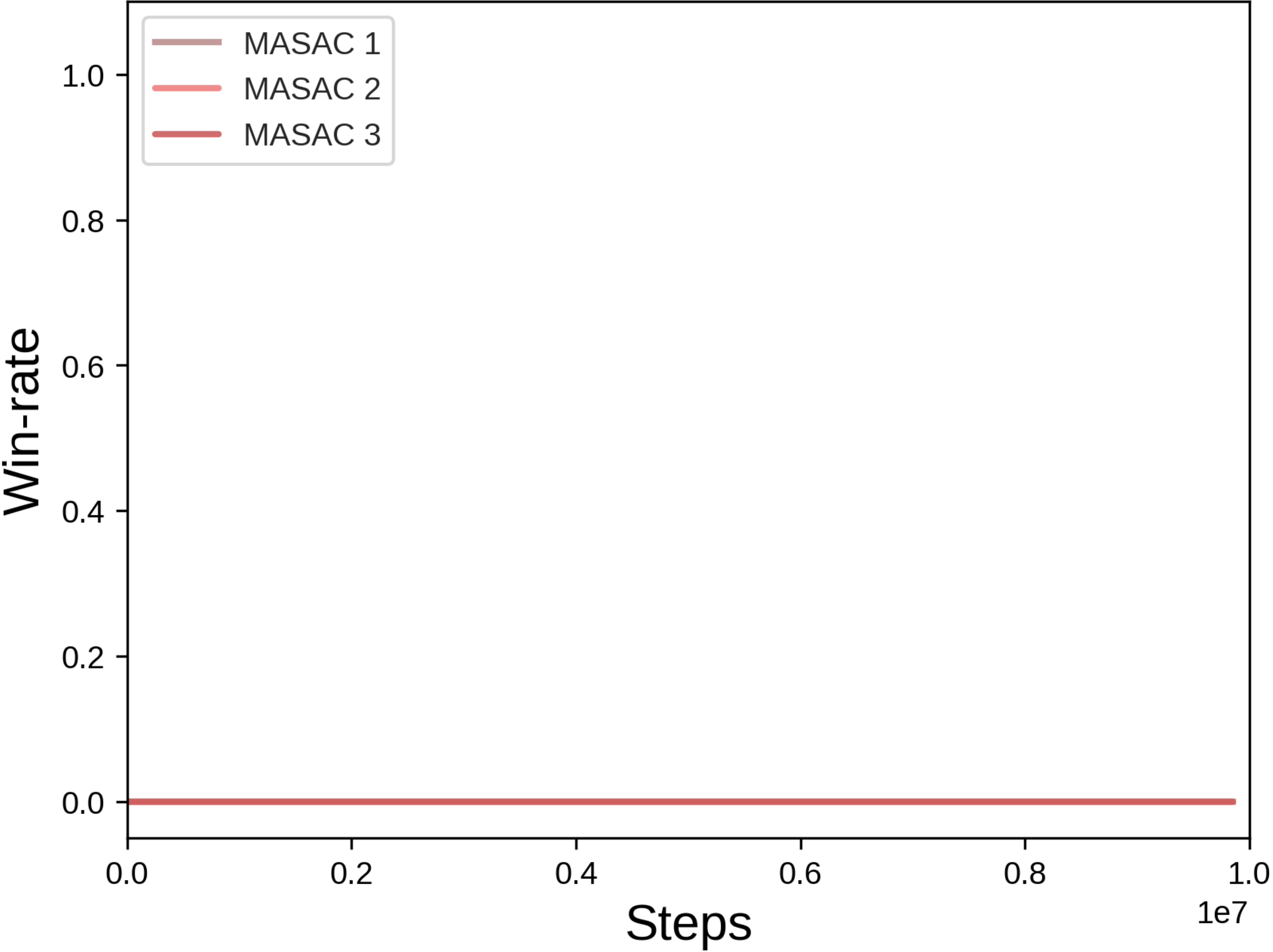}
            \caption{Offense distant}
            \label{fig:app_masac_parallel_off_dist}
        \end{subfigure}%
        \begin{subfigure}{0.26\columnwidth}
            \includegraphics[width=\columnwidth]{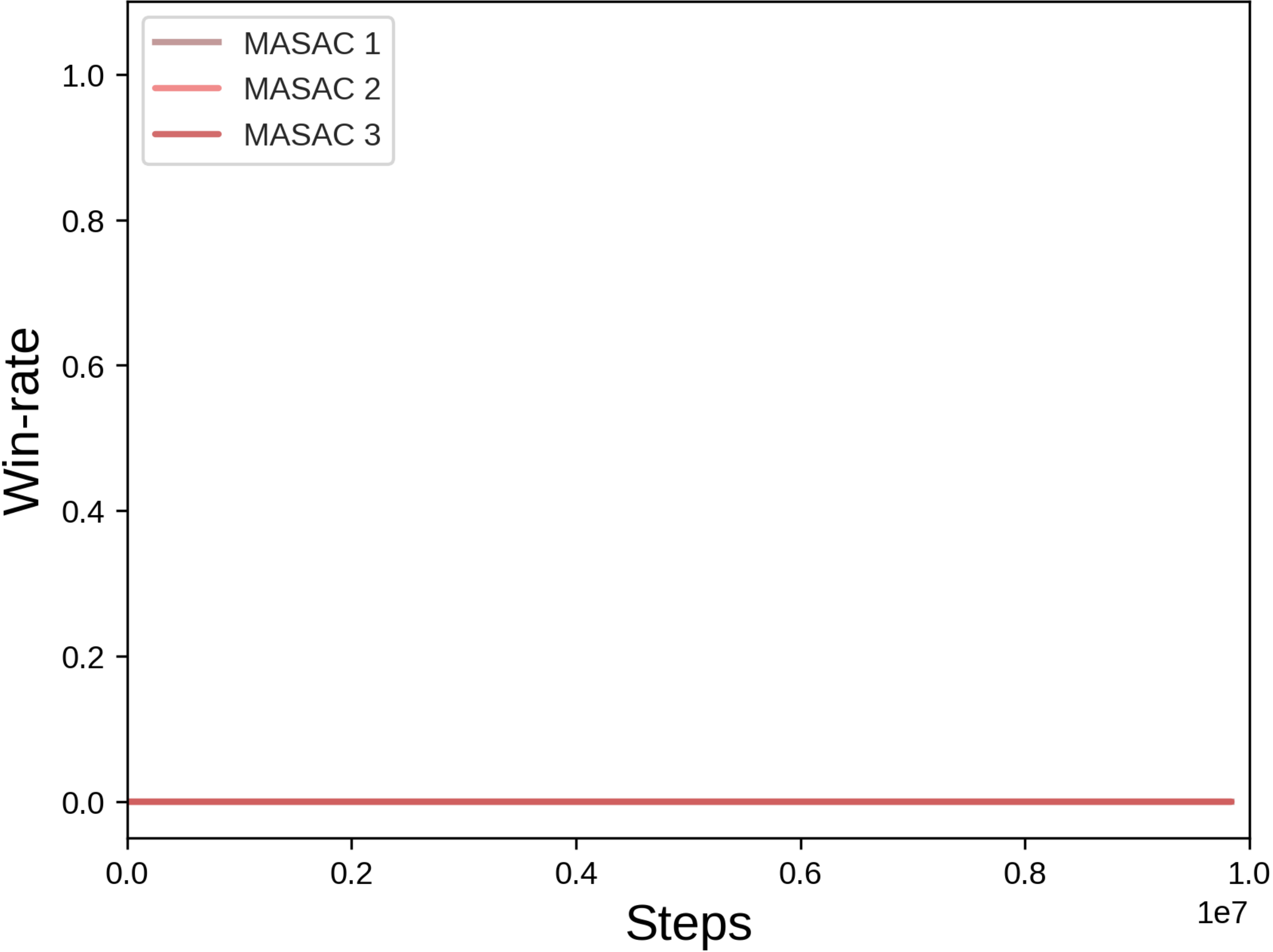}
            \caption{Offense complicated}
            \label{fig:app_masac_parallel_off_com}
        \end{subfigure}%
        
        \begin{subfigure}{0.27\columnwidth}
            \includegraphics[width=\columnwidth]{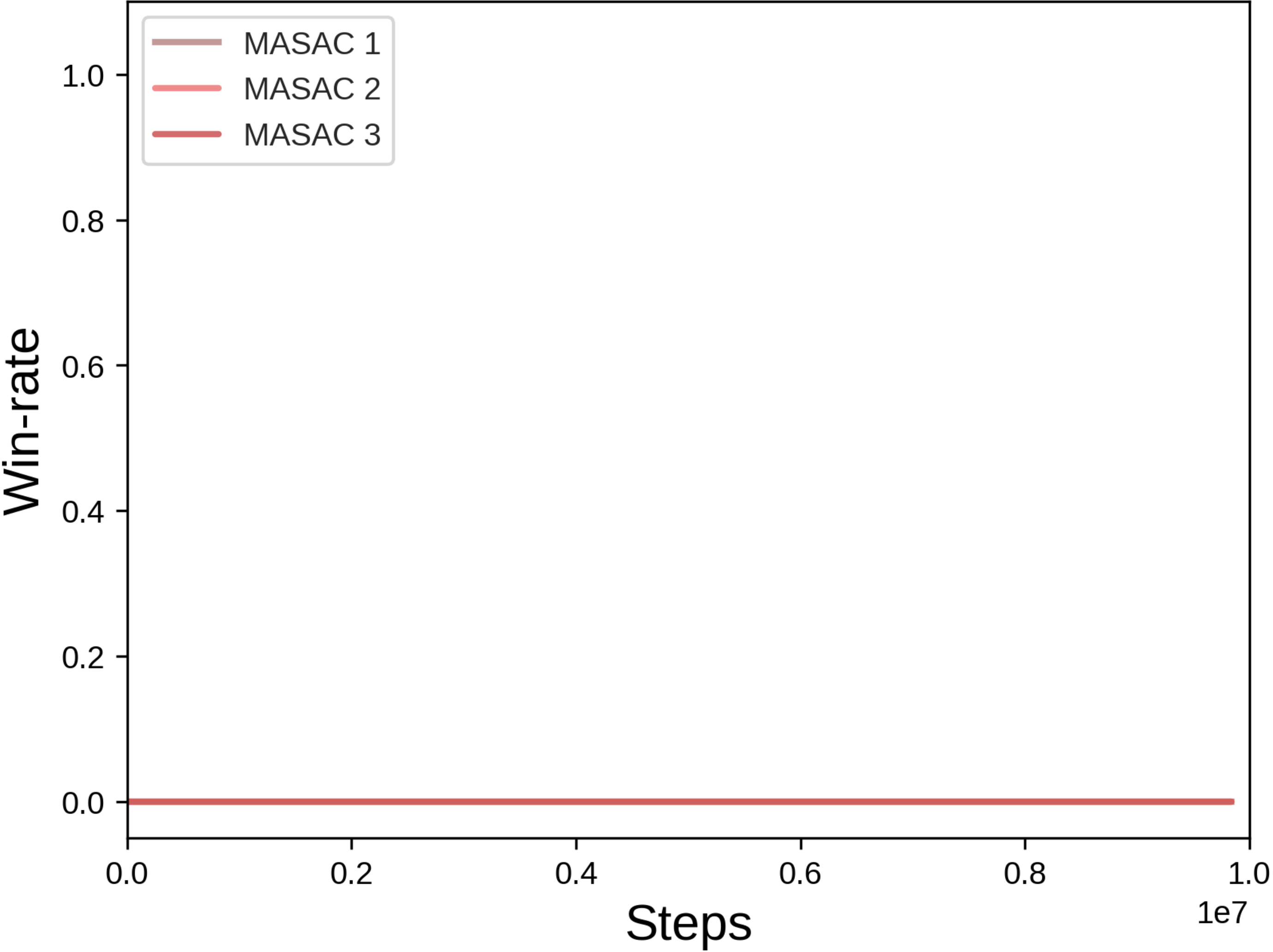}
            \caption{Offense hard}
            \label{fig:app_masac_parallel_off_hard}
        \end{subfigure}%
        \begin{subfigure}{0.27\columnwidth}
            \includegraphics[width=\columnwidth]{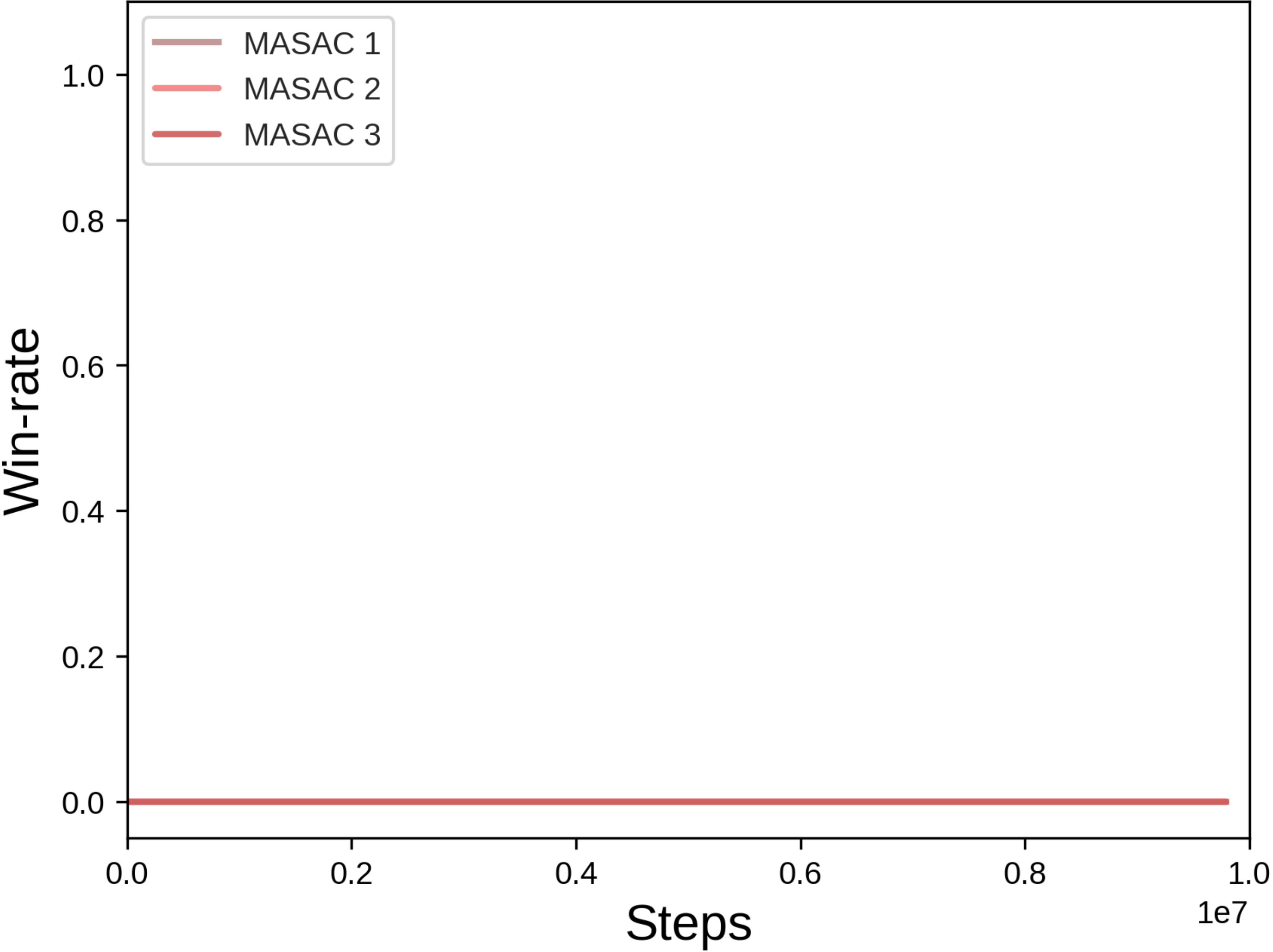}
            \caption{Offense superhard}
            \label{fig:app_masac_parallel_off_super}
        \end{subfigure}%
    \caption{MASAC trained on the parallel episodic buffer}
    \label{fig:app_masac_parallel}
}
\end{figure}

\begin{figure}[!ht]{
    \centering
        \begin{subfigure}{0.26\columnwidth}
            \includegraphics[width=\columnwidth]{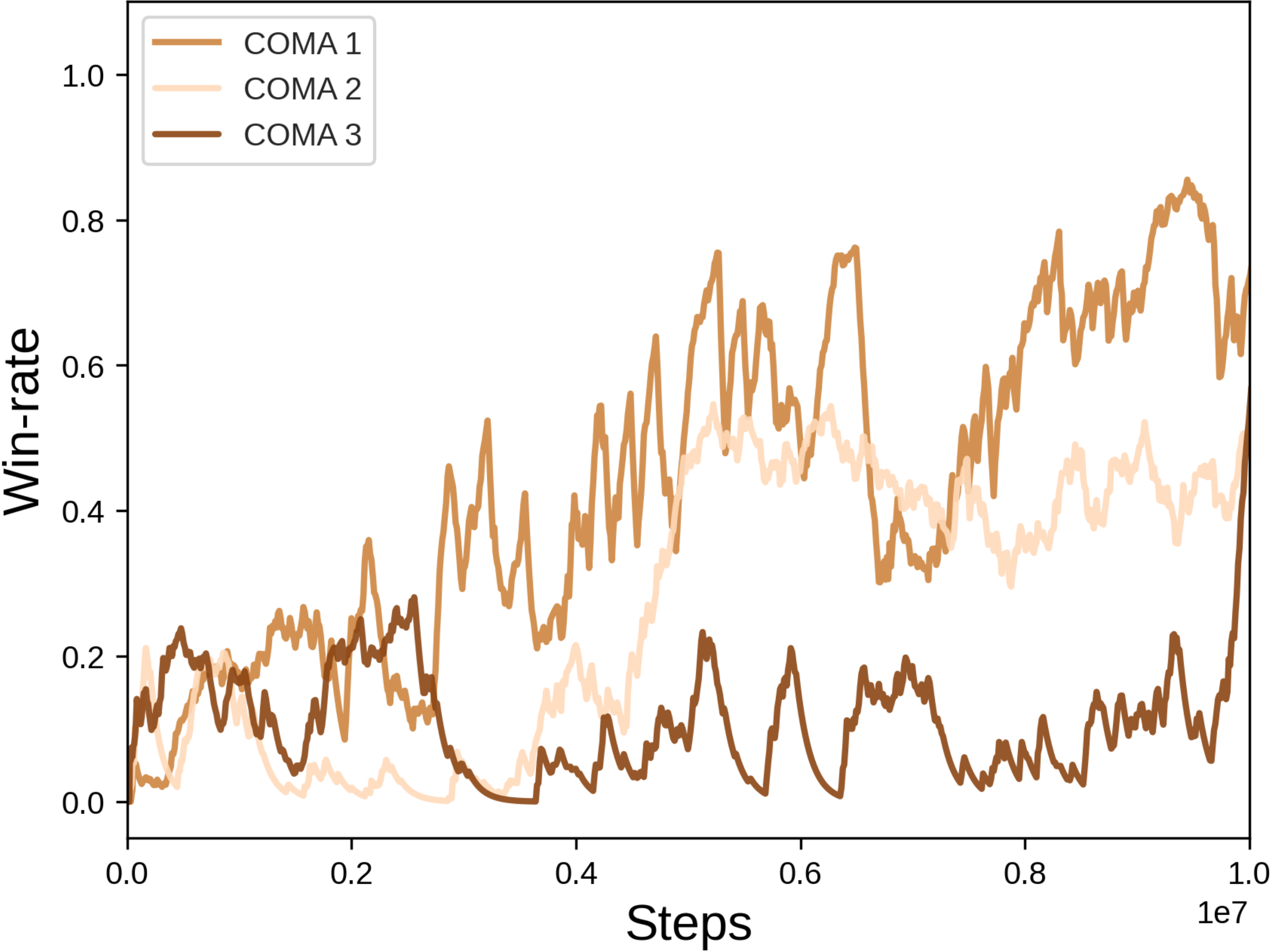}
            \caption{Defense infantry}
            \label{fig:app_coma_parallel_def_inf}
        \end{subfigure}%
        \begin{subfigure}{0.26\columnwidth}
            \includegraphics[width=\columnwidth]{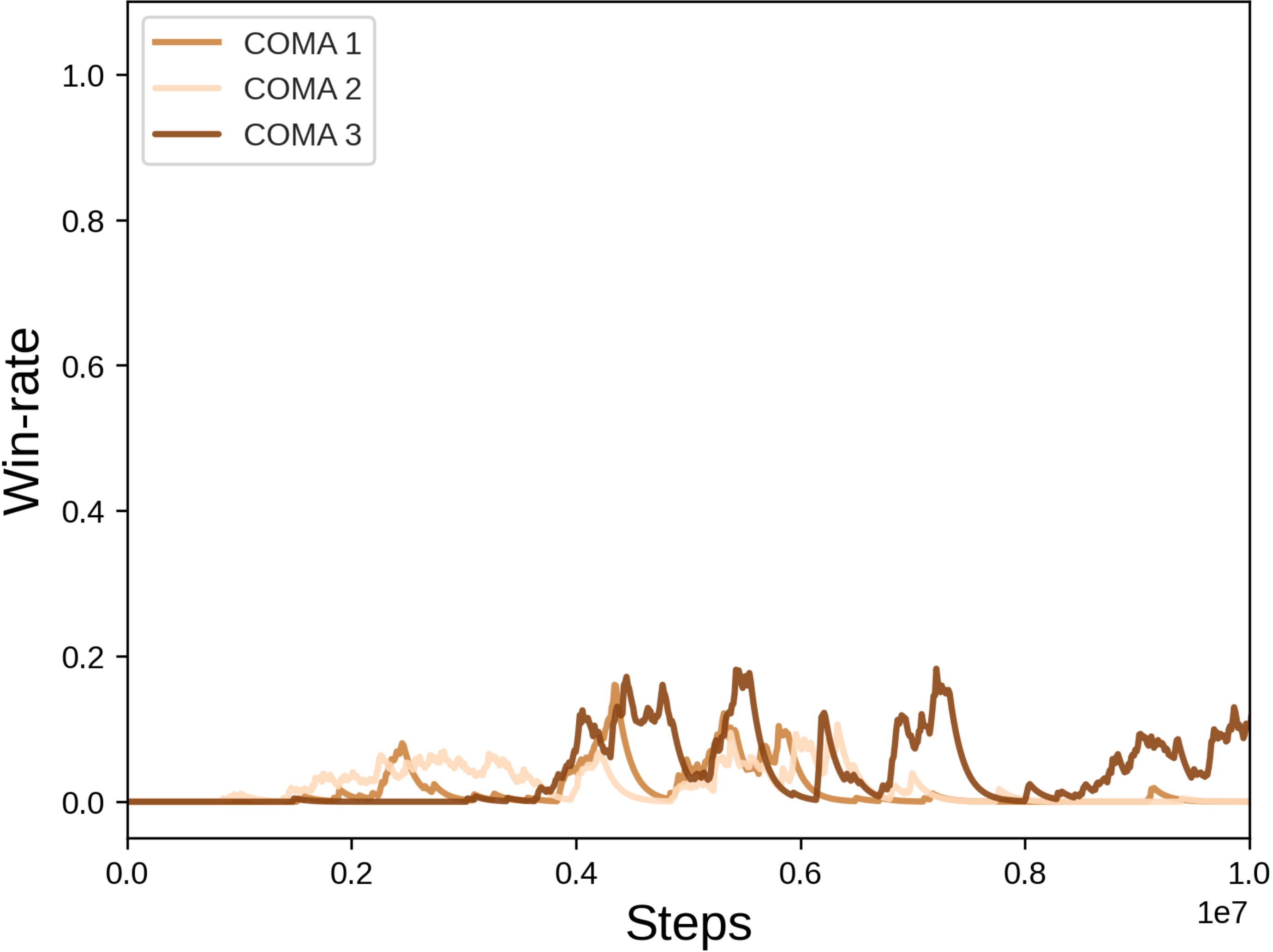}
            \caption{Defense armored}
            \label{fig:app_coma_parallel_def_arm}
        \end{subfigure}%
        \begin{subfigure}{0.26\columnwidth}
            \includegraphics[width=\columnwidth]{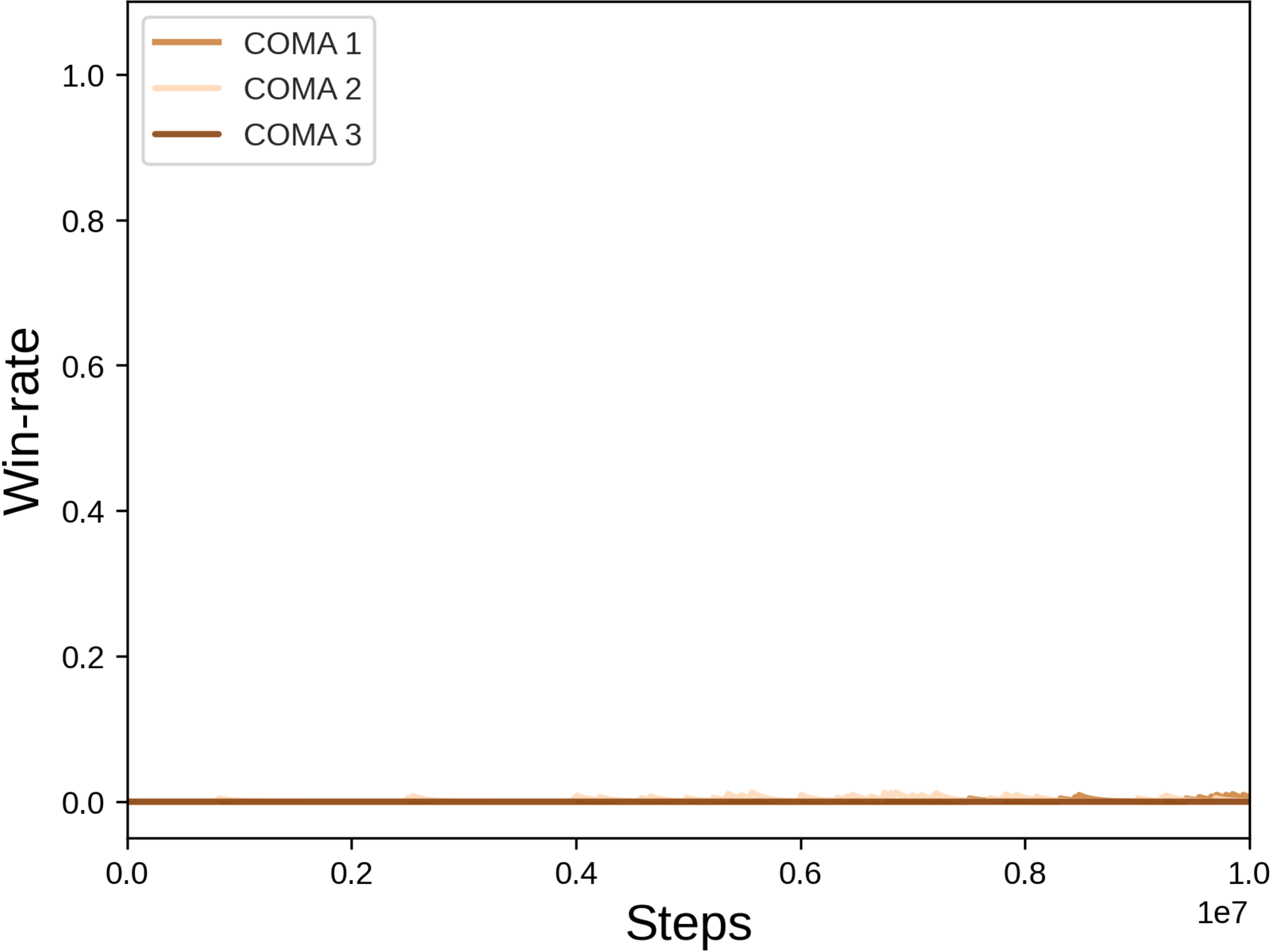}
            \caption{Defense outnumbered}
            \label{fig:app_coma_parallel_def_out}
        \end{subfigure}%
        
        \begin{subfigure}{0.26\columnwidth}
            \includegraphics[width=\columnwidth]{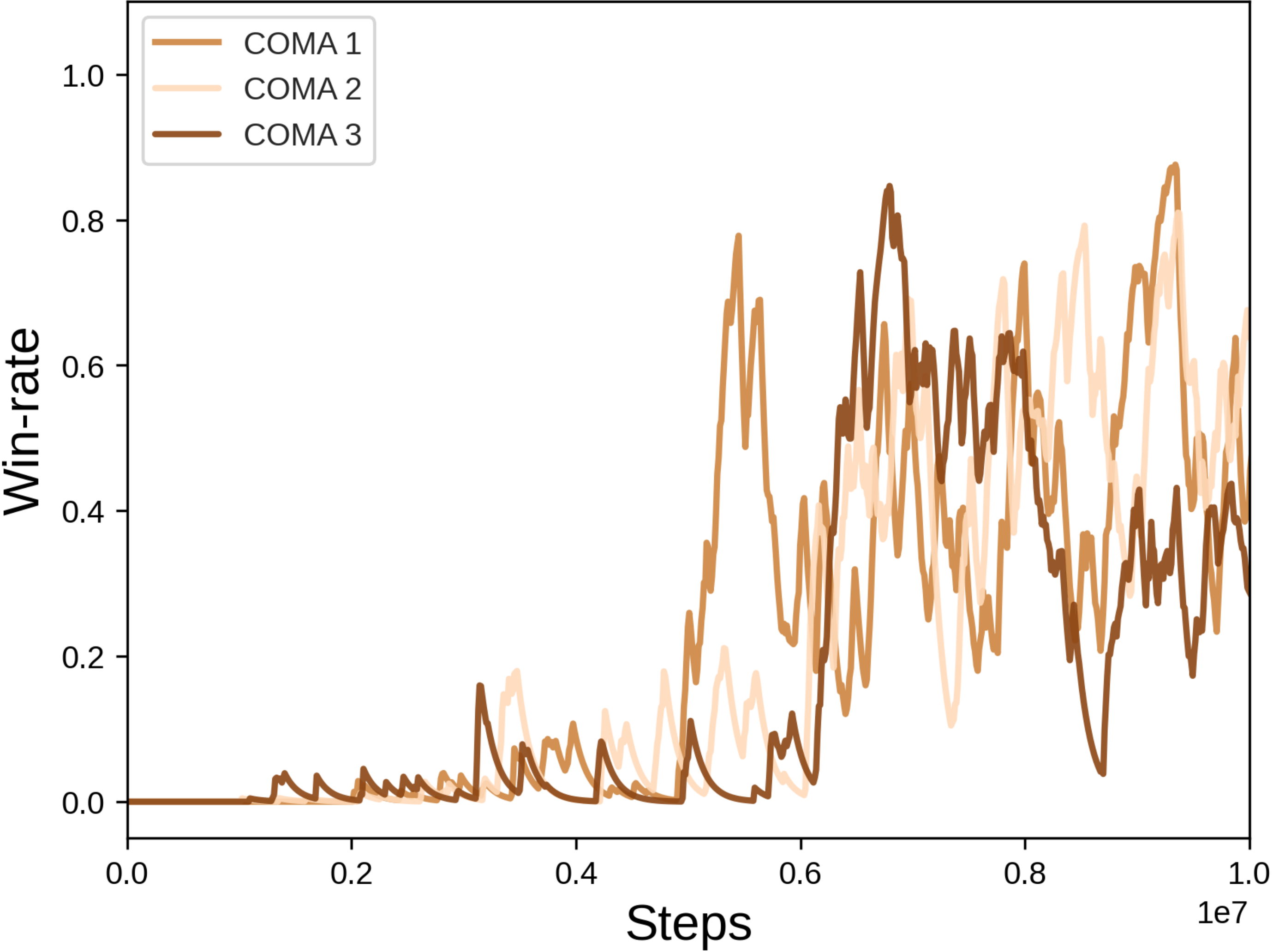}
            \caption{Offense near}
            \label{fig:app_coma_parallel_off_near}
        \end{subfigure}%
        \begin{subfigure}{0.26\columnwidth}
            \includegraphics[width=\columnwidth]{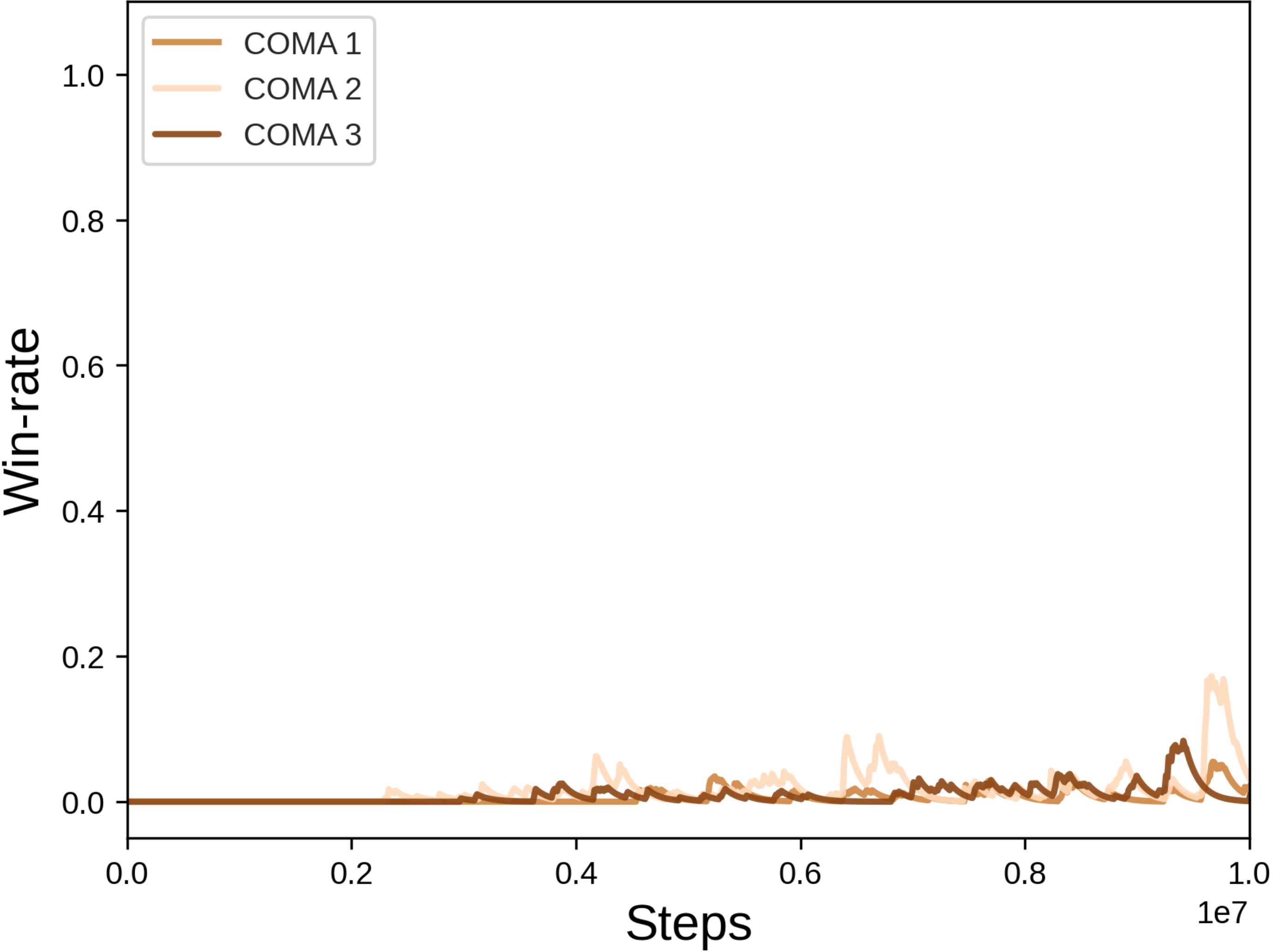}
            \caption{Offense distant}
            \label{fig:app_coma_parallel_off_dist}
        \end{subfigure}%
        \begin{subfigure}{0.26\columnwidth}
            \includegraphics[width=\columnwidth]{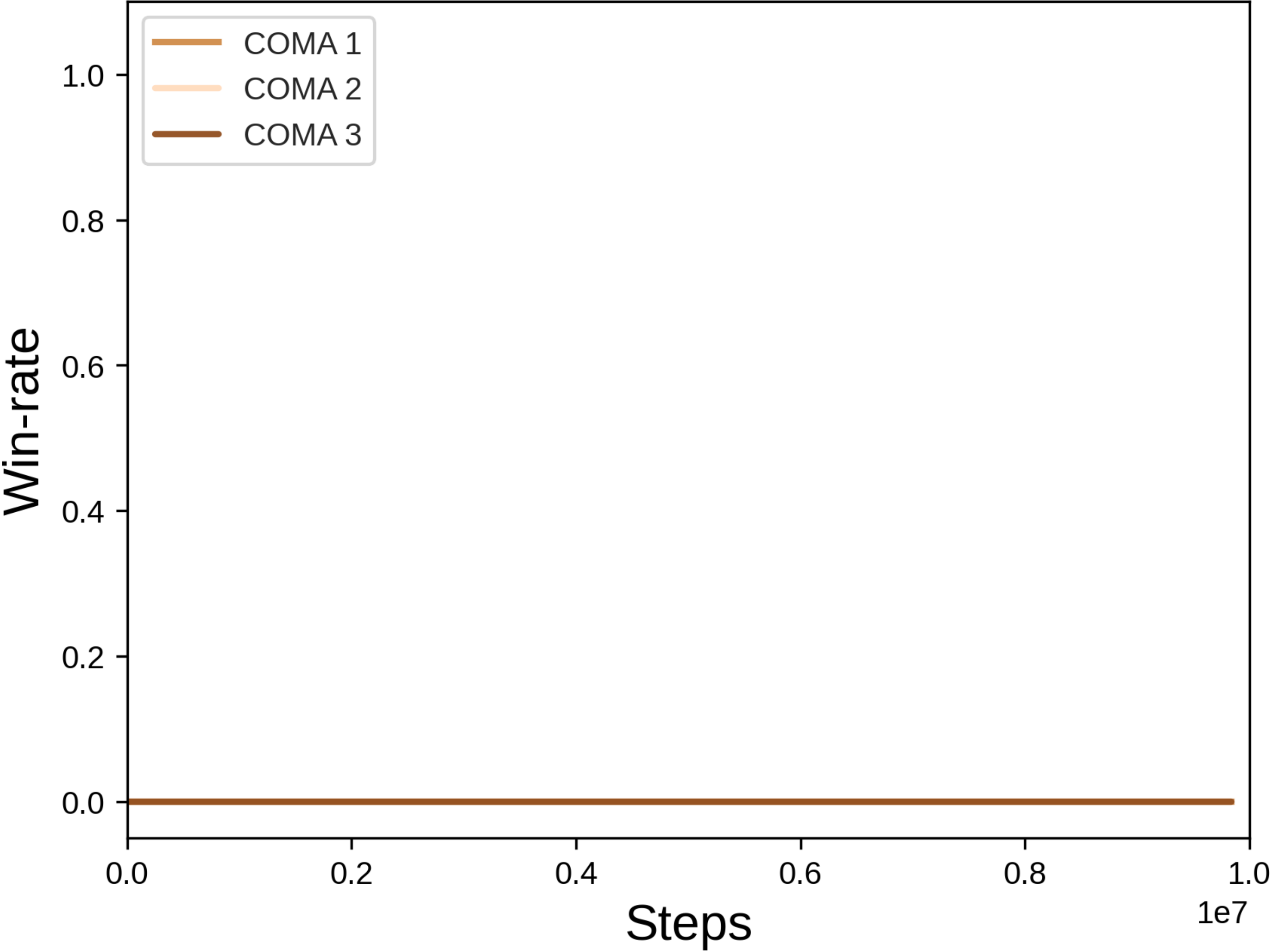}
            \caption{Offense complicated}
            \label{fig:app_coma_parallel_off_com}
        \end{subfigure}%
        
        \begin{subfigure}{0.27\columnwidth}
            \includegraphics[width=\columnwidth]{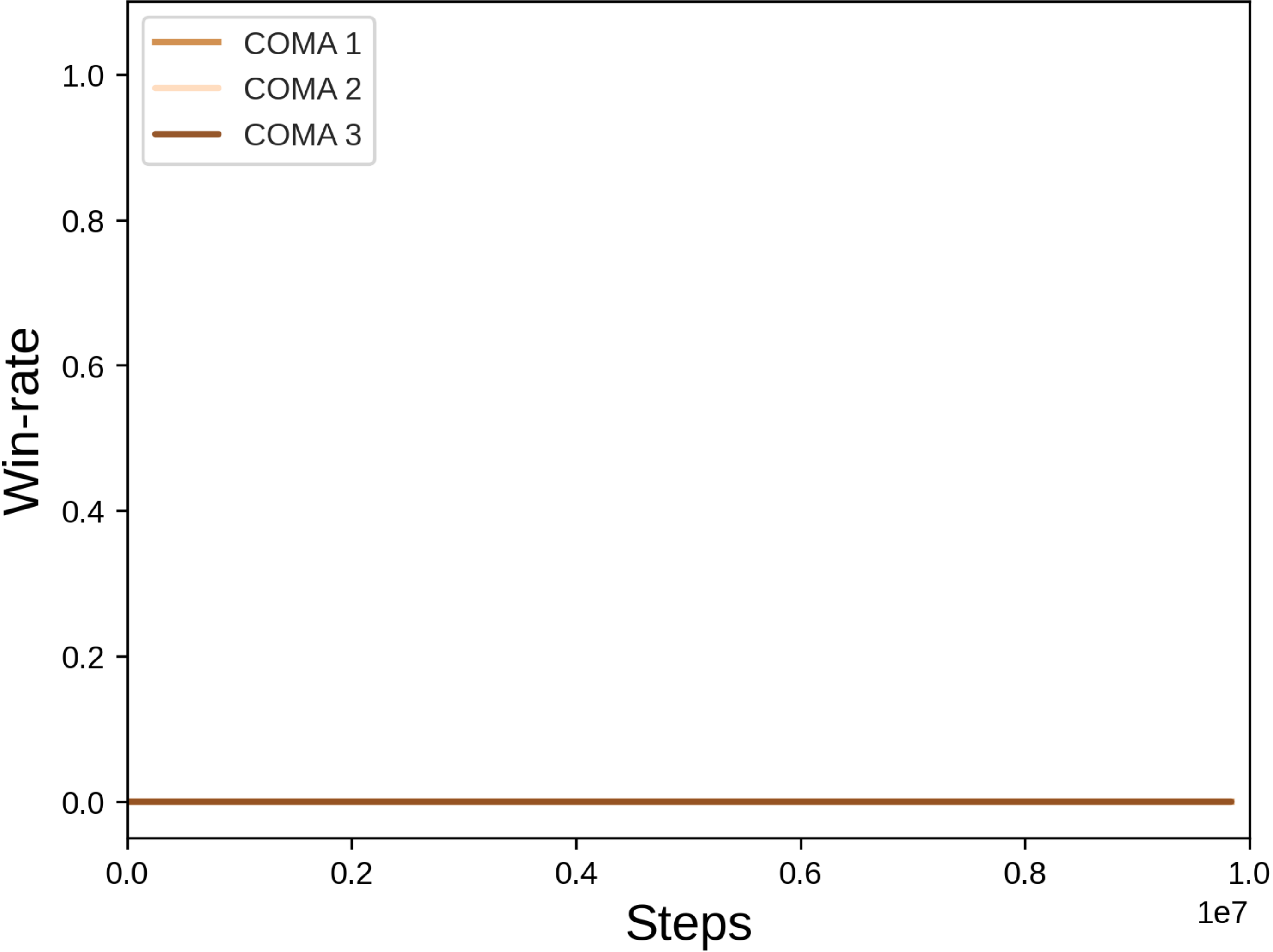}
            \caption{Offense hard}
            \label{fig:app_coma_parallel_off_hard}
        \end{subfigure}%
        \begin{subfigure}{0.27\columnwidth}
            \includegraphics[width=\columnwidth]{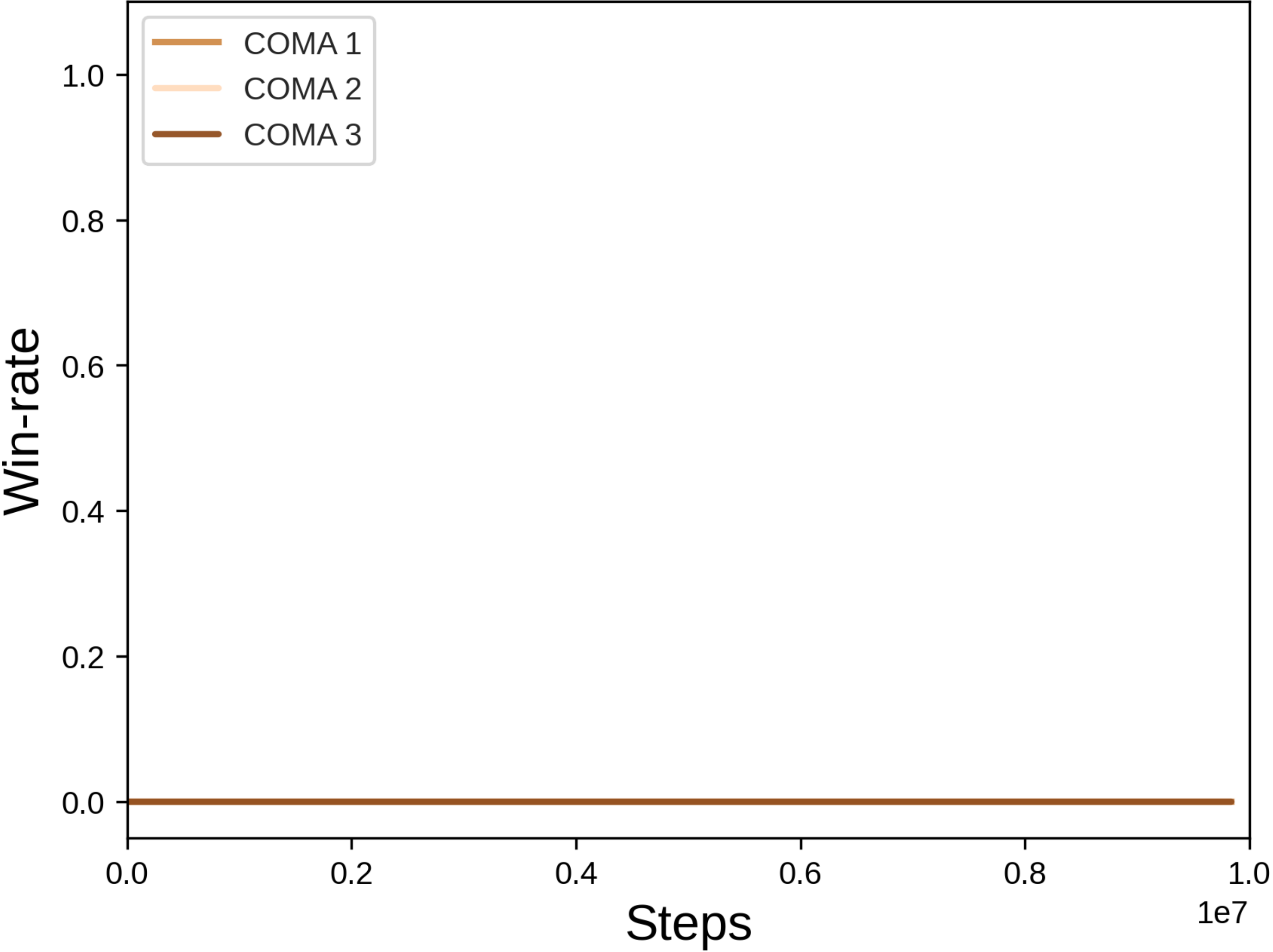}
            \caption{Offense superhard}
            \label{fig:app_coma_parallel_off_super}
        \end{subfigure}%
    \caption{COMA trained on the parallel episodic buffer}
    \label{fig:app_coma_parallel}
}
\end{figure}

\begin{figure}[!ht]{
    \centering
        \begin{subfigure}{0.26\columnwidth}
            \includegraphics[width=\columnwidth]{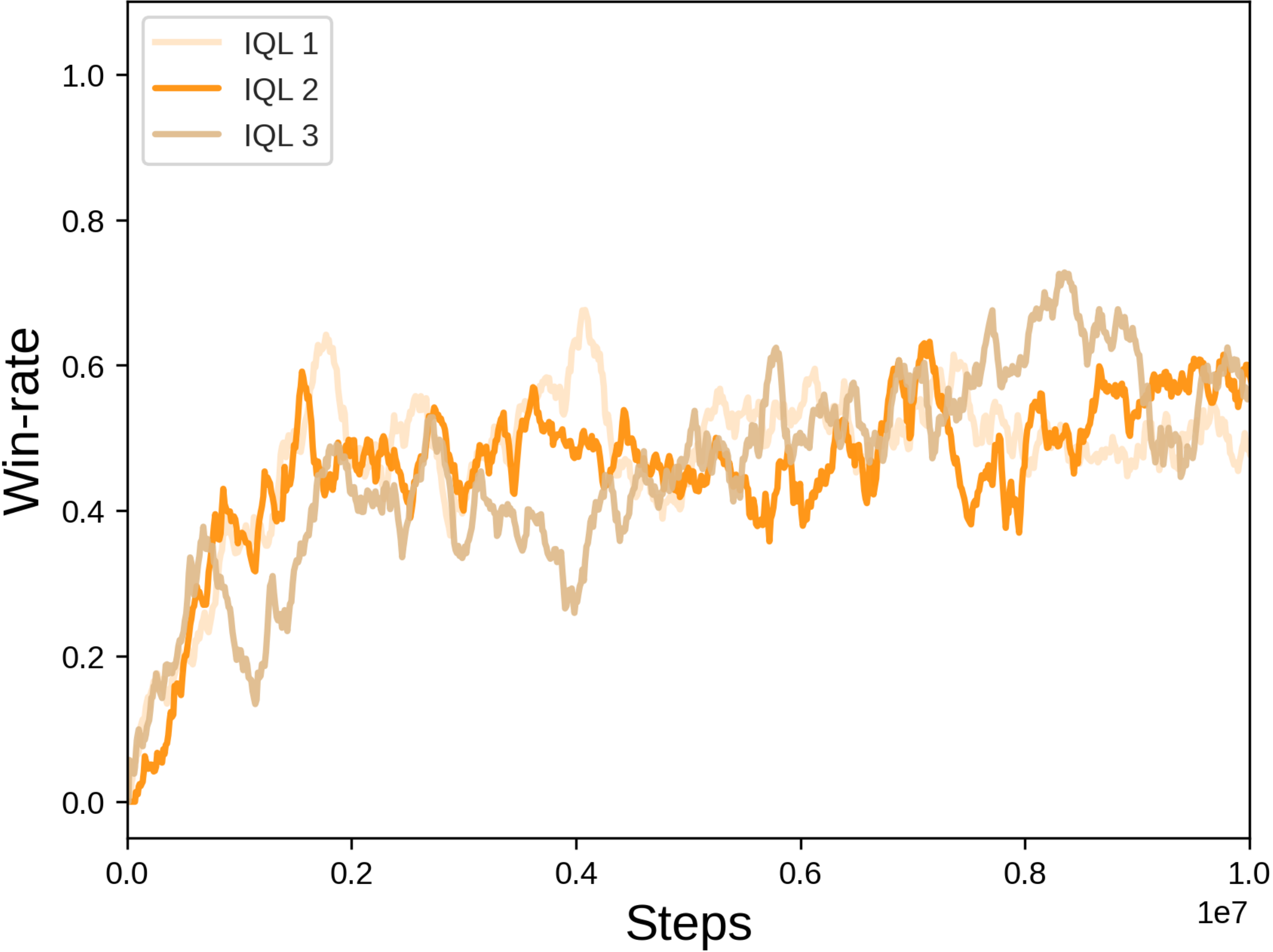}
            \caption{Defense infantry}
            \label{fig:app_iql_parallel_def_inf}
        \end{subfigure}%
        \begin{subfigure}{0.26\columnwidth}
            \includegraphics[width=\columnwidth]{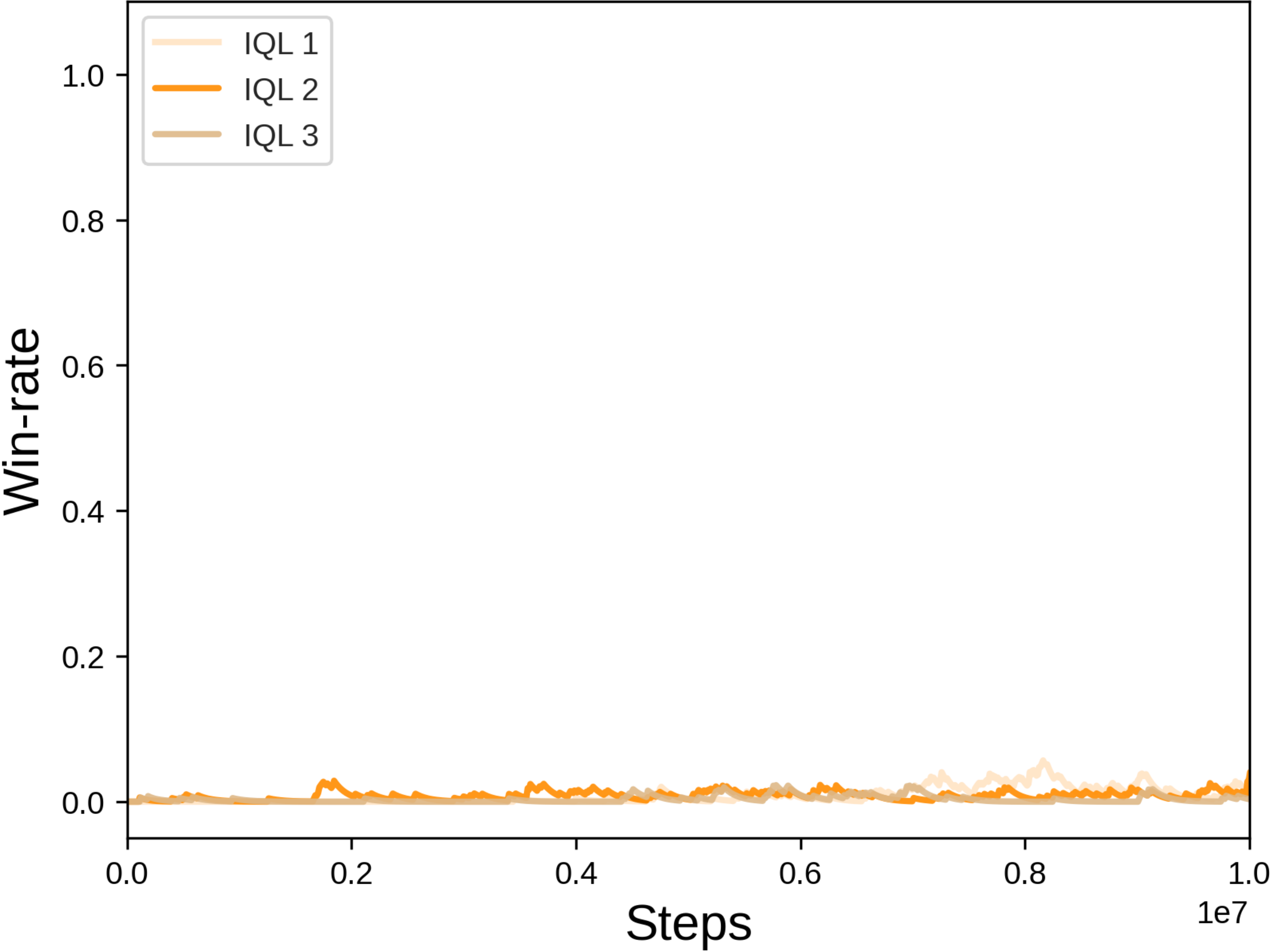}
            \caption{Defense armored}
            \label{fig:app_iql_parallel_def_arm}
        \end{subfigure}%
        \begin{subfigure}{0.26\columnwidth}
            \includegraphics[width=\columnwidth]{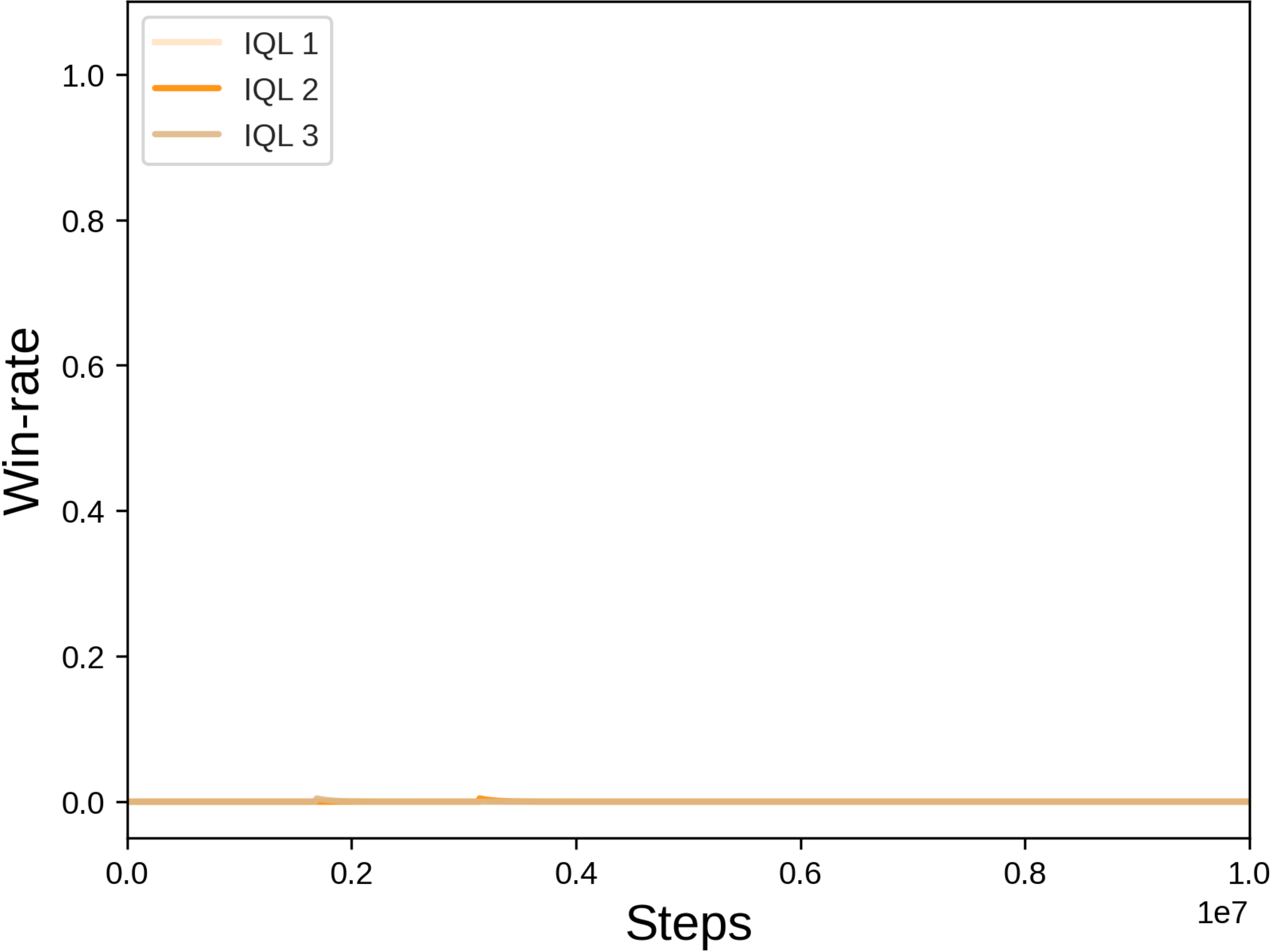}
            \caption{Defense outnumbered}
            \label{fig:app_iql_parallel_def_out}
        \end{subfigure}%
        
        \begin{subfigure}{0.26\columnwidth}
            \includegraphics[width=\columnwidth]{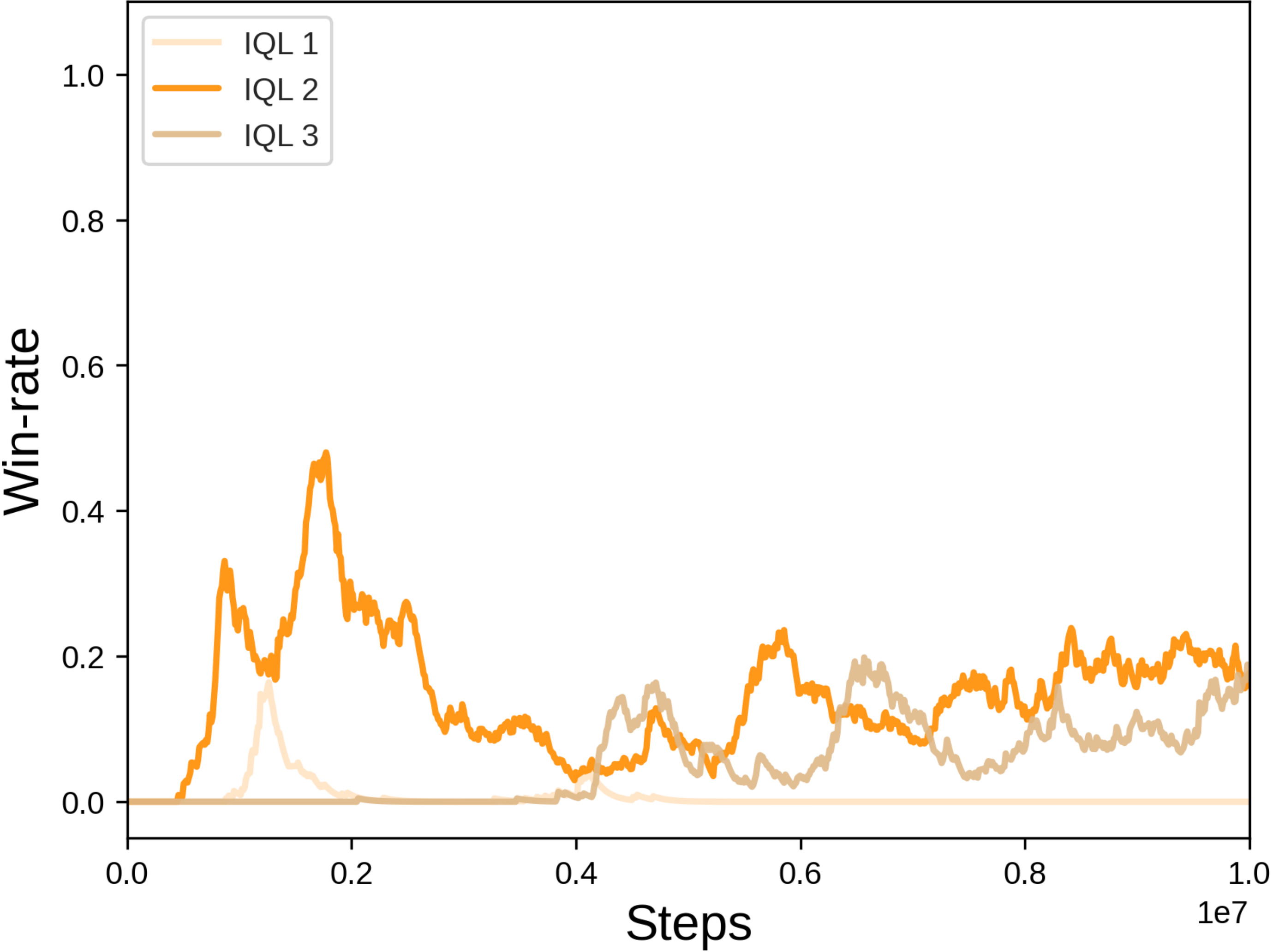}
            \caption{Offense near}
            \label{fig:app_iql_parallel_off_near}
        \end{subfigure}%
        \begin{subfigure}{0.26\columnwidth}
            \includegraphics[width=\columnwidth]{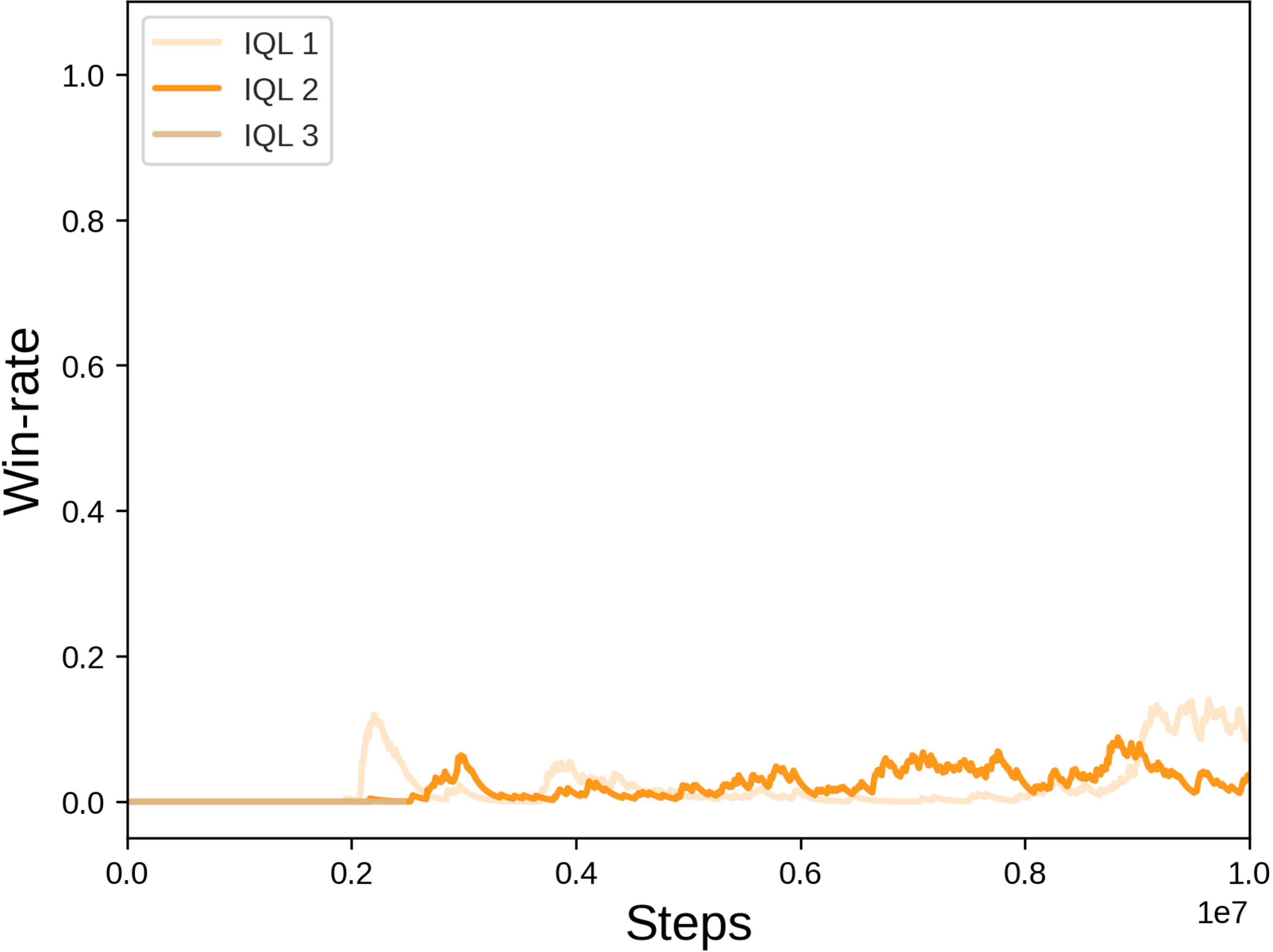}
            \caption{Offense distant}
            \label{fig:app_iql_parallel_off_dist}
        \end{subfigure}%
        \begin{subfigure}{0.26\columnwidth}
            \includegraphics[width=\columnwidth]{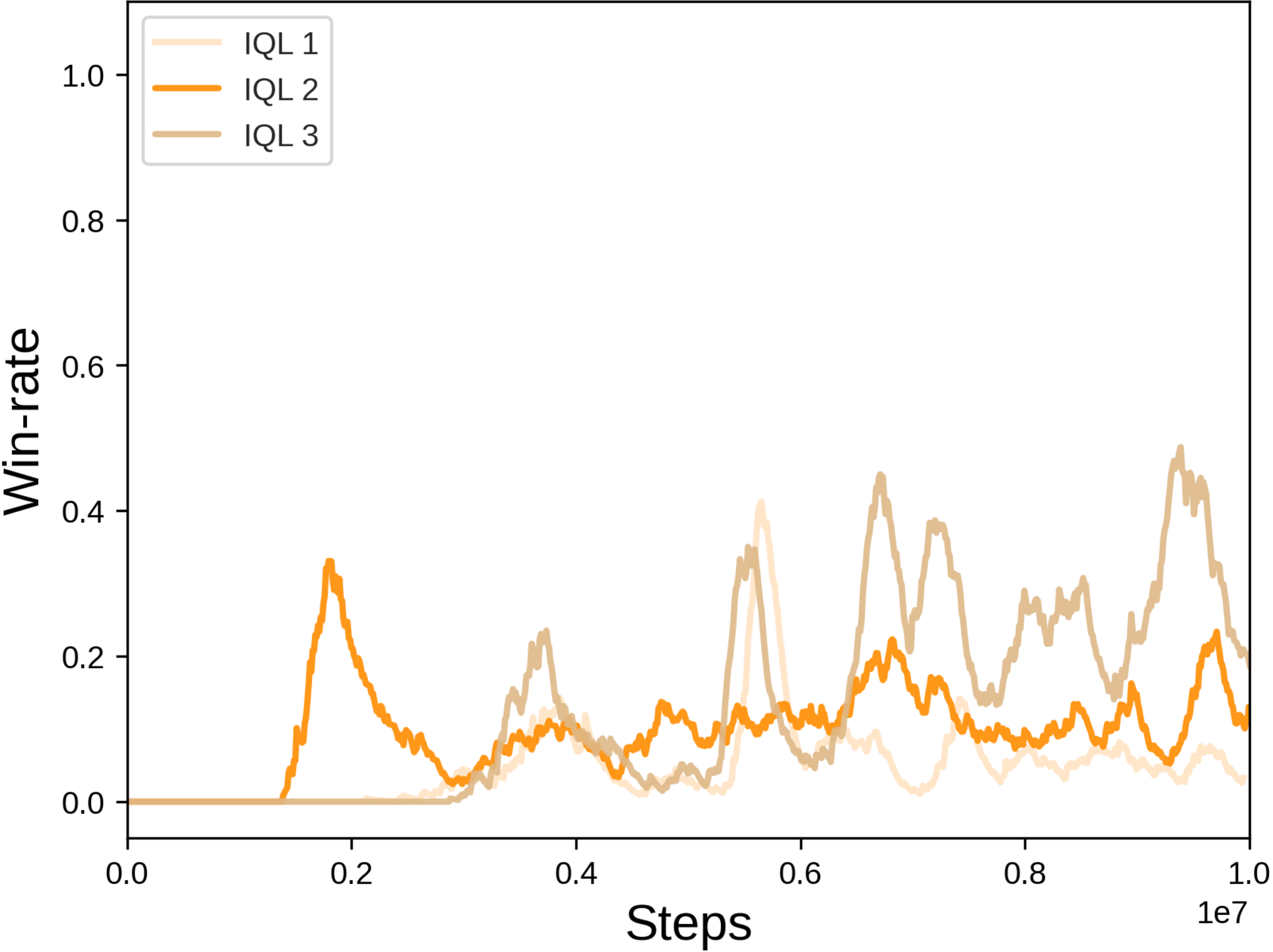}
            \caption{Offense complicated}
            \label{fig:app_iql_parallel_off_com}
        \end{subfigure}%
        
        \begin{subfigure}{0.27\columnwidth}
            \includegraphics[width=\columnwidth]{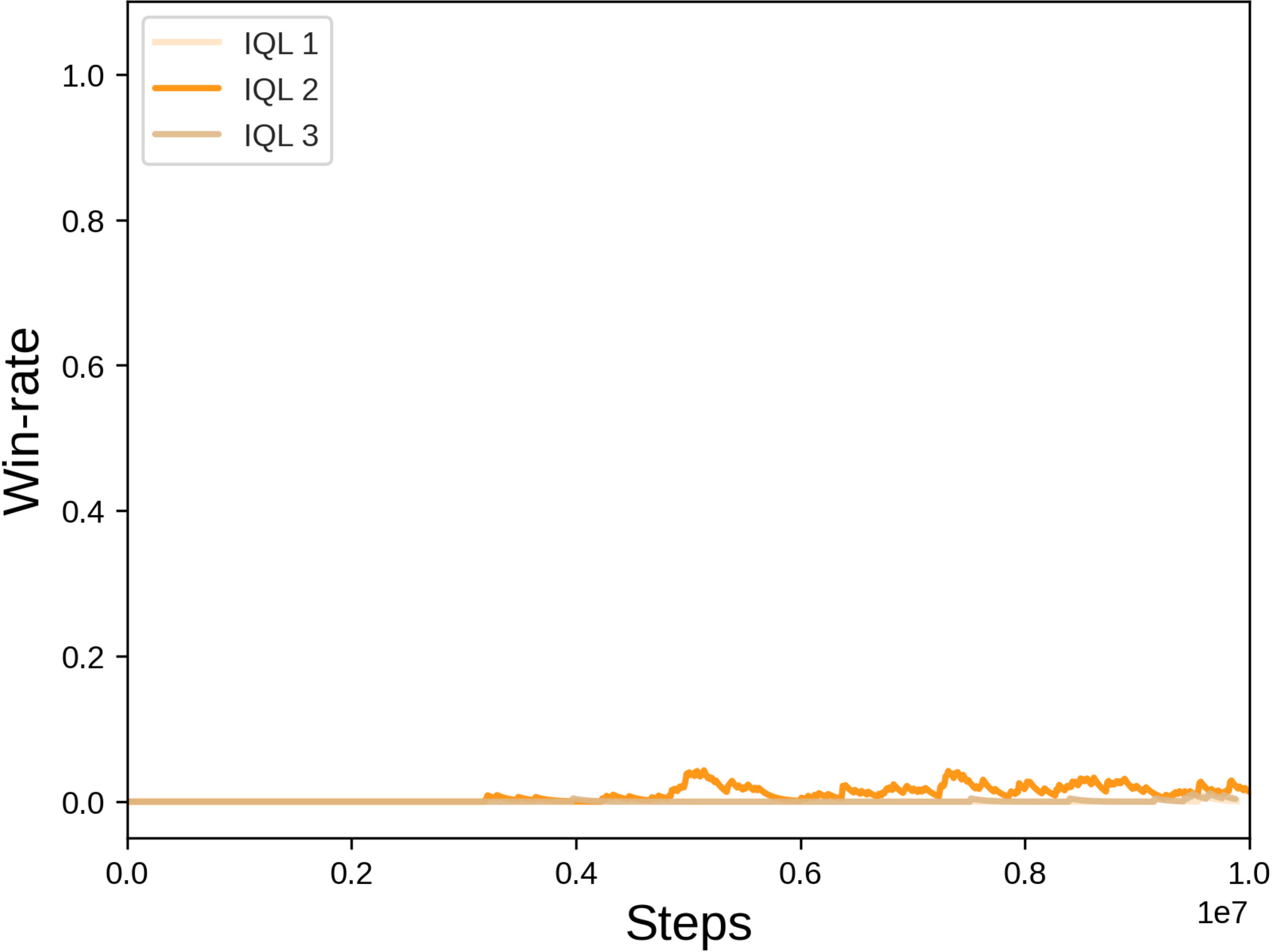}
            \caption{Offense hard}
            \label{fig:app_iql_parallel_off_hard}
        \end{subfigure}%
        \begin{subfigure}{0.27\columnwidth}
            \includegraphics[width=\columnwidth]{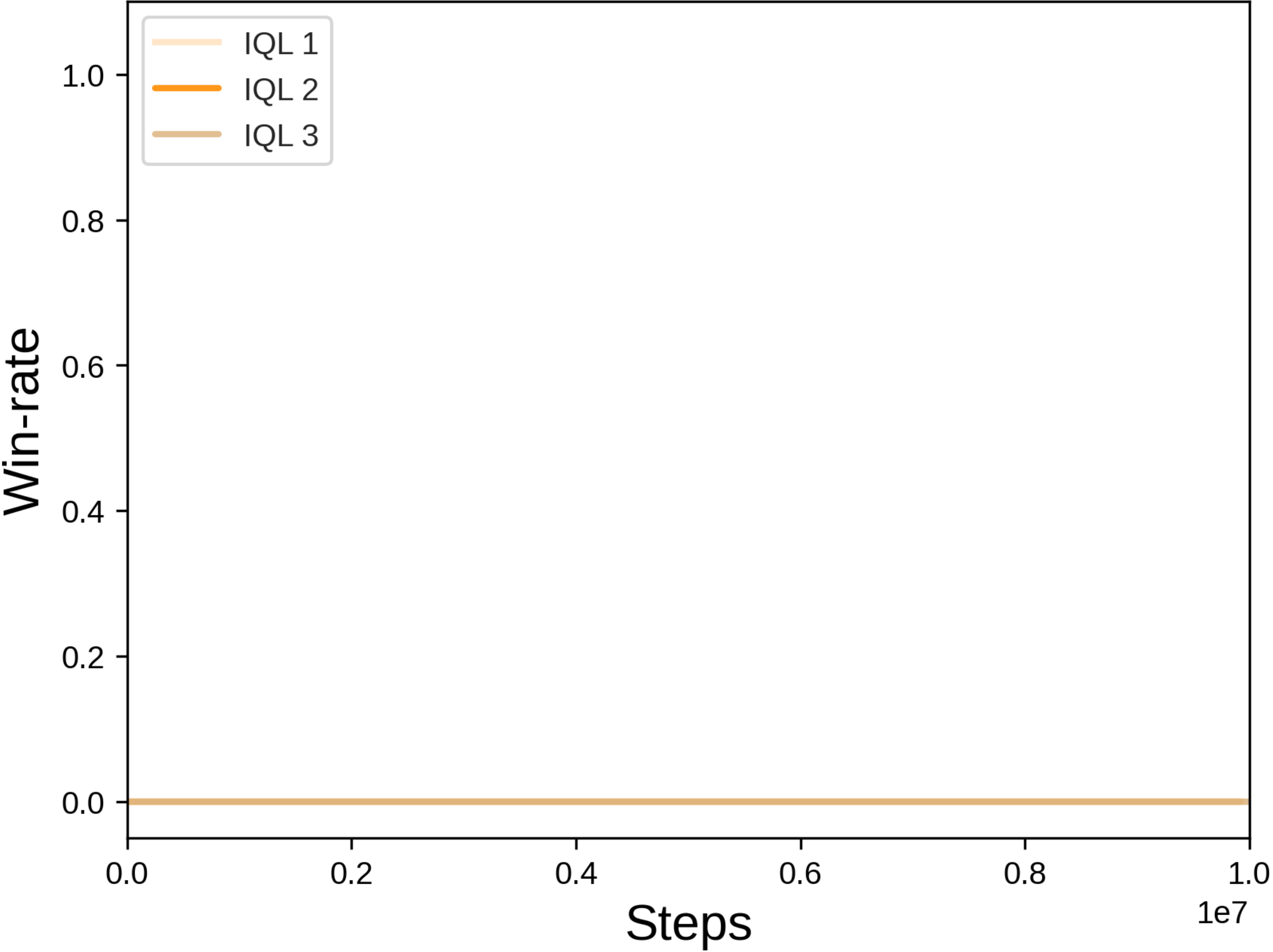}
            \caption{Offense superhard}
            \label{fig:app_iql_parallel_off_super}
        \end{subfigure}%
    \caption{IQL trained on the parallel episodic buffer}
    \label{fig:app_iql_parallel}
}
\end{figure}

\begin{figure}[!ht]{
    \centering
        \begin{subfigure}{0.26\columnwidth}
            \includegraphics[width=\columnwidth]{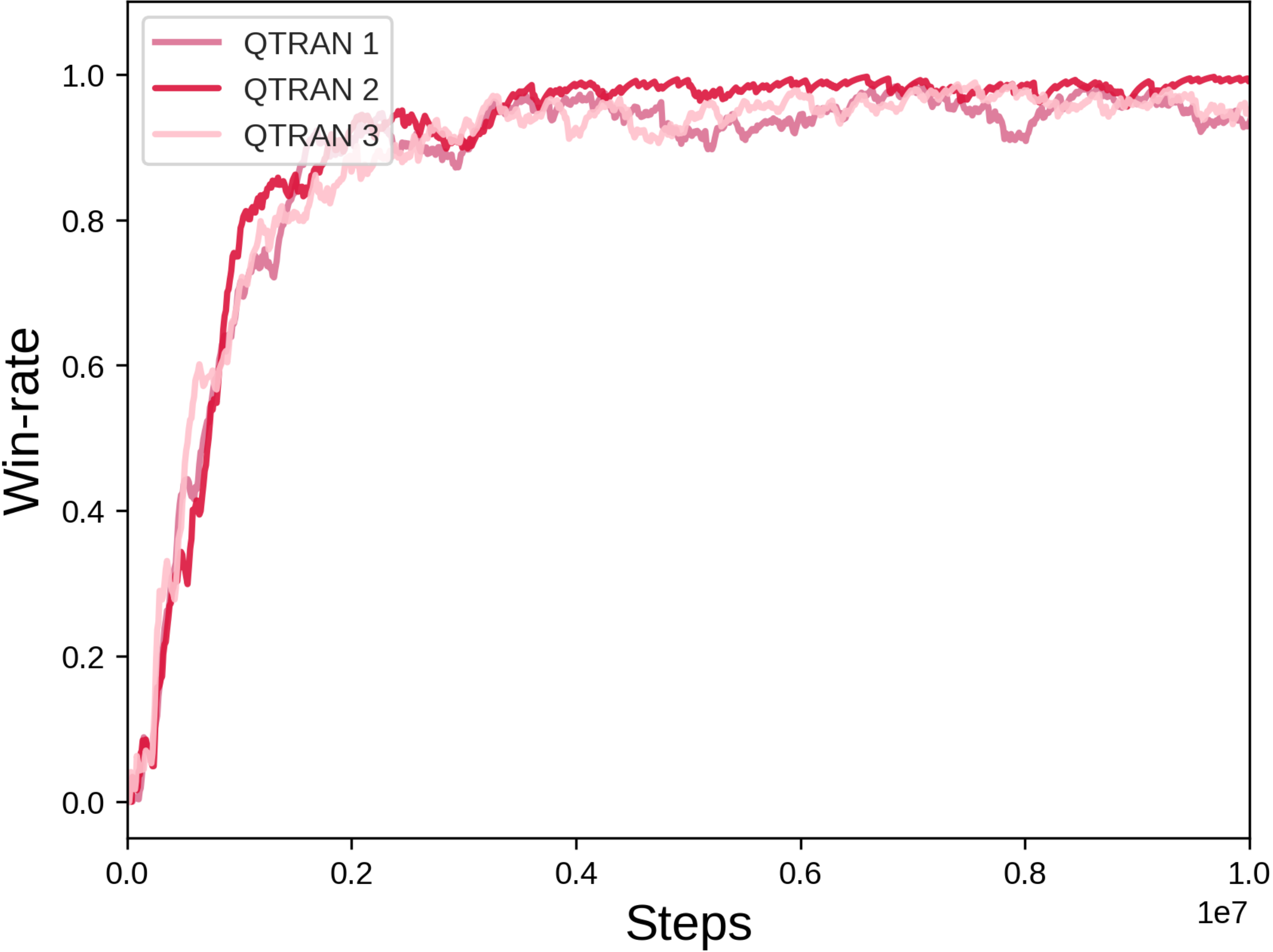}
            \caption{Defense infantry}
            \label{fig:app_qtran_parallel_def_inf}
        \end{subfigure}%
        \begin{subfigure}{0.26\columnwidth}
            \includegraphics[width=\columnwidth]{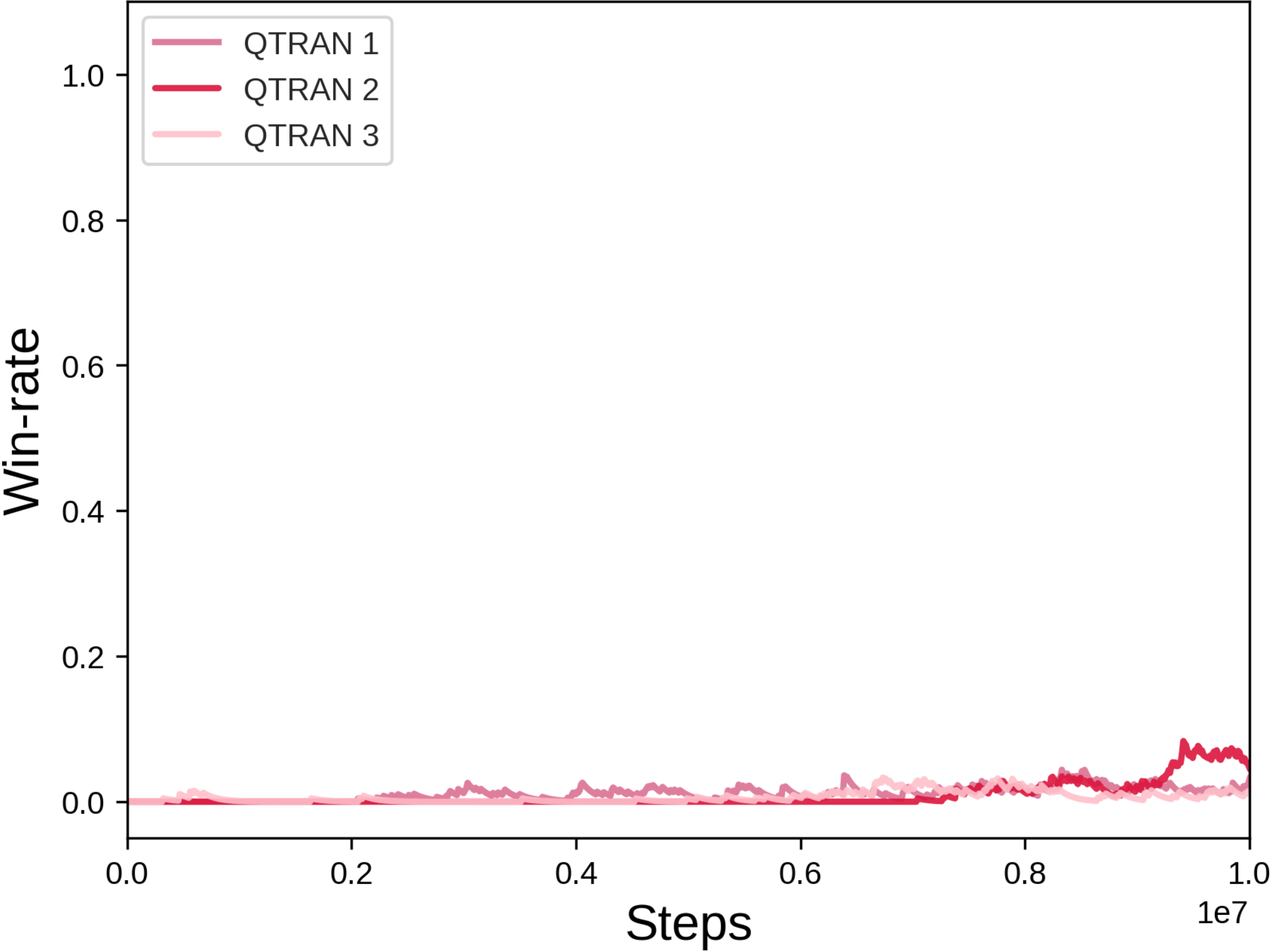}
            \caption{Defense armored}
            \label{fig:app_qtran_parallel_def_arm}
        \end{subfigure}%
        \begin{subfigure}{0.26\columnwidth}
            \includegraphics[width=\columnwidth]{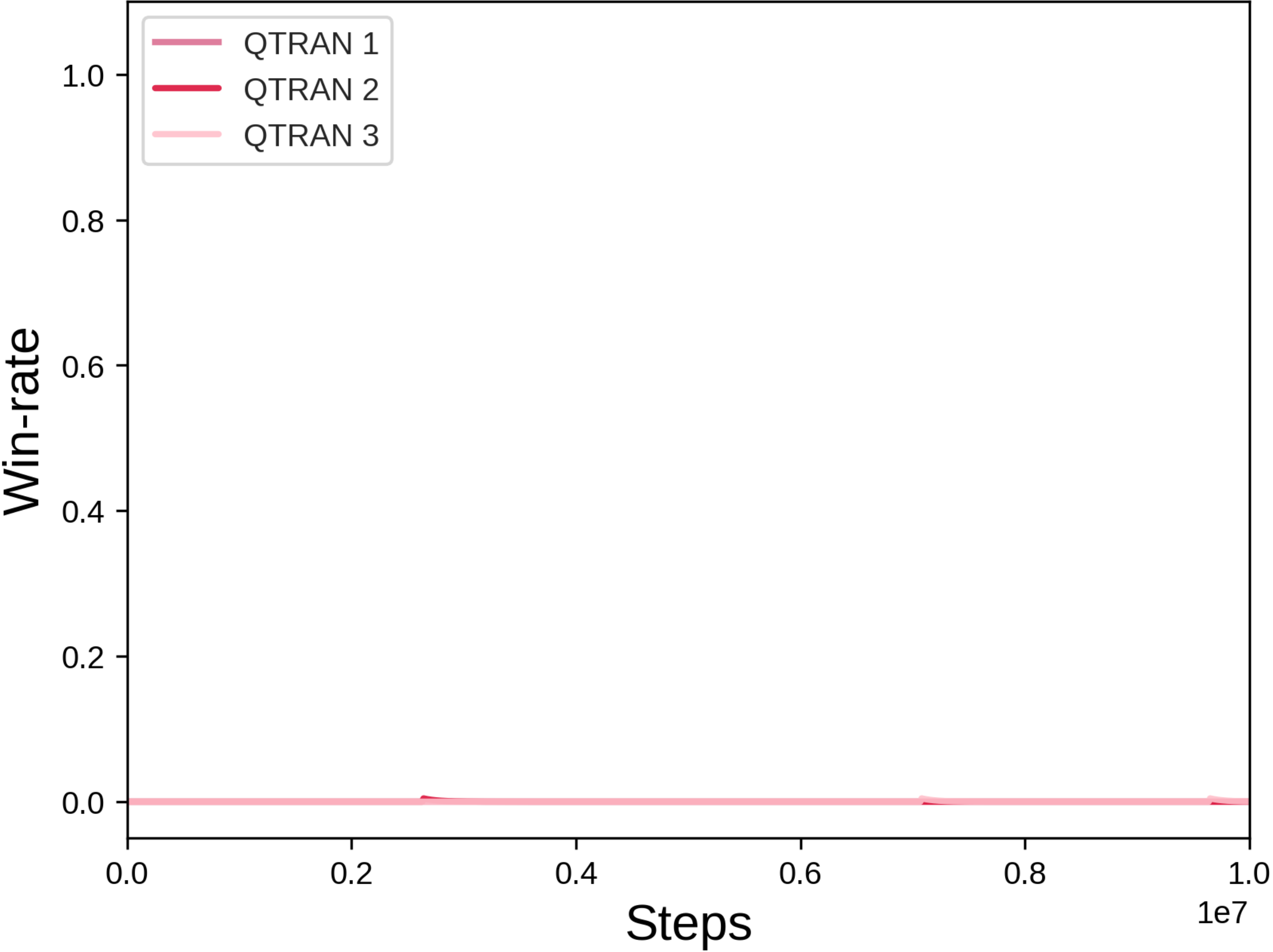}
            \caption{Defense outnumbered}
            \label{fig:app_qtran_parallel_def_out}
        \end{subfigure}%
        
        \begin{subfigure}{0.26\columnwidth}
            \includegraphics[width=\columnwidth]{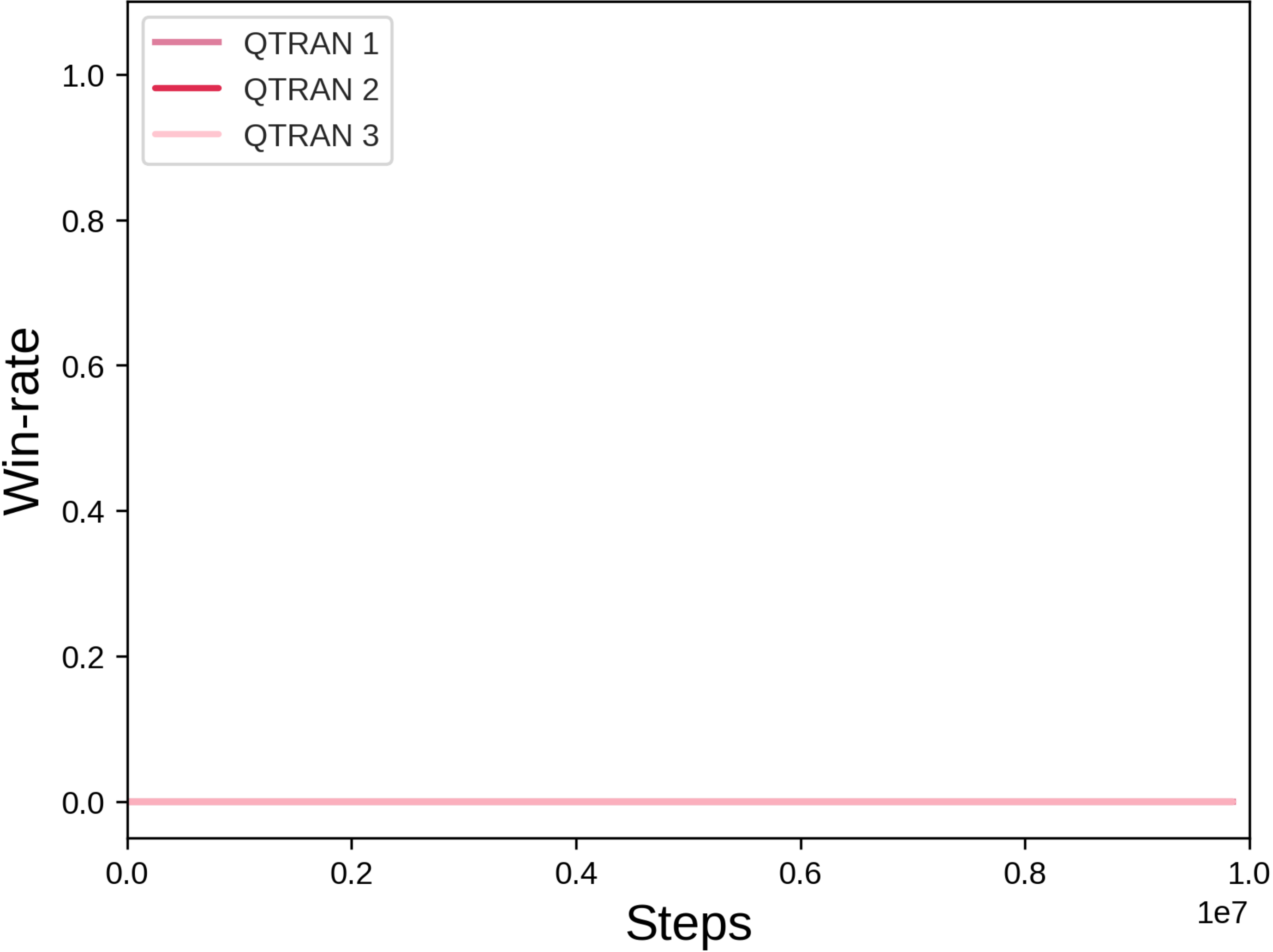}
            \caption{Offense near}
            \label{fig:app_qtran_parallel_off_near}
        \end{subfigure}%
        \begin{subfigure}{0.26\columnwidth}
            \includegraphics[width=\columnwidth]{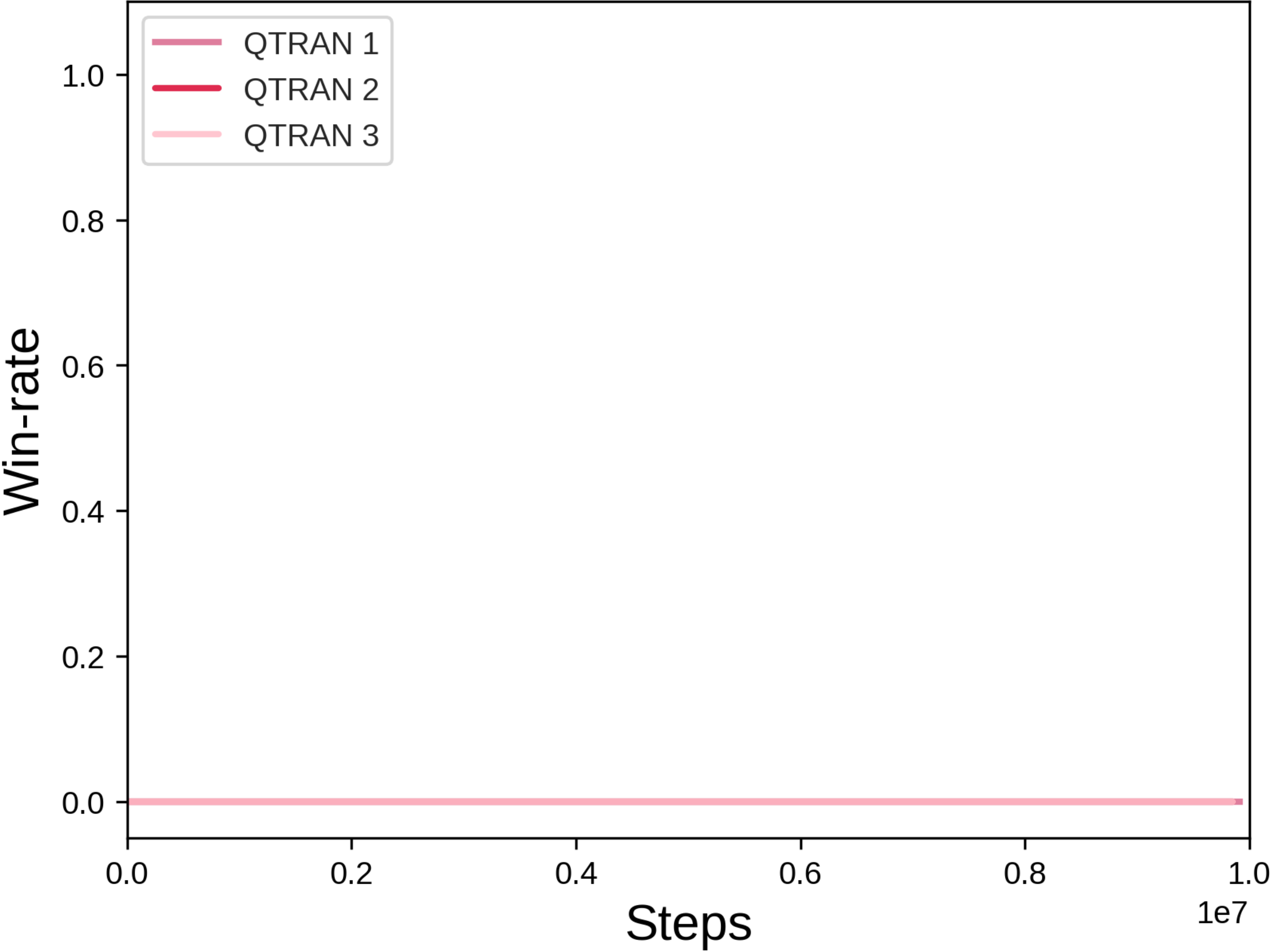}
            \caption{Offense distant}
            \label{fig:app_qtran_parallel_off_dist}
        \end{subfigure}%
        \begin{subfigure}{0.26\columnwidth}
            \includegraphics[width=\columnwidth]{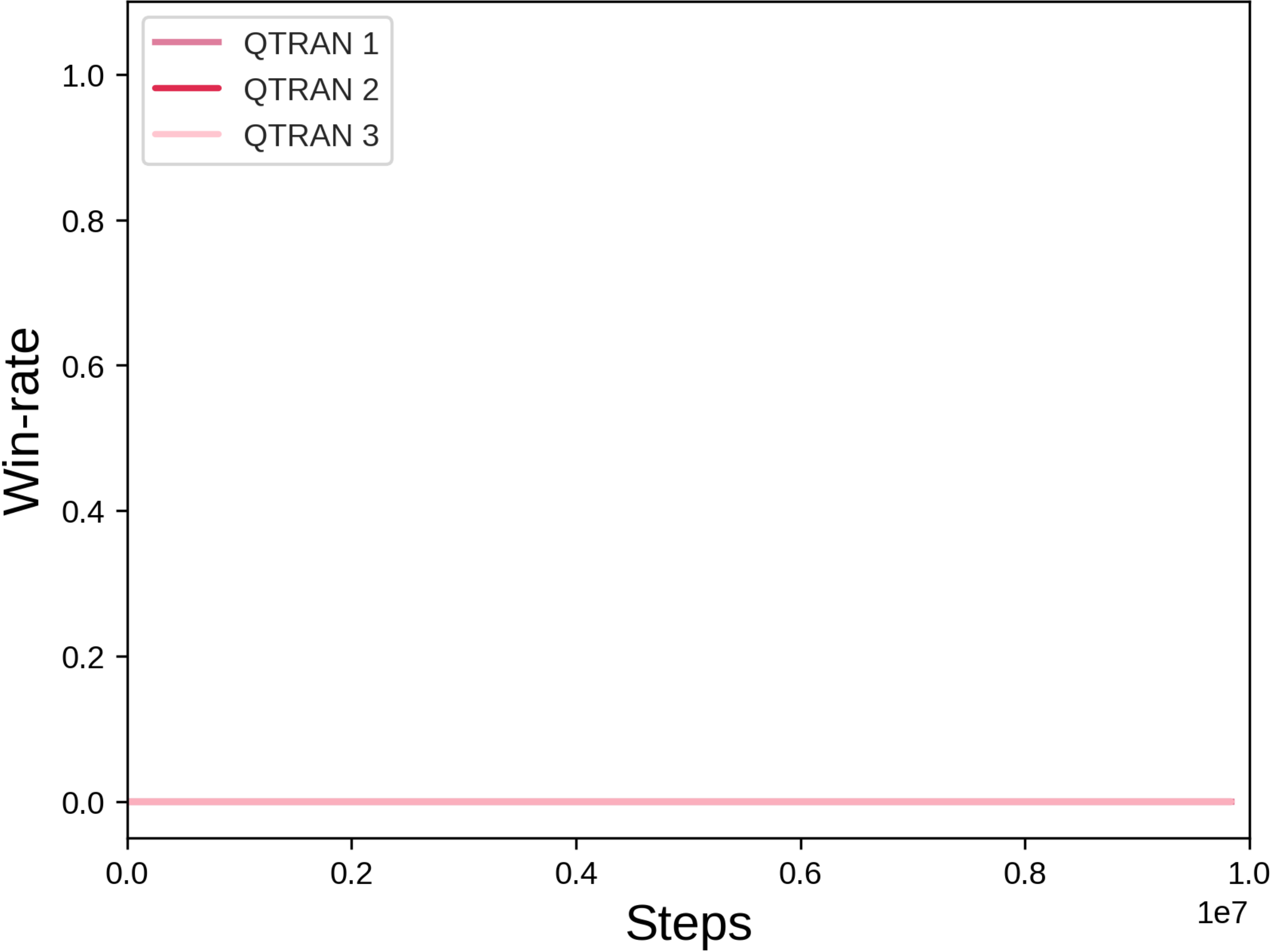}
            \caption{Offense complicated}
            \label{fig:app_qtran_parallel_off_com}
        \end{subfigure}%
        
        \begin{subfigure}{0.27\columnwidth}
            \includegraphics[width=\columnwidth]{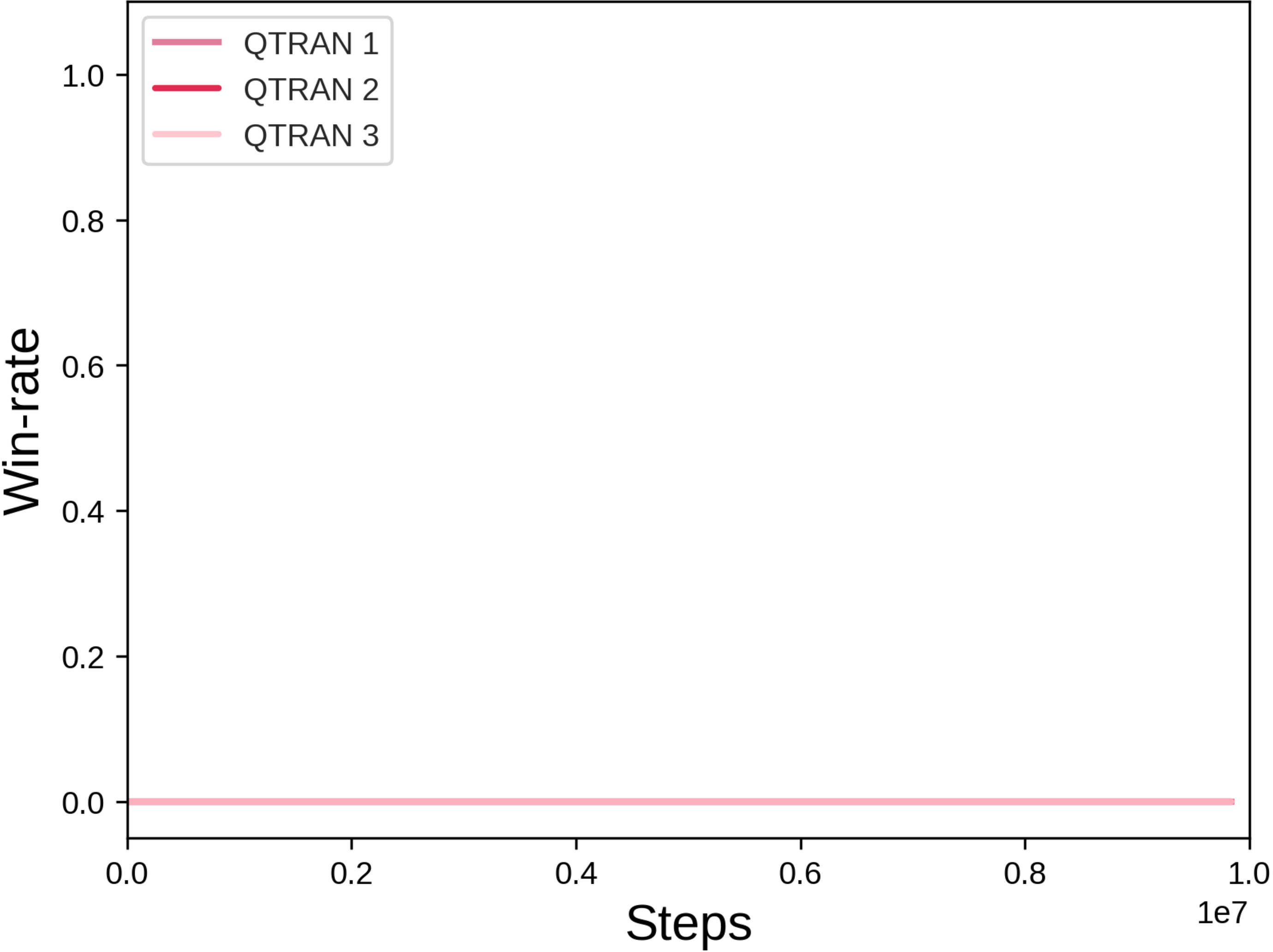}
            \caption{Offense hard}
            \label{fig:app_qtran_parallel_off_hard}
        \end{subfigure}%
        \begin{subfigure}{0.27\columnwidth}
            \includegraphics[width=\columnwidth]{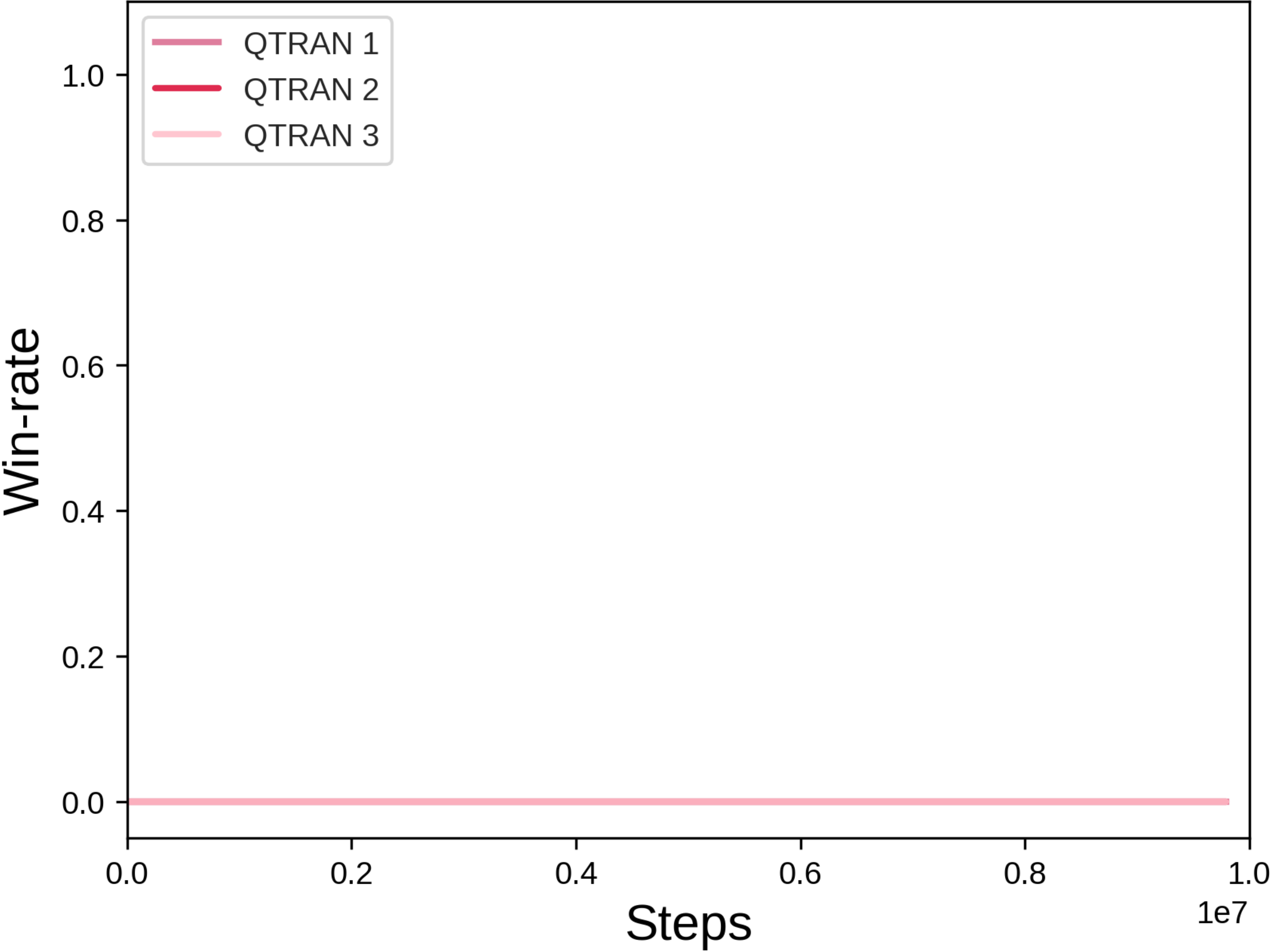}
            \caption{Offense superhard}
            \label{fig:app_qtran_parallel_off_super}
        \end{subfigure}%
    \caption{QTRAN trained on the parallel episodic buffer}
    \label{fig:app_qtran_parallel}
}
\end{figure}

\begin{figure}[!ht]{
    \centering
        \begin{subfigure}{0.26\columnwidth}
            \includegraphics[width=\columnwidth]{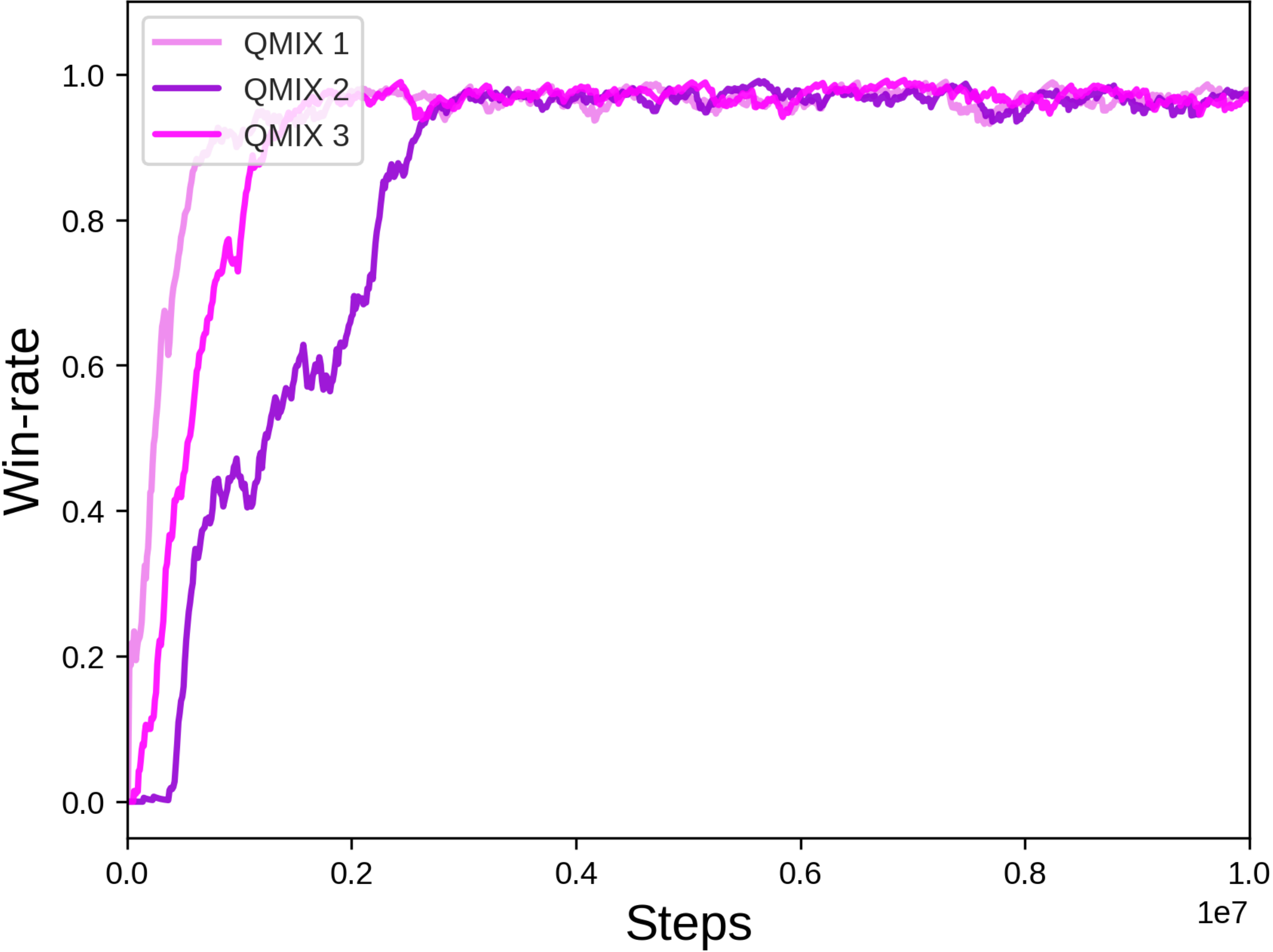}
            \caption{Defense infantry}
            \label{fig:app_qmix_parallel_def_inf}
        \end{subfigure}%
        \begin{subfigure}{0.26\columnwidth}
            \includegraphics[width=\columnwidth]{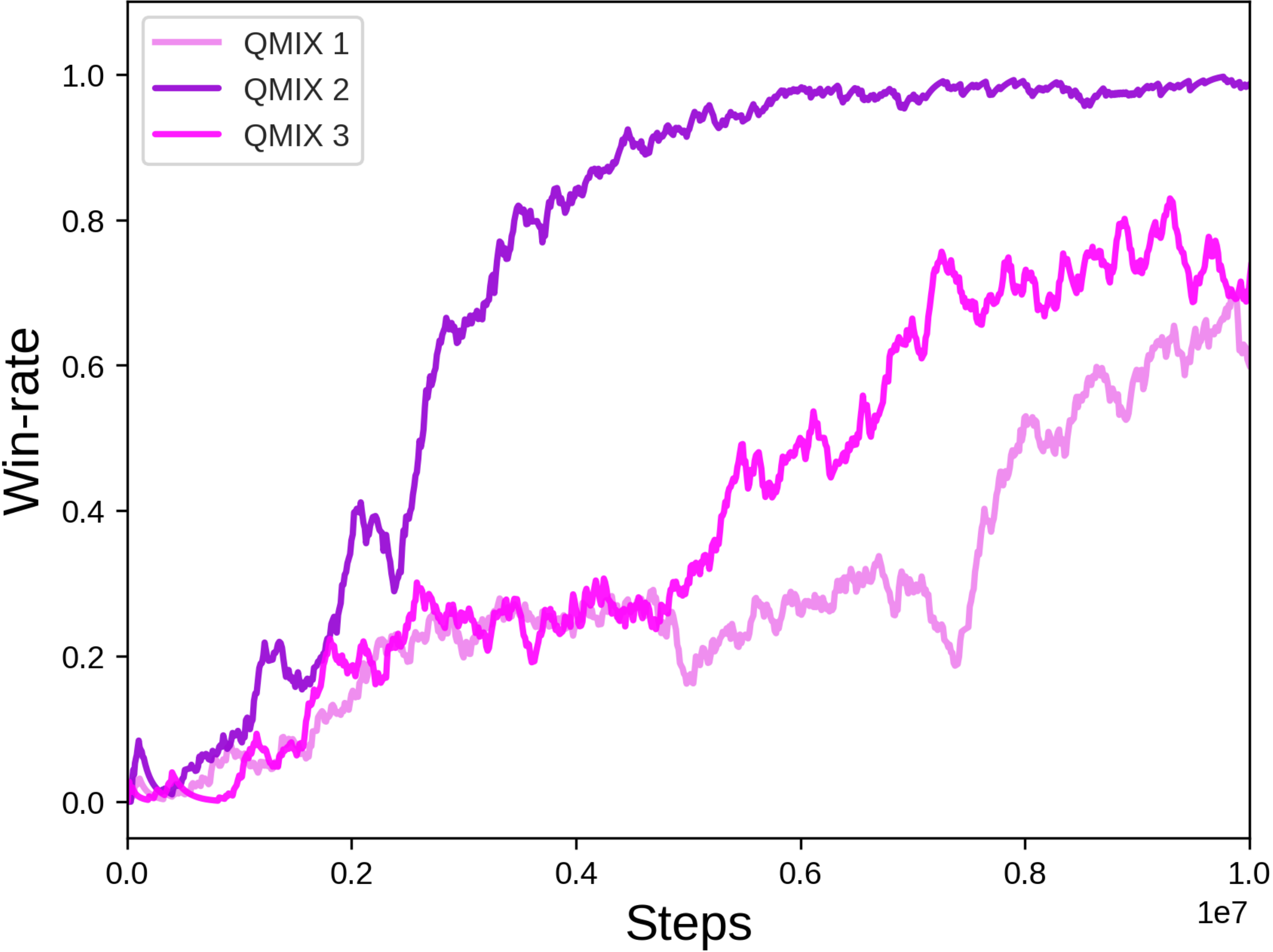}
            \caption{Defense armored}
            \label{fig:app_qmix_parallel_def_arm}
        \end{subfigure}%
        \begin{subfigure}{0.26\columnwidth}
            \includegraphics[width=\columnwidth]{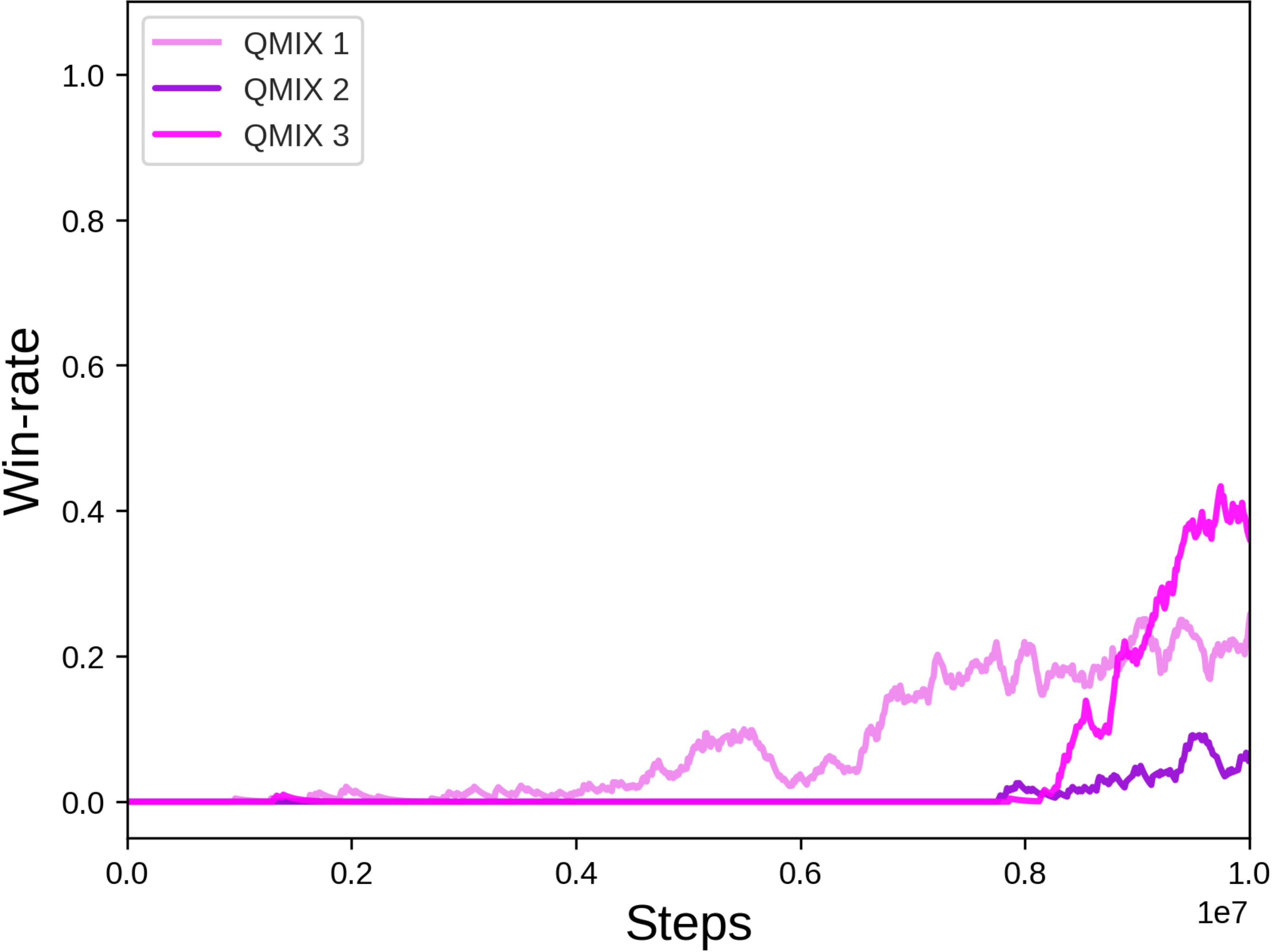}
            \caption{Defense outnumbered}
            \label{fig:app_qmix_parallel_def_out}
        \end{subfigure}%
        
        \begin{subfigure}{0.26\columnwidth}
            \includegraphics[width=\columnwidth]{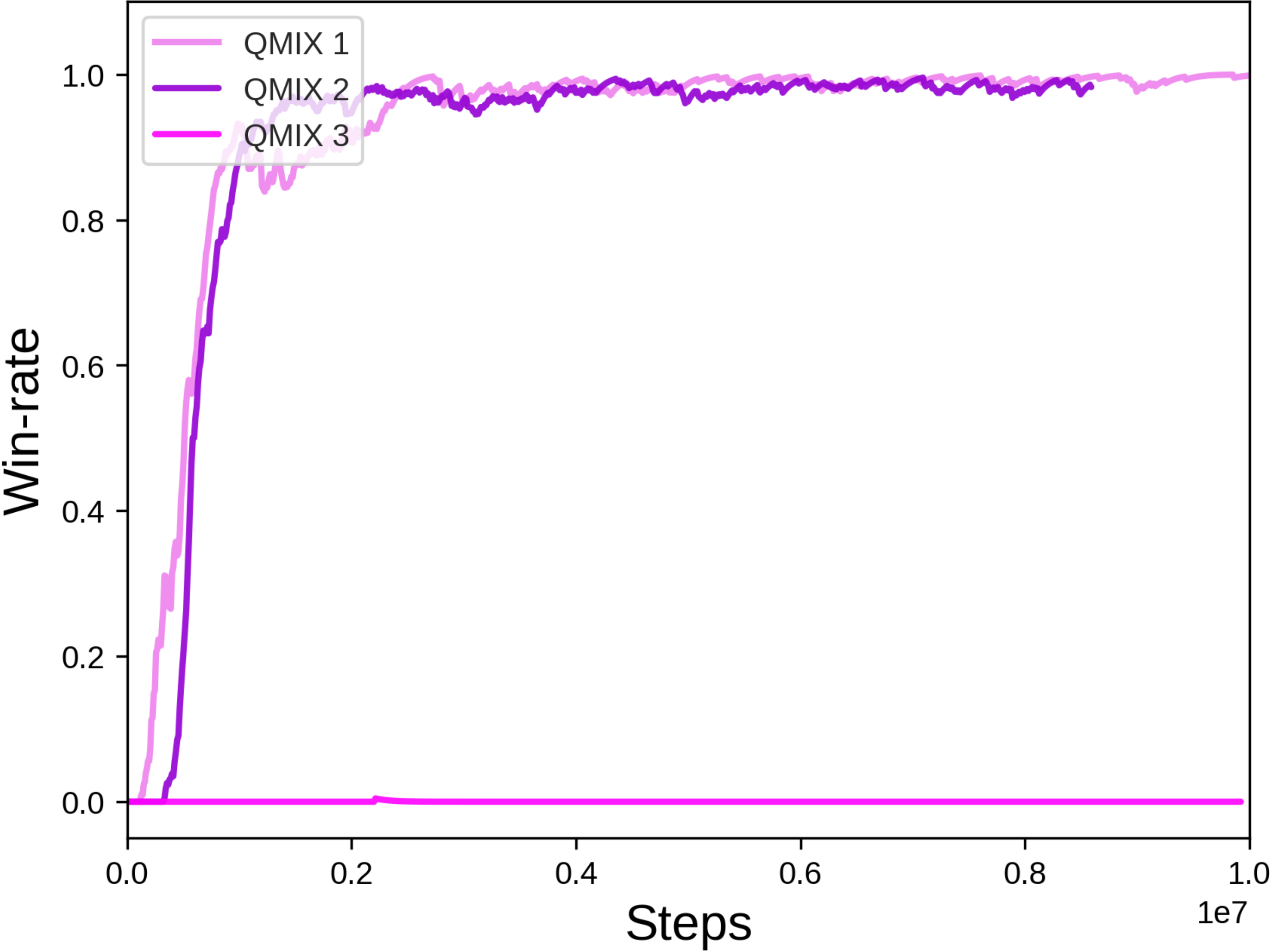}
            \caption{Offense near}
            \label{fig:app_qmix_parallel_off_near}
        \end{subfigure}%
        \begin{subfigure}{0.26\columnwidth}
            \includegraphics[width=\columnwidth]{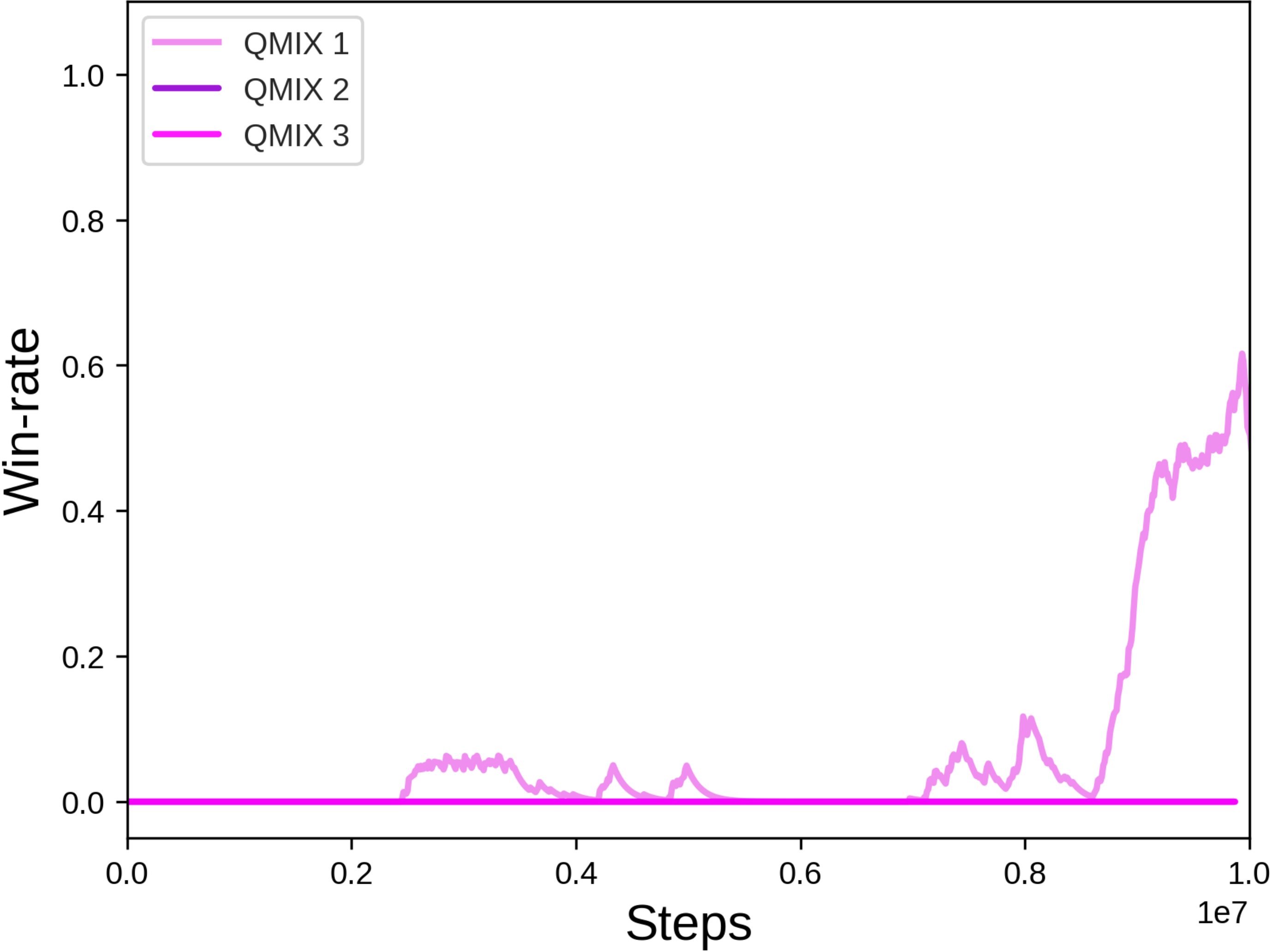}
            \caption{Offense distant}
            \label{fig:app_qmix_parallel_off_dist}
        \end{subfigure}%
        \begin{subfigure}{0.26\columnwidth}
            \includegraphics[width=\columnwidth]{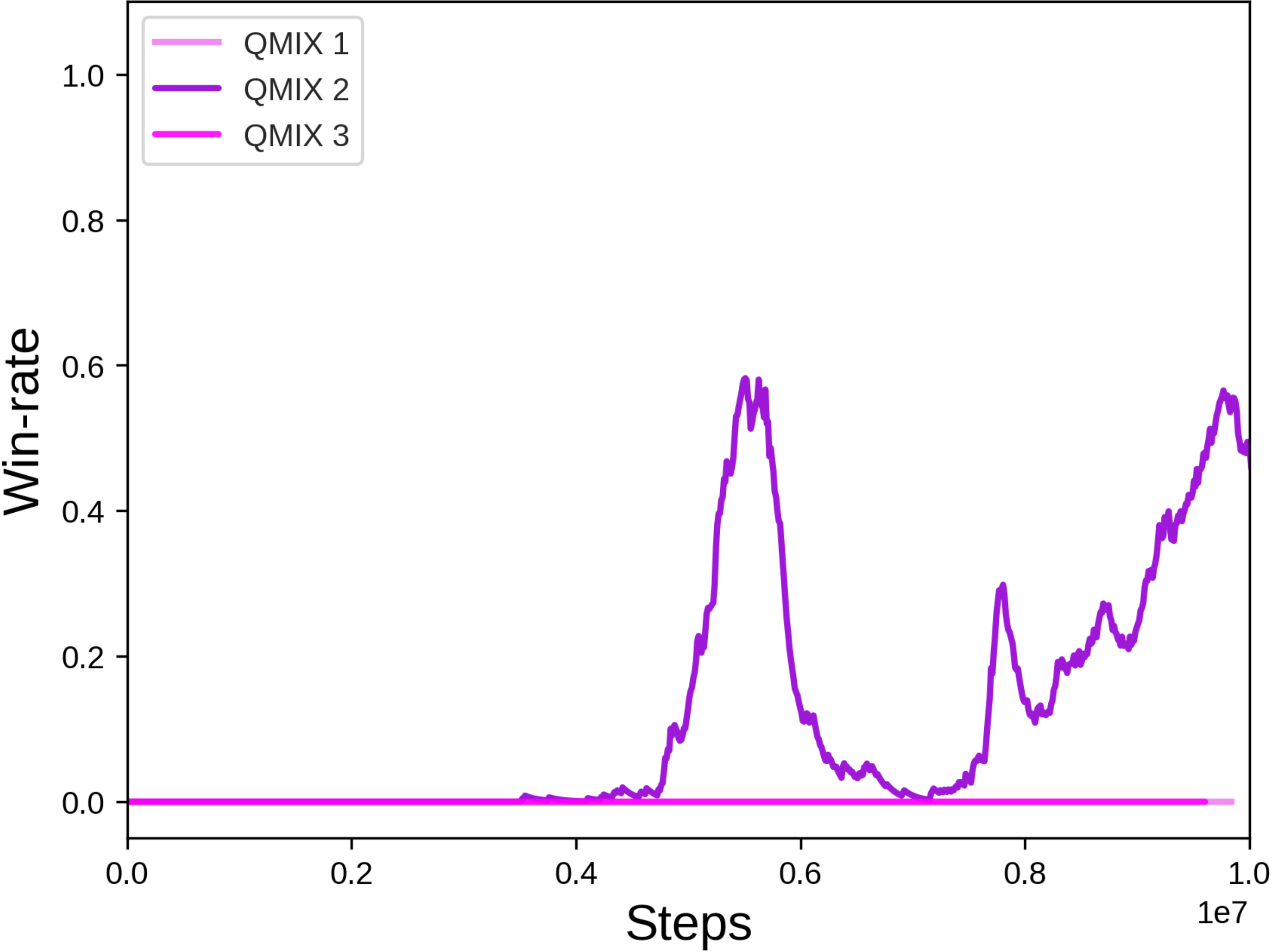}
            \caption{Offense complicated}
            \label{fig:app_qmix_parallel_off_com}
        \end{subfigure}%
        
        \begin{subfigure}{0.27\columnwidth}
            \includegraphics[width=\columnwidth]{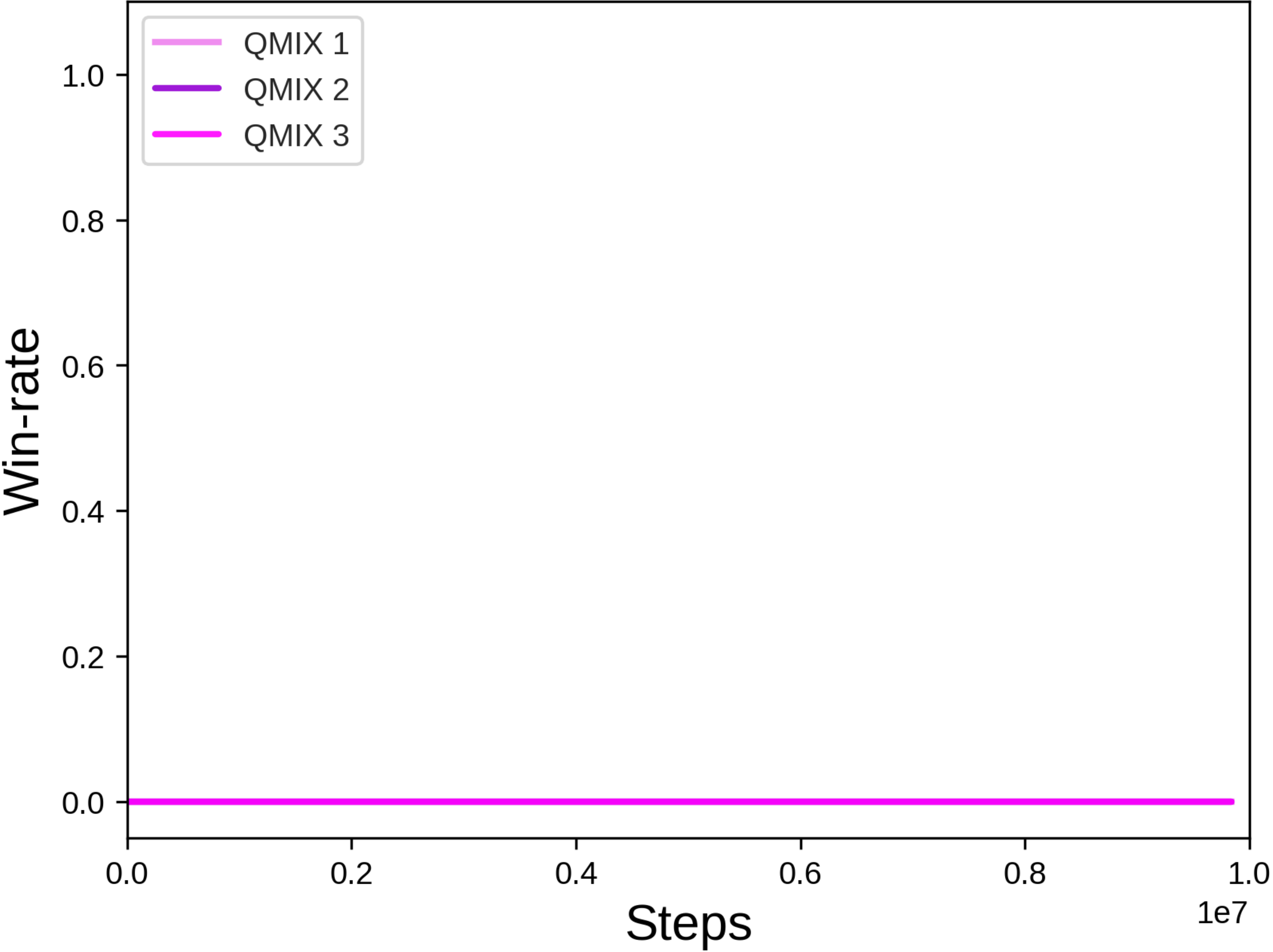}
            \caption{Offense hard}
            \label{fig:app_qmix_parallel_off_hard}
        \end{subfigure}%
        \begin{subfigure}{0.27\columnwidth}
            \includegraphics[width=\columnwidth]{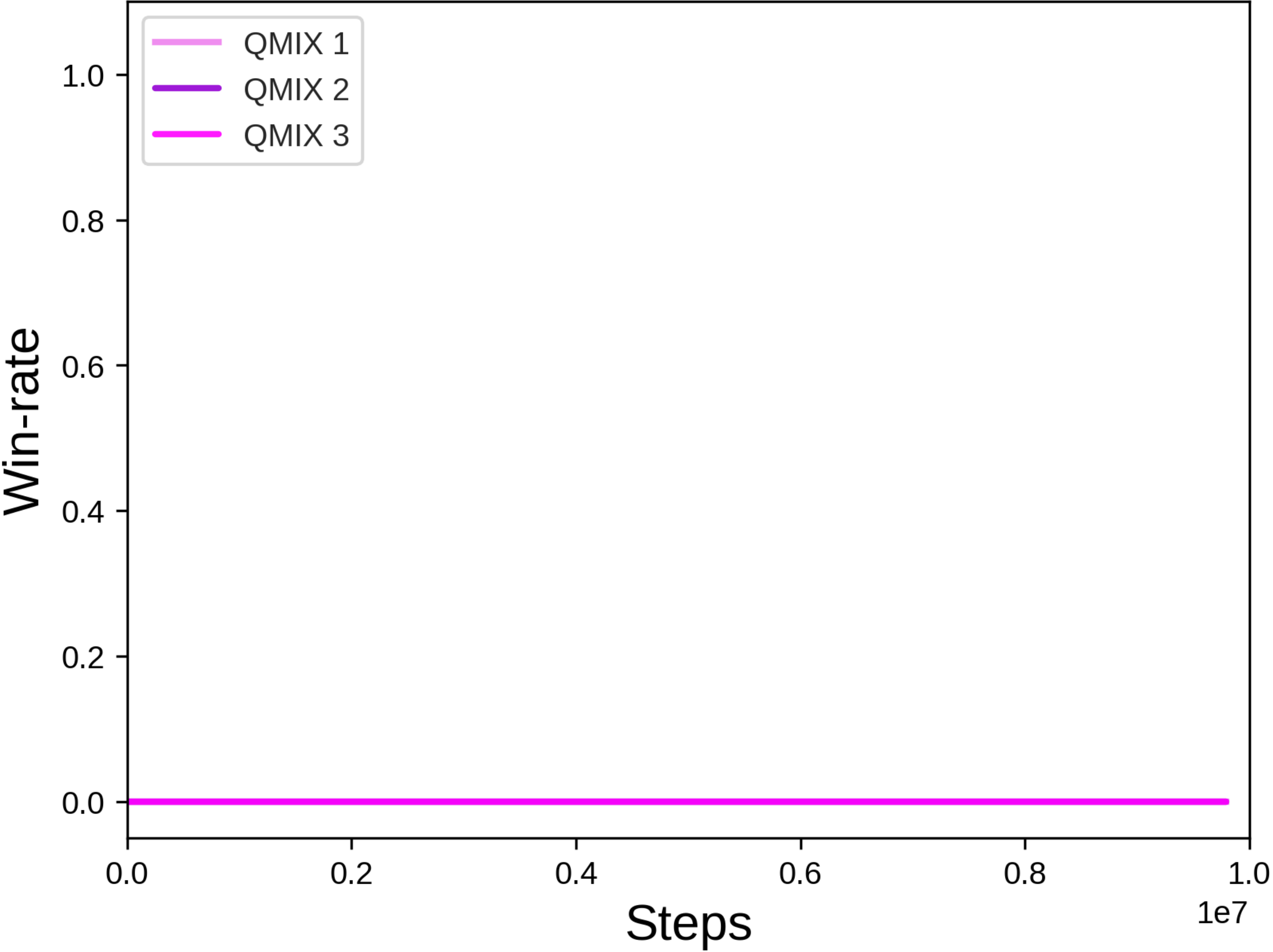}
            \caption{Offense superhard}
            \label{fig:app_qmix_parallel_off_super}
        \end{subfigure}%
    \caption{QMIX trained on the parallel episodic buffer}
    \label{fig:app_qmix_parallel}
}
\end{figure}

\begin{figure}[!ht]{
    \centering
        \begin{subfigure}{0.26\columnwidth}
            \includegraphics[width=\columnwidth]{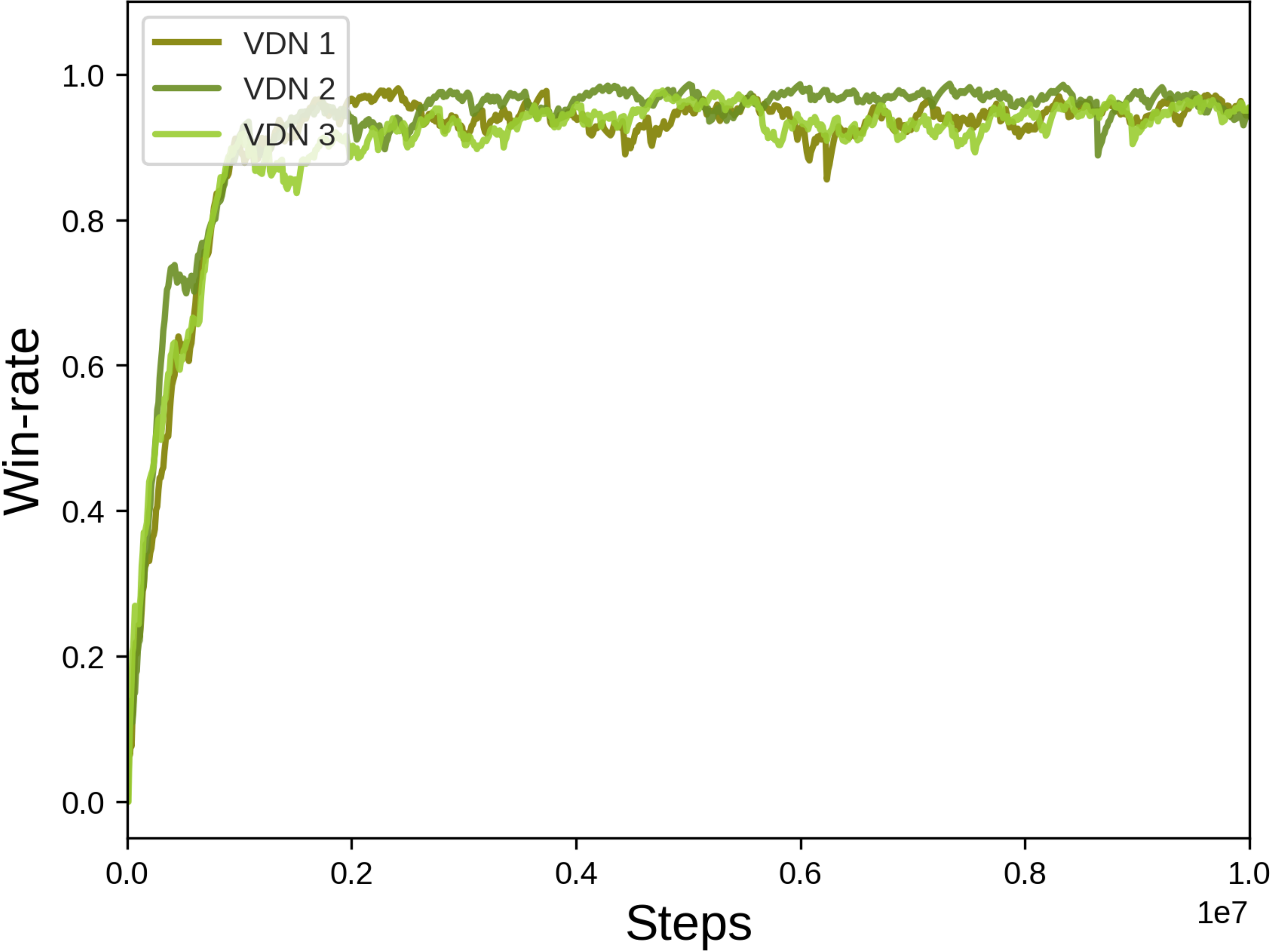}
            \caption{Defense infantry}
            \label{fig:app_vdn_parallel_def_inf}
        \end{subfigure}%
        \begin{subfigure}{0.26\columnwidth}
            \includegraphics[width=\columnwidth]{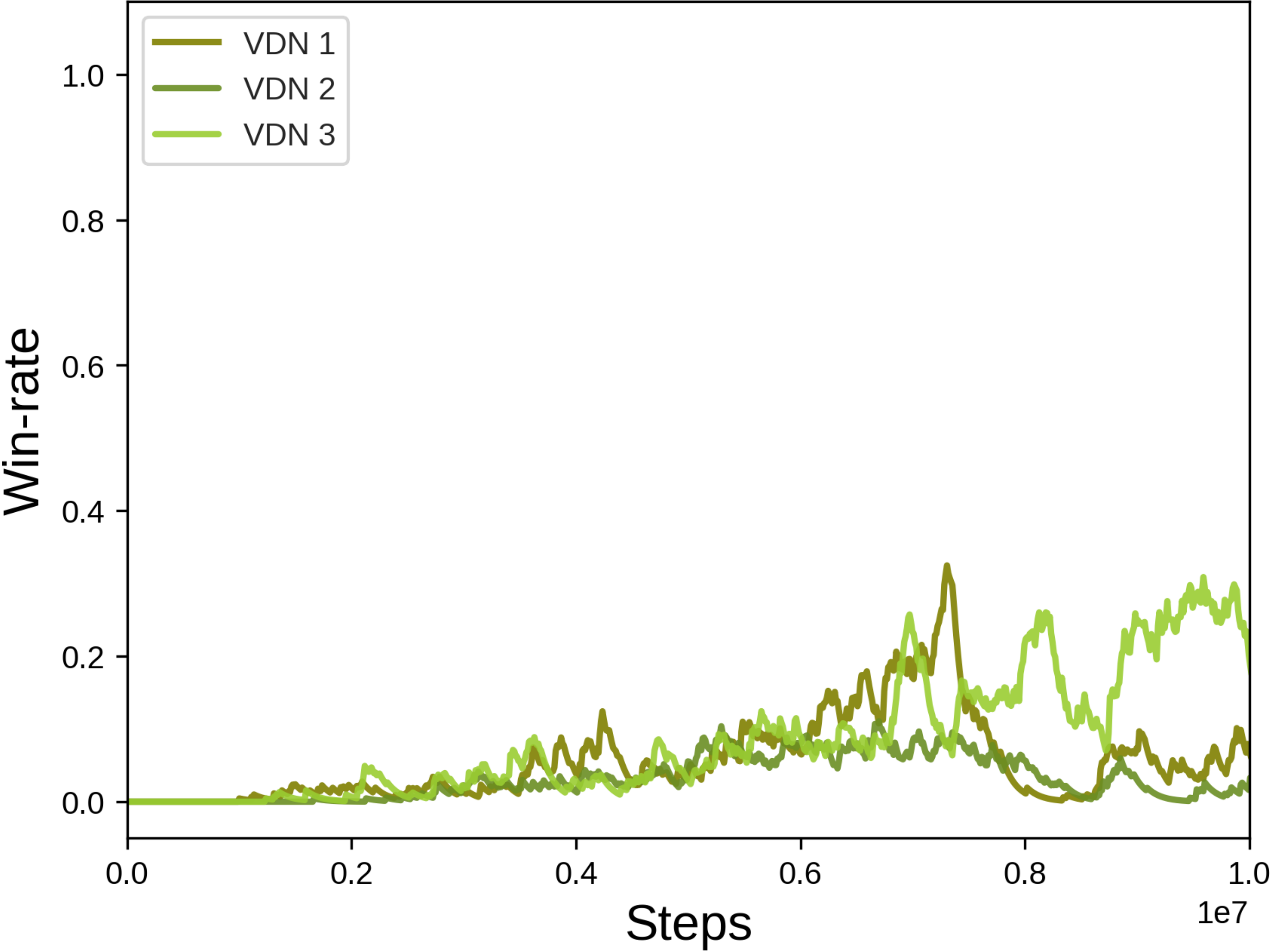}
            \caption{Defense armored}
            \label{fig:app_vdn_parallel_def_arm}
        \end{subfigure}%
        \begin{subfigure}{0.26\columnwidth}
            \includegraphics[width=\columnwidth]{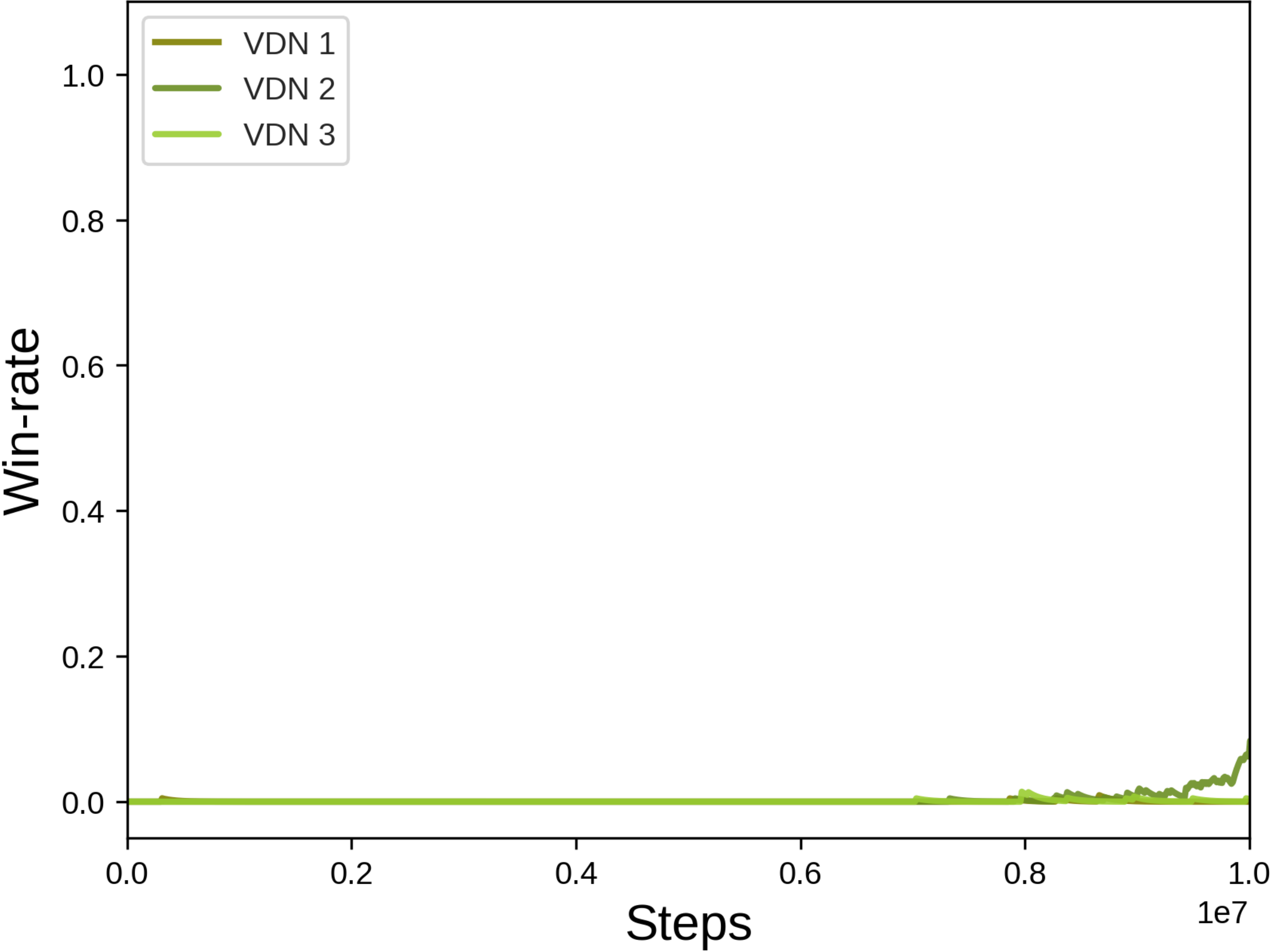}
            \caption{Defense outnumbered}
            \label{fig:app_vdn_parallel_def_out}
        \end{subfigure}%
        
        \begin{subfigure}{0.26\columnwidth}
            \includegraphics[width=\columnwidth]{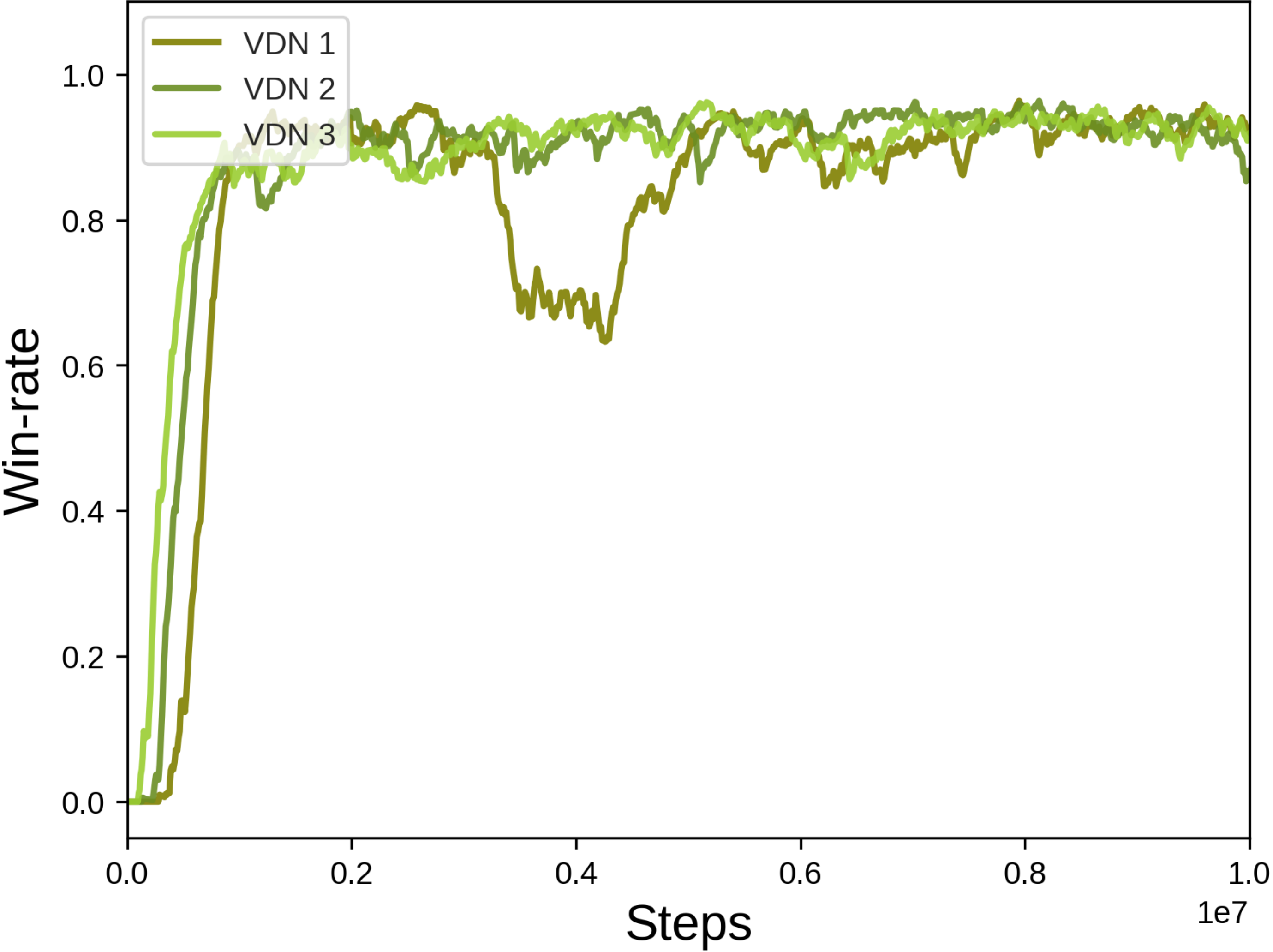}
            \caption{Offense near}
            \label{fig:app_vdn_parallel_off_near}
        \end{subfigure}%
        \begin{subfigure}{0.26\columnwidth}
            \includegraphics[width=\columnwidth]{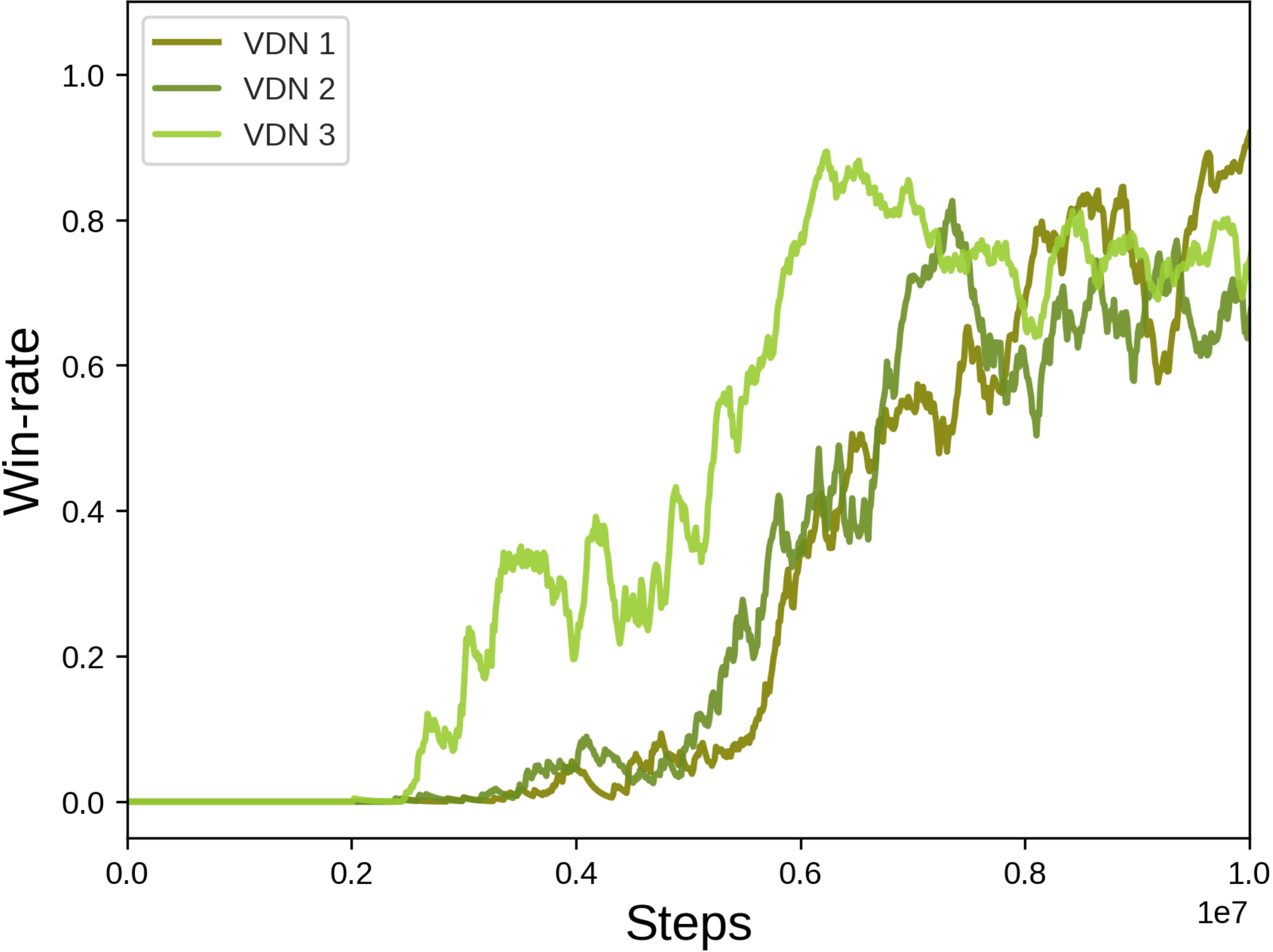}
            \caption{Offense distant}
            \label{fig:app_vdn_parallel_off_dist}
        \end{subfigure}%
        \begin{subfigure}{0.26\columnwidth}
            \includegraphics[width=\columnwidth]{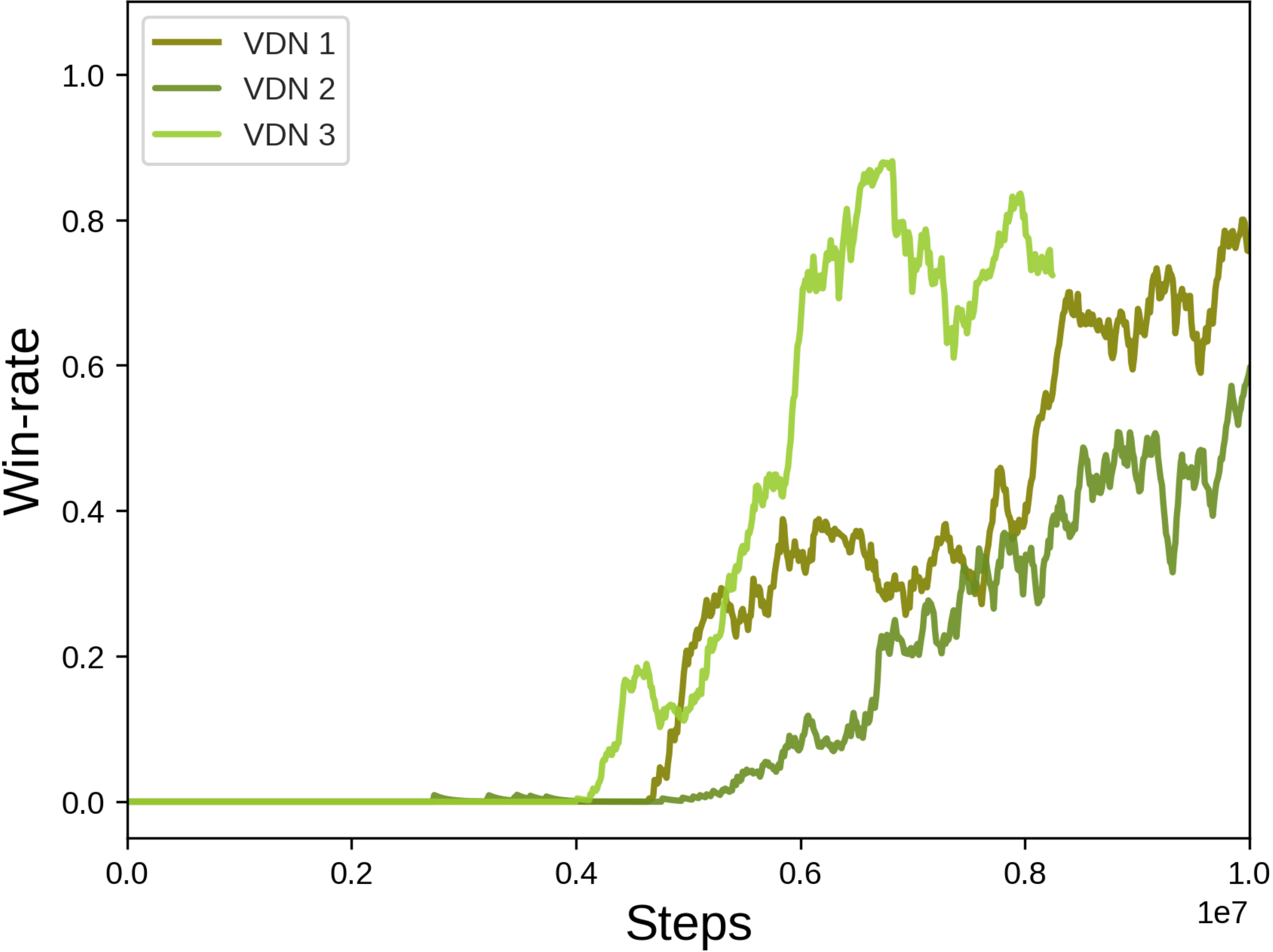}
            \caption{Offense complicated}
            \label{fig:app_vdn_parallel_off_com}
        \end{subfigure}%
        
        \begin{subfigure}{0.27\columnwidth}
            \includegraphics[width=\columnwidth]{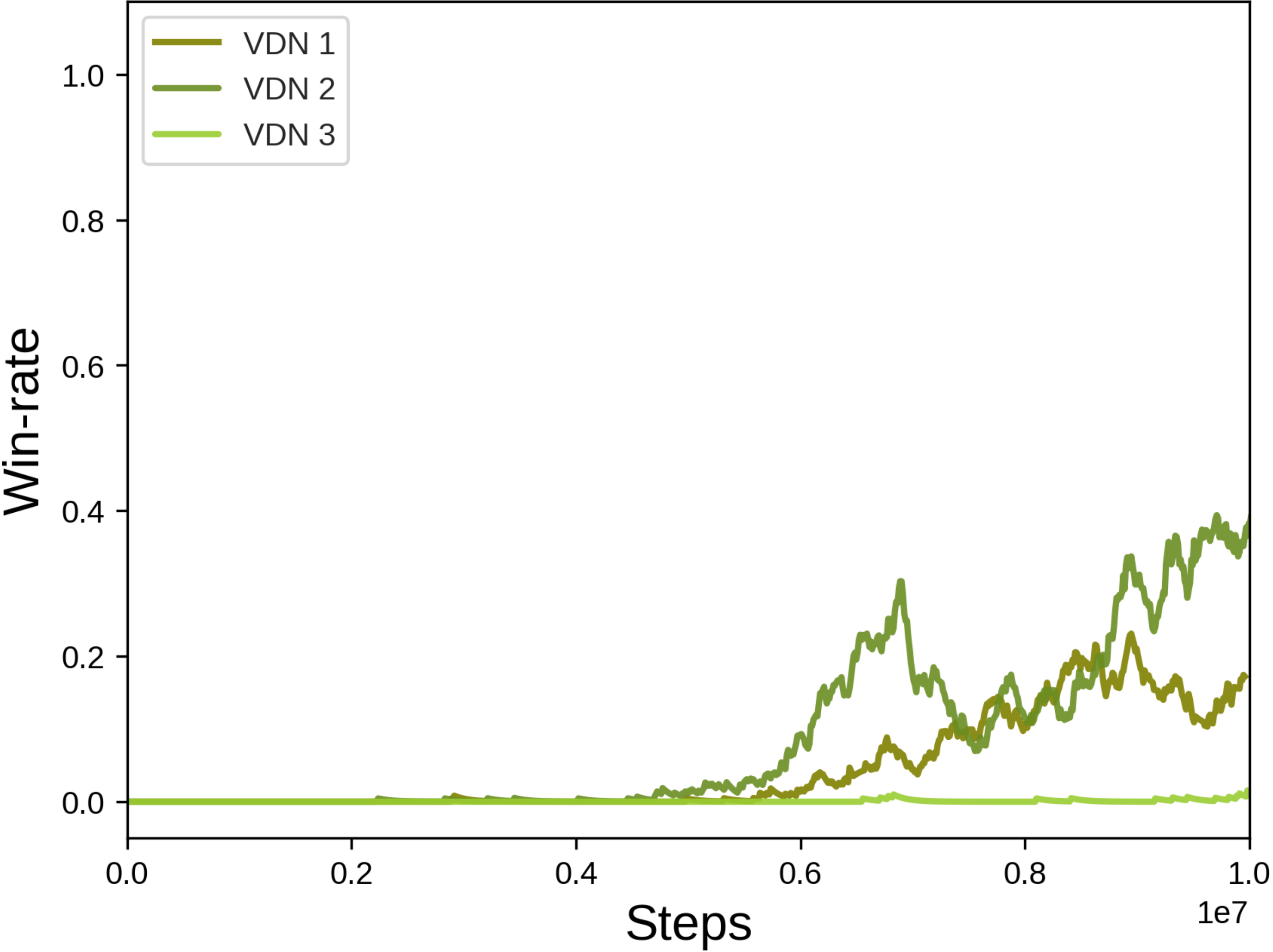}
            \caption{Offense hard}
            \label{fig:app_vdn_parallel_off_hard}
        \end{subfigure}%
        \begin{subfigure}{0.27\columnwidth}
            \includegraphics[width=\columnwidth]{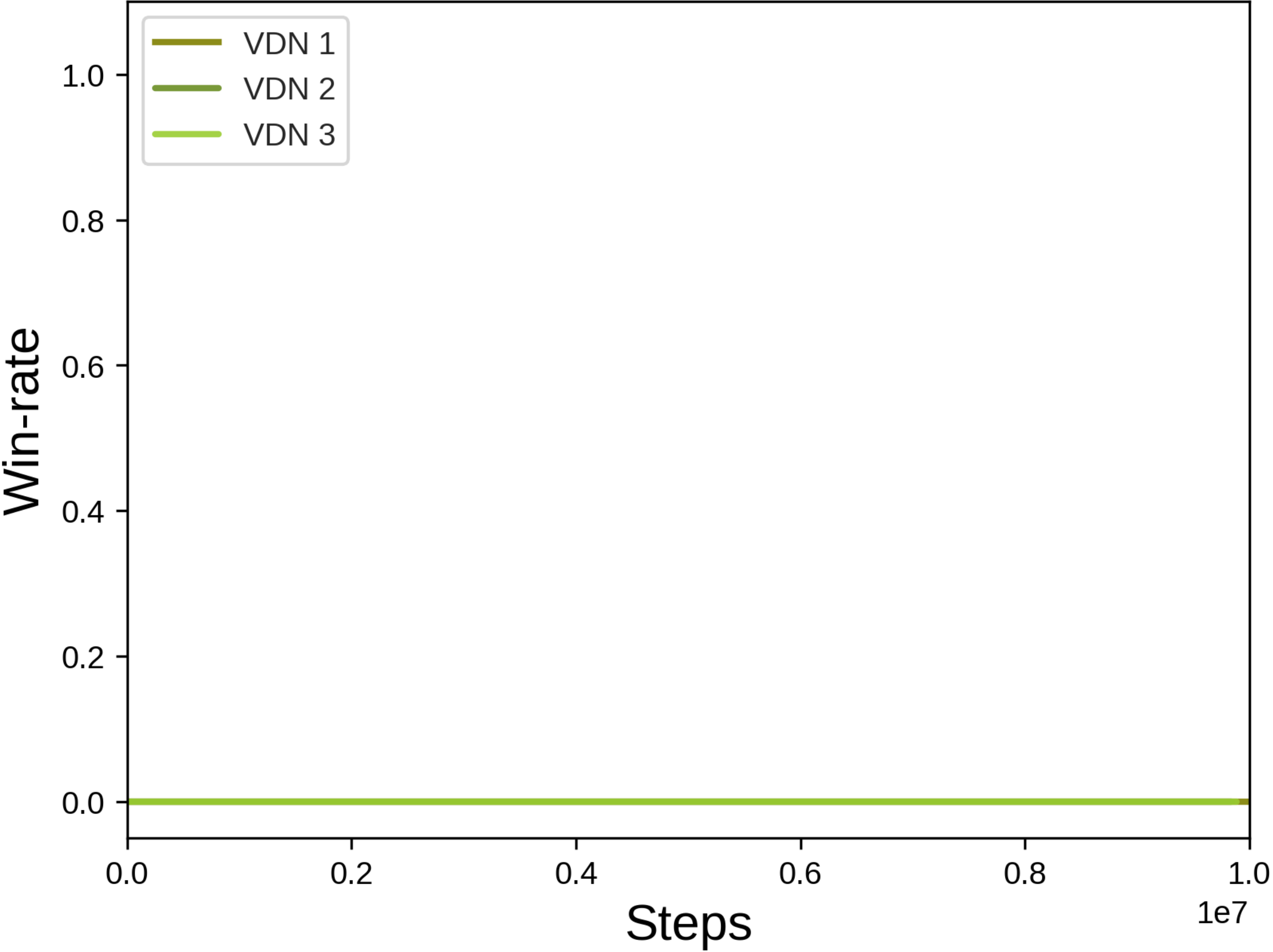}
            \caption{Offense superhard}
            \label{fig:app_vdn_parallel_off_super}
        \end{subfigure}%
    \caption{VDN trained on the parallel episodic buffer}
    \label{fig:app_vdn_parallel}
}
\end{figure}

\begin{figure}[!ht]{
    \centering
        \begin{subfigure}{0.26\columnwidth}
            \includegraphics[width=\columnwidth]{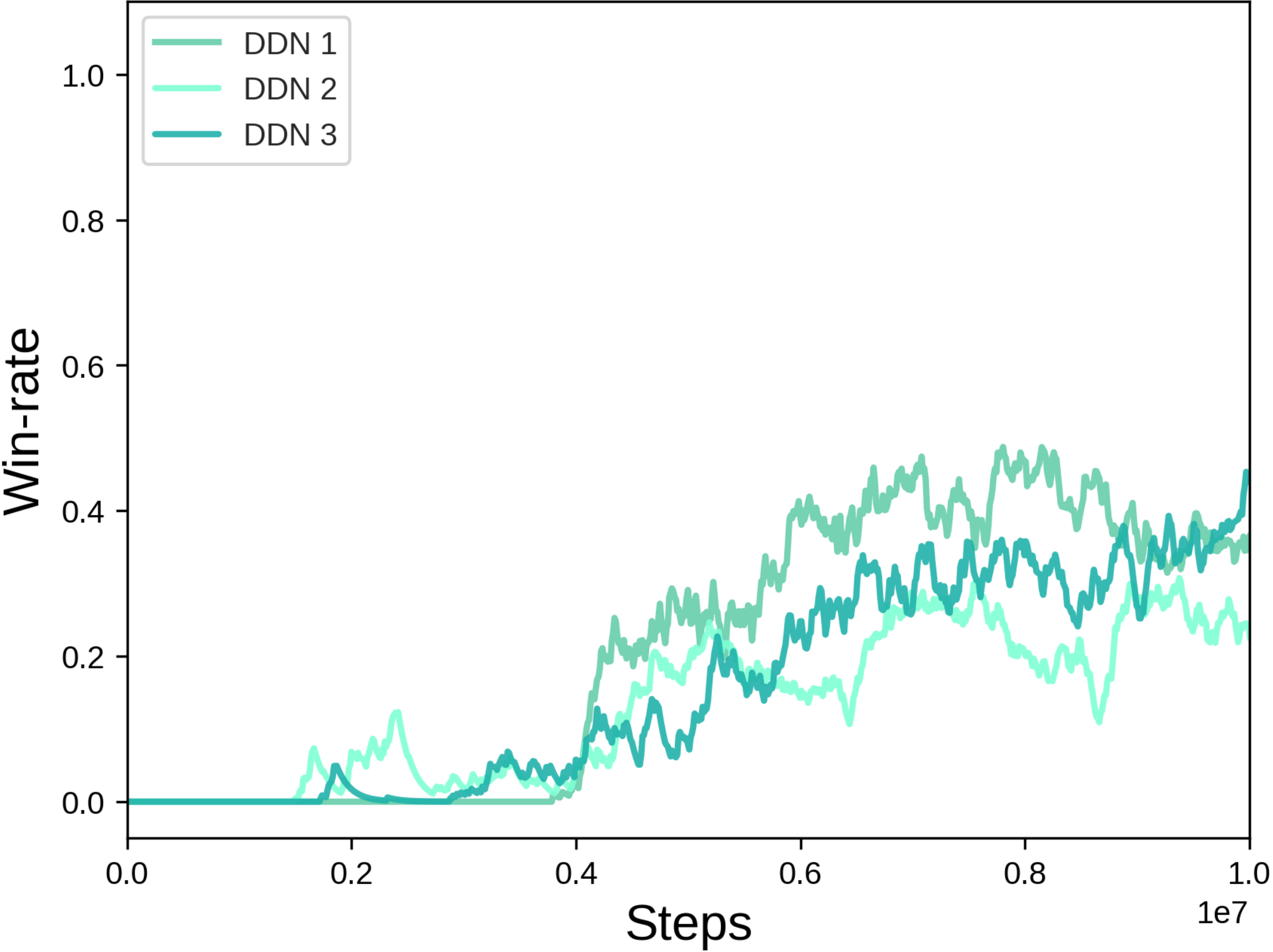}
            \caption{Defense infantry}
            \label{fig:app_ddn_parallel_def_inf}
        \end{subfigure}%
        \begin{subfigure}{0.26\columnwidth}
            \includegraphics[width=\columnwidth]{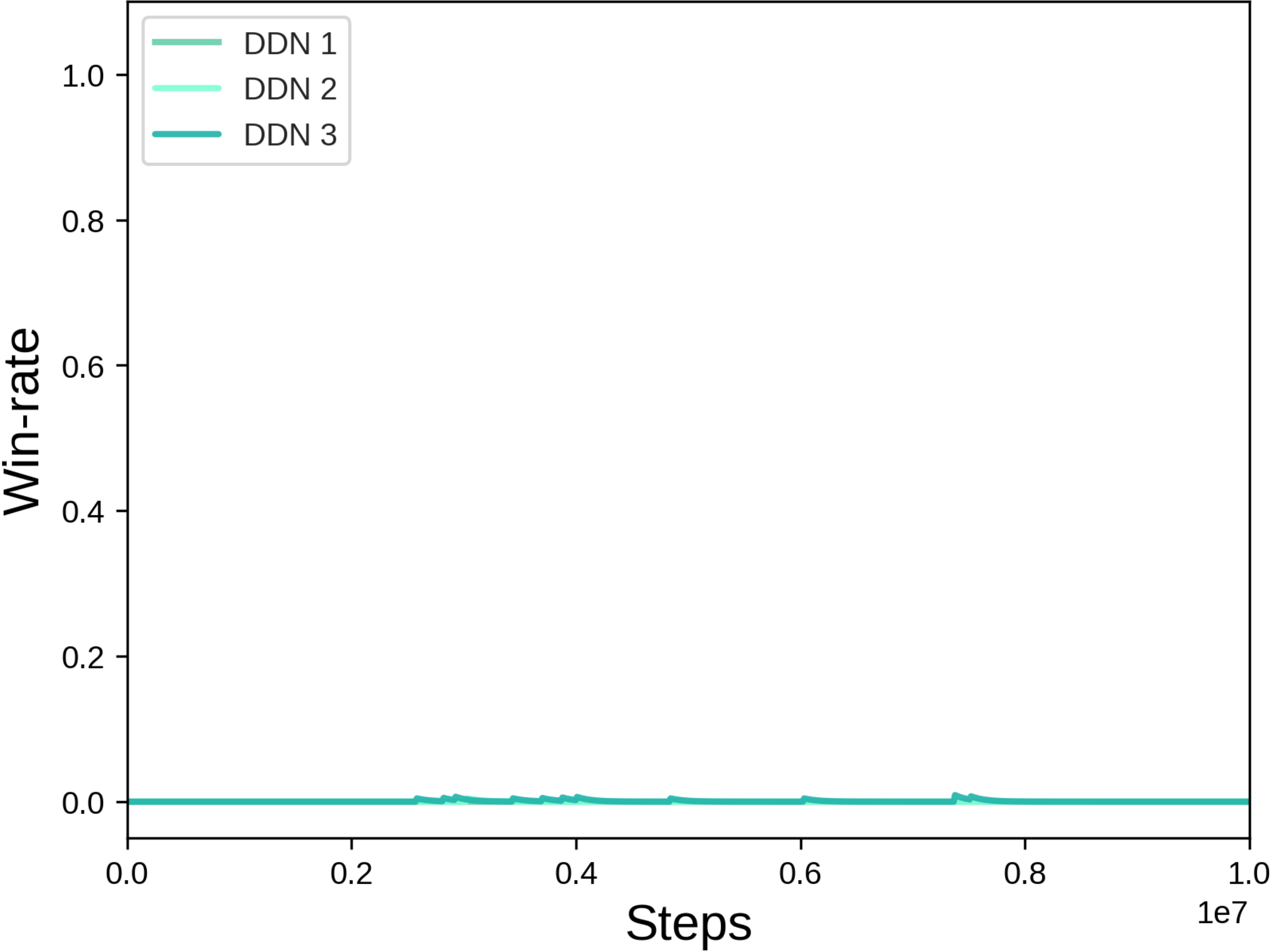}
            \caption{Defense armored}
            \label{fig:app_ddn_parallel_def_arm}
        \end{subfigure}%
        \begin{subfigure}{0.26\columnwidth}
            \includegraphics[width=\columnwidth]{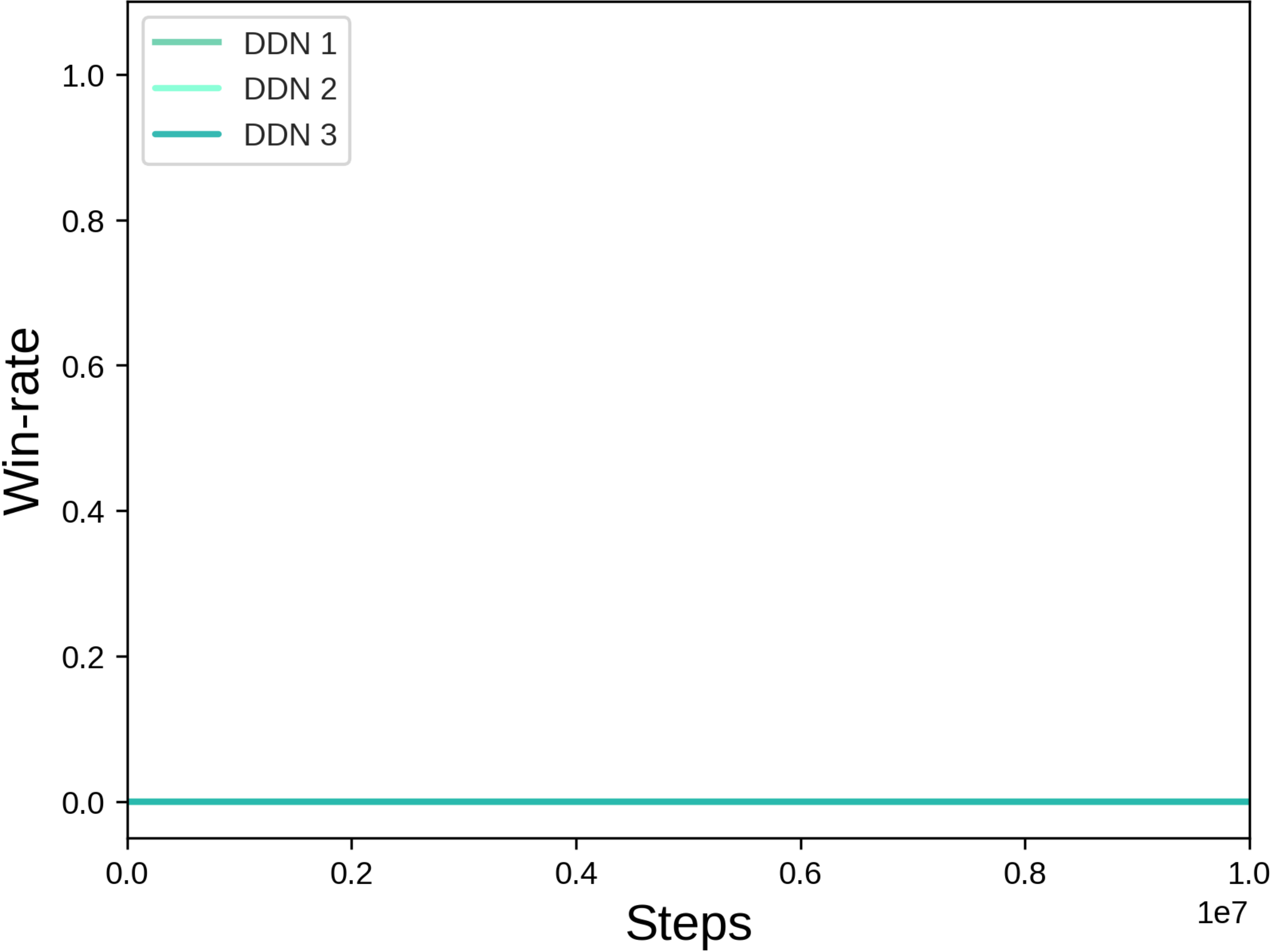}
            \caption{Defense outnumbered}
            \label{fig:app_ddn_parallel_def_out}
        \end{subfigure}%
        
        \begin{subfigure}{0.26\columnwidth}
            \includegraphics[width=\columnwidth]{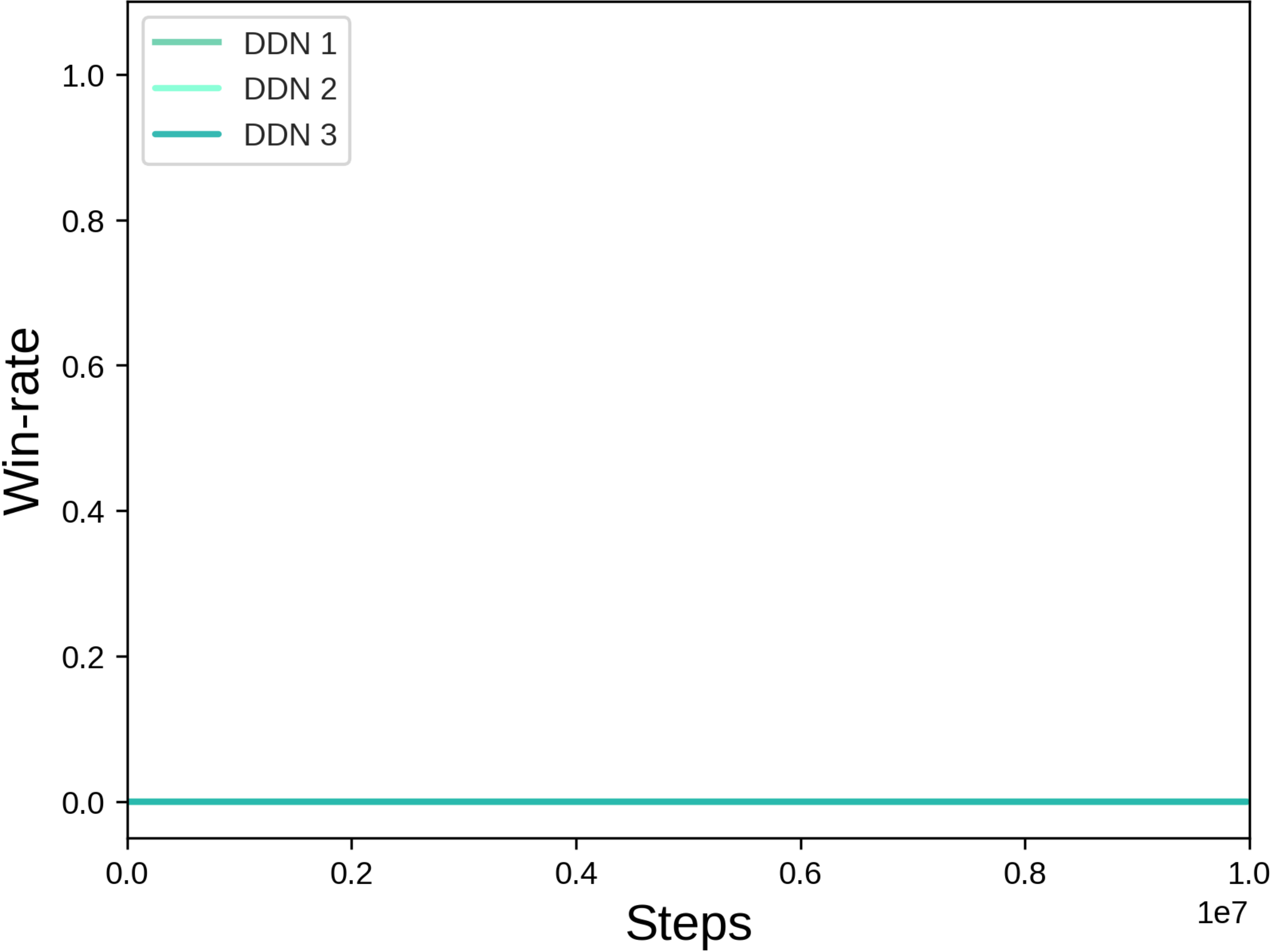}
            \caption{Offense near}
            \label{fig:app_ddn_parallel_off_near}
        \end{subfigure}%
        \begin{subfigure}{0.26\columnwidth}
            \includegraphics[width=\columnwidth]{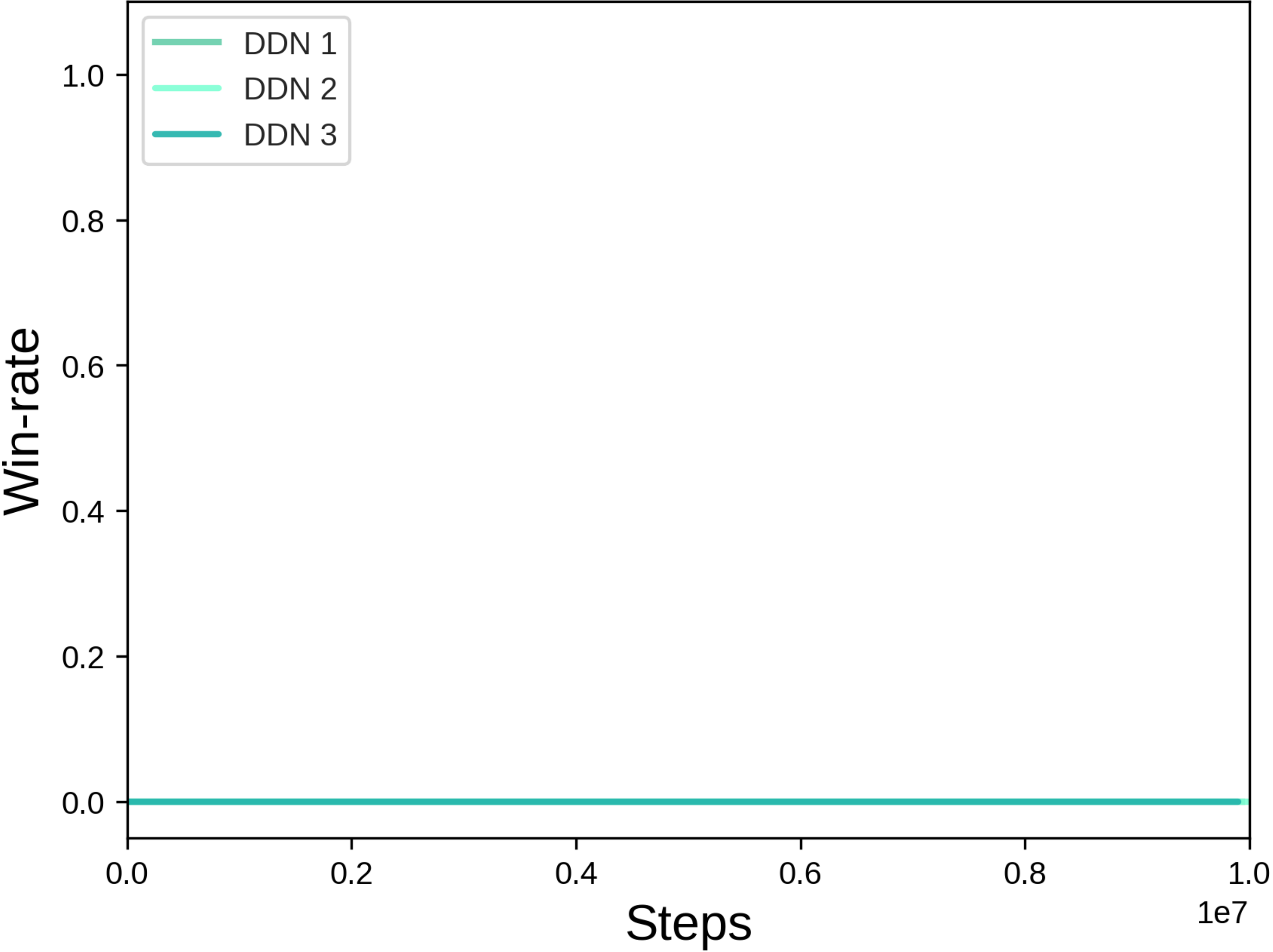}
            \caption{Offense distant}
            \label{fig:app_ddn_parallel_off_dist}
        \end{subfigure}%
        \begin{subfigure}{0.26\columnwidth}
            \includegraphics[width=\columnwidth]{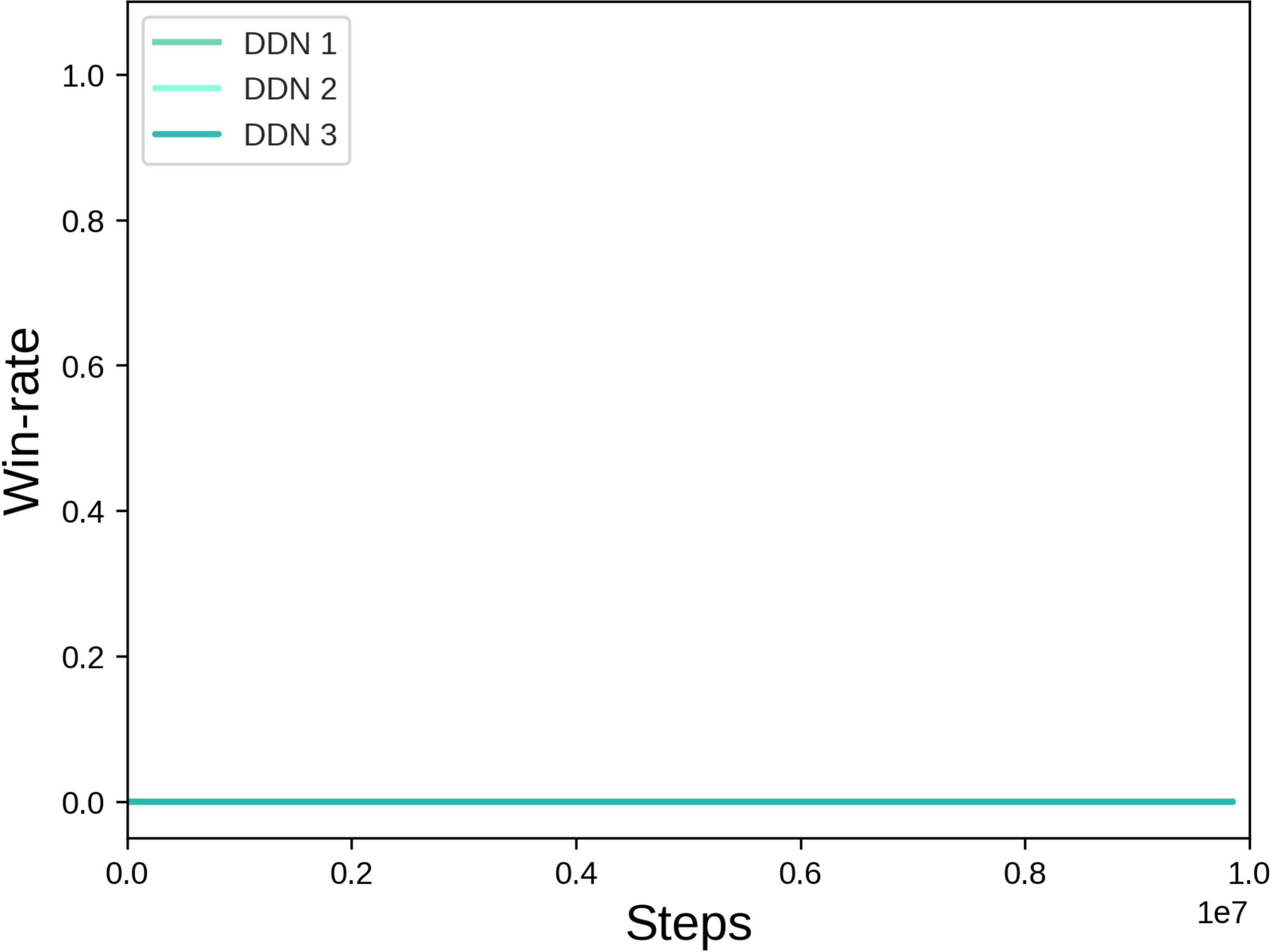}
            \caption{Offense complicated}
            \label{fig:app_ddn_parallel_off_com}
        \end{subfigure}%
        
        \begin{subfigure}{0.27\columnwidth}
            \includegraphics[width=\columnwidth]{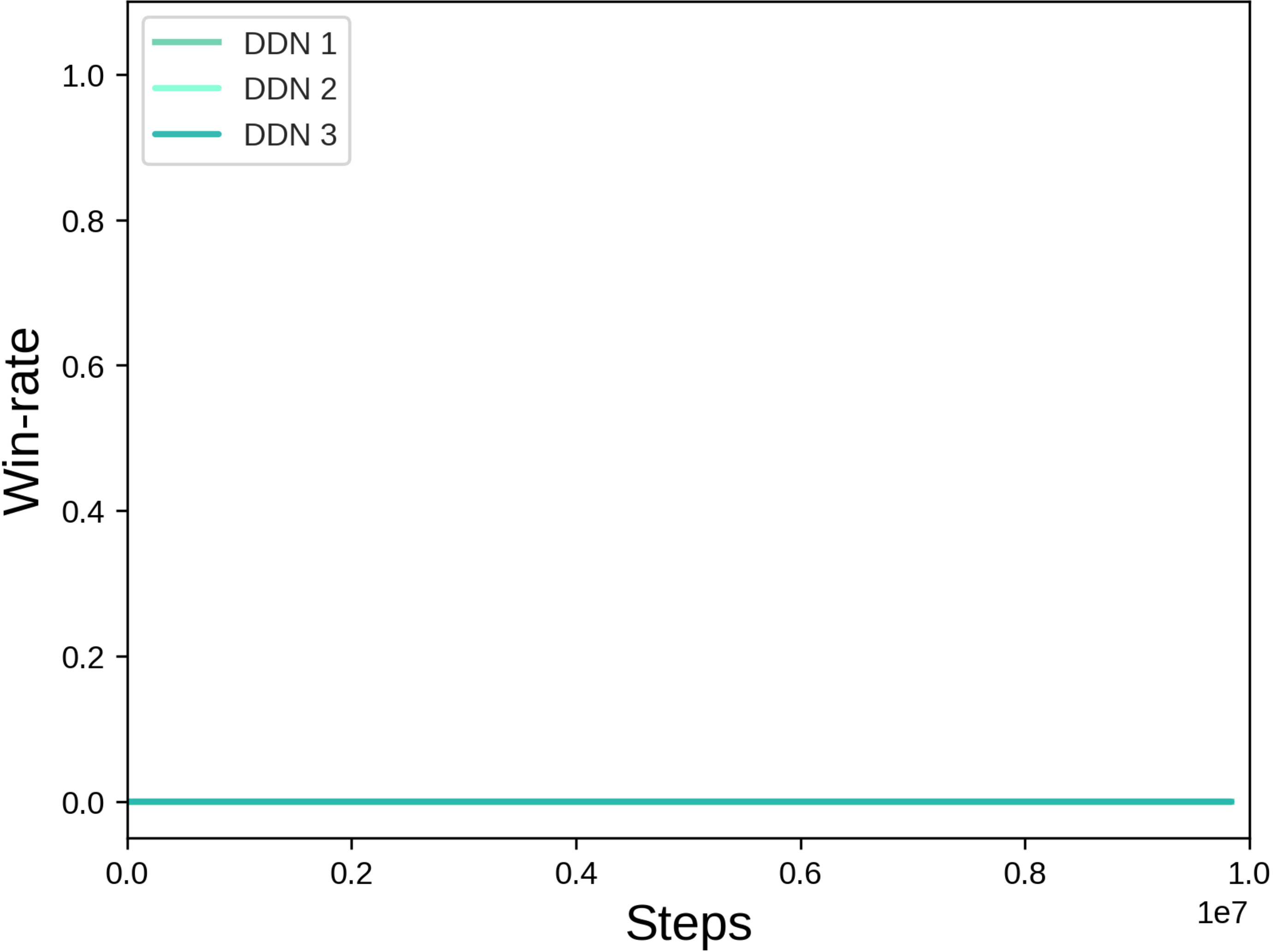}
            \caption{Offense hard}
            \label{fig:app_ddn_parallel_off_hard}
        \end{subfigure}%
        \begin{subfigure}{0.27\columnwidth}
            \includegraphics[width=\columnwidth]{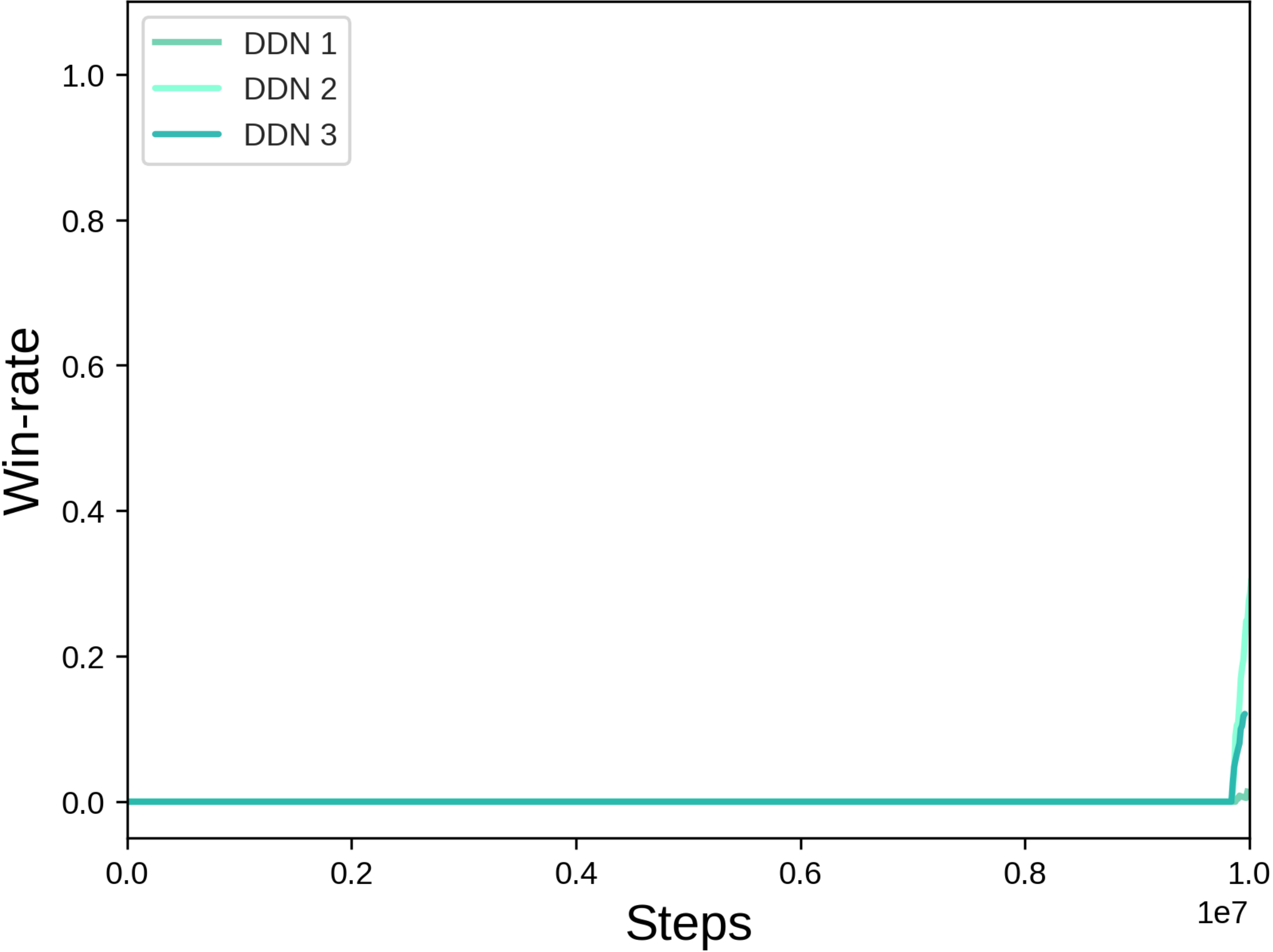}
            \caption{Offense superhard}
            \label{fig:app_ddn_parallel_off_super}
        \end{subfigure}%
    \caption{DDN trained on the parallel episodic buffer}
    \label{fig:app_ddn_parallel}
}
\end{figure}

\begin{figure}[!ht]{
    \centering
        \begin{subfigure}{0.26\columnwidth}
            \includegraphics[width=\columnwidth]{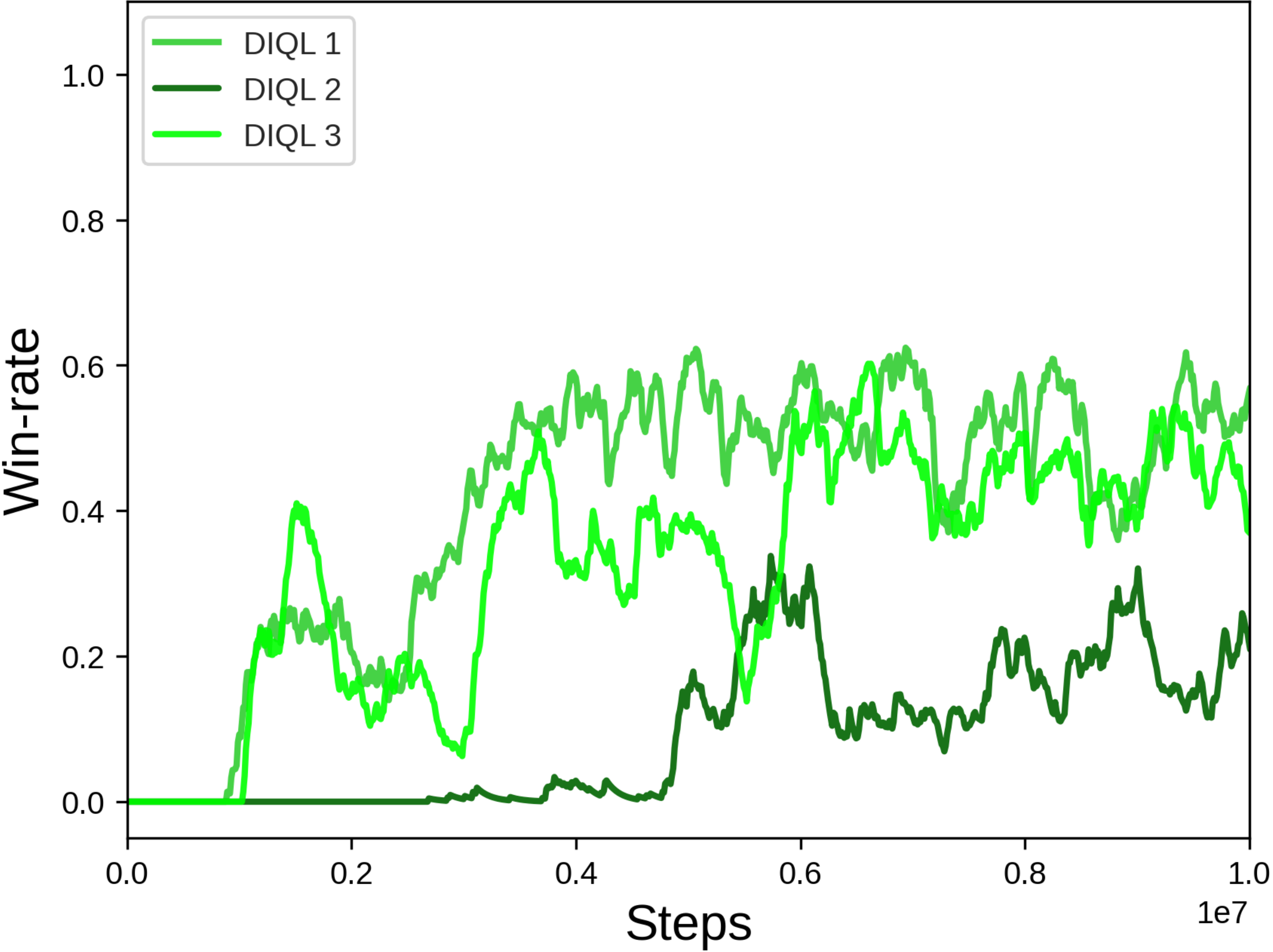}
            \caption{Defense infantry}
            \label{fig:app_diql_parallel_def_inf}
        \end{subfigure}%
        \begin{subfigure}{0.26\columnwidth}
            \includegraphics[width=\columnwidth]{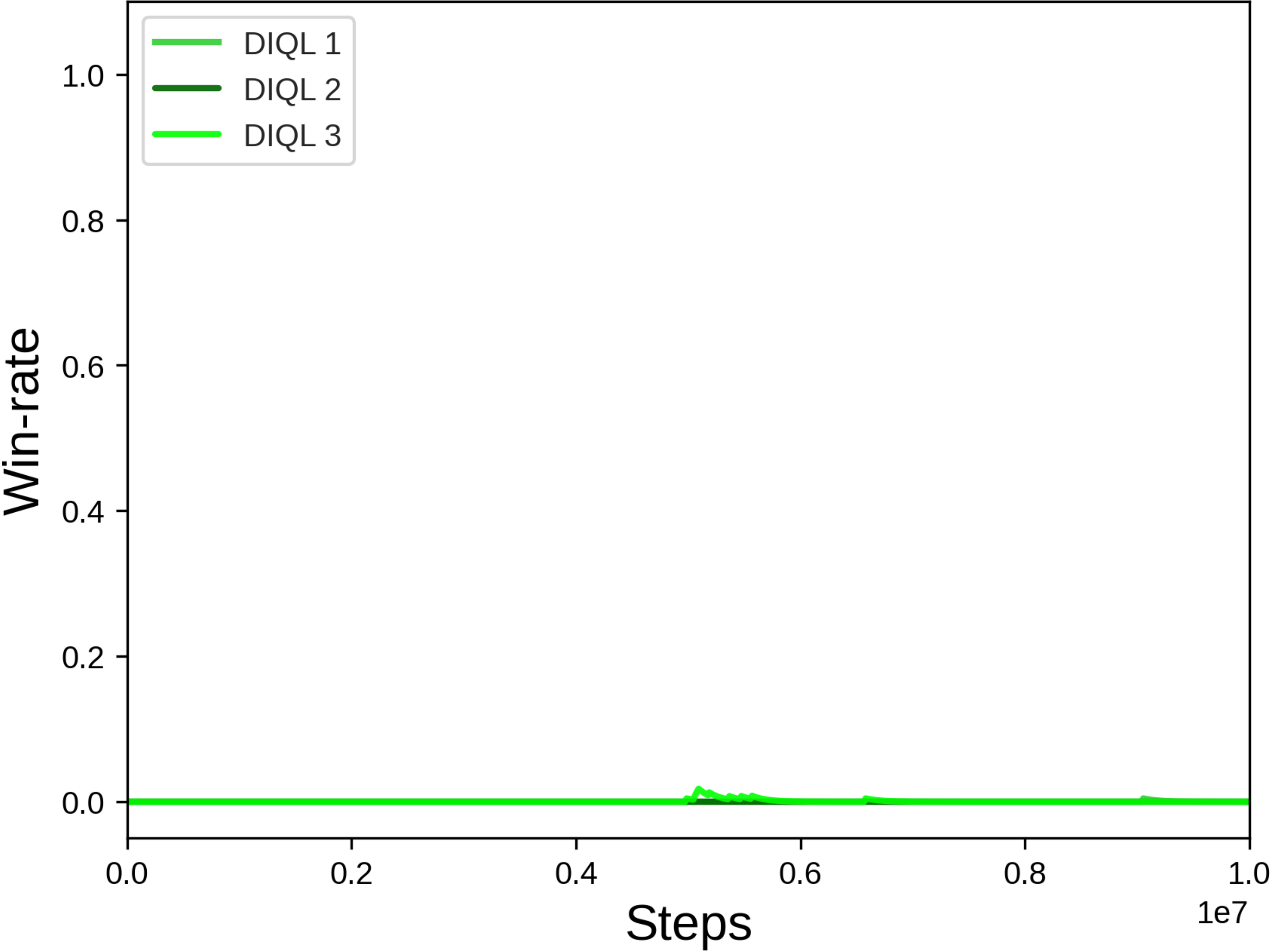}
            \caption{Defense armored}
            \label{fig:app_diql_parallel_def_arm}
        \end{subfigure}%
        \begin{subfigure}{0.26\columnwidth}
            \includegraphics[width=\columnwidth]{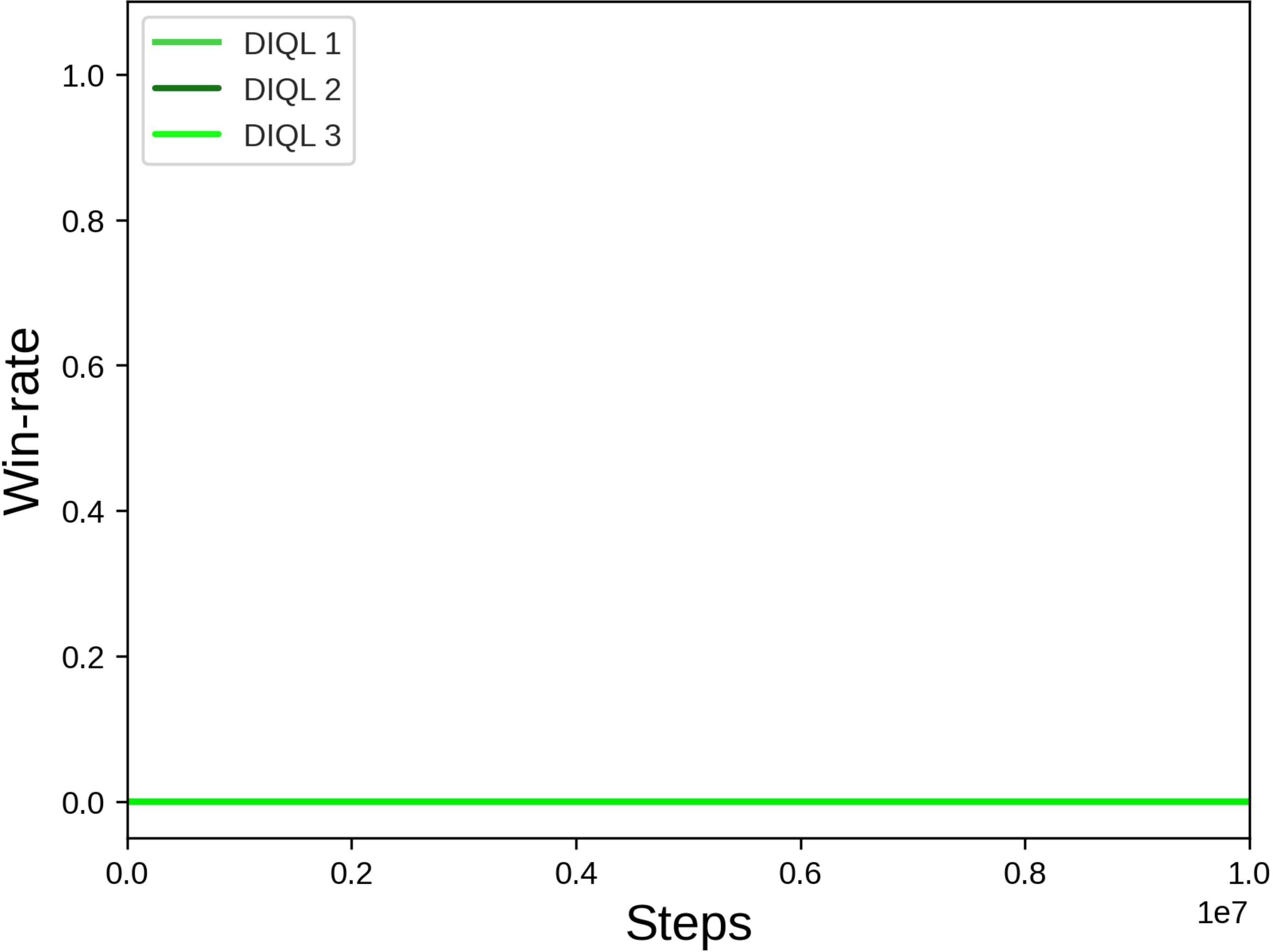}
            \caption{Defense outnumbered}
            \label{fig:app_diql_parallel_def_out}
        \end{subfigure}%
        
        \begin{subfigure}{0.26\columnwidth}
            \includegraphics[width=\columnwidth]{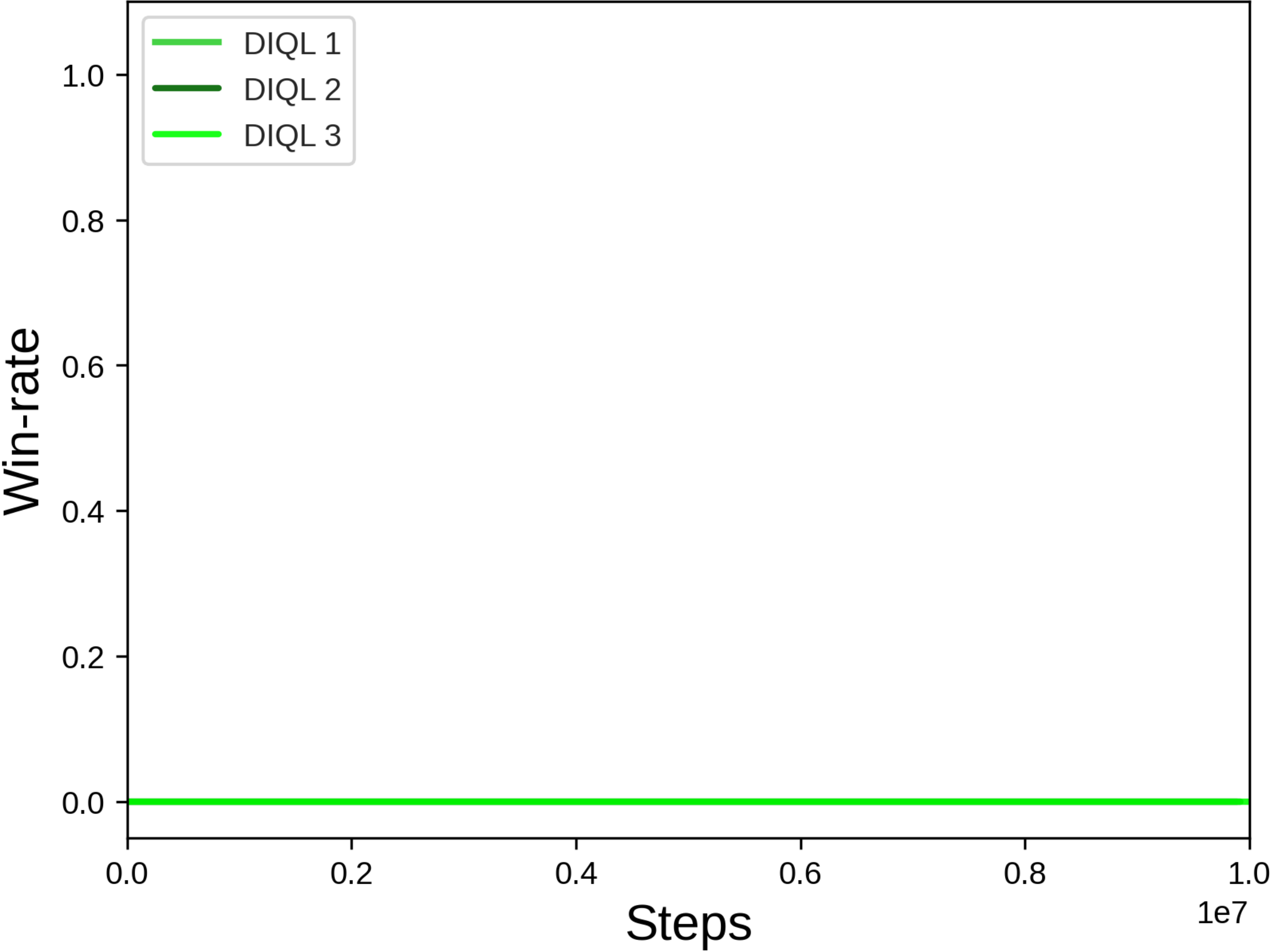}
            \caption{Offense near}
            \label{fig:app_diql_parallel_off_near}
        \end{subfigure}%
        \begin{subfigure}{0.26\columnwidth}
            \includegraphics[width=\columnwidth]{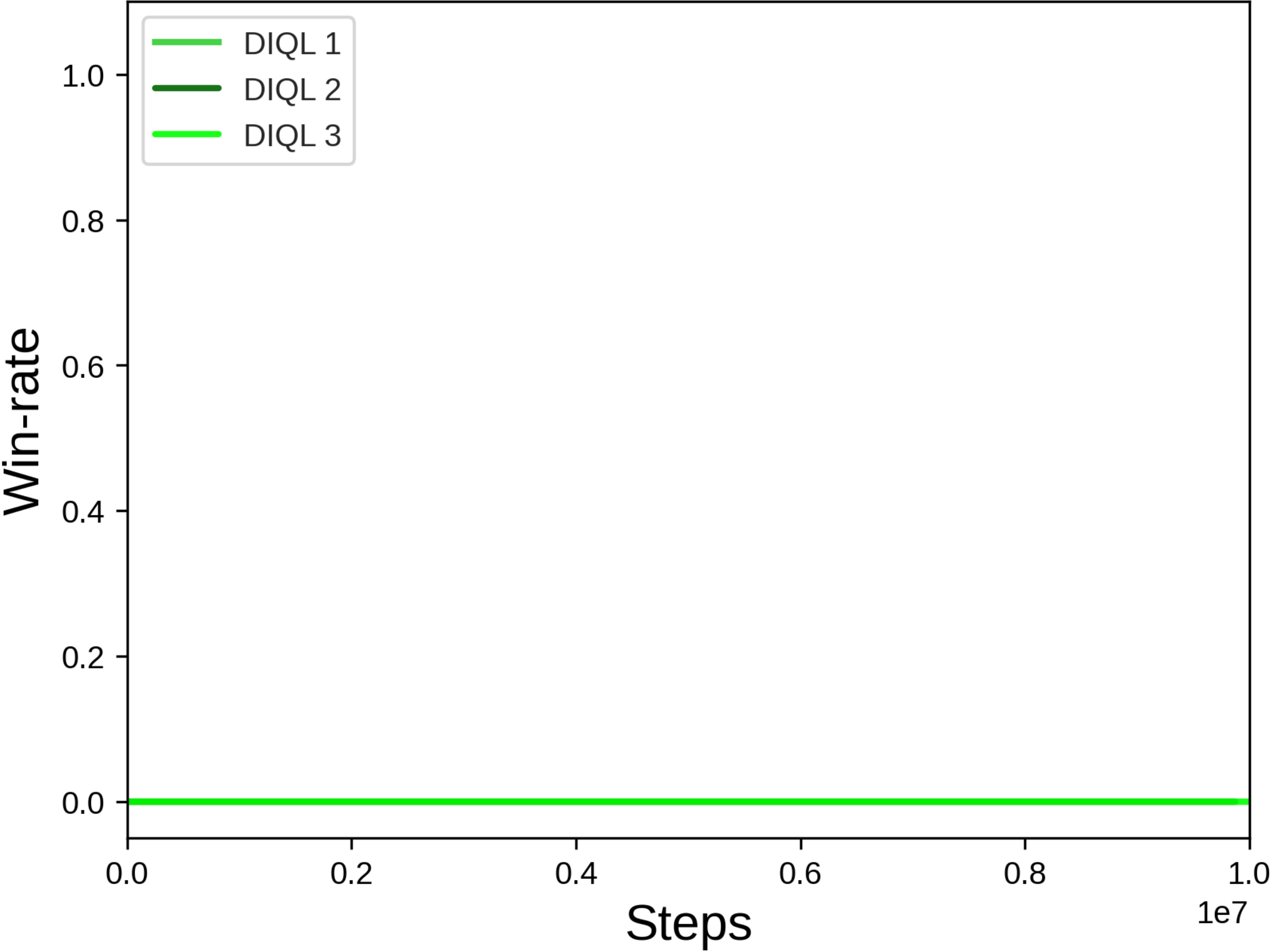}
            \caption{Offense distant}
            \label{fig:app_diql_parallel_off_dist}
        \end{subfigure}%
        \begin{subfigure}{0.26\columnwidth}
            \includegraphics[width=\columnwidth]{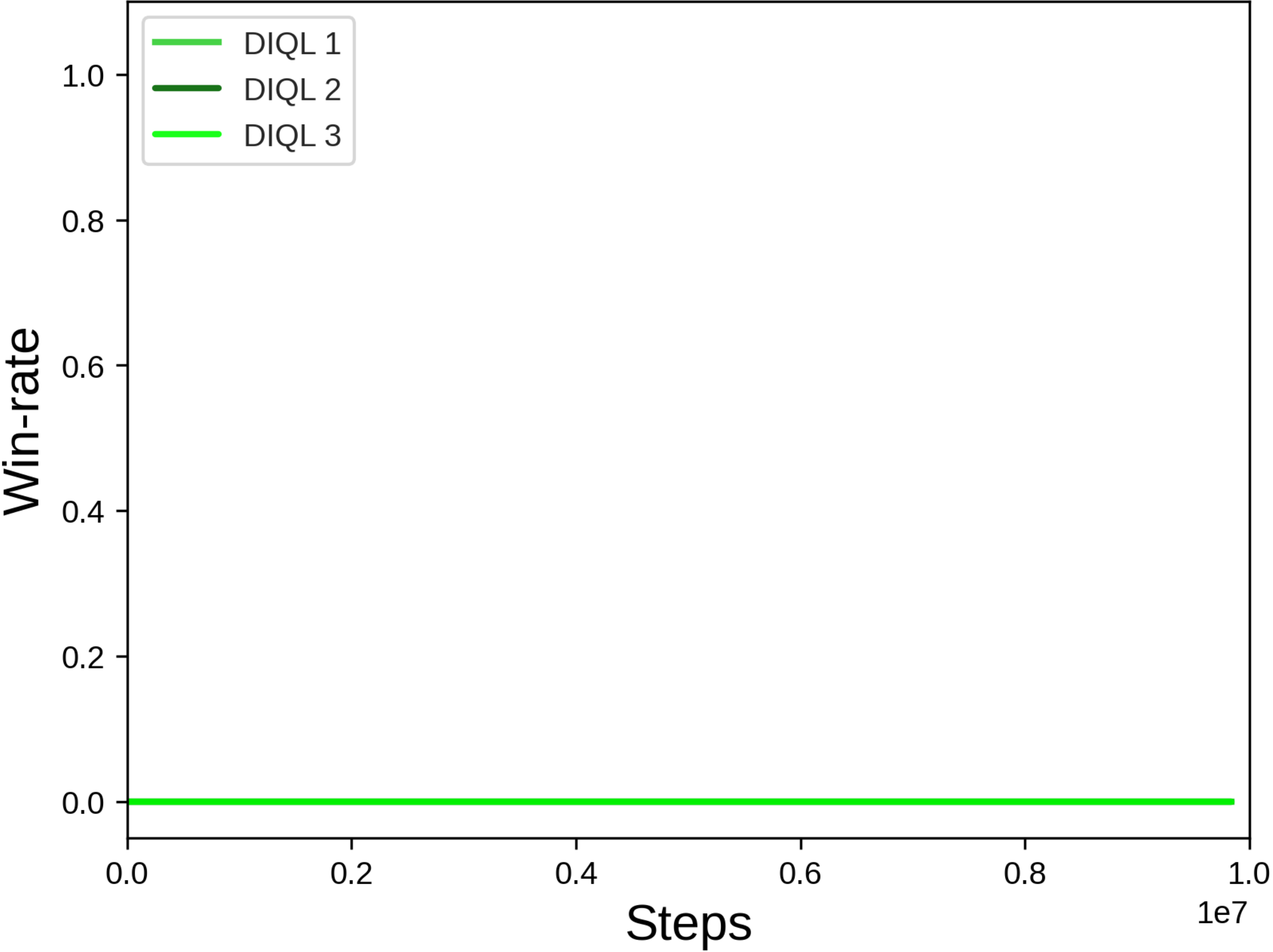}
            \caption{Offense complicated}
            \label{fig:app_diql_parallel_off_com}
        \end{subfigure}%
        
        \begin{subfigure}{0.27\columnwidth}
            \includegraphics[width=\columnwidth]{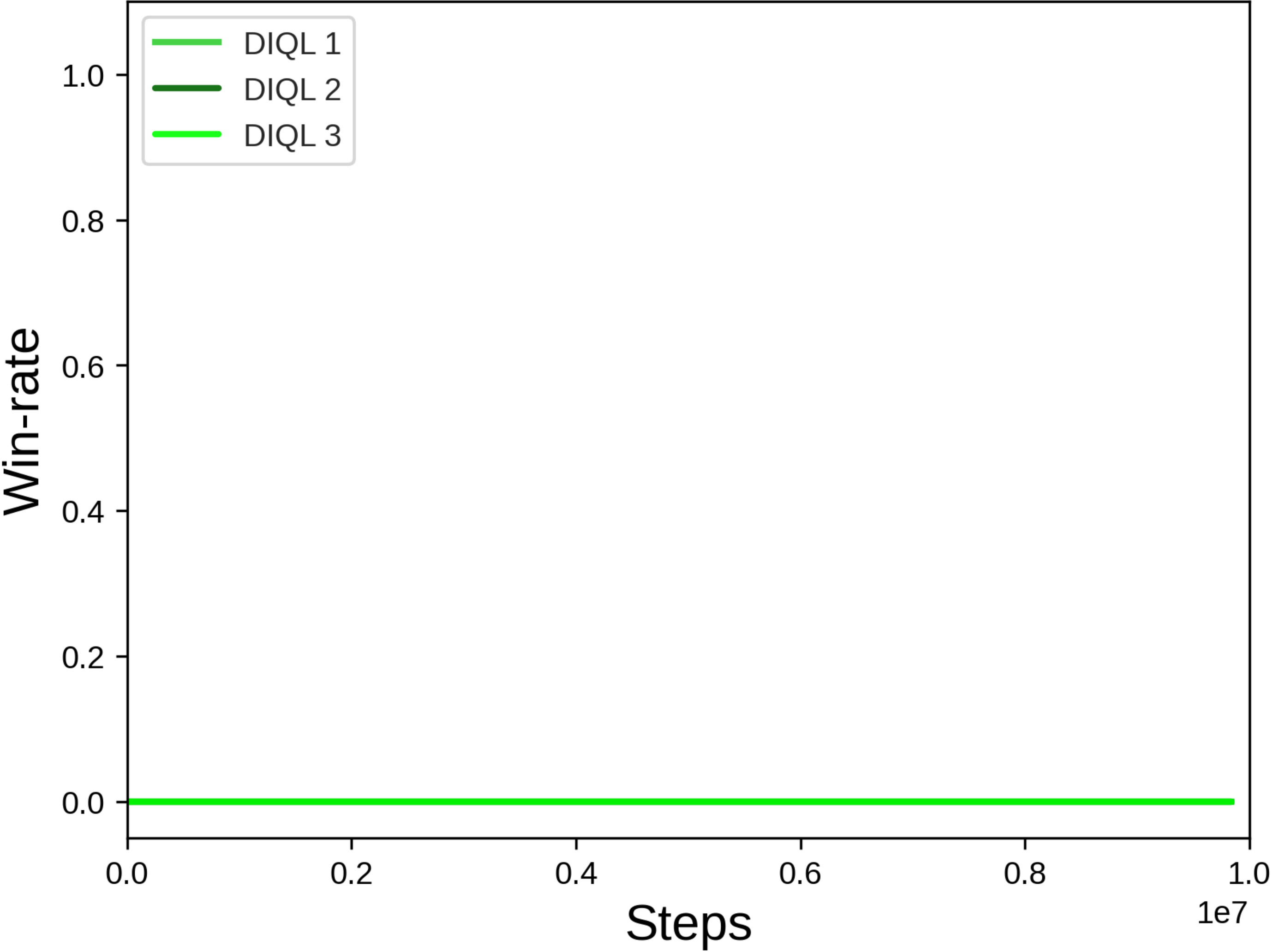}
            \caption{Offense hard}
            \label{fig:app_diql_parallel_off_hard}
        \end{subfigure}%
        \begin{subfigure}{0.27\columnwidth}
            \includegraphics[width=\columnwidth]{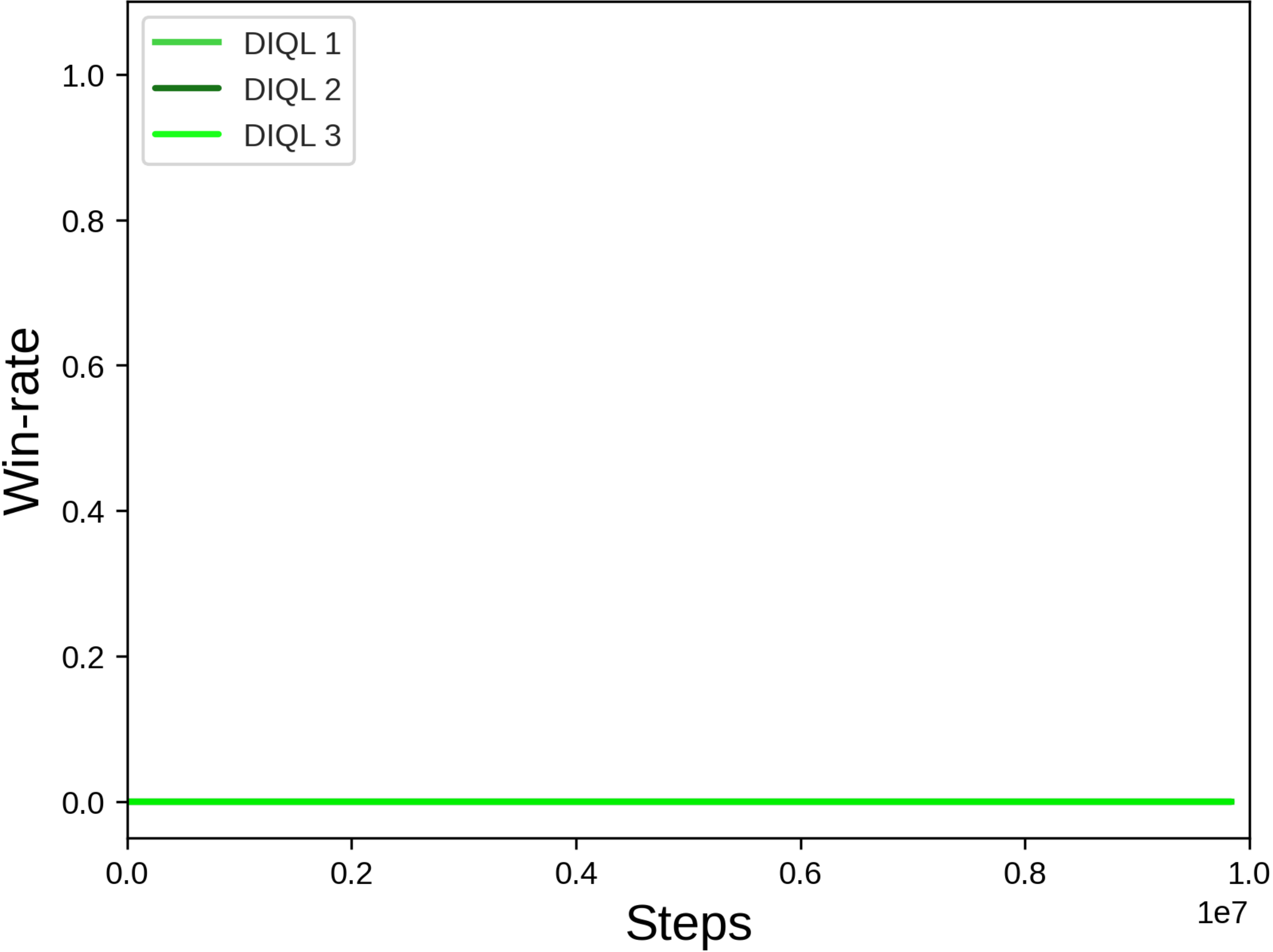}
            \caption{Offense superhard}
            \label{fig:app_diql_parallel_off_super}
        \end{subfigure}%
    \caption{DIQL trained on the parallel episodic buffer}
    \label{fig:app_diql_parallel}
}
\end{figure}

\begin{figure}[!ht]{
    \centering
        \begin{subfigure}{0.26\columnwidth}
            \includegraphics[width=\columnwidth]{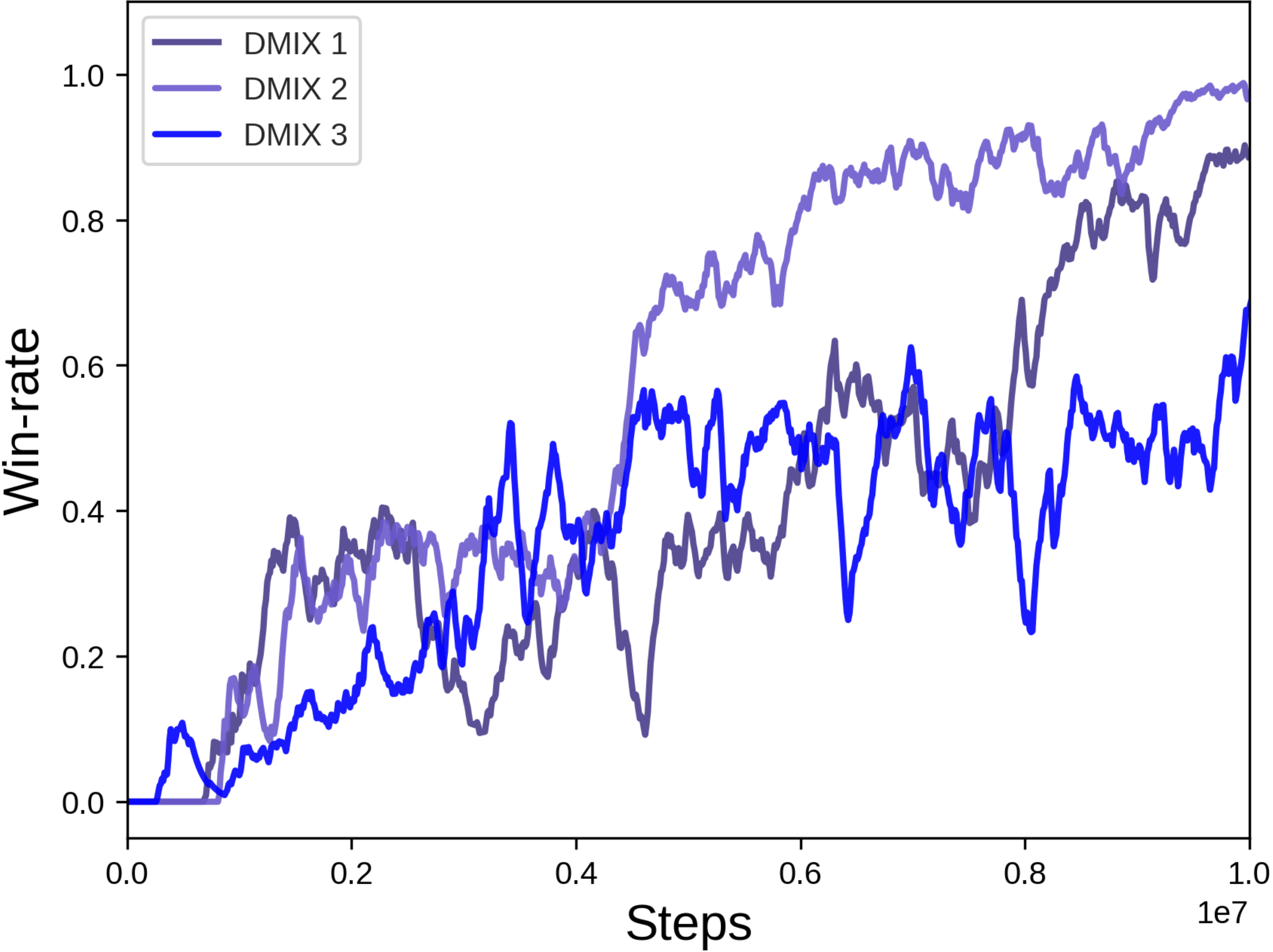}
            \caption{Defense infantry}
            \label{fig:app_dmix_parallel_def_inf}
        \end{subfigure}%
        \begin{subfigure}{0.26\columnwidth}
            \includegraphics[width=\columnwidth]{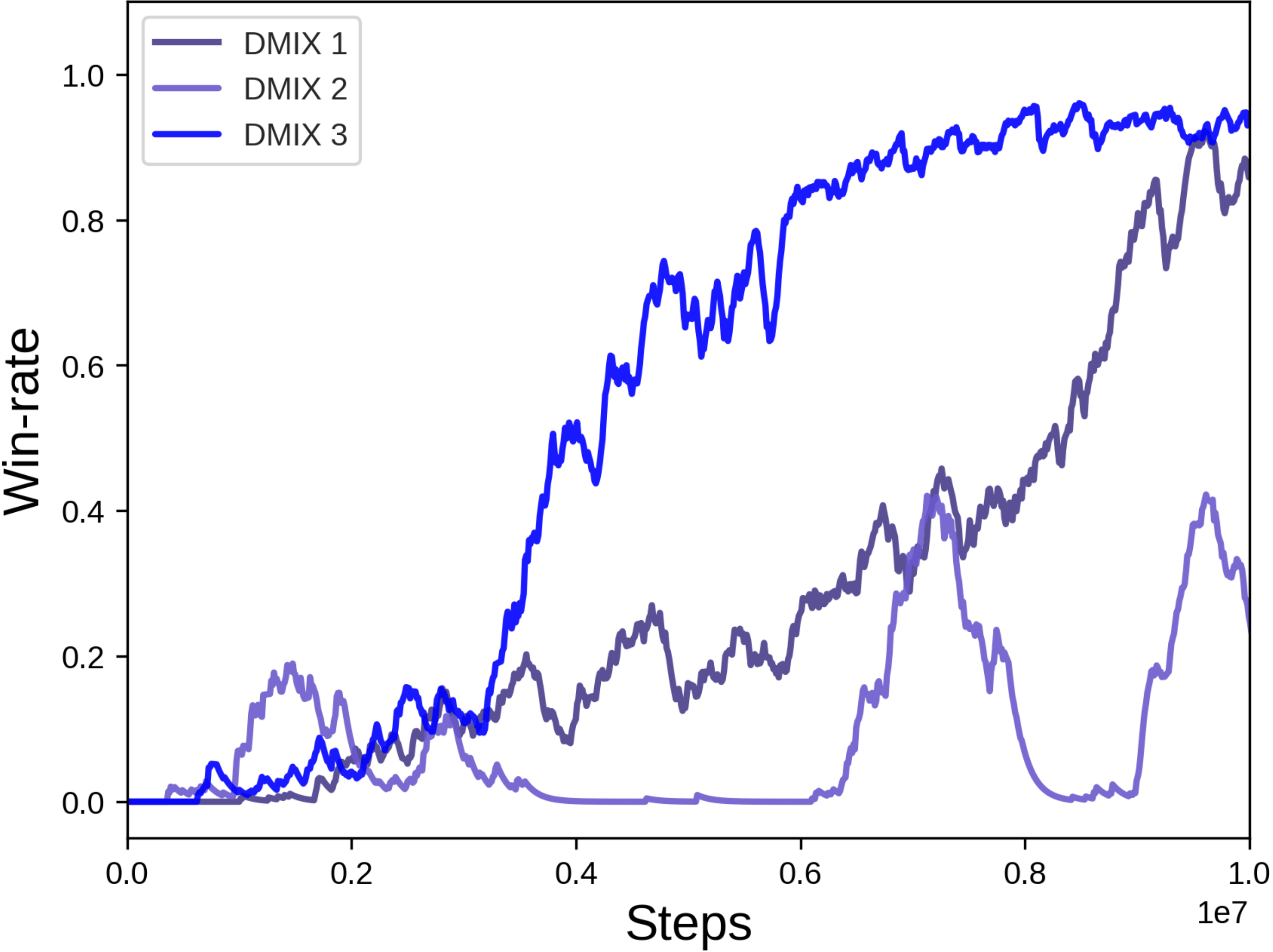}
            \caption{Defense armored}
            \label{fig:app_dmix_parallel_def_arm}
        \end{subfigure}%
        \begin{subfigure}{0.26\columnwidth}
            \includegraphics[width=\columnwidth]{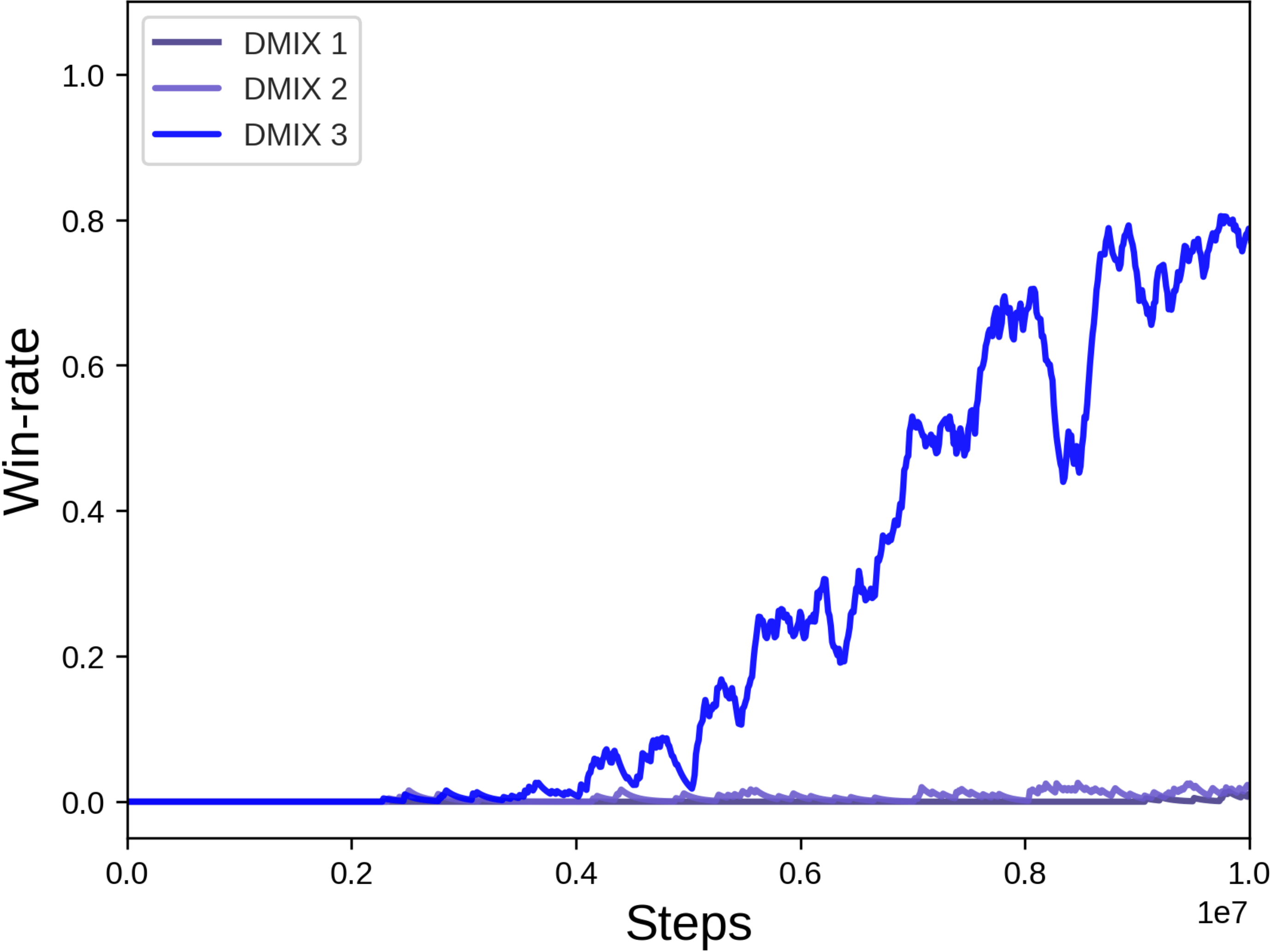}
            \caption{Defense outnumbered}
            \label{fig:app_dmix_parallel_def_out}
        \end{subfigure}%
        
        \begin{subfigure}{0.26\columnwidth}
            \includegraphics[width=\columnwidth]{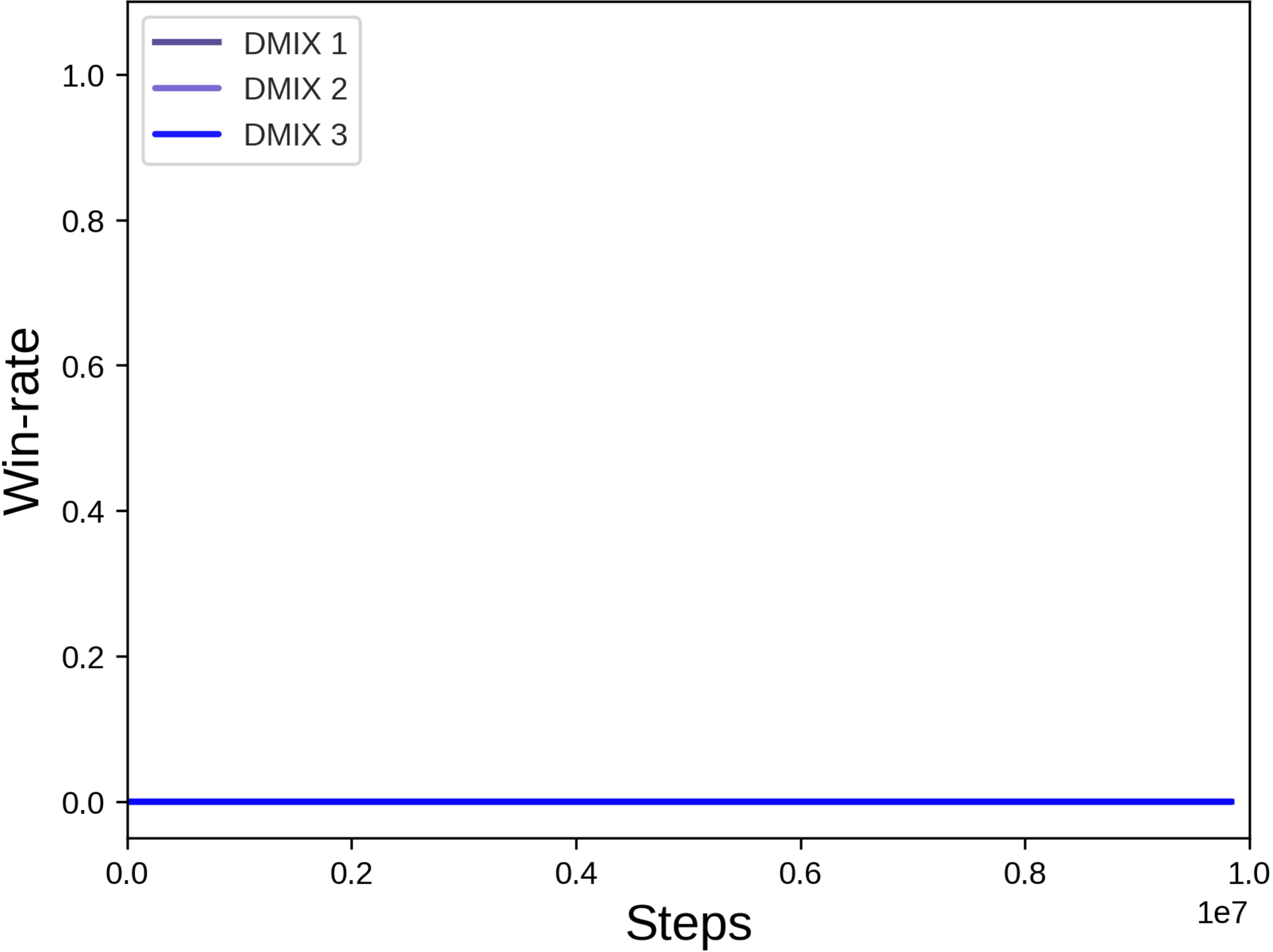}
            \caption{Offense near}
            \label{fig:app_dmix_parallel_off_near}
        \end{subfigure}%
        \begin{subfigure}{0.26\columnwidth}
            \includegraphics[width=\columnwidth]{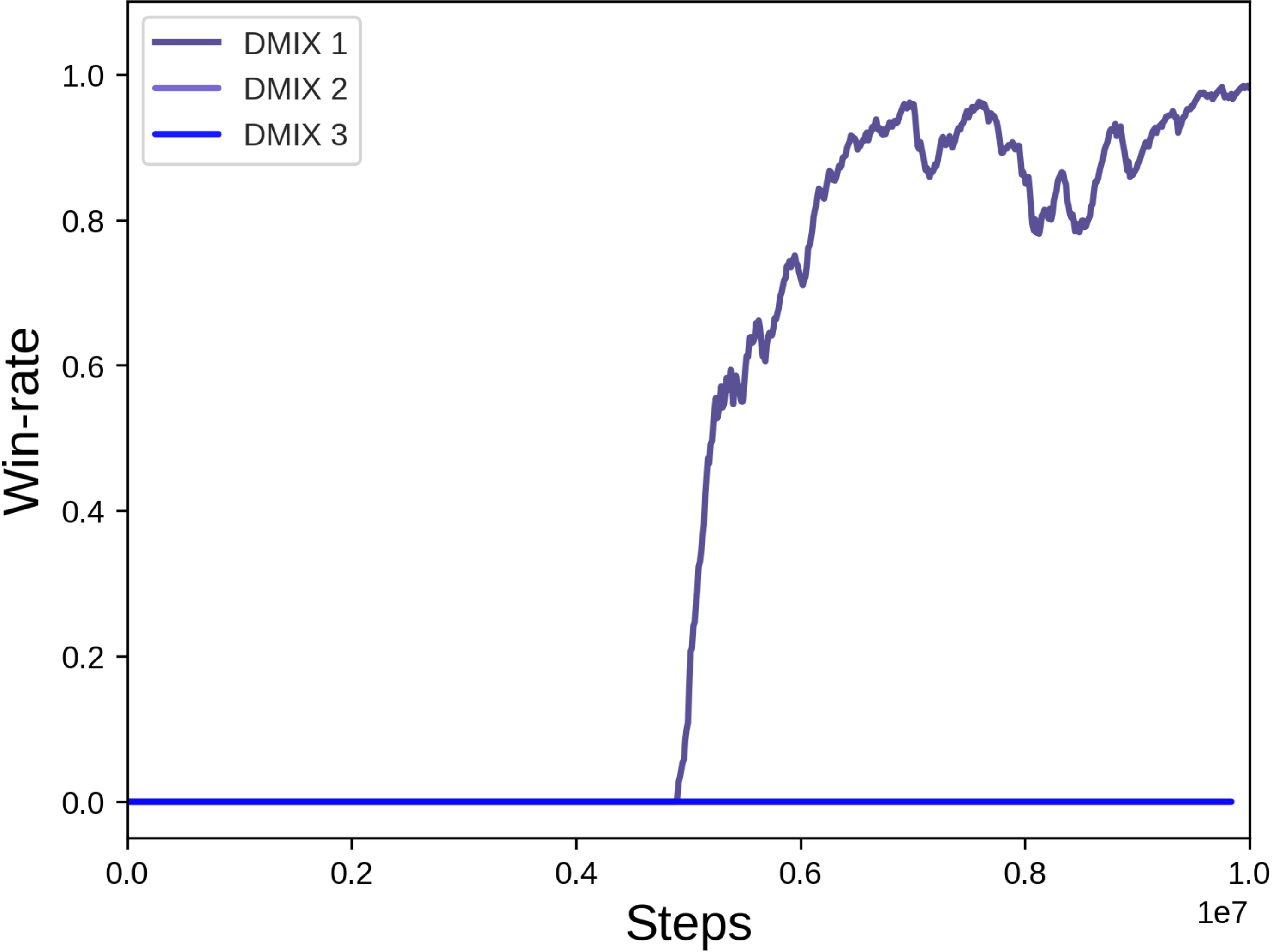}
            \caption{Offense distant}
            \label{fig:app_dmix_parallel_off_dist}
        \end{subfigure}%
        \begin{subfigure}{0.26\columnwidth}
            \includegraphics[width=\columnwidth]{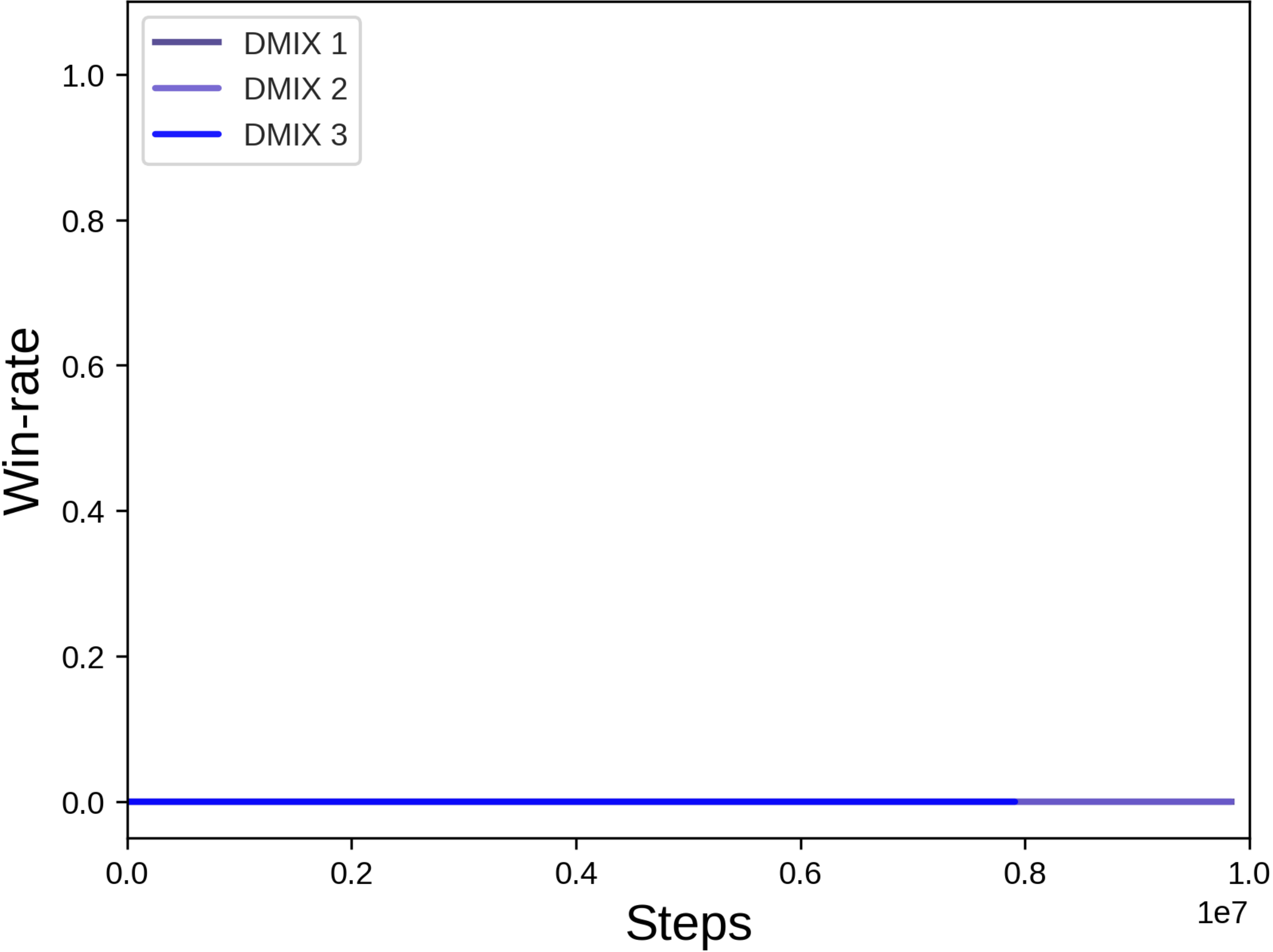}
            \caption{Offense complicated}
            \label{fig:app_dmix_parallel_off_com}
        \end{subfigure}%
        
        \begin{subfigure}{0.27\columnwidth}
            \includegraphics[width=\columnwidth]{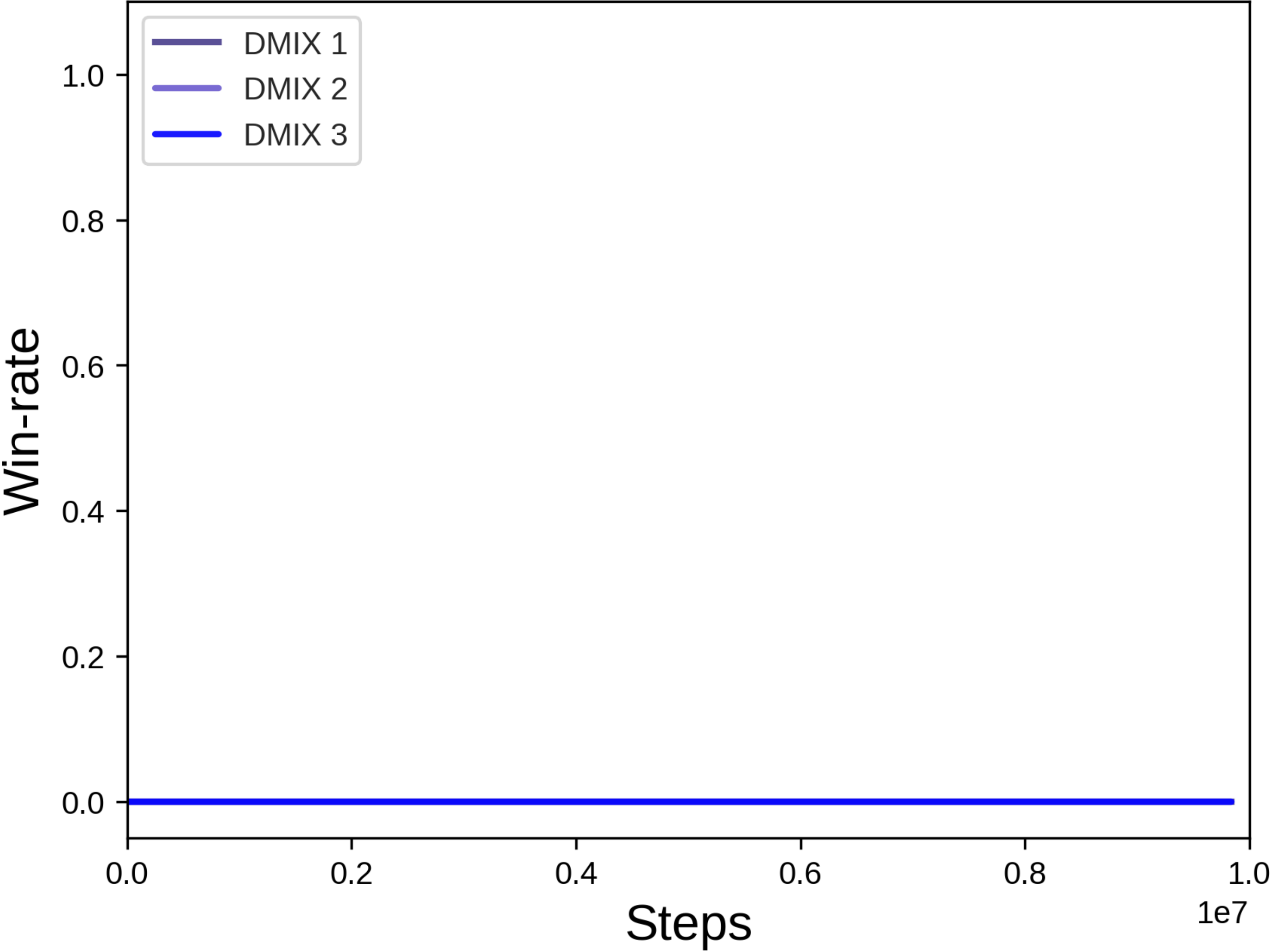}
            \caption{Offense hard}
            \label{fig:app_dmix_parallel_off_hard}
        \end{subfigure}%
        \begin{subfigure}{0.27\columnwidth}
            \includegraphics[width=\columnwidth]{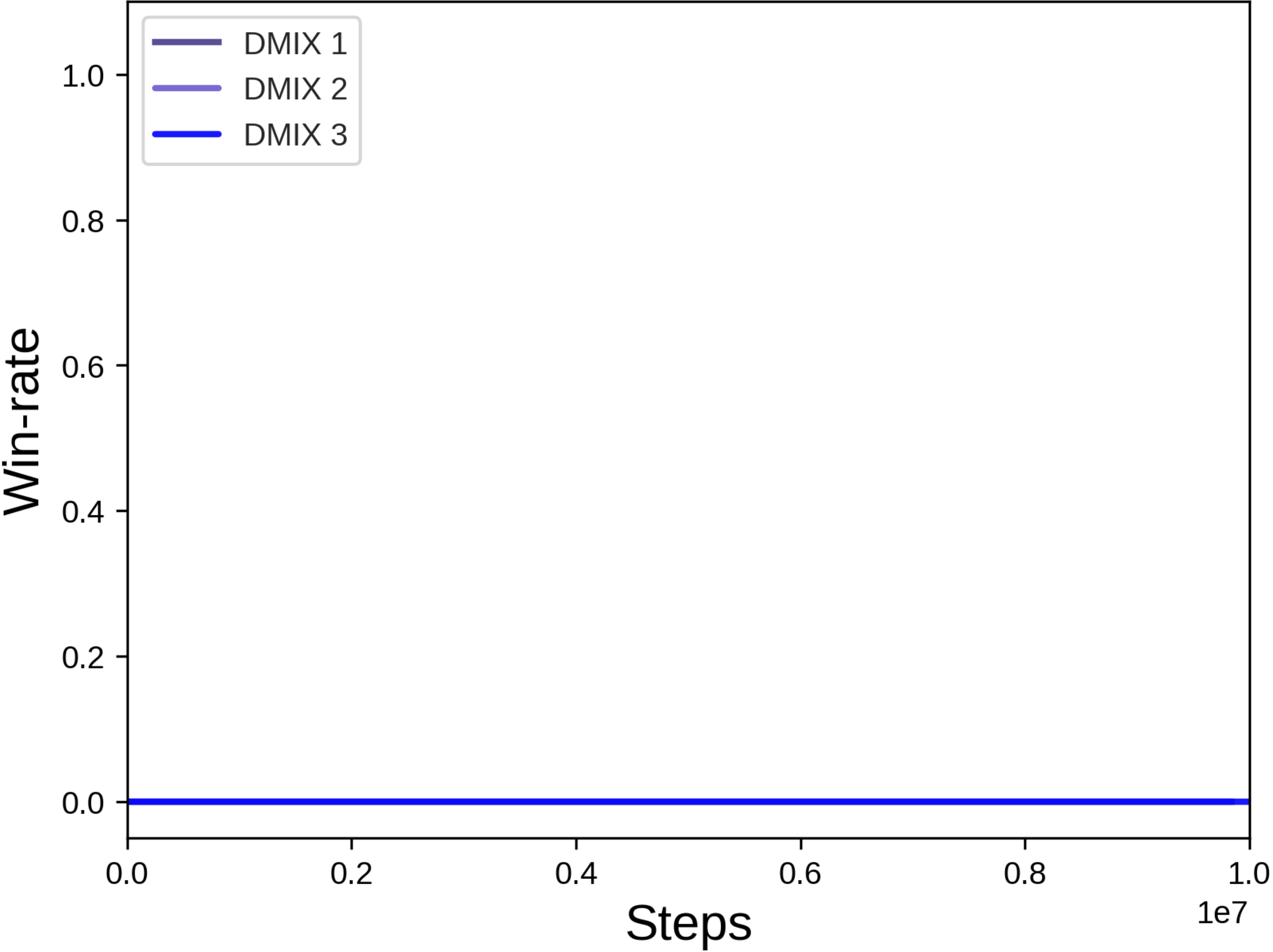}
            \caption{Offense superhard}
            \label{fig:app_dmix_parallel_off_super}
        \end{subfigure}%
    \caption{DMIX trained on the parallel episodic buffer}
    \label{fig:app_dmix_parallel}
}
\end{figure}

\begin{figure}[!ht]{
    \centering
        \begin{subfigure}{0.26\columnwidth}
            \includegraphics[width=\columnwidth]{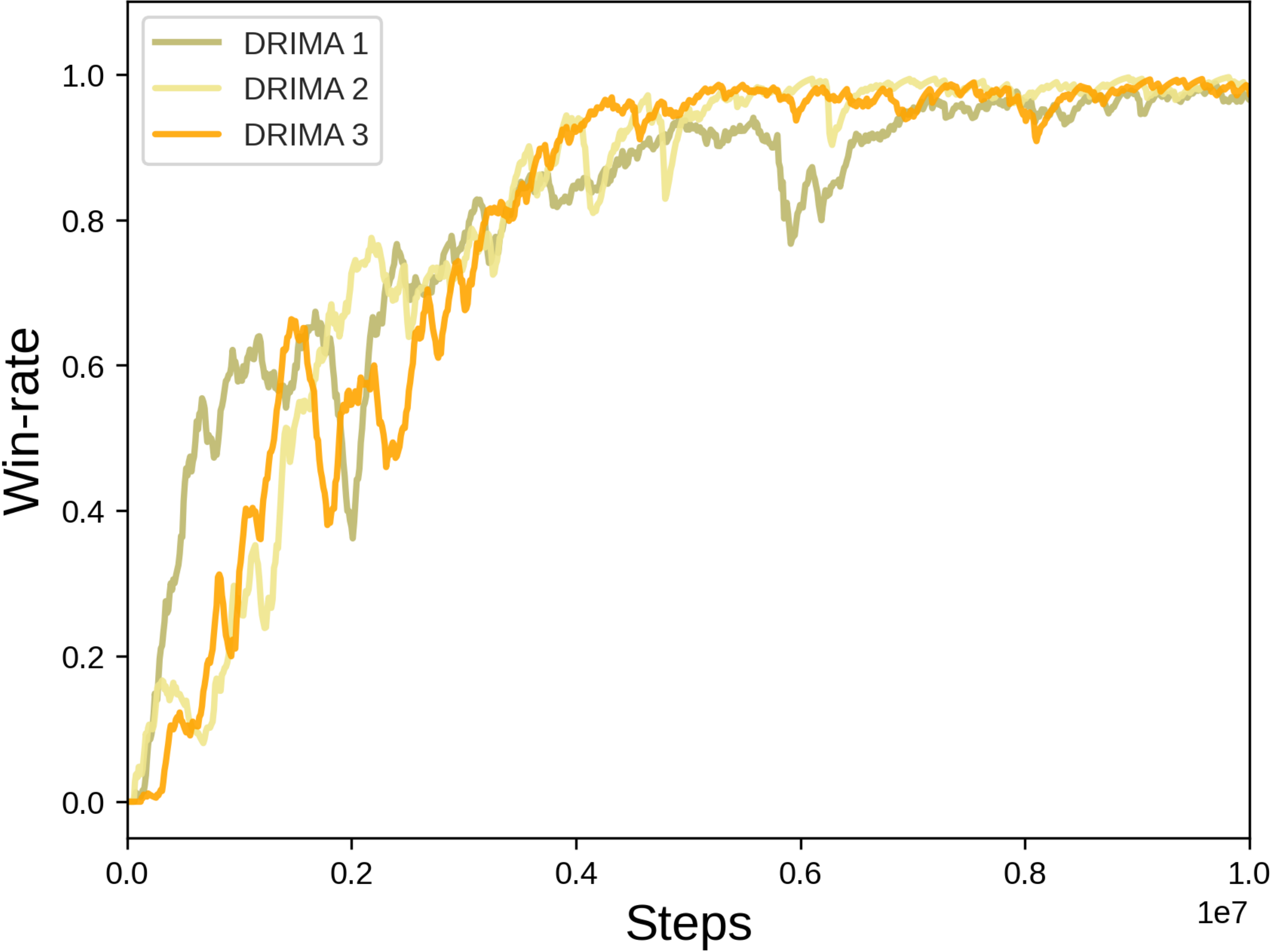}
            \caption{Defense infantry}
            \label{fig:app_drima_parallel_def_inf}
        \end{subfigure}%
        \begin{subfigure}{0.26\columnwidth}
            \includegraphics[width=\columnwidth]{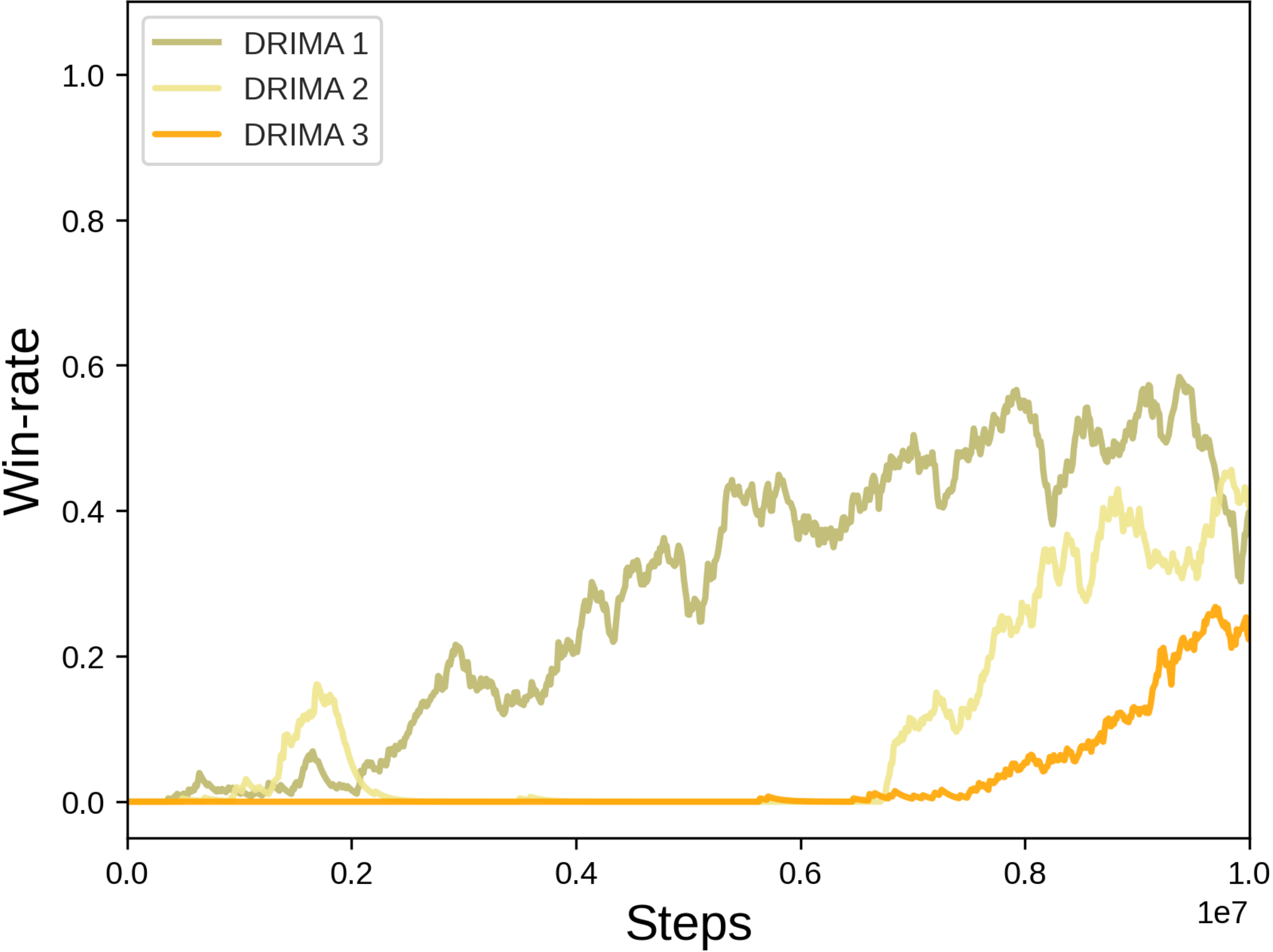}
            \caption{Defense armored}
            \label{fig:app_drima_parallel_def_arm}
        \end{subfigure}%
        \begin{subfigure}{0.26\columnwidth}
            \includegraphics[width=\columnwidth]{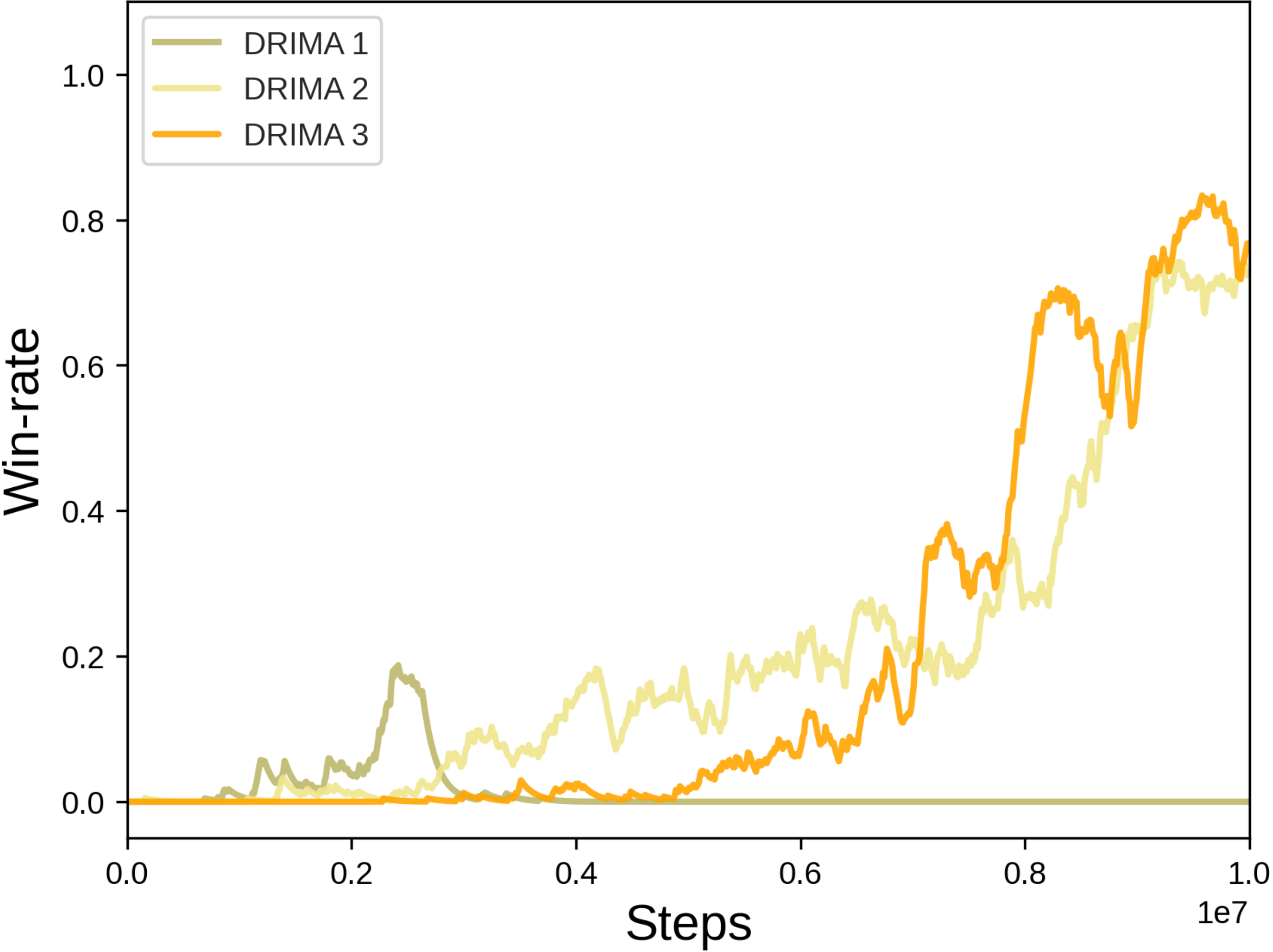}
            \caption{Defense outnumbered}
            \label{fig:app_drima_parallel_def_out}
        \end{subfigure}%
        
        \begin{subfigure}{0.26\columnwidth}
            \includegraphics[width=\columnwidth]{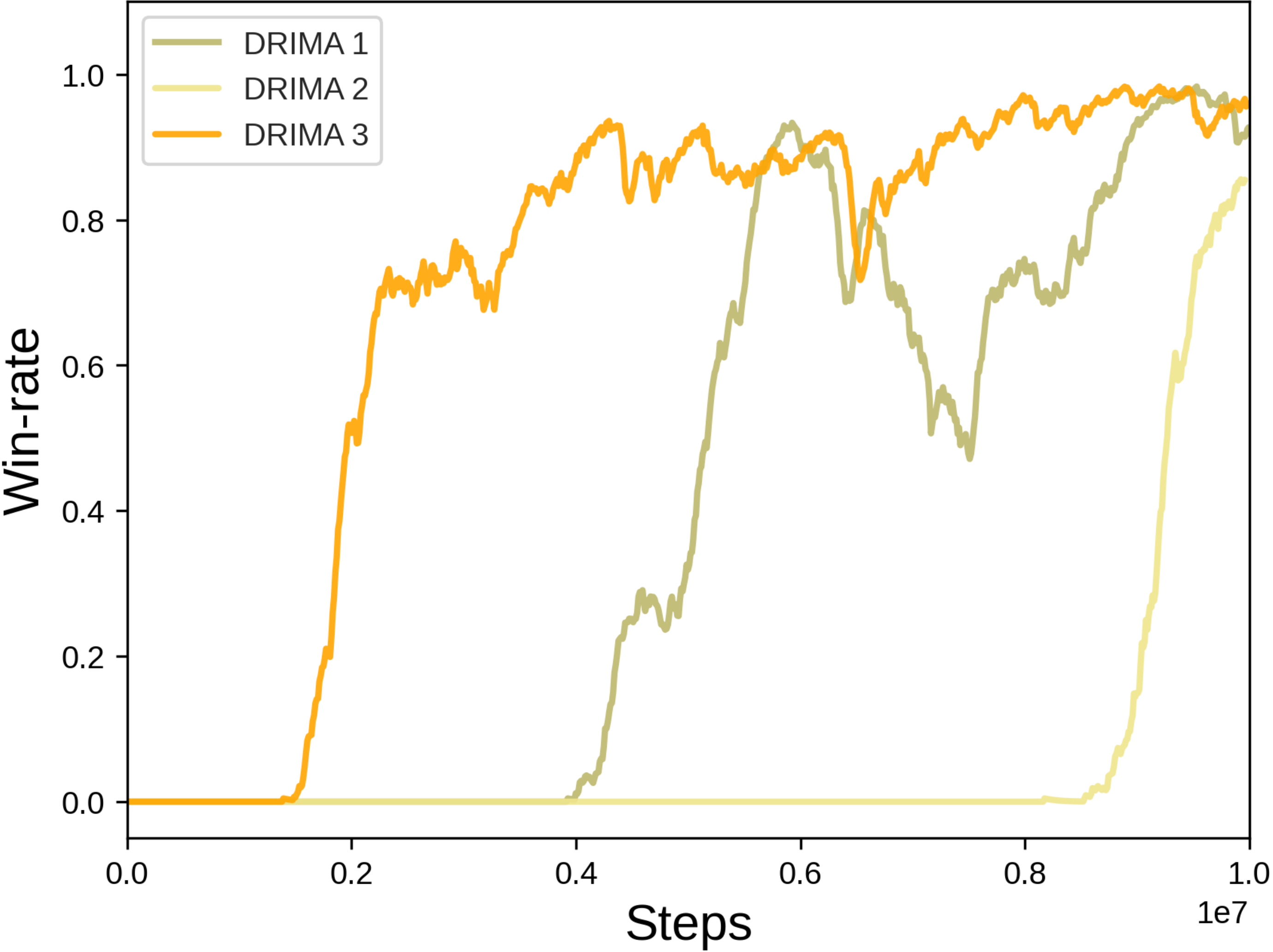}
            \caption{Offense near}
            \label{fig:app_drima_parallel_off_near}
        \end{subfigure}%
        \begin{subfigure}{0.26\columnwidth}
            \includegraphics[width=\columnwidth]{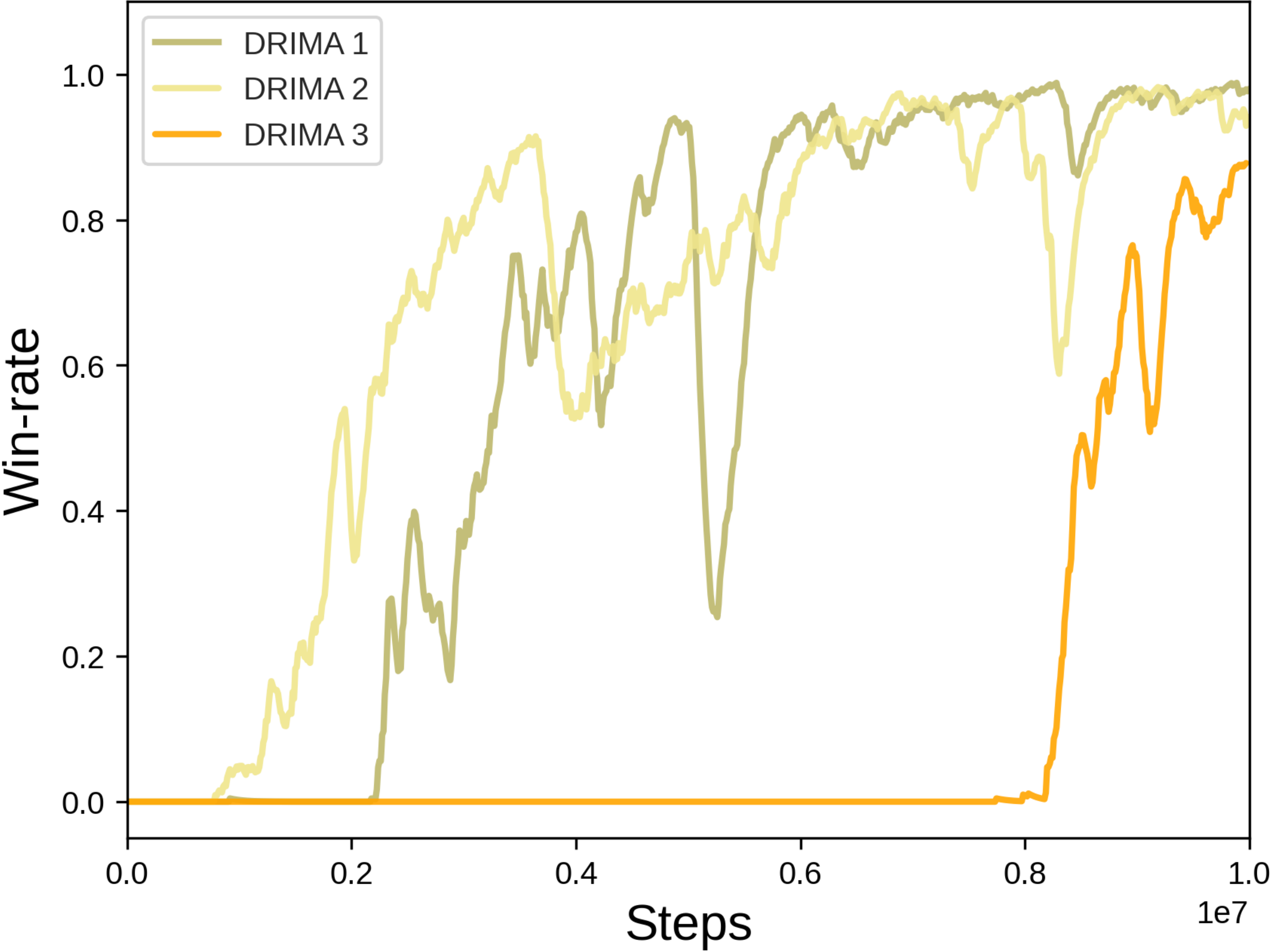}
            \caption{Offense distant}
            \label{fig:app_drima_parallel_off_dist}
        \end{subfigure}%
        \begin{subfigure}{0.26\columnwidth}
            \includegraphics[width=\columnwidth]{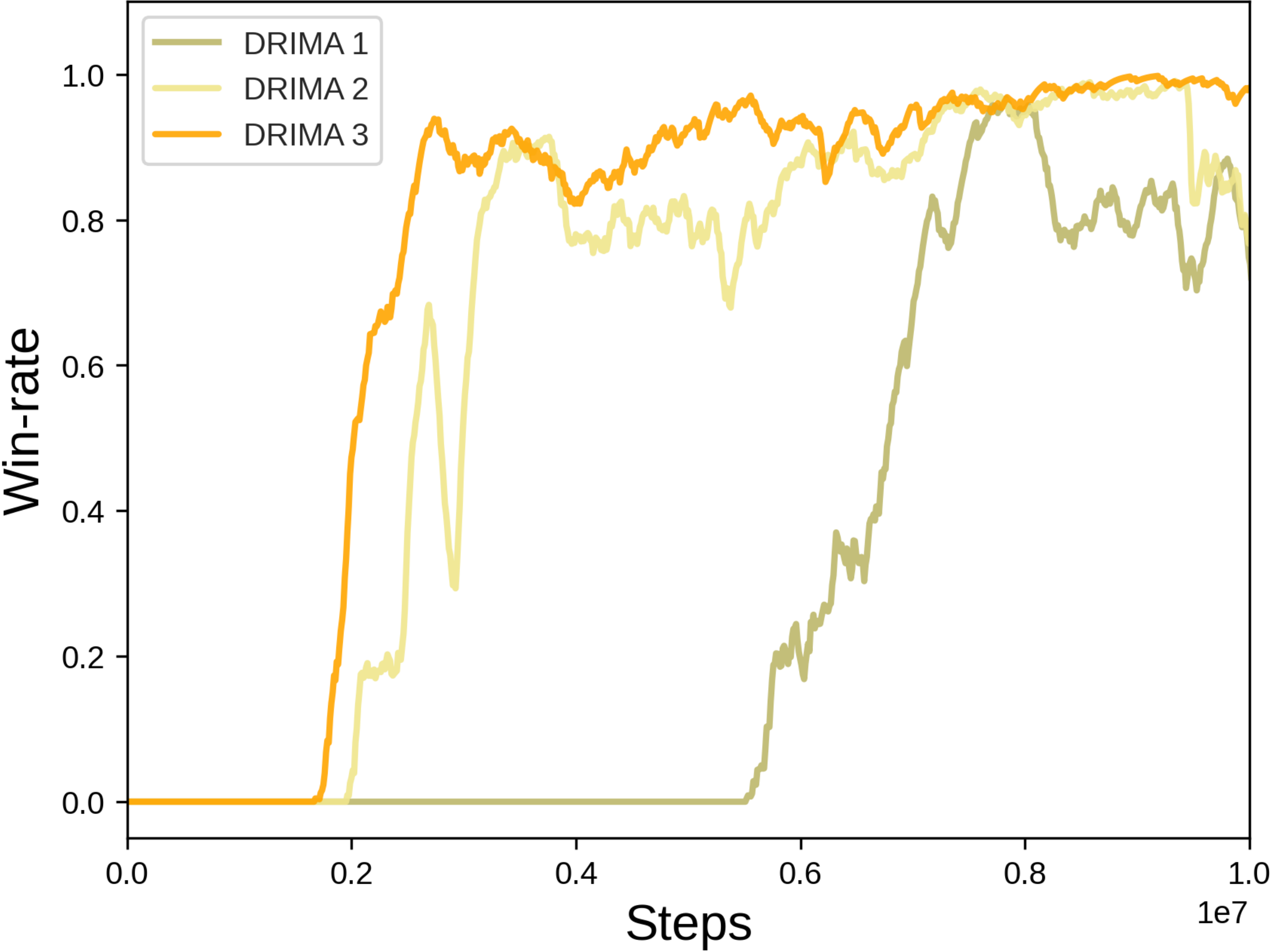}
            \caption{Offense complicated}
            \label{fig:app_drima_parallel_off_com}
        \end{subfigure}%
        
        \begin{subfigure}{0.27\columnwidth}
            \includegraphics[width=\columnwidth]{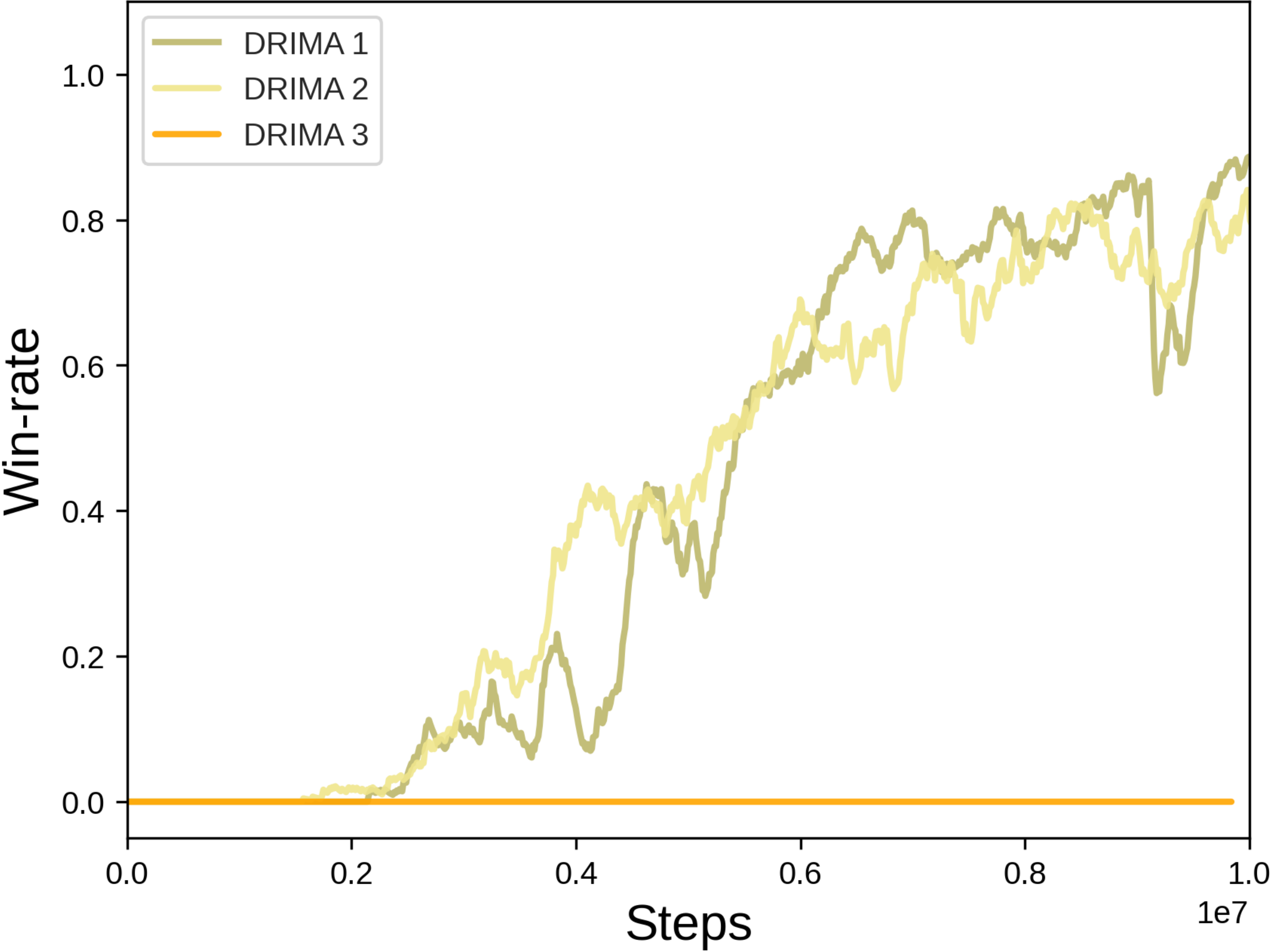}
            \caption{Offense hard}
            \label{fig:app_drima_parallel_off_hard}
        \end{subfigure}%
        \begin{subfigure}{0.27\columnwidth}
            \includegraphics[width=\columnwidth]{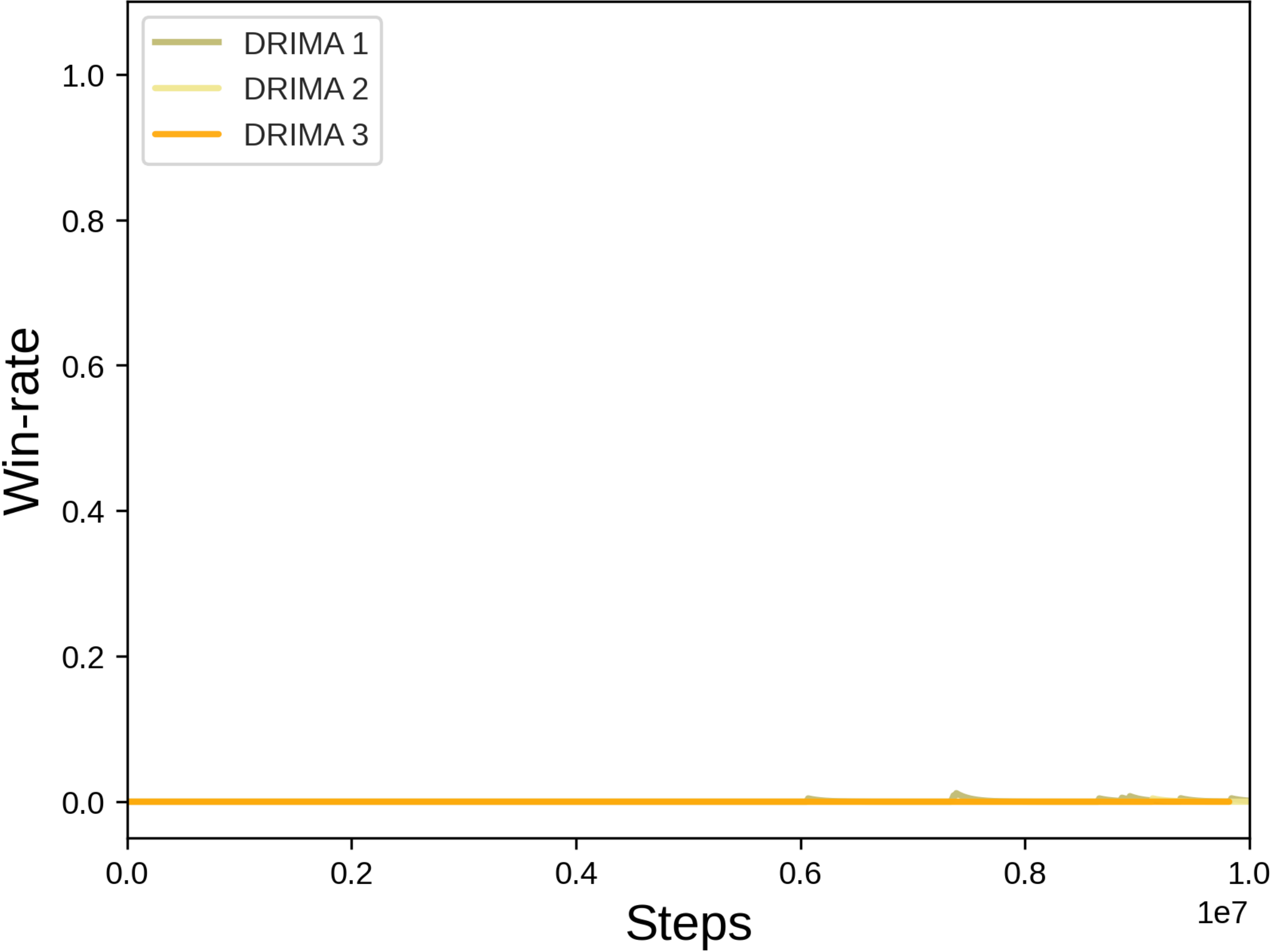}
            \caption{Offense superhard}
            \label{fig:app_drima_parallel_off_super}
        \end{subfigure}%
    \caption{DRIMA trained on the parallel episodic buffer}
    \label{fig:app_drima_parallel}
}
\end{figure}

As seen in \autoref{table:final_winrate}, and \autoref{fig:app_masac_parallel} - \ref{fig:app_drima_parallel}, for 10 million steps, we present the experimental outcomes of a parallel training setup with 20 parallel runners for each scenario and algorithms except MADDPG for not compatible with parallel episodic buffer setting. This setting contains 20 times fewer updates than episodic training, which may result in a performance decrease. With a parallel buffer, the overall tendency of performance is the same as that of episodic buffer setting, which shows that the performance in defense scenarios tends to decline as the supply differential increases. But in offensive scenarios, the pattern that the performance decreases as the complexity of the scenario increases is strongly shown than in sequential episodic settings. But likewise sequential episodic setting, none of the algorithms can solve the offense hard and superhard scenarios, which are also very challenging in a parallel setting. The only algorithm which solved the offense hard scenario are DRIMA, as shown in \autoref{fig:app_drima_parallel_off_hard}, which uses risk-based exploration.

\begin{table}[!t]
    \caption{Average win-rate (\%) performance of QMIX, DRIMA, COMA and MADDPG. All methods used sequential episodic buffers. Note that MADDPG is only compatible with the sequenced experience buffer.}
    \label{table:app_final_winrate_episode}
    \centering
    \resizebox{0.99\columnwidth}{!}{%
    \begin{tabular}{cccccccccc}
    \toprule
        & \multirow{2}{*}{Trial} & \multicolumn{3}{c}{\bf Defensive scenarios} & \multicolumn{5}{c}{\bf Offensive scenarios} \\
    \cmidrule(r){3-5} \cmidrule(r){6-10}
        & & \texttt{infantry} & \texttt{armored} & \texttt{outnumbered} & \texttt{near} & \texttt{distant} & \texttt{complicated} & \texttt{hard} & \texttt{superhard} \\
    \midrule

\multirow{3}{*}{COMA\cite{foerster2018counterfactual}}   &1  & 75.0 &  0.0 &  0.0 &  0.0 & 0.0 & 0.0 & 0.0 & 0.0 \\
                        &2  & 28.1 &  0.0 &  0.0 &  0.0 & 0.0 & 0.0 & 0.0 & 0.0 \\
                        &3  & 21.9 &  0.0$^{1)}$ &  0.0 &  0.0 & 0.0 & 0.0$^{2)}$ & 0.0 & 0.0 \\
\cmidrule(r){1-10}
\multirow{3}{*}{QMIX\cite{rashid2018qmix}}   &1  & 100 &  100 &  3.1 &  0.0 & 0.0 & 100 & 96.9 & 0.0  \\
                        &2  & 93.8 &  0.0 &  0.0 &  100.0 & 100.0 & 87.5 & 0.0 & 0.0  \\
                        &3  & 96.9 &  0.0 &  0.0 &  90.6 & 93.8 & 0.0$^{3)}$ & 96.9 & 0.0  \\
\cmidrule(r){1-10}
\multirow{3}{*}{MADDPG\cite{lowe2017multi}} &1  & 100 &  96.9 &  81.3 & 0.0 & 90.6 & 0.0 & 0.0 & 0.0  \\
                        &2  & 100 &  84.4 &  81.3 &  75.0 &  0.0 & 75.0 & 0.0 & 0.0 \\
                        &3  & 100 &  90.6 &  71.9 &  100.0 & 0.0 & 0.0 & 0.0 & 0.0  \\
\cmidrule(r){1-10}
\multirow{3}{*}{DRIMA\cite{learning5disentangling}}  &1  & 100 &  100 &  100 & 93.8 & 100$^{4)}$ & 96.9 & 96.9 & 15.6  \\
                        &2  & 100 &  96.9 &  96.9 &  93.8 & 100 & 100 & 93.8 & 3.1  \\
                        &3  & 100 &  100 &  100 & 100 & 100 & 96.9 & 93.8 & 15.6  \\
\hline\hline
\multicolumn{2}{c}{The total number of win-rate $\ge 80\%$} &9 &7 &5 &6 &6 &5 &5 &0\\ 
    \bottomrule
    \multicolumn{10}{l}{1) Takes total cumulative 3.29 million episode steps during training} \\
    \multicolumn{10}{l}{2) Takes total cumulative 4.21 million episode steps during training} \\
    \multicolumn{10}{l}{3) Takes total cumulative 4.59 million episode steps during training} \\
    \multicolumn{10}{l}{4) Takes total cumulative 2.53 million episode steps during training} \\
    \end{tabular}
     }%
\end{table}

\begin{table}[!t]
    \caption{Average win-rate (\%) performance of additional seven algorithms with sequential episodic buffers in defensive scenarios.}
    \label{table:app_final_winrate_episode}
    \centering
    \resizebox{0.65\columnwidth}{!}{%
    \begin{tabular}{ccccc}
    \toprule
        & \multirow{2}{*}{Trial} & \multicolumn{3}{c}{\bf Defensive scenarios} \\
    \cmidrule(r){3-5}
        & & \texttt{infantry} & \texttt{armored} & \texttt{outnumbered} \\
    \midrule

\multirow{3}{*}{MASAC\cite{pu2021decomposed}}   &1  & 50.0 & 0.0 & 0.0 \\
                        &2  & 37.5 & 0.0 & 0.0 \\
                        &3  & 0.0 & 0.0 & 0.0 \\
\cmidrule(r){1-5}
\multirow{3}{*}{IQL\cite{tan1993multi}}   &1  & 96.9 & 9.4 & 0.0  \\
                        &2  & 93.8 & 90.6 & 0.0  \\
                        &3  & 84.4 & 6.3 & 0.0  \\
\cmidrule(r){1-5}
\multirow{3}{*}{VDN\cite{sunehag2017value}} &1  & 96.9 & 84.4 & 15.6  \\
                        &2  & 93.8 & 96.9 & 9.4 \\
                        &3  & 100 & 100 & 37.5  \\
\cmidrule(r){1-5}
\multirow{3}{*}{QTRAN\cite{son2019qtran}}  &1  & 100 & 25.0 & 81.3  \\
                        &2  & 100 & 96.9 & 65.6  \\
                        &3  & 100 & 93.8 & 93.8  \\
\cmidrule(r){1-5}
\multirow{3}{*}{DDN\cite{sun2021dfac}}  &1  & 81.3 & 96.9 & 0.0  \\
                        &2  & 96.9 & 71.9 & 68.8  \\
                        &3  & 90.6 & 71.9 & 0.0  \\
\cmidrule(r){1-5}
\multirow{3}{*}{DIQL\cite{sun2021dfac}}  &1  & 93.8 & 68.8 & 78.1  \\
                        &2  & 93.8 & 25.0 & 0.0  \\
                        &3  & 93.8 & 53.1 & 0.0  \\
\cmidrule(r){1-5}
\multirow{3}{*}{DMIX\cite{sun2021dfac}}  &1  & 100 & 81.3 & 0.0  \\
                        &2  & 100 & 93.8 & 0.0  \\
                        &3  & 96.9 & 46.9 & 53.1  \\
\hline\hline
\multicolumn{2}{c}{The total number of win-rate $\ge 80\%$} &18 &9 &2\\ 
    \bottomrule
    \end{tabular}
     }%
\end{table}

\subsection{Benchmark on Parallel Episodic Buffer}
While the majority of studies have reported performance using the sequential episodic buffer, as we mentioned above, the parallel episodic buffer would be more practical due to reduced training time. For this reason, we provide a comprehensive benchmark of MARL algorithms in this setting. Since the previous challenge \cite{samvelyan2019starcraft} already provided technical implementation of the parallel setup, we merely determine the parallelism level. With consideration of computation resources, we intend to run twenty simulators simultaneously to gather episodes. In these experiments, we evaluate 10 MARL algorithms on SMAC$^+$. Overall, we observe that the tendency of experimental results is identical to those of the results from the episodic buffer as seen in \autoref{table:final_winrate}. In addition, this result supports that both defensive and offensive scenarios can also be adjusted by the complexity of multi-tasks and environmental factors. Instead, as stated, due to the reduced update frequency, we observe that the performance of all baselines is marginally decreased and learning instability is increased. For example, in relatively simple scenarios like \texttt{off\_near} and \texttt{off\_distant}, typical value-based algorithms such as VDN and QMIX outperform other algorithms, whereas as complexity increases, DRIMA with an enhanced exploration capability by risk-based information attains the highest scores only in \texttt{off\_complicated} and \texttt{def\_outnumbered}. When we look closely at distributional value-based algorithms, we find that enhanced exploration by distributional value functions positively affects the performance on SMAC$^+$, meanwhile the DMIX shows unstable outcomes despite that it occasionally gains the best scores in complex scenarios. Contrary, we see that DRIMA captures robust scores compared to other distributional algorithms.

In another aspect, we find the moment at which win-rate begins to rise is intriguing. As you can see \autoref{fig:off_results}, the learning curves of QMIX obviously depicts distinct tendency according to the buffer setting, on the other hand, the learning curves of DRIMA are comparable in both settings if convergence occurs. Therefore, we claim that learning multi-stage tasks and environmental factors without direct incentives are influenced by the exploration capability of MARL algorithms.

\begin{table}[!t]
    \caption{Average win-rate (\%) performance of 10 algorithms in both defensive and offensive scenarios using parallel episodic buffers with 20 parallel simulators.}
    \label{table:final_winrate}
    \centering
    \resizebox{0.99\columnwidth}{!}{%
    \begin{tabular}{ccccccccccc}
    \toprule
        \multirow{3}{*}{Category} &  \multirow{3}{*}{Algorithm} & \multirow{3}{*}{Trial} & \multicolumn{3}{c}{\bf Defensive scenarios} & \multicolumn{5}{c}{\bf Offensive scenarios} \\
    \cmidrule(r){4-6} \cmidrule(r){7-11} 
        & & & \texttt{infantry} & \texttt{armored} & \texttt{outnumbered} & \texttt{near} & \texttt{distant} & \texttt{complicated} & \texttt{hard} & \texttt{superhard} \\
    \midrule
    \multirow{6}{*}{\makecell{Policy\\gradient}}& 
    \multirow{3}{*}{MASAC\cite{pu2021decomposed}} &1 & 40.0 &  0.0 &  0.0 & 0.0 & 0.0 & 0.0 & 0.0 & 0.0 \\
    & &2 & 25.0 &  0.0 &  0.0 & 0.0 & 0.0 & 0.0 & 0.0 & 0.0 \\
    & &3 & 30.0 &  0.0 &  0.0 & 0.0 & 0.0 & 0.0 & 0.0 & 0.0\\
    \cmidrule(r){2-11}&
    \multirow{3}{*}{COMA\cite{foerster2018counterfactual}} &1 & 85.0 &  0.0 &  0.0 & 20.0 & 0.0 & 0.0 & 0.0 & 0.0\\
    & &2 & 50.0 &  0.0 &  0.0 & 80.0 & 0.0 & 0.0 & 0.0 & 0.0\\
    & &3 & 5.0 &  0.0 &  0.0 & 20.0 & 10.0 & 0.0 & 0.0 & 0.0\\
    \cmidrule(r){1-11}
    \multirow{13}{*}{\makecell{Value\\based}} & \multirow{3}{*}{IQL\cite{tan1993multi}} &1 & 20.0 &  5.0 &  0.0 & 0.0 & 25.0 & 10.0 & 0.0 & 0.0\\
    & &2 & 40.0 &  0.0 &  0.0 &  5.0 & 0.0 & 35.0 & 0.0 & 0.0\\
    & &3 & 45.0 &  0.0 &  0.0 &  10.0 & 0.0$^{1)}$ & 40.0 & 0.0 & 0.0\\
    \cmidrule(r){2-11} &
    \multirow{3}{*}{QTRAN\cite{son2019qtran}} &1 & 100 &  5.0 &  0.0 & 0.0 & 0.0 & 0.0 & 0.0 & 0.0\\
    & &2 & 80.0 & 25.0 &  0.0 & 0.0 & 0.0 & 0.0 & 0.0 & 0.0 \\
    & &3 & 100 &  0.0 &  0.0 & 0.0 & 0.0 & 0.0 & 0.0 & 0.0 \\
    \cmidrule(r){2-11} &
    \multirow{3}{*}{QMIX\cite{rashid2018qmix}} &1 & 100.0 &  75.0 &  30.0 & 95.0 & 20.0 & 0.0 & 0.0 & 0.0 \\
    & &2 & 95.0 &  100 &  0.0 & 95.0$^{2)}$ & 0.0 & 25.0 & 0.0 & 0.0 \\
    & &3 & 95.0 &  75.0 &  65.0 & 0.0 & 0.0 & 0.0 & 0.0 & 0.0 \\
    \cmidrule(r){2-11} &
     \multirow{3}{*}{VDN\cite{sunehag2017value}} &1 & 100 &  0.0 &  0.0 & 100 & 90.0 & 85.0 & 15.0 & 0.0 \\
    & &2 & 95.0 &  5.0 &  20.0 & 90.0 & 70.0 & 55.0 & 50.0 & 0.0 \\
    & &3 & 95.0 &  5.0 &  0.0 & 85.0 &  85.0 & 70.0$^{3)}$ & 10.0 & 0.0 \\
    \cmidrule(r){1-11}
    \multirow{13}{*}{\makecell{Distributional\\value based}} & 
    \multirow{3}{*}{DDN\cite{sun2021dfac}} &1 & 20.0 &  0.0 &  0.0 & 0.0 & 0.0 & 0.0 & 0.0 & 0.0\\
    & &2 & 30.0 &  0.0 &  0.0 & 0.0 & 0.0 & 0.0 & 0.0 & 0.0\\
    & &3 & 10.0 &  0.0 &  0.0 & 0.0 & 0.0 & 0.0 & 0.0 & 0.0\\
    \cmidrule(r){2-11} &
    \multirow{3}{*}{DIQL\cite{sun2021dfac}} &1 & 70.0 &  0.0 &  0.0 & 0.0 & 0.0 & 0.0 & 0.0 & 0.0\\
    & &2 & 5.0 &  0.0 &  0.0 & 0.0 & 0.0 & 0.0 & 0.0 & 0.0\\
    & &3 & 45.0 &  0.0 &  0.0 & 0.0 & 0.0 & 0.0 & 0.0 & 0.0\\
    \cmidrule(r){2-11} &
    \multirow{3}{*}{DMIX\cite{sun2021dfac}} &1 & 95.0 &  100 &  0.0 & 0.0 & 100 & 0.0 & 0.0 & 0.0\\
    & &2 & 90.0 &  55.0 &  5.0 & 0.0 & 0.0 & 0.0 & 0.0 & 0.0 \\
    & &3 & 85.0 &  90.0 &  90.0 & 0.0 & 0.0 & 0.0$^{4)}$ & 0.0 & 0.0 \\
    \cmidrule(r){2-11} &
    \multirow{3}{*}{DRIMA\cite{learning5disentangling}} &1 & 100 &  70.0 &  0.0 & 95.0 & 95.0 & 85.0 & 100 & 0.0\\
    & &2 & 100 &  60.0 &  70.0 & 90.0 & 100 & 100 & 80.0 & 0.0\\
    & &3 & 95.0 &  50.0 &  80.0 & 95.0 & 90.0 & 100 & 0.0 & 0.0\\
    \hline \hline
    \multicolumn{3}{c}{The total number of win-rate $\ge 80\%$}  &  16 &  3 &  2 & 9 & 6 & 4 & 2 & 0 \\
    \bottomrule
    \multicolumn{10}{l}{1) Takes total cumulative 2.41 million episode steps during training} \\
    \multicolumn{10}{l}{2) Takes total cumulative 7.85 million episode steps during training} \\
    \multicolumn{10}{l}{3) Takes total cumulative 8.15 million episode steps during training} \\
    \multicolumn{10}{l}{4) Takes total cumulative 8.10 million episode steps during training} \\
    \end{tabular}
    }%
\end{table}

\newpage

\newpage
\section{Ablation Study}
\label{app:Ablation_study}

We additionally investigate the reason why some algorithms perform worse in offensive scenarios than the defensive scenarios. We point out that the exploration is a main issue to increase performance in offensive scenarios. For this reason, we try to find the optimal hyper-parameter setting with respect to exploration perspective by sweeping out $\epsilon$-greedy steps and adjusting risk-sensitiveness.

\subsection{Optimization for Exploration}
To determine the optimal hyperparameters for algorithms that do not solve scenarios requiring exploration, we conducted multiple types of exploration-exploitation trade-off balancing in value-based, distribution-based, and policy-based algorithms with $\epsilon$-greedy decaying, various risk levels, and entropy term control. We choose QMIX, DFAC (DMIX, DDN), and MASAC for each category that can be trained in parallel.

\subsubsection{Value-based Algorithms}
\begin{figure}[!h]{
    \centering
        \begin{subfigure}{0.32\columnwidth}
            \includegraphics[width=\columnwidth]{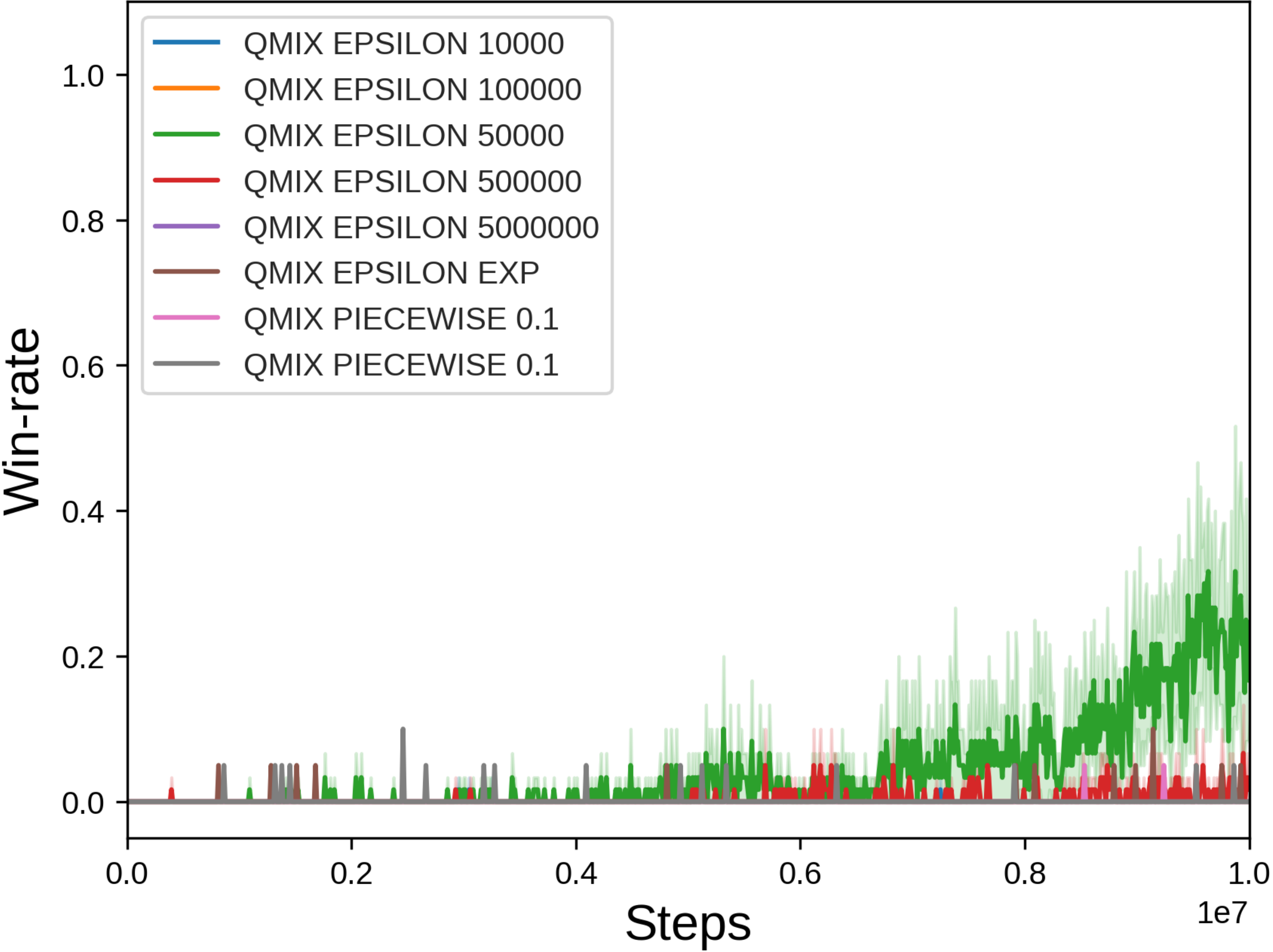}
            \caption{Defense outnumbered}
            \label{fig:eps_def-shd}
        \end{subfigure}%
        \hspace{0.05cm}
        \begin{subfigure}{0.32\columnwidth}
            \includegraphics[width=\columnwidth]{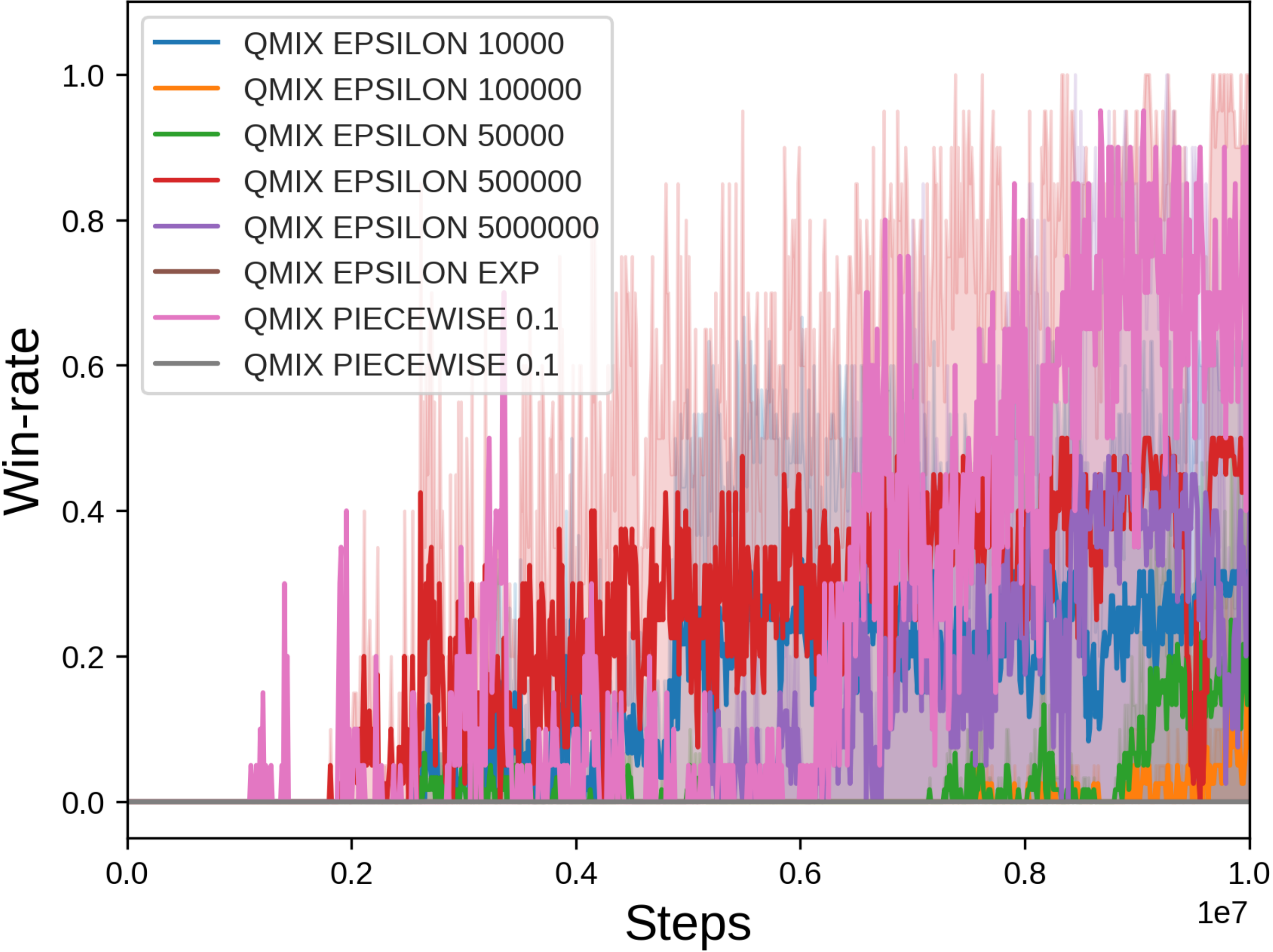}
            \caption{Offense distant}
            \label{fig:eps_off-shd}
        \end{subfigure}%
        \hspace{0.05cm}
        \begin{subfigure}{0.32\columnwidth}
            \includegraphics[width=\columnwidth]{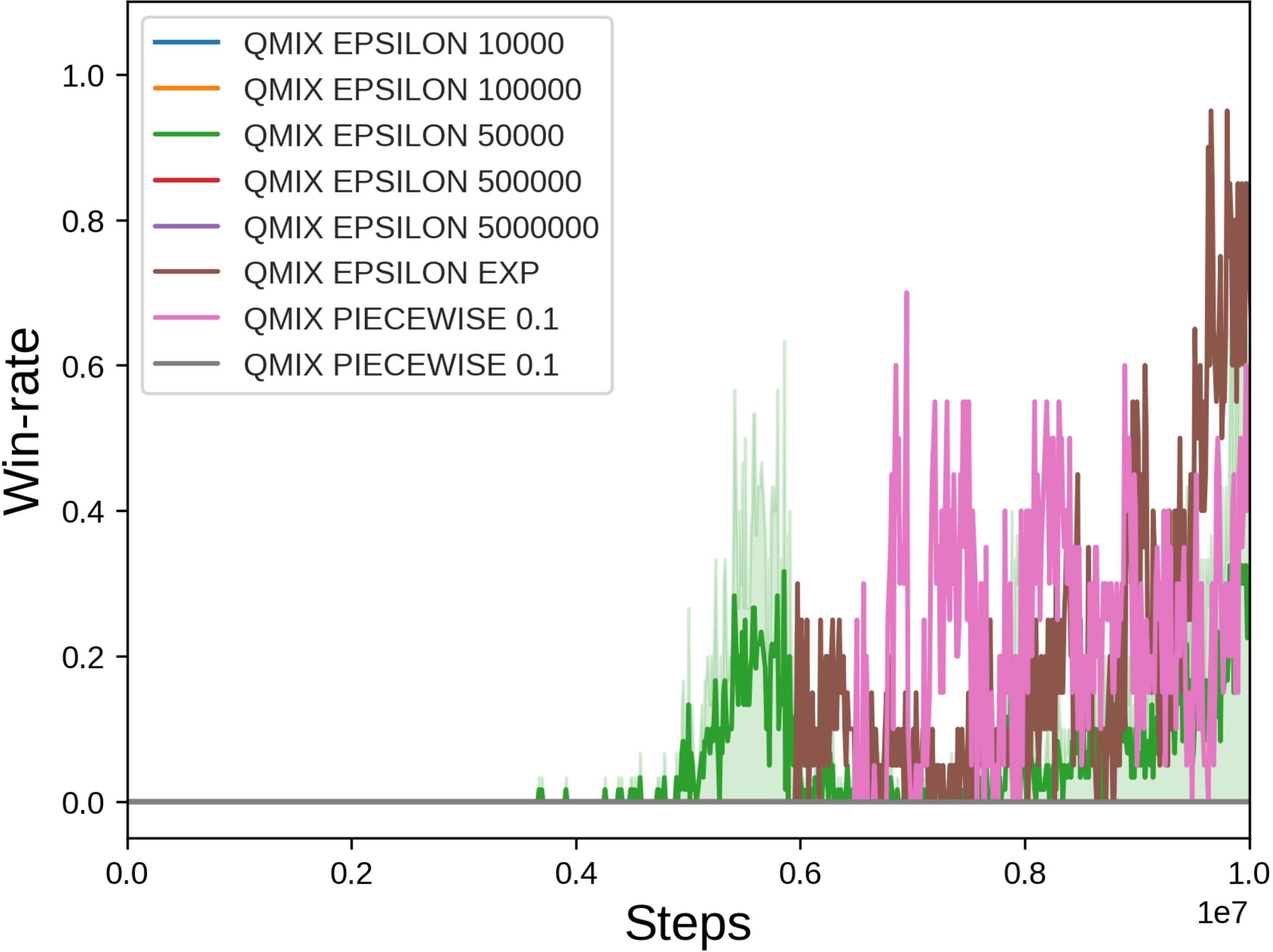}
            \caption{Offensive complicated}
            \label{fig:eps_off-uhd}
        \end{subfigure}%
    \caption{Sweeping out $\epsilon$-decaying steps with QMIX algorithm. The numbers next to the EPSILON letter means the decaying steps. In piece-wise setting, decaying line consist of combination of two linear line and set $\epsilon$ decaying point to 0.1. Arriving that point, again decay epsilon to 0.5. The number `10000 to 50000' in label means how to use piece-wise decaying strategy That is, for example, `10000 to 50000' means decay epsilon from 1.0 to 0.1 for 10000 steps and then decay from 0.1 to 0.05 for 40000 steps}
    \label{fig:eps}   
    }
\end{figure}

We explore for the optimal hyperparamater for exploration with $\epsilon$-greedy in space \{10k, 50k, 100k, 500k, 5000k\} for QMIX, particularly in difficult defensive and offensive scenarios that demand extensive explorations. In addition to the linear decaying approach, we utilize the exponential and piece-wise decaying methods for a total of 50k steps. We discover that there is no discernible trend in adjusting $\epsilon$-greedy, exponential, and piece-wise decaying steps for exploration in scenarios requiring exploration. We find that more exploratory factors are required for value-based algorithms to win offensive scenarios. The results are depicted in Figure \autoref{fig:eps}.

\subsubsection{Distribution-based Algorithms}
In this section, we discuss the main concept of Distribution-based algorithms, followed by the outcomes of controlling risk-sensitivity.
\label{app:exp_risk_based}
\paragraph{Quantile function} 
Distributioanl deep reinforcement learning gained popularity when \cite{bellemare2017distributional} proposed the C51 method, which generates output as a distribution of return given state and action. In this domain, algorithms compute TD-error between distributional bellman's updated distribution and the current distribution of return using the Wasserstein metric. It differs among methods, but the majority of them estimate the return distribution in order to compute the Wasserstein distance between the present distribution and the desired distribution (which is updated by distributional bellman update). Consequently, we may pretend that we have a model that approximates a quantile function with domain [0, 1] and return value range $(-\inf, +\inf)$. A model's output may approximate the inverse of the cumulative density function.

\paragraph{Risk-sensitive criteria} 
We can observe that utilizing return distributions can result in risk-sensitive reinforcement learning. When a model approximates the quantile function, sampling $\tau$ uniformly from $\mathcal{U}[0, 0.25]$ yields relatively low output values. This is known as risk-averse behavior since we anticipate low rewards. If, on the other hand, we randomly pick $\tau$ from $\mathcal{U}[0.75, 1]$, we will obtain relatively high values, which may be viewed as an optimistic behavior. This is referred to as risk-seeking behavior. If we randomly choose $\tau$ from $\mathcal{U}[0, 1]$, we get a risk-neutral distribution that is not skewed toward risk-averse or risk-seeking criteria. Also, we may consider risk-sensitive behavior from the perspective of variance, also known as uncertainty. Lower variance of an action's return distribution indicates lower uncertainty, which indicates risk-averse action, whereas higher variance of an action's return distribution indicates greater uncertainty, which indicates risk-seeking action, because it is extremely difficult to predict what return we will receive. With these distributional RL characteristics, we may train a model with risk-sensitive behavior criteria.

\begin{figure}[ht!]{
    \centering
        \begin{subfigure}{0.33\columnwidth}
            \includegraphics[width=\columnwidth]{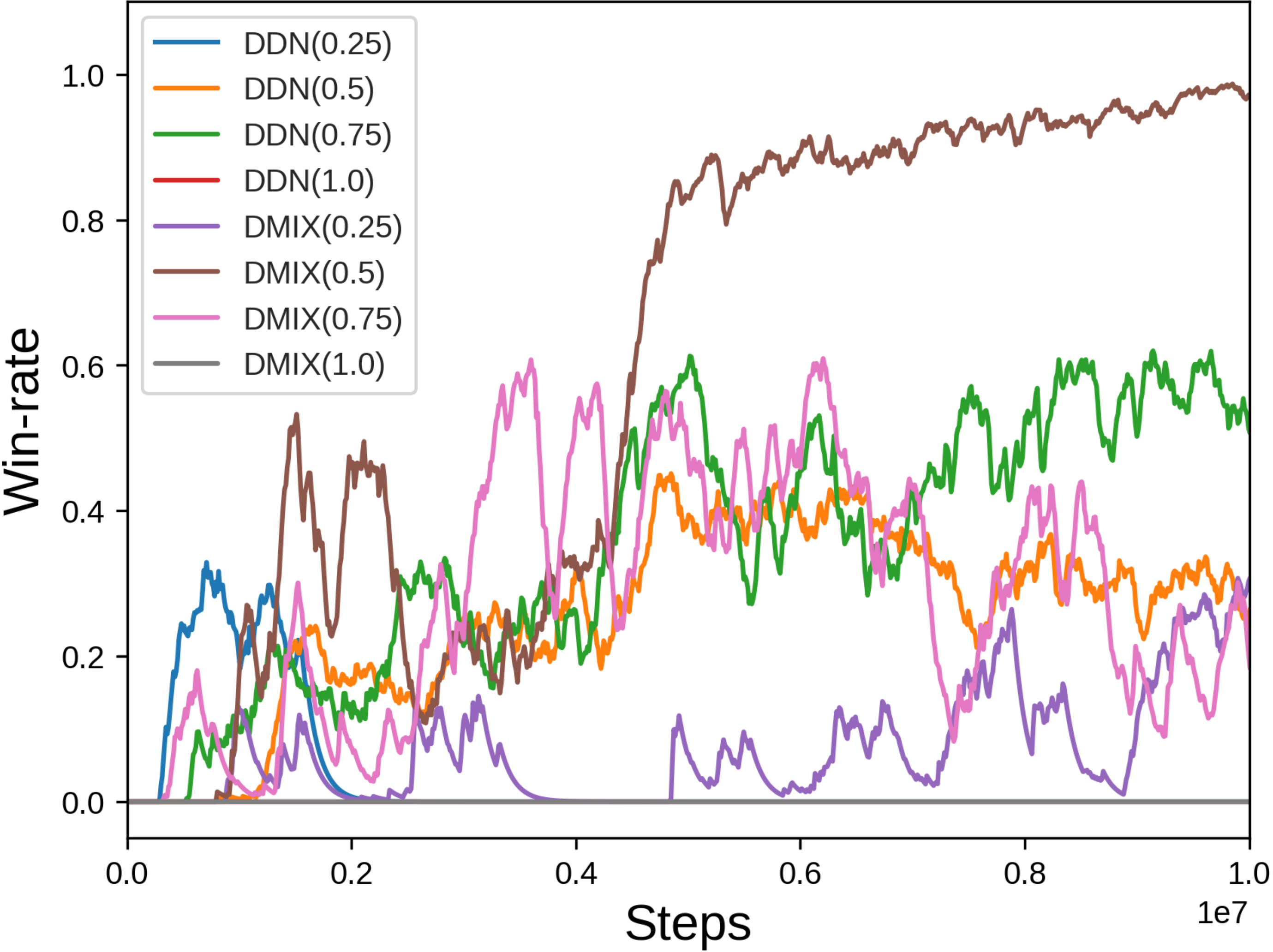}
            \caption{Defense Infantry}
            \label{fig:dfac_Defense Easy}
        \end{subfigure}%
        \begin{subfigure}{0.33\columnwidth}
            \includegraphics[width=\columnwidth]{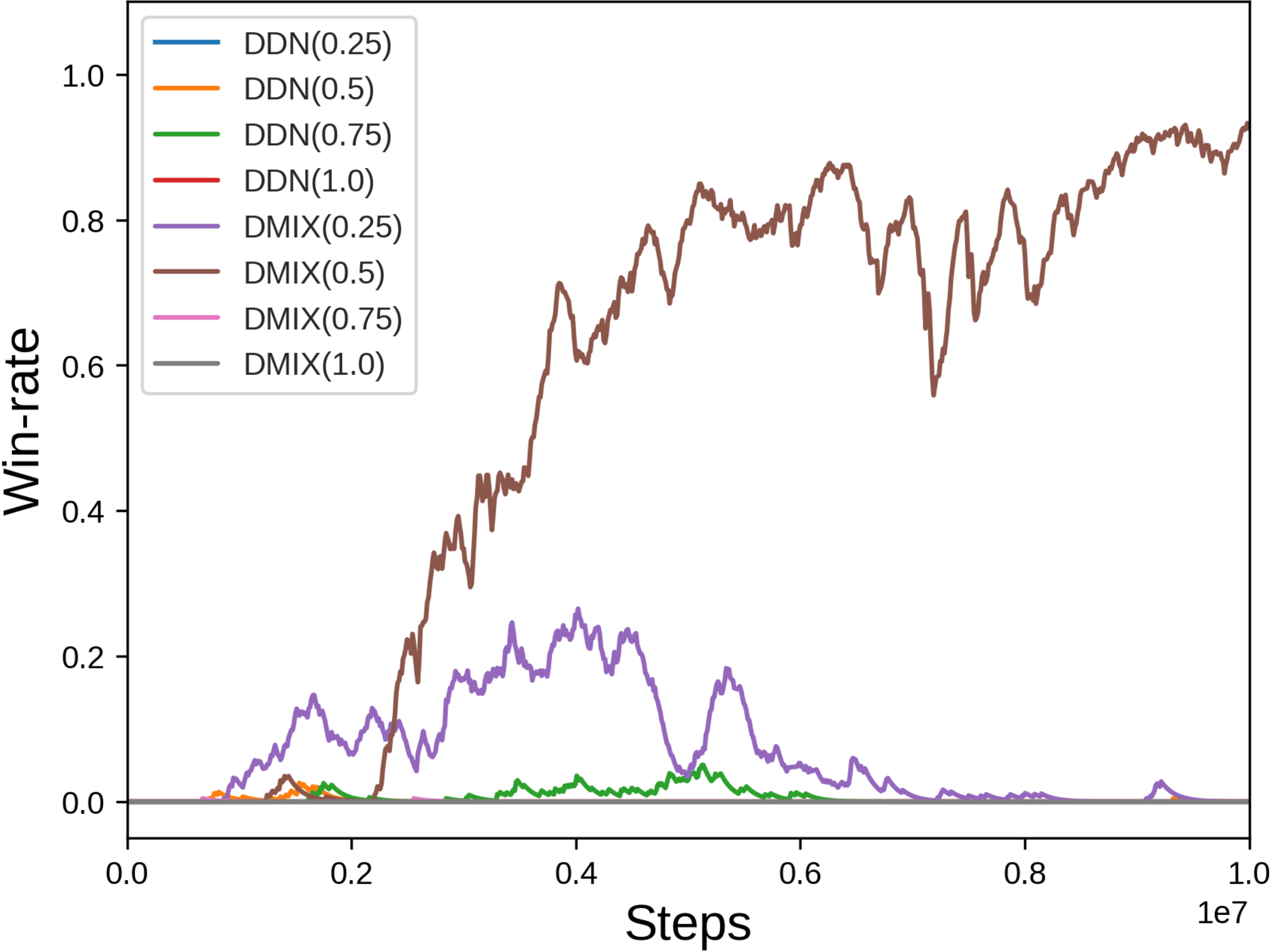}
            \caption{Defense Armored}
            \label{fig:dfac_Defense Hard}
        \end{subfigure}%
        \begin{subfigure}{0.33\columnwidth}
            \includegraphics[width=\columnwidth]{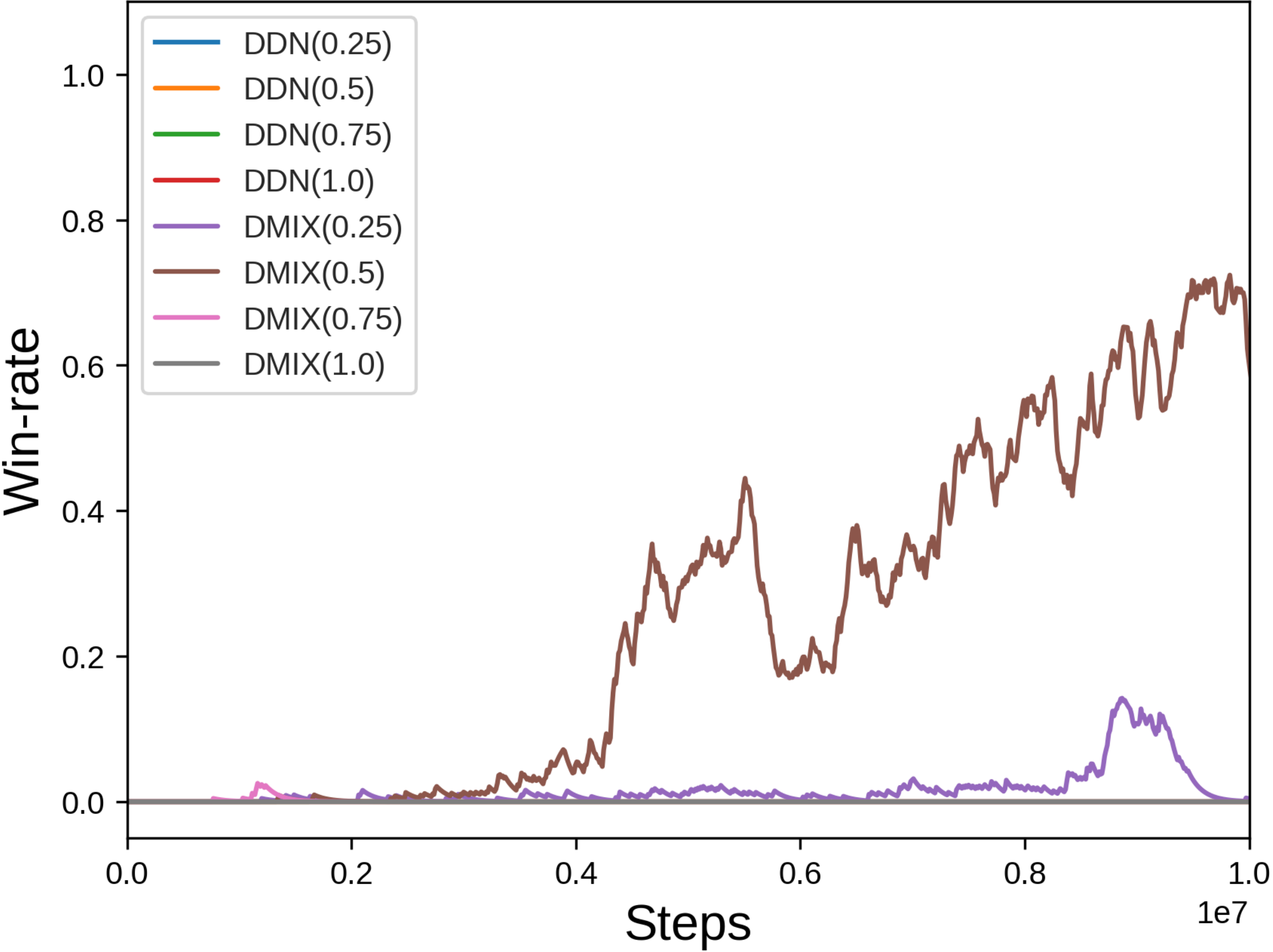}
            \caption{Defense Outnumbered}
            \label{fig:dfac_Defense Super Hardr}
        \end{subfigure}%

        \begin{subfigure}{0.33\columnwidth}
            \includegraphics[width=\columnwidth]{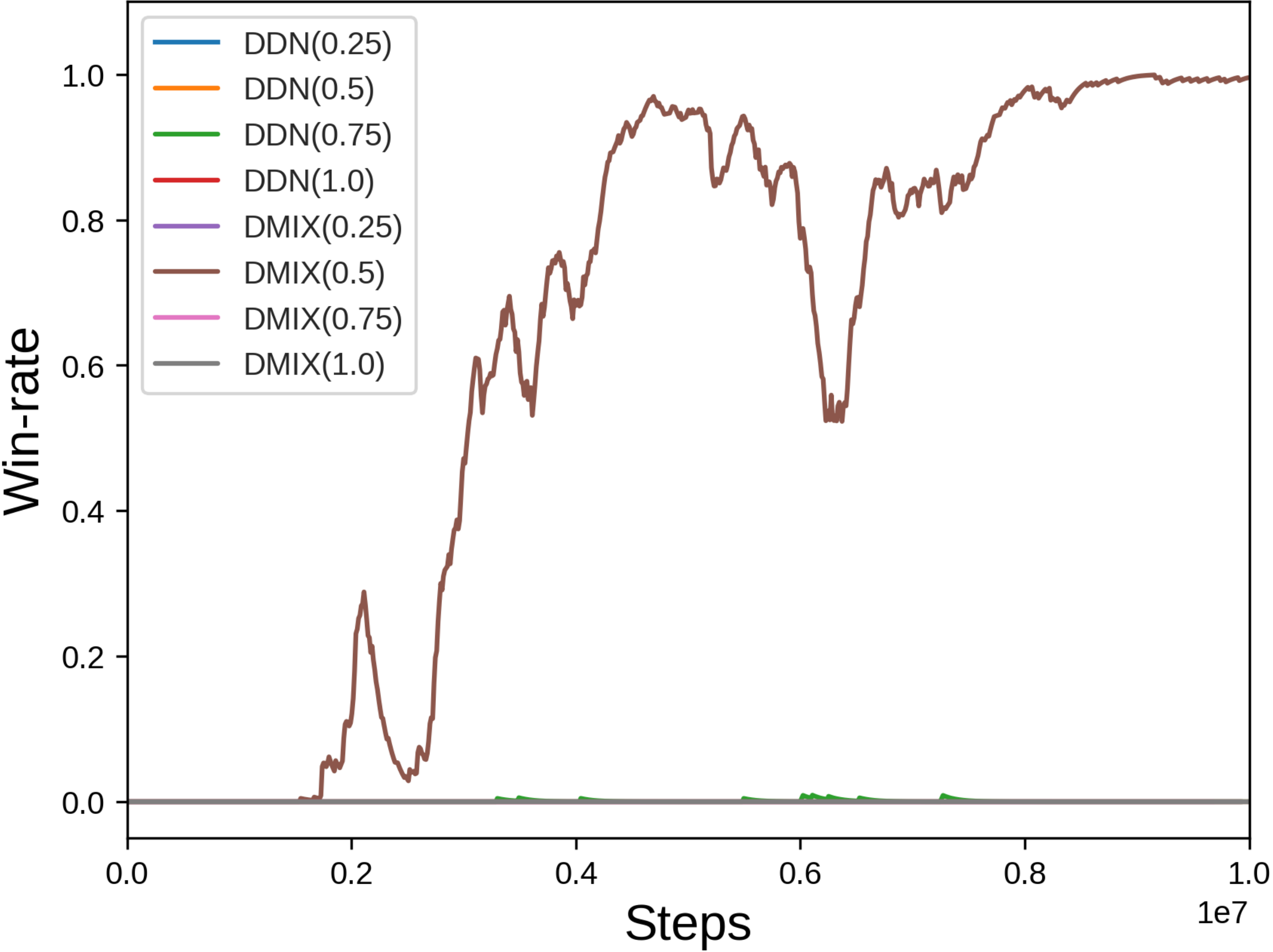}
            \caption{Offense Near}
            \label{fig:dfac_offense_easy}
        \end{subfigure}%
        \begin{subfigure}{0.33\columnwidth}
            \includegraphics[width=\columnwidth]{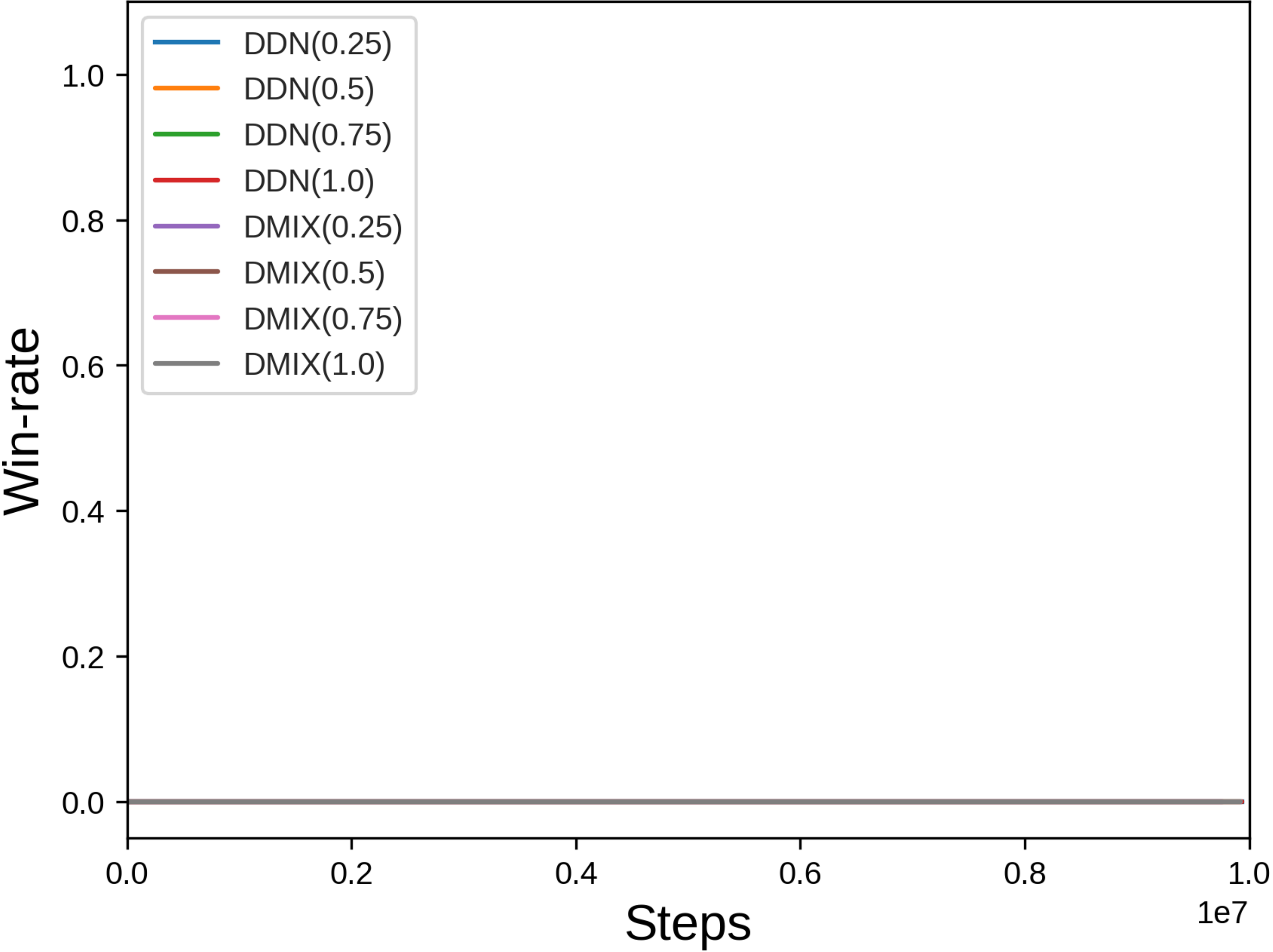}
            \caption{Offense Distant}
            \label{fig:dfac_offense_superhard}
        \end{subfigure}%
        \begin{subfigure}{0.33\columnwidth}
            \includegraphics[width=\columnwidth]{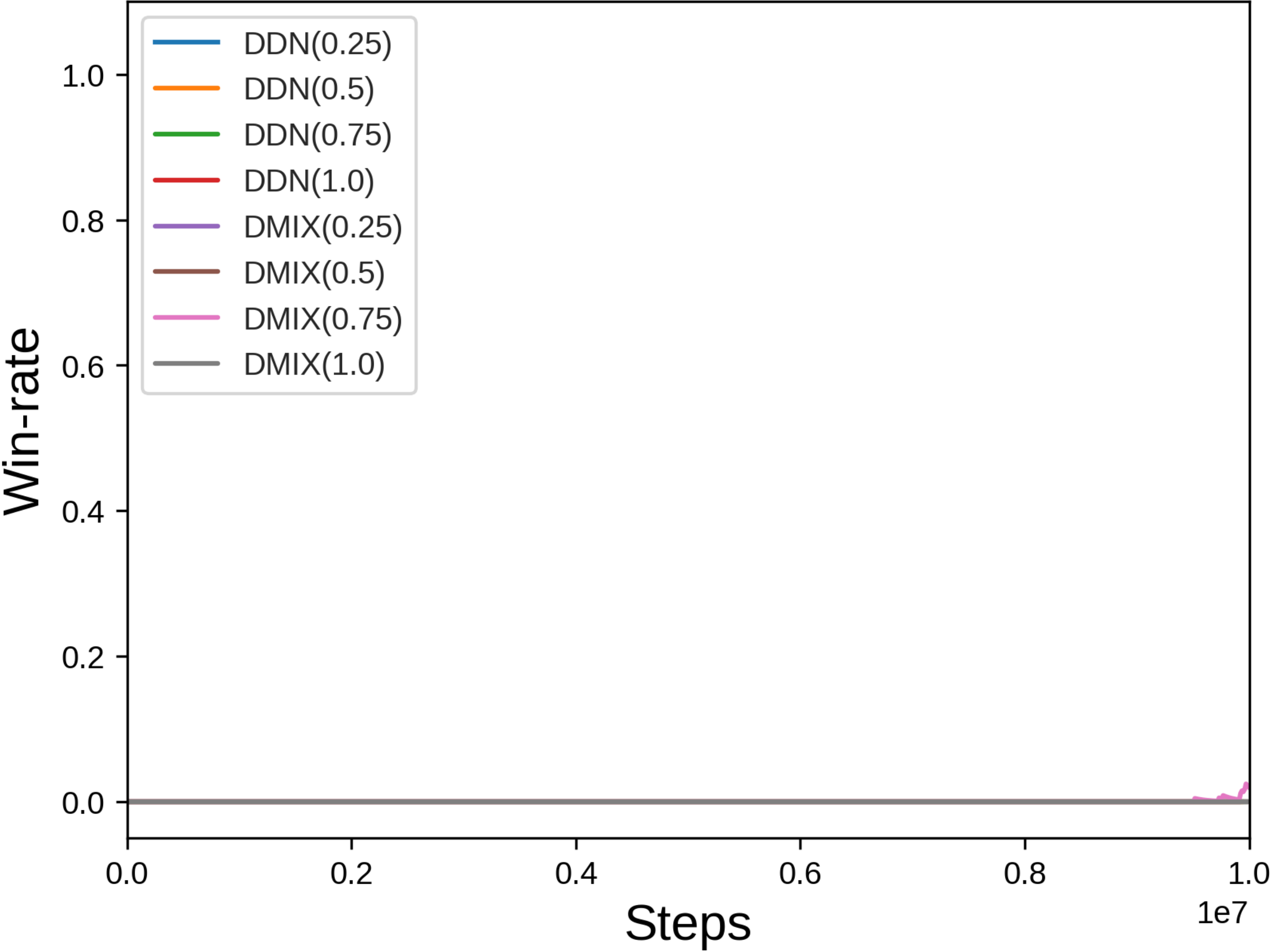}
            \caption{Offense Complicated}
            \label{fig:dfac_offense_ultrahard}
        \end{subfigure}%

    \caption{Risk-sensitive experiments with DFAC algorithm according to the sampling methods. DDN(0.25) and DDN(0.5) means sampling $\tau$ $\sim$ $\mathcal{U}([0, 0.25])$, $\mathcal{U}([0.25, 0.5])$ each using DDN algorithm.}
    \label{risk-sensitive_exp}
}
\end{figure}

\paragraph{Risk-sensitive experiments} We conducted risk-sensitive criteria experiments in our setting, and we adapted DFAC variants for distributional RL named DDN and DMIX. In DRIMA algorithm which is also risk-based distributional RL, we just follow the default setting of the hyperparameter not conducted with further experiments. We divide the sampling quantile fractions into 4 portion ([0, 0.25], [0.25, 0.5], [0.5, 0.75], [0.75, 1.0]) from uniform distribution each $\tau \sim \mathcal{U}([\cdot, \cdot])$. We name each portion as risk-averse, risk-neutral-averse, risk-neutral-seeking and risk-seeking sampling. The results are shown in Figure \ref{risk-sensitive_exp}. Looking the results, we notice that risk-neutral-averse criteria with DMIX work best in most of the setting except some scenarios that no training has developed. Also, we can find that risk-seeking criteria in DMIX and DDN never worked in this setting. With these experiments, we knew that distributional RL is very fragile to the sampling $\tau$ or risk-sensitive criteria so as to make the needs for research about the risk-sensitive behavior in distributional reinforcement learning which is very critical in performance. 

\subsubsection{Policy-based algorithms}
\label{app:policy_based_entropy}
\begin{figure}[!h]{
    \centering
        \begin{subfigure}{0.32\columnwidth}
            \includegraphics[width=\columnwidth]{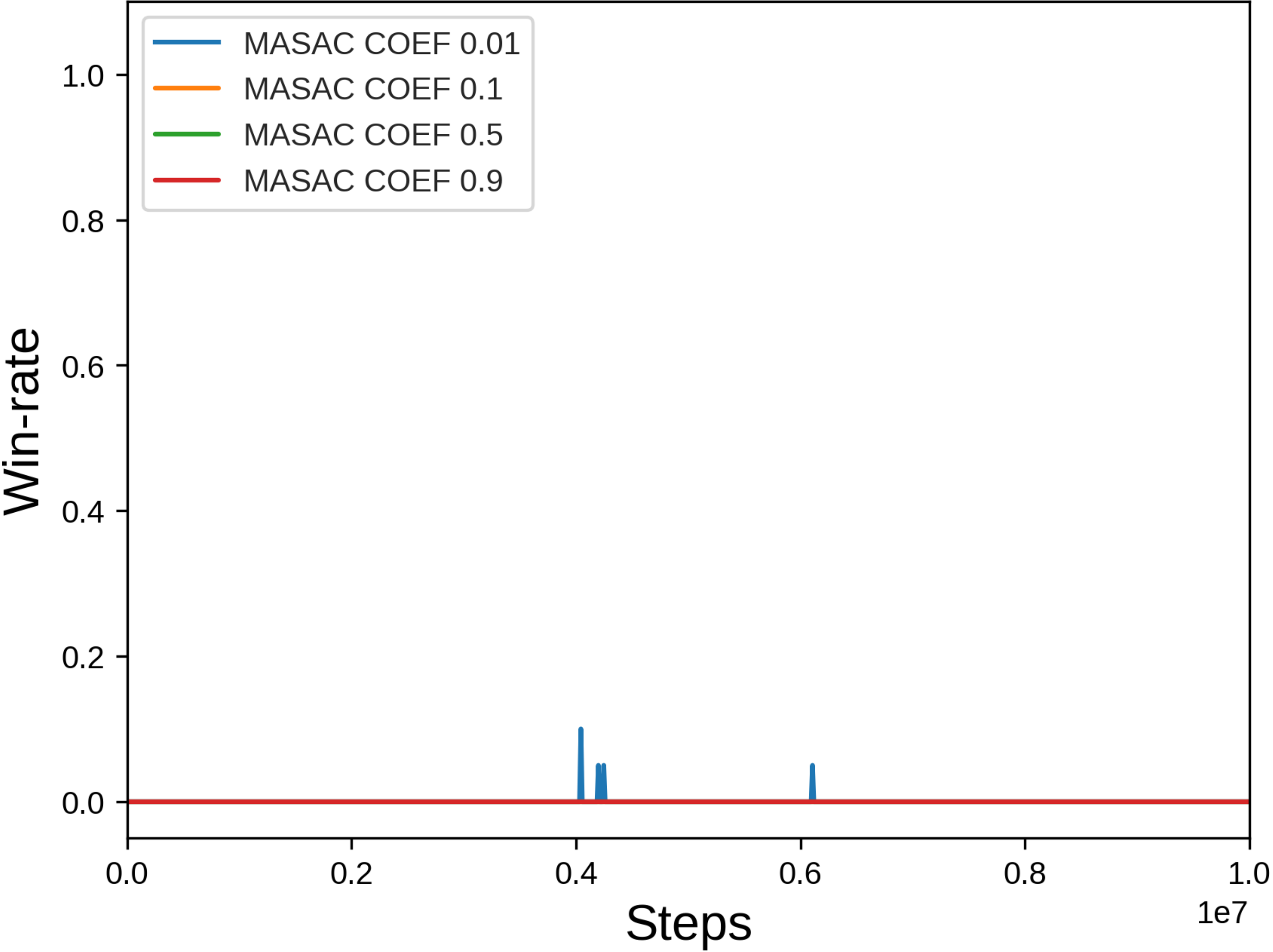}
            \caption{Defense Armored}
            \label{fig:coef_def-shd}
        \end{subfigure}%
        \hspace{0.05cm}
        \begin{subfigure}{0.32\columnwidth}
            \includegraphics[width=\columnwidth]{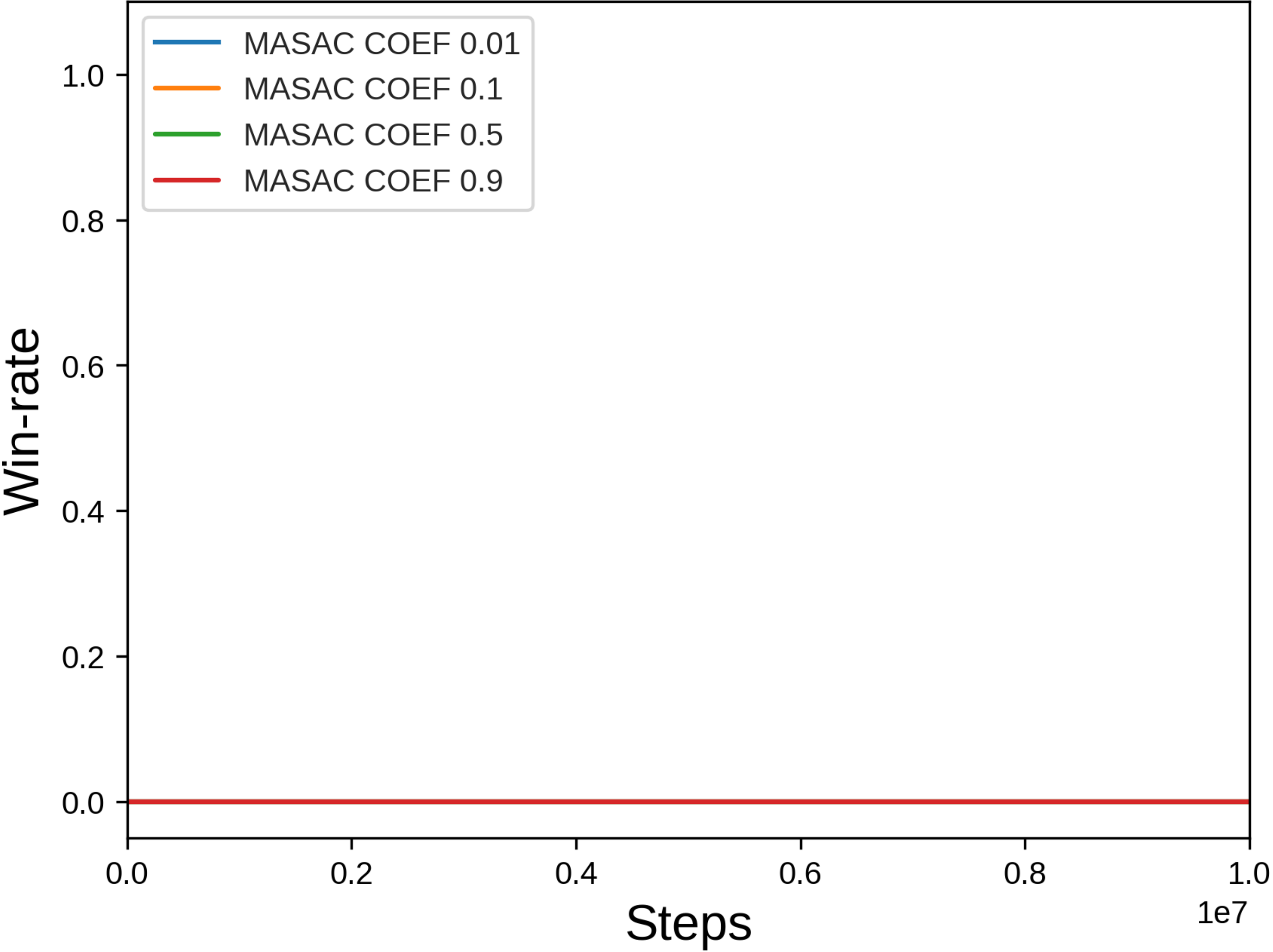}
            \caption{Offense Distant}
            \label{fig:coef_off-shd}
        \end{subfigure}%
        \hspace{0.05cm}
        \begin{subfigure}{0.32\columnwidth}
            \includegraphics[width=\columnwidth]{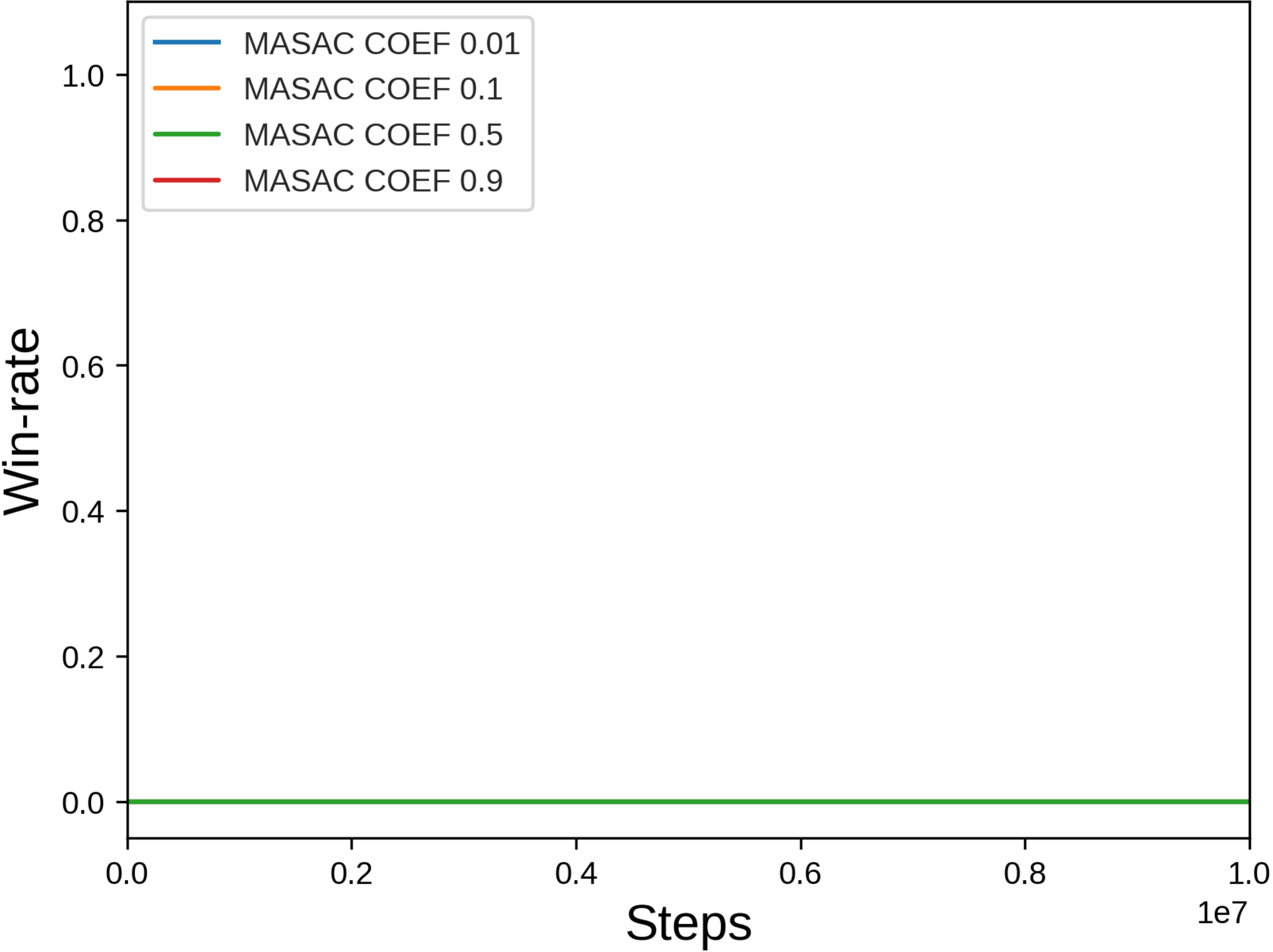}
            \caption{Offense Complicated}
            \label{fig:coef_off-uhd}
        \end{subfigure}%
    \caption{Sweeping out coefficient $\alpha$ of entropy term in loss function with MASAC algorithm. The numbers next to the COEF letter means the $\alpha$. As the coefficient is close to the 0 it indicates less exploration, vice versa.}
    \label{fig:coef}   
    }
\end{figure}

We choose MASAC as a representative model among policy-based algorithms since SAC was designed to improve exploration strategy by adding an entropy term to the loss function with coefficient $\alpha$, as described in \autoref{eqn:masac_policy}. The default value for $\alpha$ is 0.01, which favors exploration over exploitation. We set $\alpha$ space to the values \{0.01, 0.1, 0.5, 0.9\}. Results indicate that increasing the entropy term for exploration in our scenario has little effect, and algorithms may require additional parameters for exploration. The outcome is in \autoref{fig:coef}

\subsection{Reward Engineering}
\label{app:reward_engineering}

We evaluate the extent to the reward function engineering can solve the offense scenarios. The alternative reward function is designed to reward as agents get closer to the enemies. Since the offensive scenarios require agents to find the enemies before defeating them, agents with the explicit reward for finding the enemies to agents early in training can solve tasks sequentially. In detail, agents are rewarded by how much they get close to the enemies with respect to euclidean distance in the map until the 100k training time step. After that, they get the basic reward that is used in SMAC, SMAC$^{+}$. The equation of the alternative reward function is as follows

\begin{equation}
    \begin{aligned}
    \label{eqn:reward_function}
        R_{alt}(h, u) = \Delta \frac{1}{|e|}\sum_{a}\sum_{e}distance(pos_{a}-pos_{e})
    \end{aligned}
\end{equation}
where pos is 2D cartesian coordinates, $a$ and $e$ denote agents and enemies respectively.

\begin{figure}[!h]{
    \centering
        \begin{subfigure}{0.30\columnwidth}
            \includegraphics[width=\columnwidth]{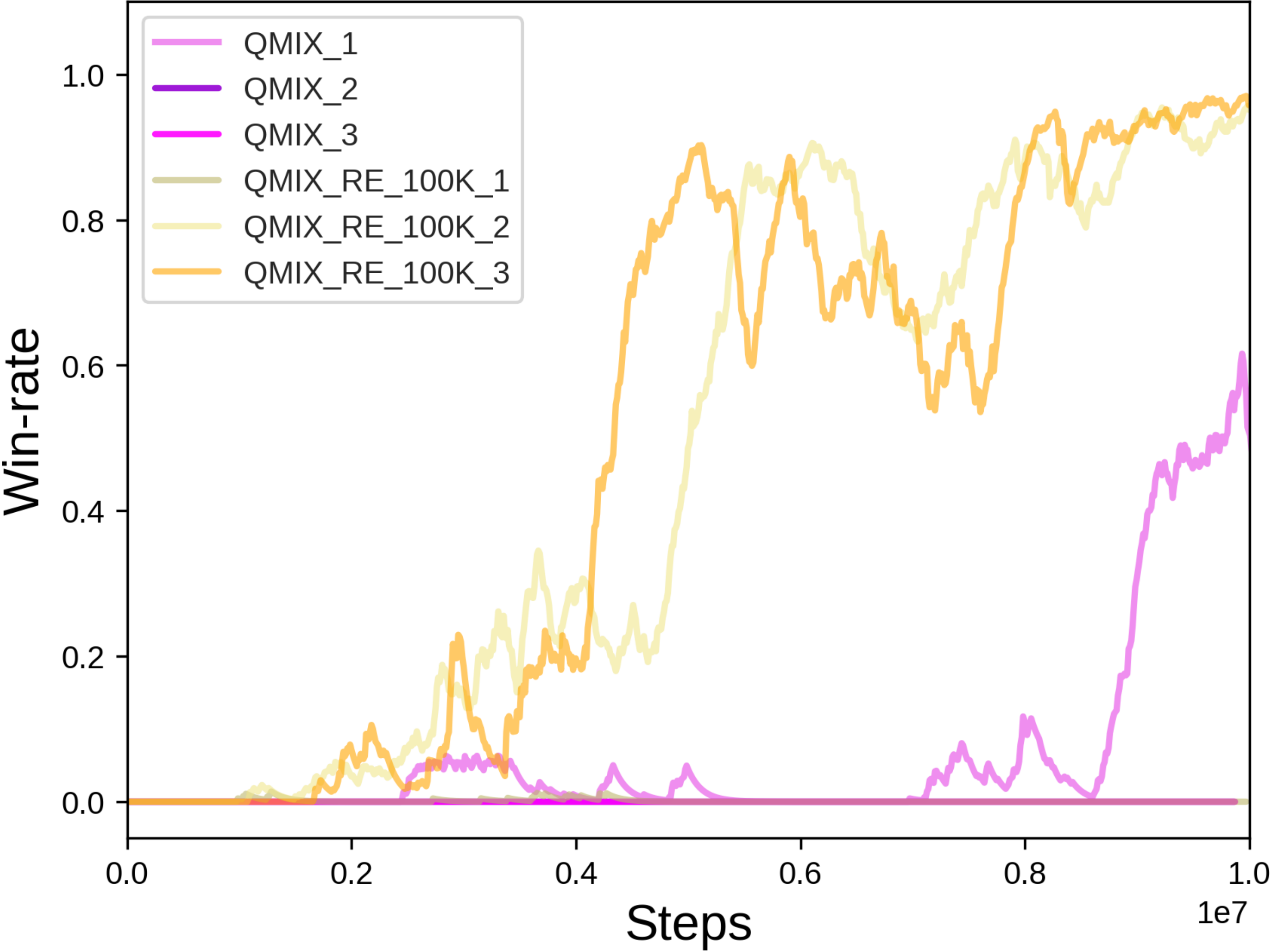}
            \caption{QMIX Training Curve}
            \label{fig:reward_engineered_qmix}
        \end{subfigure}%
        \hfill
        \begin{subfigure}{0.33\columnwidth}
            \includegraphics[width=\columnwidth]{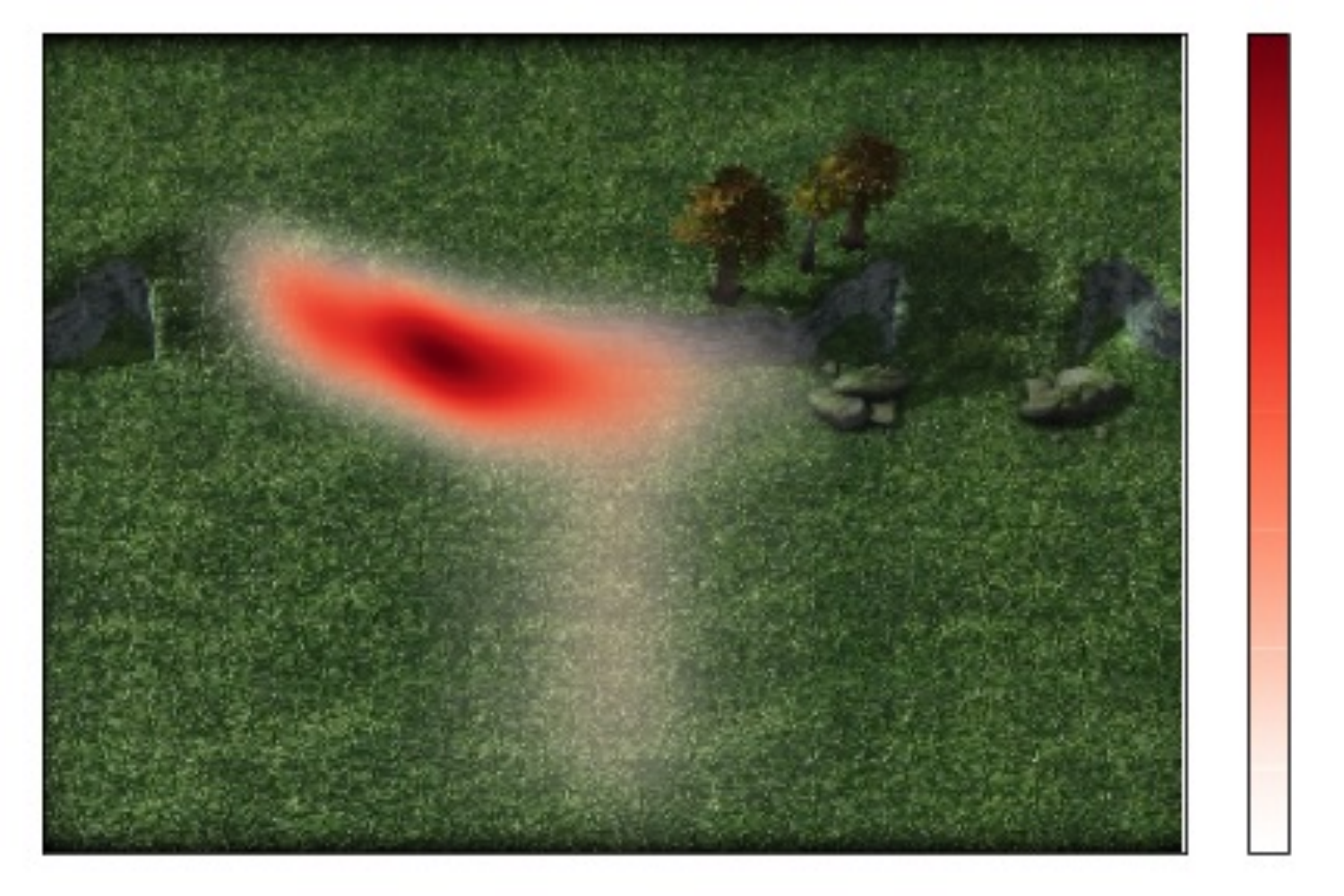}
            \caption{Early stage}
            \label{fig:reward_engineered_qmix_early}
        \end{subfigure}%
        \hfill
        \begin{subfigure}{0.33\columnwidth}
            \includegraphics[width=\columnwidth]{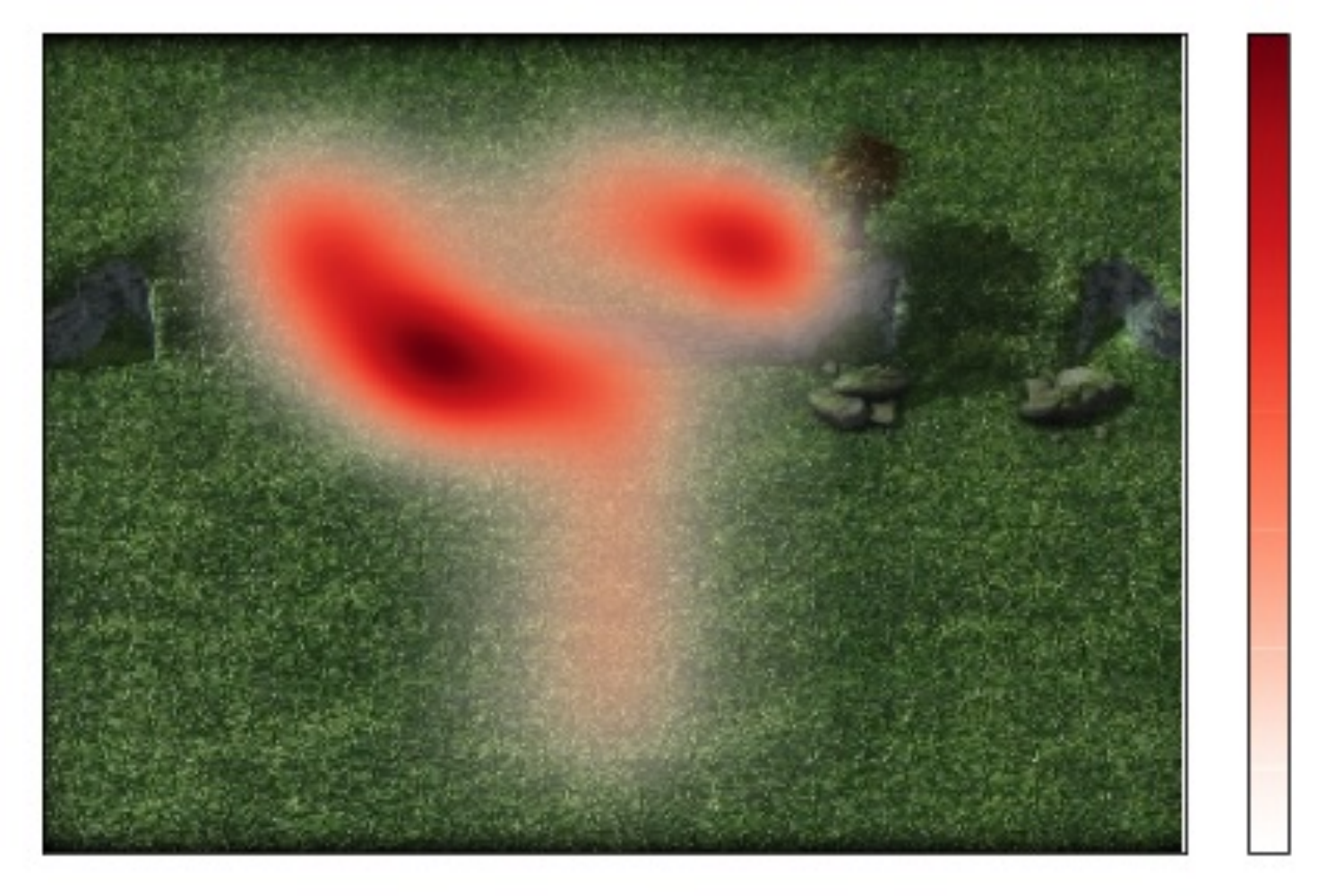}
            \caption{Last stage}
            \label{fig:reward_engineered_qmix_late}
        \end{subfigure}%
    \caption{(a) is the result of Reward engineering experiment. The line QMIX RE 100K indicate QMIX trained with reward engineering until 100k training time step. (b), (c) are heat-maps of all agents movements at 500k, 10m training time step each.}
    \label{fig:reward_engineering}   
    }
\end{figure}

We test QMIX in \texttt{Off\_distant} with parallel episodic buffer and identical hyper-parameter setting. As in \autoref{fig:reward_engineered_qmix}, the results show that reward engineering makes significant improvements both in final win-rate and convergence speed compared to QMIX with a basic reward function. \autoref{fig:reward_engineered_qmix_early} illustrate agents find the enemies properly in the early stage of training. Even though the reward function is changed drastically at 100k time step, the agents utilize the information learned through the alternative reward function until the end of training \autoref{fig:reward_engineered_qmix_late}.

\subsection{Heat-map Analysis of Offensive Scenarios}

\begin{figure}[!h]{
    \centering
        \begin{subfigure}{0.65\columnwidth}
            \includegraphics[width=\columnwidth]{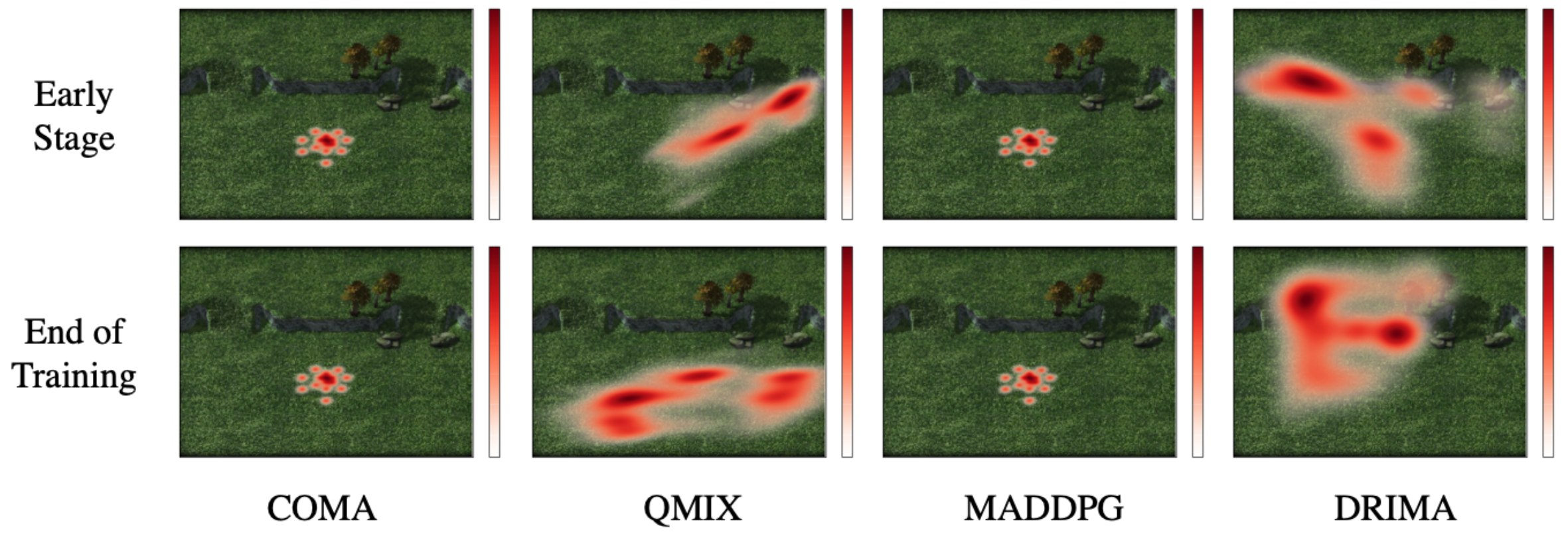}
            \caption{\texttt{Off\_near} scenario}
            \label{app_heatmap_off_near}
        \end{subfigure}%
        
        \begin{subfigure}{0.65\columnwidth}
            \includegraphics[width=\columnwidth]{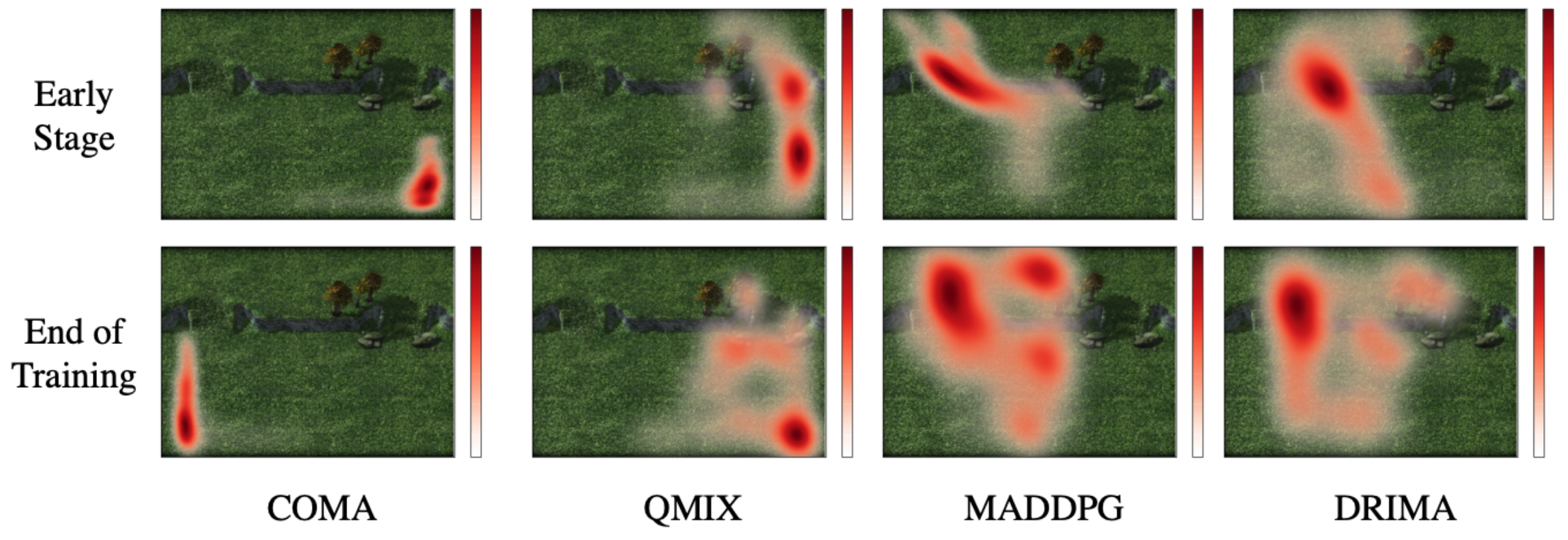}
            \caption{\texttt{Off\_distant} scenario}
            \label{app_heatmap_off_distant}
        \end{subfigure}%
        
        \begin{subfigure}{0.65\columnwidth}
            \includegraphics[width=\columnwidth]{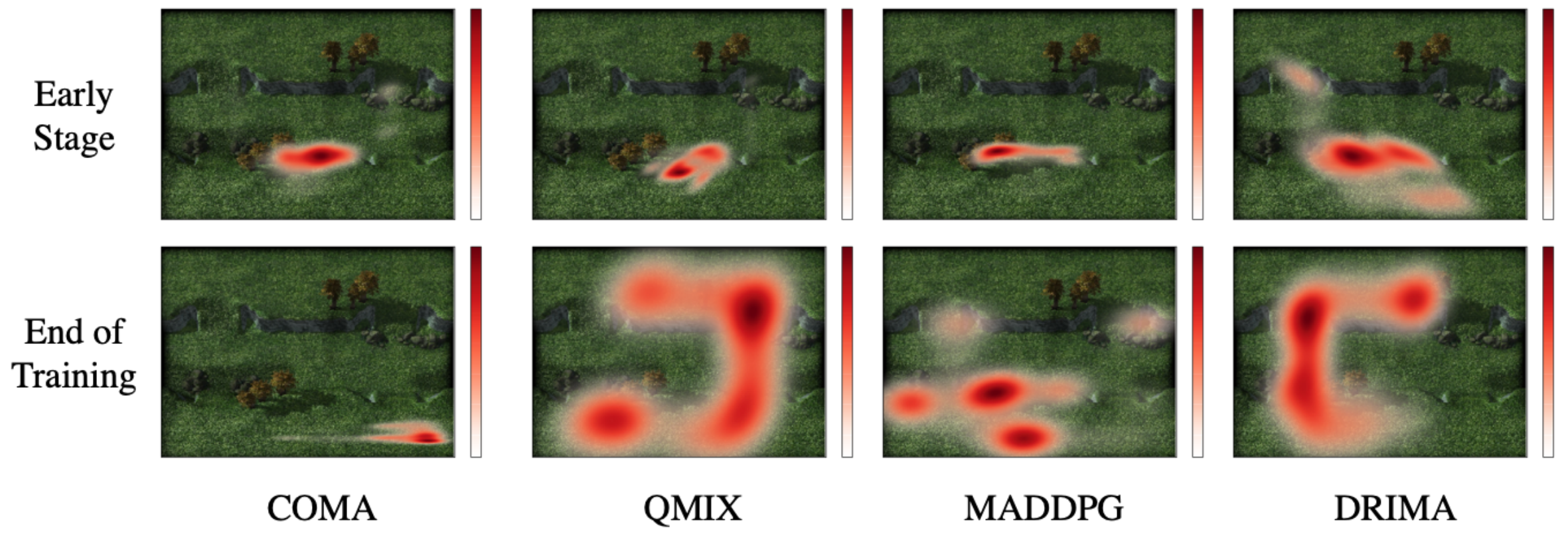}
            \caption{\texttt{Off\_complicated} scenario}
            \label{app_heatmap_off_complicated}
        \end{subfigure}%
        
        \begin{subfigure}{0.65\columnwidth}
            \includegraphics[width=\columnwidth]{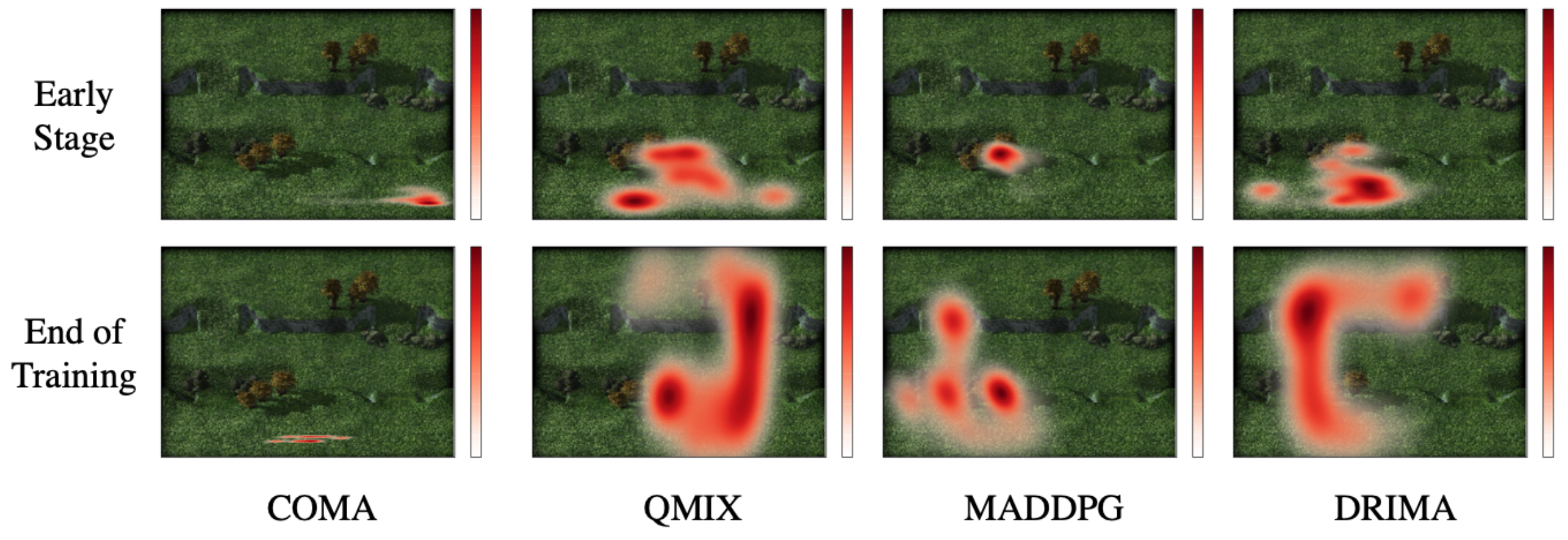}
            \caption{\texttt{Off\_hard} scenario}
            \label{app_heatmap_off_hard}
        \end{subfigure}%
        
        \begin{subfigure}{0.65\columnwidth}
            \includegraphics[width=\columnwidth]{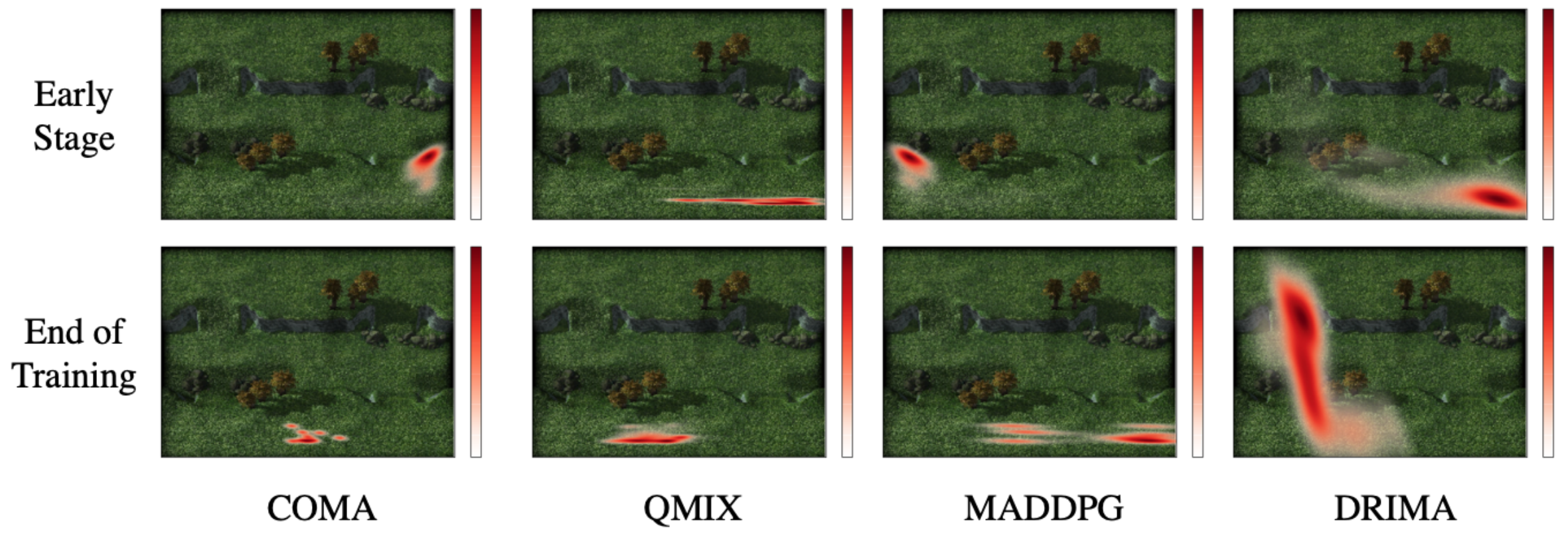}
            \caption{\texttt{Off\_superhard} scenario}
            \label{app_heatmap_off_superhard}
        \end{subfigure}%

    \caption{Heat-maps of agents movement by COMA, QMIX, MADDPG and DRIMA in all offensive scenarios. All baselines are trained on the sequential episodic buffer. The result of early stage is captured at six hundred thousands steps while The end of training is 5 million cumulative steps.}
    \label{app_heatmap_off}
}
\end{figure}

We argue that heat-maps of agent movement in offensive scenarios prove efficient exploration of MARL algorithms in \autoref{off_complicated_heatmap}. In this subsection, we show all heat-maps generated by four algorithms like COMA, QMIX, MADDPG, and DRIMA trained on the sequential episode buffer. As previously indicated, as the number of training steps increases, all agents (lower side) must move forward to the enemies positioned on the hill (upper side). As shown in \autoref{app_heatmap_off}, we identify that DRIMA can make all agents go for the enemies in all scenarios meanwhile, the other algorithms do not succeed in making agents move forward in some scenarios. When compared of \autoref{table:app_final_winrate_episode} and \autoref{app_heatmap_off}, we can see that the win-rate performance at the test phase is highly correlated to agents going for enemies without direct incentives. Therefore, we claim that the exploration capability of MARL algorithms enables agents to train to discover enemies even though an algorithm at early stage does not move agents forward to enemies. 

Additionally, we observe that these heat-maps might be used to determine the complexity of offensive scenarios. In the \texttt{Off\_near}, \texttt{Off\_near}, two algorithms, at least, seems to find enemies, however all algorithms except DRIMA fail to find enemies in the \texttt{Off\_complicated}, \texttt{Off\_hard} and \texttt{Off\_superhard}. When you see the results of QMIX in the  \texttt{Off\_complicated} and \texttt{Off\_hard}, the agents appear to be approaching enemies, but they are not capable of precisely locating enemies. In these situations, where the enemies are scattered two-sided in front of each entrance of the hill, QMIX removes just one-sided opponents and does not find the remaining regions. On the other hand, DRIMA perfectly complete to locate enemies scattered and eliminate them in this situation. Especially in the \texttt{Off\_superhard}, the other algorithms do not completely train, but DRIMA makes agents move forward to the opponents and begin to beat them. Nonetheless, as seen in \autoref{table:app_final_winrate_episode}, DRIMA attains about 10\% win-rate performance on \texttt{Off\_superhard}, leading us to conclude that a longer training horizon is necessary in order to simultaneously learn fine-manipulation and enemy detection.

\end{document}